\documentclass[fleqn,11pt]{book}
\usepackage{geometry}

\usepackage[T1]{fontenc}
\usepackage{algorithm}
\usepackage{algpseudocode}
\usepackage{amsmath}
\usepackage{amssymb}
\usepackage{amsfonts}
\usepackage{array}
\usepackage[round,sort]{natbib}
\usepackage[french,english]{babel}
\usepackage{cases}
\usepackage{enumitem}
\usepackage{bibentry}
\usepackage{booktabs}

\usepackage{eurosym}
\usepackage{graphicx}
\usepackage{listings}
\usepackage{multicol}
\usepackage{multirow}
\usepackage{tikz}
\usepackage{nomencl}
\usepackage{nth}
\usepackage{pgfplotstable}
\usepackage{placeins}
\usepackage{longtable}
\usepackage{xcolor}
\usepackage{xspace}
\usepackage{setspace}
\usepackage[bookmarksnumbered=true,hidelinks,pdftex]{hyperref}
\usepackage{tabularx}
\usepackage{ragged2e}
\usepackage{blindtext}
\usepackage[thmmarks,amsmath]{ntheorem}
    \RequirePackage{latexsym}
    \RequirePackage{amssymb}
    \theoremnumbering{arabic}
    \theoremstyle{break}
    \theoremsymbol{}
    \theorembodyfont{\itshape}
    \theoremheaderfont{\normalfont\bfseries}
    \theoremseparator{}

\usepackage{algpseudocode}
\newcommand\Algphase[1]{%
\vspace*{-.7\baselineskip}\Statex\hspace*{\dimexpr-\algorithmicindent-2pt\relax}\rule{\textwidth}{0.4pt}%
\Statex\hspace*{-\algorithmicindent}\textbf{#1}%
\vspace*{-.7\baselineskip}\Statex\hspace*{\dimexpr-\algorithmicindent-2pt\relax}\rule{\textwidth}{0.4pt}%
}
\makeatletter
\newcommand{\algmargin}{\the\ALG@thistlm}
\makeatother
\newlength{\whilewidth}
\algdef{SE}[parWHILE]{parWhile}{EndparWhile}[1]
  {\parbox[t]{\dimexpr\linewidth-\algmargin}{%
     \hangindent\whilewidth\strut\algorithmicwhile\ #1\ \algorithmicdo\strut}}{\algorithmicend\ \algorithmicwhile}%
\newlength{\ifwidth}
\algdef{SE}[parIF]{parIf}{EndparIf}[1]
  {\parbox[t]{\dimexpr\linewidth-\algmargin}{%
     \hangindent\ifwidth\strut\algorithmicif\ #1\ \algorithmicthen\strut}}{\algorithmicend\ \algorithmicif}%
\newlength{\loopwidth}
\algdef{SE}[parLOOP]{parLoop}{EndparLoop}[1]
  {\parbox[t]{\dimexpr\linewidth-\algmargin}{%
     \hangindent\loopwidth\strut\algorithmicloop\ #1\ \strut}}{\algorithmicend\ \algorithmicloop}%
\algnewcommand{\parState}[1]{\State%
  \parbox[t]{\dimexpr\linewidth-\algmargin}{\strut #1\strut}}

\makeatletter
\newenvironment{breakablealgorithm}
  {
   \begin{center}
     \refstepcounter{algorithm}
     \hrule height.8pt depth0pt \kern2pt
     \renewcommand{\caption}[2][\relax]{
       {\raggedright\textbf{\fname@algorithm~\thealgorithm} ##2\par}%
       \ifx\relax##1\relax 
         \addcontentsline{loa}{algorithm}{\protect\numberline{\thealgorithm}##2}%
       \else 
         \addcontentsline{loa}{algorithm}{\protect\numberline{\thealgorithm}##1}%
       \fi
       \kern2pt\hrule\kern2pt
     }
  }{
     \kern2pt\hrule\relax
   \end{center}
  }
\makeatother

\newcommand{\outrank}{\succcurlyeq}

\newcommand{\Ck}[1][k]{\ensuremath{\mathcal{C}_{#1}}}

\newcommand{\dm}{DM}

\newcommand{\electre}{\textsc{Electre}}

\newcommand{\mcda}{MCDA}

\newcommand{\mrsort}{MR-Sort\xspace}

\newcommand{\msc}{MSC}

\newcommand{\ncs}{NCS}

\newcommand{\ie}{i.e.,\xspace}
\newcommand{\eg}{e.g.,\xspace}
\newcommand{\wrt}{w.r.t.\xspace}
\newcommand{\electreI}{\textsc{Electre~I}\xspace}
\newcommand{\etriI}{\textsc{Electre~I-Tri}\xspace}
\newcommand{\electreII}{\textsc{Electre~II}\xspace}
\newcommand{\electreIII}{\textsc{Electre~III}\xspace}
\newcommand{\etri}{\textsc{Electre-Tri}\xspace}

\newcommand{\epr}{\hfill \ensuremath{\Box}\mbox{}\\ }
\newcommand{\bpr}{\noindent \textbf{Proof.} \hspace{1 em}}

\newcommand{\bsk}{\bigskip}
\newcommand{\ssk}{\smallskip}
\newcommand{\msk}{\medskip}

\newcommand{\calA}{\ensuremath{\mathcal{A}}}
\newcommand{\calU}{\ensuremath{\mathcal{U}}}
\newcommand{\calF}{\ensuremath{\mathcal{F}}}

\newcommand{\calAU}{\ensuremath{\langle\mathcal{A,U}\rangle}}

\newcommand{\Fmin}{\ensuremath{\mathcal{F}_{min}}\xspace}
\newcommand{\Gmax}{\ensuremath{\mathcal{G}_{max}}\xspace}
\newcommand{\suffco}{sufficient coalitions}

\newcommand{\resp}{resp.\ }

\newcommand{\Mi}[1][i]{\ensuremath{M_{#1}}}
\newcommand{\mmi}[1][i]{\ensuremath{m_{#1}}}

\newcommand{\vsucc}{\ensuremath{V_{succ}(x)}}
\newcommand{\suclex}{\ensuremath{\succ_{lex}}}
\newcommand{\Bnull}[1][n]{\ensuremath{B_{#1}^0}}
\newcommand{\Bone}[1][n]{\ensuremath{B_{#1}^1}}

\newcommand{\ensn}{\ensuremath{\{1, \ldots, n\}}\xspace}

\newcommand{\R}{\ensuremath{\, R}}

\newcommand{\Pn}{\ensuremath{P(N)}}

\newcommand{\calG}{\ensuremath{\mathcal{G}}}

\newcolumntype{P}[1]{>{\centering\arraybackslash}p{#1}}
\newcolumntype{M}[1]{>{\centering\arraybackslash}m{#1}}

\newtheorem{theorem}{Theorem}
\newtheorem{definition}[theorem]{Definition}
\newtheorem{proposition}{Proposition}
\newtheorem{corollary}{Corollary}
\newtheorem{lemma}[theorem]{Lemma}
\newtheorem{hypothesis}{Hypothesis}
    \theoremheaderfont{\normalfont\bfseries}
    \theorembodyfont{\upshape}
\newtheorem{example}[theorem]{Example} 
\newtheorem{remark}[theorem]{Remark}



\pgfplotsset{compat=1.18}

\begin{document}

\frontmatter
\thispagestyle{empty}
\begin{center}
\begin{tabularx}{\textwidth}
	{>{\centering\arraybackslash}m{3.5cm}
	X
	>{\centering\arraybackslash}m{9.5cm}}
{
\includegraphics[width=3cm]{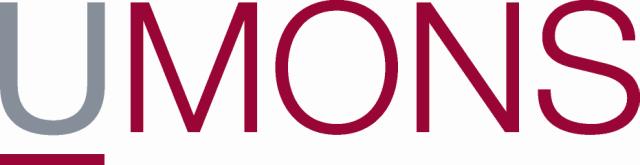}
}
&     &  
{\hrule width 0pt height 0.5cm depth 0cm \includegraphics[width=3cm]{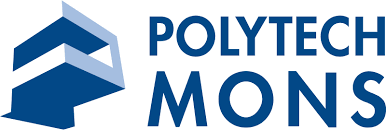}}

\end{tabularx}

\vspace{1cm}
\hrule
\vspace{1cm}
\textbf{\huge Elicitation of the parameters of a multi-criteria sorting model}
\vspace{1cm}
\hrule
\vspace{1cm}
by\\[0.5cm]
{\Large Eda Ersek Uyan{\i}k}\\[1cm]

{A dissertation submitted in partial fulfillment
of the requirements for the degree of
``PhD in Engineering Sciences and Technology'' of the University of Mons} \\

\vspace{1cm}
{\large July 2023}
\end{center}
\vspace{1.5cm}
Members of the jury:\\[0.2cm]
\begin{tabular}{ll >{\em}l}
    Prof. Yves De Smet & Universit\'e libre & Reviewer\\ 
    &  de Bruxelles & \\
    Prof. Pierre Manneback & Universit\'e de Mons & Reviewer\\    
    Prof. Thierry Marchant & Ghent University & Reviewer\\
    Prof. Vincent Mousseau & Universit\'e Paris-Saclay & Reviewer\\
    Prof. Xavier Siebert & Universit\'e de Mons & Reviewer,   President\\
    Prof. Philippe Fortemps & Universit\'e de Mons & Supervisor\\
    Prof. Marc Pirlot & Universit\'e de Mons & Co-Supervisor
\end{tabular}
\cleardoublepage
\thispagestyle{plain}
\begin{center}
    \textbf{Abstract}
\end{center}

\vspace{2cm}
In order to help decision-makers in making up their mind in difficult decision problems, multiple-criteria decision aiding (MCDA) proposes models and methods for taking into account several, generally conflicting, objectives. MCDA methods provide aid for different types of decision-making problems such as choice, ranking or sorting. No matter the type of decision making model, all of them involve various parameters that have to be elicited. 

\ssk The elicitation of the parameters of MCDA methods is not straightforward. All necessary information has to be taken from the decision-maker. For this, several direct and indirect elicitation methods are suggested in the literature. The most delicate point is eliciting these parameters in a rigorous way, with procedures that are in accordance with the spirit of the model used and do not put an excessive cognitive burden on the decision-maker. In this work, we concentrate on direct elicitation: we aim at providing a rigorous elicitation procedure for the parameters of the Non-Compensatory Sorting model (NCS), which is the only one that has been characterized by an axiomatic analysis.

\ssk Our study focuses on a basic version of NCS model, \ie a model with two categories and without vetoes. The parameters to be determined are the set of sufficient coalitions of criteria and the profile separating the two categories. Regarding the sufficient coalitions, we propose several questioning algorithms that engraft questions on classical graph exploration algorithms such as depth-first search. The DFS3 algorithm, which is the most promising, asks the questions in the middle of the branches of the depth-first search arborescence. This algorithm is also implemented and evaluated by simulation. Even though it is not optimal, the possibility of inserting information about the relative importance of the criteria, if available, is the main advantage of this questioning algorithm. For eliciting the limit profile, we design a questioning algorithm using dichotomy.

\ssk Questioning algorithms about sufficient coalitions of criteria can be optimal in the worst case or on average. An optimal algorithm in the worst case has been proposed by Hansel in the framework of monotone Boolean functions and guarantees that the number of questions never surpasses a lower bound. We implemented and evaluated this algorithm as a variant of the depth-first search approaches.

\ssk In order to find an optimal questioning strategy on average, we evaluated decision trees for up to four criteria. To achieve these computations we used several simplifying tools such as symmetries, recognition of already computed patterns, branch and bound, etc. We identified optimal questioning trees that minimize the number of questions on average.

\ssk On the other hand, we assess how more general is NCS as compared to MR-Sort, which is a particular case of the NCS model in which sufficient coalitions of criteria can be represented by criteria weights and a threshold. We do this by (exhaustive) simulation for up to six criteria and compute the proportion of the sets of sufficient coalitions that can be represented by 2-additive and 3-additive models.

\ssk The main difficulty for applying these tools to the real-world decision making, is the number and the difficulty of the questions to be asked. We suggest different trails for using or adapting the proposed rigorous elicitation methods to real decision aiding processes.

\cleardoublepage

\thispagestyle{plain}
\begin{otherlanguage}{french}
\begin{center}
    \textbf{Résumé}
\end{center}

\vspace{2cm}
Afin d'aider les décideurs face à des problèmes de décision difficiles, l'aide multicritère à la décision (AMCD) propose des modèles et des méthodes permettant de prendre en compte plusieurs objectifs, généralement contradictoires. De telles méthodes existent pour différents types de problèmes de décision comme le choix, le rangement ou le tri. Quel que soit le type de modèle de décision, tous contiennent des paramètres à éliciter.

\ssk L'élicitation des paramètres des méthodes AMCD n'est pas simple. Toutes les informations nécessaires doivent être obtenues du décideur. Pour cela, plusieurs méthodes d'élicitation directes et indirectes sont proposées dans la littérature. Le point le plus délicat est d'éliciter ces paramètres de manière rigoureuse, avec des procédures conformes à l'esprit du modèle utilisé et sans augmenter la charge cognitive du décideur. Dans ce travail, nous nous concentrons sur les méthodes d'élicitation directe pour une méthode de tri particulière : nous visons à fournir une procédure d'élicitation rigoureuse pour les paramètres du modèle de tri non-compensatoire (NCS) car c'est le seul qui a été caractérisé par une analyse axiomatique.

\ssk Notre étude concerne une version simple du modèle NCS, à savoir le tri en deux catégories et sans veto. Les paramètres à déterminer sont l'ensemble des coalitions suffisantes de critères et le profil séparant les catégories. En ce qui concerne les coalitions suffisantes, nous proposons plusieurs algorithmes de questionnement qui greffent des questions sur des algorithmes classiques d'exploration de graphes tels que la recherche en profondeur. L'algorithme DFS3, qui semble le plus prometteur, pose des questions au milieu des branches de l'arborescence de recherche en profondeur. Cet algorithme est également implémenté et évalué par simulation. Même s'il n'est pas optimal, la possibilité d'insérer des informations sur l'importance relative des critères, est l'avantage principal de cet algorithme de questionnement. Pour éliciter le profil limite, nous avons conçu un algorithme utilisant la dichotomie.  

\ssk Des algorithmes de questionnement pour déterminer les coalitions suffisantes de critères peuvent être optimaux dans le pire des cas ou en moyenne. Un algorithme optimal dans le pire des cas a été proposé par Hansel dans le cadre des fonctions booléennes monotones et garantit que le nombre de questions ne dépasse jamais une borne inférieure. Nous implémentons et évaluons cet algorithme comme une variante des approches de recherche en profondeur d'abord.

\ssk Afin de trouver une stratégie de questionnement optimale en moyenne, des arbres de décision ont été évalués pour quatre critères au maximum. Ces calculs ont été faits grâce à plusieurs outils simplificateurs tels que les symétries, les motifs déjà calculés, le ``branch \& bound'', etc. Nous avons identifié des chemins de questionnement optimaux qui minimisent le nombre de questions en moyenne.

\ssk D'autre part, nous évaluons à quel point le modèle NCS est plus général que le modèle MR-Sort, qui en est un cas particulier, dans lequel des coalitions suffisantes de critères peuvent être représentées par des poids de critères et un seuil. Nous le faisons pour jusqu'à six critères, par une simulation (exhaustive) et calculons la proportion des ensembles de coalitions suffisantes qui peuvent être représentées par des modèles 2- et 3-additifs.

\ssk La difficulté principale pour appliquer nos outils à la prise de décision dans le monde réel, est le nombre et la difficulté des questions à poser au décideur. Nous suggérons différentes pistes pour utiliser ou adapter nos méthodes rigoureuses d'élicitation à des processus réels d'aide à la décision.

\end{otherlanguage}
\cleardoublepage
%
\begin{center}
    \textbf{Remerciements}
\end{center}

\vspace{2cm}

C’est sans doute la partie la plus difficile à écrire\ldots Mais aussi la plus importante\ldots

\ssk
Quand je regarde en arrière aujourd’hui: une thèse de doctorat évoluant sur trois continents, qui a été témoin d'une partie très importante de ma vie et de nombreuses belles personnes qui m'ont été d’un grand soutien inoubliable\ldots

\ssk
Je dois avouer qu’au tout début de ce parcours, concernant le choix de directeur de thèse, j'ai fait preuve d’un talent naturel dans le domaine de la décision multicritère\ldots J'ai vu et ressenti à chaque instant de mes études doctorales, à quel point cette « décision multicritère » était cruciale.

\ssk
Prof.\ Marc Pirlot\ldots Un professeur extraordinaire, un guide perspicace, un très bon scientifique, un esprit incroyablement clair et analytique, une personne \og complète \fg qui s'intéresse à tous les sujets, de l'art à la littérature, de la philosophie à la nature\ldots Les mots dans toutes les langues restent insuffisants pour exprimer mes pensées. Je voudrais lui dire dans ma langue maternelle et de la manière la plus simple : « Sonsuz teşekkürler Marc\ldots »

\ssk
Prof.\ Philippe Fortemps a toujours été à mes côtés, toujours prêt à m'aider, des premiers jours jusqu’aujourd'hui. Ses aides sur la programmation étaient primordiales et ont servi de base à toutes les programmations ultérieures. Merci beaucoup pour tout son soutien et tous ses efforts destinés à la réalisation de cette thèse. Je le remercie également pour avoir accepter d'être le promoteur officiel de cette thèse. 

\ssk
Je tiens à exprimer ma gratitude au Prof.\ Vincent Mousseau pour ses contributions précieuses et ses remarques pertinentes dans plusieurs parties de cette thèse. Je le remercie pour avoir accepté de faire partie de mon jury.

\ssk
Prof.\ Yves de Smet avait un rôle très important dans l’ambiance conviviale de l’ULB que j’avais beaucoup appréciée au début de mon parcours. Un chercheur toujours efficace, axé sur les résultats et qui m’a guidé avec ses conseils réalistes et constructifs. Je le remercie sincèrement pour tout son soutien et son amitié.

\ssk
Je voudrais remercier les Prof.\ Denis Bouyssou et Prof.\ Thierry Mar\-chant, pour avoir créé de nouveaux champs de recherche pour plusieurs chercheurs. Cette thèse s’est basée sur leurs recherches axiomatisées. C'est un vrai plaisir pour moi que Thierry Marchant ait accepté de faire partie de ce jury.

\ssk
Je remercie également le Prof.\ Pierre Manneback pour ses remarques lors du comité d'accompagnement et le Prof.\ Xavier Siebert pour avoir accepté de présider le jury de thèse. 

\ssk
Un grand merci au Dr. Olivier Sobrie avec qui nous avons travaillé en parallèle, sur des sujets complémentaires et qui a toujours été prêt à apporter son aide.

\ssk
Je tiens également à remercier le personnel du secrétariat de l'UMONS pour avoir compris et supporté les difficultés que j’ai rencontrées à distance lors des démarches administratives.

\ssk
Je ne peux pas m'empêcher de mentionner quelques personnes qui m'ont accueillies d’une façon exceptionnelle quand je suis arrivée en Belgique. Prof.\ Martine Labbé, Prof.\ Philippe Vincke et Prof.\ Bertrand Maréchal ont été les premiers à me guider dans mon cheminement. Ma chère Martha avec qui j’ai partagé le bureau à l’ULB, et tous les chercheurs de ce couloir qui ont été les créateurs d’une ambiance très chaleureuse\ldots

\ssk
Je voudrais remercier le Prof.\ Ertuğrul Karsak de l'Université Galata\-saray, un professeur à qui je dois beaucoup\ldots C’est aussi grâce à lui que j’ai aimé ce domaine et que j’ai voulu l’approfondir.

\ssk
Je commémore une très chère professeure, amie, Prof.\ Esra Albayrak avec respect et gratitude. Son côté humain, ses aides pour ma thèse de DEA, ainsi que son soutien et son amitié resteront toujours gravés dans ma mémoire\ldots

\ssk
Et les membres des familles Çiçekçiler et Pirlot\ldots Je leur suis très reconnaissante pour le temps qu'ils m’ont consacré même pendant leurs vacances, l'ambiance familiale qu'ils m'ont fait ressentir et leurs traces inoubliables dans ma vie\ldots Je voudrais commémorer la merveilleuse grand-mère de la famille Çiçekçiler avec qui j’ai beaucoup de beaux souvenirs.

\ssk
Je voudrais remercier tous les membres de ma famille et mes amis, dont j’ai toujours ressenti la présence, même quand nous étions à des milliers de kilomètres\ldots Banu, Pınar, Gülfem, Tuncay, Didem, Anne-Marie\ldots Merci à Özgür pour tout son soutien, à mes amis Feyhan \& Erdem Canbay, à Berkay et à Ahmet, pour leur aide dans la programmation.
Je remercie les grand-parents pour avoir gardé mes enfants quand j’en ai eu besoin et à Murat qui est un vrai dépanneur des moments difficiles.

\ssk
Et mon frère extraordinaire qui a toujours été présent pour moi avec un grand sacrifice. Merci pour son existence dans ma vie\ldots

\ssk
Bien que cette thèse ait apporté beaucoup de belles personnes dans ma vie, elle a aussi été le témoin de la perte de ce que j'avais de plus précieux dans ma vie, ma très chère maman\ldots
J’aurais beaucoup aimé que les plus grandes chances de ma vie, mes parents exceptionnels, lisent cette thèse\ldots

\ssk
Venons-en tout au début de l'aventure\ldots Ma raison d'aller en Belgique\ldots \   
Merci Sait, pour tout et de tout cœur\ldots 

\ssk
À mes deux trésors, Nehir et Orhan Kaan : Je suis très reconnaissante pour votre existence et toute la beauté que vous m'amenez. Je souhaite que vous poursuiviez les chemins de votre choix avec courage et patience, quels que soient les obstacles\ldots

\cleardoublepage

\thispagestyle{plain}
\begin{center}
\vspace*{\fill}

\begin{flushright}
\textit{à Bilge \& Orhan Ersek, Nehir \& Orhan Kaan  $\heartsuit$} 

\end{flushright}
\vspace*{\fill}
\end{center}

\clearpage

\tableofcontents

\mainmatter


\chapter{Introduction}\label{chap:intro}
In sorting problems each alternative is considered independently from the others and its category is determined, by comparing it to profiles, references or norms. Different methods or models have been proposed in order to compare alternatives to such norms. Certain of these methods define norms as desirable minimal requirements to be attained on a set of criteria. Rules establishing whether or not such norms are fulfilled often consist of specifying minimal subsets of criteria (\textit{sufficient coalitions of criteria}) on which the requirements should be attained. 

\ssk Several sorting models have been proposed. UTADIS is one of them which is based on an additive value function and thresholds associated with each category. \etri{} is based on an outranking relation and category limit profiles.

\ssk  Multi-criteria decision aiding (MCDA) aims to create an interaction between the decision analyst and the decision-maker (DM) in order to reflect the preferences of the DM relative to the decision problem. The DM's preferences are formalised in a model which requires to determine several parameters. Elicitation of these parameters is a very delicate step for MCDA models. It cannot be expected that the DM has a deep understanding of the technical aspects of the decision model and of the precise significance of the model's parameters. Hence asking the DM to assess the parameters values may cause hesitations and induce the assignment of arbitrary values to these parameters. In a decision aiding process, one should preferably avoid to ask such questions to the DM. 

\ssk Direct elicitation techniques ask questions to the decision-maker in order to determine model parameters. The questions should not refer to  parameters. They should rather be formulated in as concrete as possible terms, having a direct relevance to the decision problem and the DM's preferences. Indirect elicitation collect examples of assignments and use them to infer the parameters of a model instance compatible with the examples.

\ssk Why being interested in direct elicitation? Directly requesting values of parameters to a decision-maker is not a good practice. For instance, questioning in terms of the importance of the criteria is not reliable because we can use the same weights in different models and they do not have the same meaning in all these models. In principle, a rigorous direct elicitation must question in terms of the primitives of the model, \ie in terms of assignment to classes. Instead of asking the parameters' values directly, providing preference judgements like ``alternative $a$ is classified in the worst category'' or ``alternative $a$ is preferred to $b$'' is much more understandable for a DM.

\ssk There are not many rigorous methods of direct elicitation proposed in the literature, especially for sorting problems. The goal of this thesis is to propose a direct elicitation method for a multi-criteria sorting model. A multi-criteria sorting problem consists in alternatives described by a vector of $n$ criteria values, $p$ pre-defined and ordered categories (\eg unacceptable, acceptable, good, very good). Every alternative is assigned to one and only one category. The constraint of the sorting model is that an alternative better than another one on all criteria, cannot be assigned to a worse category. To use a model, you have to adjust its parameters. For example, for \etri{}, it is necessary to determine the limit profiles, the weights of the criteria and a preference (outranking) relation allowing to compare alternatives to limit profiles. 

\ssk To develop a direct elicitation method for a model, it is useful that the model is characterized by axioms. Which means that, an existence theorem saying: if the assignment of alternatives to categories has such and such properties (axioms), then there is one and only one instance (set of parameters) of the model that represents the assignment. As the model instance   is unique, we can aim to elicit it. 

\ssk At the time we started to work on this thesis, \citet{bouyssoumarchant2007a,bouyssoumarchant2007b} had just characterized a simplified variant of \etri{}, called the Non-Compensatory Sorting (NCS) model. In the simple case where there are two categories of alernatives, say the ``good'' and the ``bad'', the scale of each attribute is divided in two subsets, one containing values typically good, and the other, values typically bad. The model assigns an alternative to the ``good'' category if it has good values on a sufficient subset (coalition) of attributes. Otherwise the alternative is assigned to the bad category. The model addresses the case of more than two categories by distinguishing more sets of values on each attribute scale and defining sufficient coalitions of attributes for each category.  

\ssk \citet{bouyssoumarchant2007a,bouyssoumarchant2007b}  describe the NCS model in the framework of conjoint measurement models, in the spirit of what has been done in the classical theory of additive value functions (or utility) models. Such approaches usually reveal objects that are determined by the output of the method (preference relation, assignment, \ldots) and on which an elicitation procedure is based. These objects are interpretable in terms of the preference relation or the assignment. In the case of the NCS model with two categories, these objects are:
\begin{itemize}
    \item the set of ``good'' (and ``bad'') levels on each criterion;
    \item the sufficient (and insufficient) coalitions of criteria.
\end{itemize}

\ssk To compute the ``strength'' of coalitions of criteria, many multi-criteria methods, including \electre{} methods, require the determination of weights attached to the criteria, used additively and compared to some threshold. These procedures determine whether the coalitions are sufficient or not. A particular case of the NCS model in which sufficient coalitions are determined by weights and a threshold is known as \mrsort{} \citep{leroyetal2011}. The version of \etri{} characterized by \citet{bouyssoumarchant2007a,bouyssoumarchant2007b} does not suppose the existence of additive weights. It is therefore more general than \mrsort{}.


\ssk As mentioned previously the main objective of this thesis is the development of a rigorous methodology for determining the set of sufficient coalitions and the limit profile of a NCS model with two categories and no veto. This question is not easy, especially for the determination of the set of sufficient coalitions. The problem is highly combinatorial. It is equivalent to determine the set of minimal sufficient coalitions which is an antichain in the set of subsets of criteria. For $n$ criteria, the number of antichains is the Dedekind number $D_n$. This number grows very rapidly with $n$. For instance for $n=4$, $D_4=168$ and for $n=7$, it grows up to $D_7=2414682040998$ ! Therefore, identifying the set of sufficient coalitions of $n$ criteria amounts to find an object in a set of $D_n$ objects, which is a challenging task.

\ssk As a consequence, we need to build efficient questioning algorithms that identify the sufficient coalitions and the limit profiles delimiting the categories while minimizing the number of questions to be asked to the decision-makers. What is an efficient questioning strategy? Minimizing the number of questions? Minimizing the computing time of the algorithm? Taking into account the difficulty of the questions? Different objectives can be selected to establish such strategies. On the other hand, it is obvious that we cannot hope to use a really optimal questioning algorithm due to several constraints like time pressure, information availability, cognitive burden on the decision-maker, etc. Focusing on the elicitation of the set of minimally sufficient coalitions of criteria in a Non Compensatory Sorting (NCS) model, our aim is to establish user-friendly elicitation algorithms minimizing the number of questions. Note that the application of such  methods are limited to small number of criteria due to the explosion of the number of possible antichains. For dealing with real-world problems, one can combine our methods with indirect elicitation techniques.

\ssk First of all, questioning algorithms are proposed in order  to use as few questions as possible, without giving a precise definition of an optimal questioning procedure. These algorithms are thus questioning heuristics. The most promising one is called \textit{DFS3}. It combines depth-first search in the coalitions graph and a questioning strategy based on dichotomy.  This algorithm is not optimal; however it can be used if some additional information on the relative importance of the criteria can be provided by the decision-maker.

\ssk Regarding optimality, there are several ways of precisely defining the goal which consists of  ``minimizing'' the number of questions. For instance, the number of questions can be minimised in the worst case or on average. Besides heuristic algorithms,  we developed an optimal algorithm in the worst case based on Hansel's theory. 
A study on how to ensure that the average number of questions is minimized, is also realized. Depending on the DM's answers, optimal questioning paths can be defined. Such optimal paths are known only up to four criteria. For more criteria, a heuristic algorithm has been proposed. 

\ssk A particular case of the \ncs{} model is used in this work. It is the case in which the set of sufficient coalitions can be represented by weights attached to the criteria and a threshold. In this representation, a coalition is sufficient if the sum of the weights of the criteria in the coalition is equal or greater than the threshold. This property is characteristic of the \mrsort{} sorting model. 

\ssk The elicitation methods proposed in this work aim at being rigorous in the sense of ``theoretically correct''. Applying them in practice may require adaptations. By the end of this study, we investigate how they can be inspiring in real decision aiding processes. 

\clearpage
\subsection*{Organisation of the thesis} 

This thesis is organized as follows.

\paragraph{Chapter 2 - Sorting methods in Multi-Criteria Decision Aiding } 
\mbox{}

\ssk The main notions of Multi-Criteria Decision Aiding (MCDA) are introduced in this chapter. We mention the problem types existing in this field and the methods dealing with them. With a brief overview of MCDA sorting methods it focuses on methods based on outranking relations, namely \etri, \mrsort and the \ncs{} model which are fundamental in our study. For these models, the methods existing in literature for eliciting either directly or indirectly their parameters are reviewed. In particular, there are no direct methods for eliciting the parameters of a \ncs{} model, which motivates the development of such methods in this thesis.

\paragraph{Chapter 3 - Algorithms for direct elicitation of NCS parameters} 
\mbox{}

\ssk In this chapter, by describing a simple and axiomatized  model, the NCS model is used for sorting alternatives in pre-defined ordered categories based on multiple criteria. Determining the parameters of such a model by questioning the DM is an important practical issue. Several questioning algorithms for finding the parameters that we need are introduced. These parameters are the sufficient coalitions and the limit profiles of each category. The ultimate aim is to find an optimal algorithm minimizing the number of questions asked to the decision-maker.

\paragraph{Chapter 4 - A program for eliciting the sufficient coalitions of criteria} 
\mbox{}
Chapter 4 describes the implementation of an algorithm, DFS3, based on depth first search, which questions  ``in the middle of the branches''. For this, two programs are developed. The ``Main Program'' is interactive. It questions the DM in order to find the sufficient coalitions of criteria. The second one, ¨Questioner'' is a simulation module which considers all possible sets of sufficient coalitions for a given number of criteria. It computes the number of questions for each of them and then returns the average number of questions for a given number of criteria. 

\clearpage
\paragraph{Chapter 5 - Worst case optimal algorithm} 
\mbox{}
Two different concepts of optimality can be used to assess elicitation procedures: the number of questions can be minimized in the worst case or on average. Chapter~5 describes in detail Hansel's theory which guarantees to  minimize the number of questions in the worst case. Based on this theory an algorithm, DFSHansel, which is optimal in the worst case is designed and implemented. This algorithm can be viewed as a modification of the algorithms based on depth first search. The performances of DFSHansel and DFS3 are finally compared.

\paragraph{Chapter 6 - Efficient strategy on average} 
\mbox{}
The aim of the chapter 6 is to minimize the average number of questions. Decision trees under uncertainty to model the problem are used and the expected number of questions is computed. Optimal questioning strategies are derived. This approach is feasible only for a small number of criteria. A heuristic algorithm is also suggested.

\paragraph{Chapter 7 - Compact representation of sufficient coalitions} 
\mbox{}
In some cases, sufficient coalitions of criteria can be described by weights attached to criteria and a threshold. This chapter tries to assess the share of the sets of sufficient coalitions that can be represented by weights and a threshold. This is done by simulation for small values of the number of criteria. 

\paragraph{Chapter 8 - Taking a step back} 
\mbox{}
In chapter 8, we take a step back and resume what has been done. The usage of rigorous simplifying tools are introduced. Suggested methods are illustrated in an example of car selection problem. Guidance for different possible situations (\ie the case of incomplete or imperfect information, the case with veto and with more than 2 categories, the elicitation of vetoes, mix of direct and indirect elicitation, working with criteria hierarchies etc.) is also treated.

\chapter{Sorting methods in Multi-Criteria Decision Aiding}
\chaptermark{Sorting methods in MCDA}
\label{chap:SortingMCDA} 

This chapter gives an overview of the key-stones of multi-criteria decision aiding (MCDA) and its framework. With this objective, some basic notions of MCDA and the types of decision problems that MCDA can solve are recalled. We introduce different methods that can be used in MCDA and explain how to elicit the parameters of these MCDA methods. 

\ssk In the first section of this chapter, the definition of MCDA is recalled and its purpose is given. Different types of decision problems are detailed and important notions of MCDA methodologies are given.

\ssk In the second section, the sorting problem in the framework of MCDA and the types of MCDA sorting methods are explained. We also recall the types of elicitation methods that can be used for the parameters of the MCDA models.

\ssk The third section focuses on \etri{} which is the main MCDA sorting method used in this thesis. The \etri{} method is explained in detail. The existing elicitation methods for the parameters are treated and the difficulties of the elicitation procedures are mentioned. 

\ssk The fourth section explains the usage of simpler methods in terms of the elicitation of the parameters. The Non Compensatory Sorting (NCS) model and \etriI{} (\mrsort{}) are recalled in detail and the elicitation of the parameters is described.

\section{Multi-criteria decision aiding (MCDA)}\label{sec:MCDA}

In real-world, many people face various decision-making problems and need help to cope with them. In multi-criteria decision aiding (MCDA), the person who has to make a decision is called decision maker (DM). MCDA is a field of operational research providing tools and methods to the decision makers confronted to different types of decision problems. The purpose is to provide methodologies and techniques to help to choose, to rank or to sort the alternatives taking into account all the important aspects of a decision-making situation. In order to help the DM(s) in taking their decisions, MCDA methods may rely on several models. The process of decision aiding is based on interactions between the DM and the analyst, that helps to construct the decision model and formulate a recommendation. The analyst is a person who knows decision models and uses them to facilitate the decision aiding process. Once the information given by DM have been validated, the analyst uses them to formulate recommendations to the DM. The MCDA analyst guides the DM at each step of the process. From the identification of the parameters of the decision problem, to the final decisions, the analyst leads the process in order to reflect the preferences of the DM as faithfully as possible. 

\ssk The aim of MCDA is not to tell the DM the decision he/she should make but to build a model that represents the preferences of DM(s) and reduce the influence of their arbitrary judgements. 

\ssk According to \citet{roy1985, roy1996} decision aid can be defined as follows: 
``Decision aiding is the activity of the person who, through the use of explicit but not necessarily completely formalized models, helps obtain elements of responses to the questions posed by a stakeholder of a decision process. These elements work towards clarifying the decision and usually towards recommending, or simply favoring, a behavior that will increase the consistency between the evolution of the process and this stakeholder's objectives and value system.''

\subsection{Problem setting}\label{ssec:problemSetting}
Different types of decision problems can be treated by MCDA. \citet{roy1985, roy1996} classified these decision problems in three categories: Choice Problems, Ranking Problems and Sorting Problems. 

\ssk In choice problems, MCDA helps the DM(s) to choose the best alternative(s) among existing alternatives according to their evaluations on multiple criteria. A student who wants to choose the best university to study among the universities to which he has been accepted can be an example of a choosing problem.

\ssk A ranking problem consists to rank the alternatives from the best to the worst. MCDA can solve this problem with a partial or complete ranking of the set of alternatives. Some alternatives may be tied in the ranking; they then form equivalence classes which are ordered but not predefined. An exam organised by an academic program in order to rank the applicants in the decreasing order is a real-world ranking problem. 

\ssk The third type of decision problem, sorting, consists in 
assigning the alternatives to predefined and ordered categories, from the worst to the best. A car selection problem that categorizes a set of cars as ``good'' or ``bad'' is an example of sorting.

\ssk Two main types of MCDA methodologies can be used to deal with the different kinds of decision problems. The first type of methods is based on multi-attribute value theory (MAVT) \citep{keeneyraiffa1976}. The objective of MAVT approaches is to give a score to each alternative, then use that score to compare the alternatives to one another or to a threshold. An alternative's score is computed by summing up its weighted scores on the set of evaluated criteria. The weighted sum method is a basic example of a MAVT method. AHP (analytic hierarchy process) \citep{saaty1980analytic}, UTA (additive utility function) \citep{jaquetlsiskos1982} are also well-known decision methods based on MAVT.

\ssk The second category consists in the methods based on outranking relations \citep{roy1991}. The \electreI{} (élimination et choix traduisant la réalité) method is developed by \citet{roy1968} in order to deal with choice problems. For ranking problems \electreII{} is developed by \citet{roybertier1973}. Then to deal with sorting problems \etri{} is developed by \citet{yu1992} \citep[see also][]{roybouyssou1993}. \electre, \textsc{Promethee} \citep{bransetal1984} and \textsc{Rubis} \citep{BisdorffMeyerRoubens08Rubis} are some examples of outranking methods.

\ssk This work is focused on \mcda{} sorting methods and particularly on \linebreak \etri{} based on outranking relation.

\subsection{Alternative and criterion}
An alternative is a possible solution to the decision problem. For instance, the brand of a car, possible places for a plant or candidate students for a master program can be the alternatives of different MCDA problems. 

\ssk The alternatives of a MCDA problem have to be evaluated on all criteria relevant for the \dm. 

\ssk All criteria have a monotone preference scale which means that the values of the criteria will be maximised or minimised depending on the criteria.

\subsection{Preference relation}
A preference relation, called outranking relation in outranking methods, is created between pairs of alternatives evaluated on multiple criteria. 
According to \citet{roybouyssou1993}, four types of preference can be identified:
\begin{enumerate}
    \item Strict preference: Occurs when one alternative is strictly preferred to another.
    \item Weak preference: Occurs when the set of arguments in favor of one alternative against another is not strong enough to consider that the first alternative is preferred to the second.
    \item Indifference: Occurs when the two alternatives are considered as equivalent by the DM.
    \item Incomparability: Occurs when the evaluations are contrasted and the alternatives cannot be compared.
\end{enumerate}

\section{Sorting Methods}\label{sec:sorting}
In this section MCDA sorting algorithms which are necessary for the framework of this thesis are recalled. This work deals only with sorting problems with ordered and predefined categories. $C^h$, for $h= 1, \ldots, p$,  denotes the set of categories; the preference increases with the parameter $h$. Note that, MCDA clustering techniques are used for the situation in which categories are neither predefined, nor ordered \citep[see e.g.][]{desmetetal2012,meyerolteanu2013}.  

\subsection{Classification of \mcda{} sorting methods}
Two main families of \mcda{} sorting methods are as following:
\begin{itemize}
    \item Methods based on a value function and thresholds as for example \textsc{Utadis} \citep{jaquetlsiskos1982}. Each alternative is associated a value, \eg obtained by an additive value function based on the alternative evaluations.  
    The idea is that an alternative is assigned to a category $C^h$ if the alternative value is equal or bigger than a minimal threshold $\lambda_h$ associated to the category and smaller than the minimal threshold $\lambda_{h+1}$ of the category $C^{h+1}$.
    
    \item Methods based on a preference relation as, for example \etri{}. An alternative is assigned into category $C^h$ if it is preferred to the lower profile $b^h$ of the category and is not preferred to the lower profile $b^{h+1}$ of category $C^{h+1}$. A profile is a vector of evaluations; it can be viewed as a distinguished alternative.  
\end{itemize}

\ssk In this work, we concentrate on the latter type of sorting methods. We are mainly interested in models derived from \etri{}, namely, the Non Compensatory Sorting model (NCS) and \mrsort{}. The latter can be seen as a version of \etri{} using the concordance relation of \electreI{}.

\ssk The models \etri{}, \electreI{} and MR-Sort, the Non Compensatory Sorting model (NCS) are described below, respectively in Sections~\ref{sec:ETRI}, \ref{ssec:EtriI}, \ref{ssec:ncs}.  

\subsection{Elicitation of a model's parameters}\label{ssec:elicitationDirectIndirect}

A general problem of MCDA is to determine the parameters of a model once it has been selected. There are basically two ways of eliciting MCDA models parameters, namely the so-called direct and indirect methods. 

\paragraph{Direct methods}
Direct methods are interactive and determine the parameters of a model in an interaction process between an analyst and the DM. In the case of sorting models, such methods proceed by questioning the decision maker directly (\eg asking for the weights of the criteria or by asking to assign some alternatives to a category) in order to reduce the indeterminacy on the parameters. The questions are designed by the analyst, taking into account the specificity of the model, in order to bring as much useful information as possible to determine the parameters values. Furthermore, the questions should be sufficiently easy to answer and not too many in order to avoid the DM's cognitive tiredness. 

\ssk After the obtention of the answers to the questions, modelling the preferences of the DM, incorporating the values and goals into the model becomes a very important task and should be done by an expert. 

\ssk A straightforward way of determining the parameters of a method is by directly questioning the DM about the value of these parameters. For instance, asking about the value of criteria weights. Such an approach is neither trustworthy nor easy. The meaning of parameters such as criteria weights depends on the selected model. It is quite different in models based on Multiattribute Value Function (MAVT) and in models of \electre{} type. It can't thus be questioned about weights without refering to the chosen model \citep[see][for discussion on this question]{PODINOVSKII1994237,BRVM96,keeney2002,bouyssouetal2006}. 

\ssk In the context of \electre{} models, particular methods for easy questioning about weights have been proposed. For instance, the \emph{method of cards} proposed by \cite{simosmaystre1990}  is a simple method which determines the criteria weights in the \electre{} methods, by using a set of cards. Then, \cite{figueiraroy2002} proposed an improvement on the method.  More recently, \cite{siskostsotsolas2015} outlined some robustness concerns on this method, which is thus not above criticism. 

\ssk Reliable direct methods rely on an in-depth understanding of the model and of the interpretation of the parameters. One way of obtaining such an understanding of a model is by characterizing it by axioms, \ie properties of a ``primitive''. For instance, in ranking models, such a primitive can be a preference relation. In sorting models, it is the assignment of an alternative to a category. Rigorous methods, in principle, only ask questions in terms of primitives. In sorting problems, the analyst should ask the DM to assign well-chosen alternatives to a category. The alternatives are selected in an organized manner, taking the characteristics of the model into account. In order to be efficient, the assignment questions should lead as quickly as possible to a sufficiently precise determination of the model's parameters.  For instance, the method of indifference judgments \citep{vonWinterfeldEdwards1986} is classically used to elicit an additive value function. At the end of the process, an approximation of marginal utility functions is obtained, without questioning about parameters values. 

\ssk In summary, direct elicitation is used in the framework of a decision-aiding process asking appropriate questions to the DM in order to elicit the parameters of a multi-criteria sorting model \citep{BelahceneMOPS23}.

\ssk Because of the limited availability of the DM or to avoid putting an excessive cognitive burden on the DM, indirect elicitation is more used in practice. 

\paragraph{Indirect methods}
Indirect methods consist to infer the parameters of the model from a set of already known assignments. These assignments are taken from DM's judgements. The assignment examples are usually represented as constraints in an optimization problem, which yields the parameters of a compatible sorting model, when it exists. Usually, the expression ``indirect elicitation'' is used in a decision-aiding context based on a small number of assignment examples; a DM may be available (or not) to answer a small number of additional questions.

\ssk On the other hand, the term ``Learning'' which is more common, should be used when the parameters of the model are determined on the basis of large sets of assignment examples. 

\ssk A number of methods have been proposed to determine parameters of a Multiple Criteria Sorting (MCS) models like the \etri{} and Additive Value Function (AVF) in order that these models reproduce a set of assignment examples. \textsc{Utadis} is used for eliciting the parameters of an AVF model and the thresholds of the  categories on the basis of already known assignments. 

\ssk An important issue with indirect elicitation is that many model instances fit the examples, because the known examples are too scarce and not designed to bring as much information as possible.  Two main strategies exist in the literature to deal with model indeterminacy. One is to select one of the compatible model
examples, based on some heuristic principle. For example, one chooses the model whose parameters are  ``central'' in the feasible set of parameters. Another is to work with all compatible model examples and make ``robust'' conclusions, \ie conclusions supported by all compatible models, involving for example, imprecise assignments \citep[see][for more details]{BelahceneMOPS23}.

\section{\etri{}}\label{sec:ETRI}

In this section we focus on \etri{} model \citep{yu1992,roybouyssou1993} which is a MCDA method dealing with assigning alternatives into predefined and ordered categories. 

\subsection{The framework of \etri{}}\label{ssec:informalEtri}
First \cite{moscarolaroy1977}, \cite{Roy81Trichotomic} and \cite{massagliaostanello1991} proposed particular sorting procedures based on a preference (outranking) relation and limiting profiles. The former two partition the set of alternatives into three categories (``trichotomic segmentation''). Then \etri{}, a more general approach to the problem, was proposed by \citet{yu1992} as an outranking sorting procedure  \citep[see][for a complete presentation]{roybouyssou1993}. \etri{} has no limitation on the number of categories.

\ssk Assume that each object is characterized by the vector of its evaluations w.r.t. $n$ criteria, an object $x$ outranks an object $y$ if $x$ is at least as good as $y$ on a “sufficiently important” subset of criteria without being unacceptably worse than $y$ on any criterion. These two conditions are called, respectively, concordance and non-discordance conditions. This is the main idea of outranking.

\ssk Several versions of the outranking relation have been proposed and implemented, particularly in various \electre{} methods \citep{roybouyssou1993, vincke1992}. The outranking relation in \etri{} is the same as in the \electreIII{} ranking method. It involves the computation of the concordance and discordance indexes and a degree of credibility of outranking. A simpler and more easily interpretable version uses the outranking relation of \electreI{}.

\ssk \etri{} is an outranking sorting method which assigns each alternative, to a category from a set of ordered and predefined categories. The alternatives are compared to the profiles of the categories and they are assigned to some category on this basis. All alternatives are evaluated on multiple attributes which have a monotone preference scale. This means that for a criterion to be maximized (\resp minimized), a higher (\resp lower) value of an alternative on a criterion $i$ is considered as better than a lower (\resp higher) one. 

\ssk \etri{} does not require to compare every pair of alternatives, like other outranking methods as \electreII{} used for ranking. 
The alternatives only have to be compared to $p-1$ profiles.

\ssk \etri{} provides also directly a sorting of the
alternatives in ordered classes while the outcome of a ranking method as \electreII{} is usually not a complete ranking of the alternatives. The difficulties due to the lack of transitivity and/or completeness of the outranking relation of \electreII{} don't show up with \etri{} which gives a complete preorder of the alternatives by assigning them into ordered classes. Contrary to ranking methods, the number of equivalent classes is fixed in advance. 

\subsection{Formal presentation of \etri{}}\label{ssec:formalEtri}
We note $A$, the set of alternatives, $C^h$, $h=1, \ldots, p$, the categories ordered by preference (in increasing order of the index $h$), \ie such that $C^p \succ C^{p-1} \succ \ldots \succ C^1$.
The set of alternatives are evaluated on multiple criteria. The preference scale is monotone (\ie depending on the type of the criterion; the higher the value of an alternative on criterion $i$ the better it is, or the inverse). $N = \{1, \ldots, n\}$ is the set of criteria and $i$ is a criterion in the set of criteria $N$.
To determine the category of an alternative, its performances are compared to the $p-1$ profiles which identify $p$ categories. Then the alternative is assigned to the suitable category. $b^h$ is the profile determining the category $C^h$ from $C^{h+1}$.

\begin{figure}[h]
	\centering
	\includegraphics{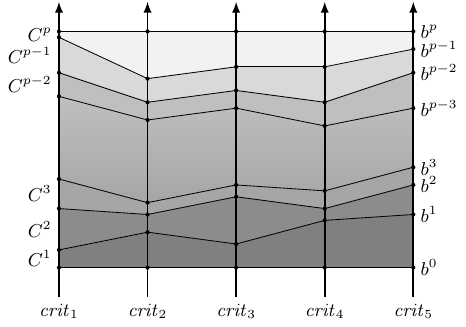}
	\caption{An example of \etri{} model showing the categories, profiles and criteria.}
	\label{fig-mrsort_model}
\end{figure}

\ssk Figure \ref{fig-mrsort_model} shows the categories, criteria and the profiles of an \etri{} model.

All alternatives are compared to the profiles defining the categories. An alternative $a$ is better than or equal to a profile $b^h$ if the credibility index of the assertion ``$a$ is at least as good as $b^h$'', $\sigma(a, b^h)$, is greater than or equal to  $\lambda$, a cutting threshold.

\ssk The aggregation of the concordance and non-discordance degree values gives the credibility index.

\ssk The concordance index indicates the power of the coalition of criteria approving the assertion ``$a$ is at least as good as $b^h$'', denoted by $a \outrank b^h$.

This index is the aggregation of partial concordance indices
calculated for each criterion $i$, which represent the degree to which an alternative $a$ is considered at least as good as the profile $b^h$ on criterion $i$.

The partial concordance index, $\mathcal{C}_i(a,b^h)$ is calculated by comparing on criterion $i$ the alternative performance $a_i$ to the profile performance, denoted by $b^h_i$.

\ssk Two thresholds are required for the computation of $\mathcal{C}_i(a, b^h)$:

\begin{itemize}
\item the preference threshold, $p_i^h$, such that $p_i^h \geq q_i^h$, that represents the difference between $a_i$ and $b^h_i$, from which the assertion $a \outrank b^h$ becomes partially true on criterion $i$.

\item the indifference threshold, $q_i^h$, such that $q_i^h \geq 0$, that represents the maximal difference between $a_i$ and $b^h_i$ up to which the assertion $a \outrank b$ is completely true on criterion $i$;
\end{itemize}

\ssk In general, the indifference and preference thresholds vary depending on the criterion value of the profile.
In order to simplify the notations, we fix the thresholds $p_i, q_i$ on each criterion $i$ independently of category $C^h$.

Thus, the value of $\mathcal{C}_i(a, b^h)$ is computed as follows:
\begin{equation}
	\mathcal{C}_i(a, b^h) =
	\begin{cases}
		0 & \text{if } a_i \leq b^h_i - p_i,\\
		1 & \text{if } a_i \geq b^h_i - q_i,\\
		\frac{a_i - b^h_i + p_i}{p_i - q_i} & \text{otherwise}.
	\end{cases}
\label{eq-etri_partial_concordance}
\end{equation}

\begin{figure}[h!!!]
    \centering   \includegraphics{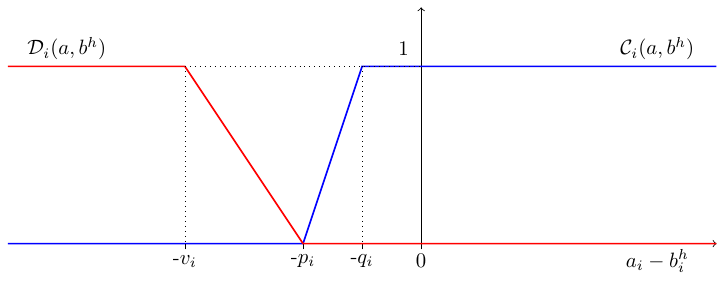}
    \caption{Concordance $\mathcal{C}_i(a, b^h)$ and discordance $\mathcal{D}_i(a, b^h)$} indices of an alternative $a$ \wrt a profile $b^h$ on criterion $i$
    \label{fig:partialConcDiscElIII}
\end{figure}

\ssk The concordance degree function $\mathcal{C}_i(a, b^h)$ is shown in Figure~\ref{fig:partialConcDiscElIII} 

\ssk The concordance degree, $\mathcal{C}(a, b^h)$ is calculated by weighting and aggregating the partial concordance indices. It can be written is as follows:

\begin{equation}
	\mathcal{C}(a, b^h) = \sum_{i = 1}^n w_i \cdot \mathcal{C}_i(a,
	b^h).
\label{eq-etri_concordance}
\end{equation}
Generally the criteria weights $w_i$ are normalized, \ie
$\sum_{i = 1}^n w_i = 1$.

\ssk The discordance degree indicates the power of the criteria coalition refusing the affirmation $a \outrank b^h$.
The complement of this value is the non-discordance degree, denoted by $\mathcal{ND}(a, b^h)$. To calculate the non-discordance degree, the performances of the alternative are compared to the profiles. A partial discordance index, denoted by $D_i(a, b^h)$ is calculated for each criterion.

\ssk To determine $\mathcal{D}_i(a, b^h)$ we need to determine two thresholds:
\begin{itemize}
	\item the preference threshold, $p_i^h$;
	\item the veto threshold, $v_i^h$, representing the maximal
	difference between $b_i^h$ and $a_i$ tolerated before refusing the
	affirmation $a \outrank b^h$ \footnote{In general, the veto threshold may depend on the value of the criterion value, but here we consider it as constant.}.
\end{itemize}

$\mathcal{D}_i(a, b^h)$ is calculated as follows:
\begin{equation}\label{eq-indiceDiscord}
	\mathcal{D}_i(a, b^h) =
	\begin{cases}
	1 & \text{if } a_i \leq b^h_i - v_i,\\
	0 & \text{if } a_i \geq b^h_i - p_i,\\
	\frac{b^h_i - p_i - a_i}{v_i - p_i} & \text{otherwise}.
	\end{cases}
\end{equation}

\ssk The discordance degree index $\mathcal{D}_i(a, b^h)$ is represented in Figure~\ref{fig:partialConcDiscElIII}. 

\ssk The non-discordance index can be obtained by the aggregation of the concordance degree and the partial discordance degrees as below:
\begin{equation}
	\mathcal{ND}(a, b^h)
		= \prod_{i \in N: \mathcal{D}_i(a, b^h)
			> \mathcal{C}(a, b^h)}
			\frac{1 - \mathcal{D}_i(a, b^h)}
			{1 - \mathcal{C}(a, b^h)}.
\label{eq-etri_nondiscordance}
\end{equation}

\ssk Then, the credibility degree of the affirmation $a$ outranks $b^h$, denoted by $\sigma(a, b^h)$, is computed by multiplying the concordance degree by the non-discordance degree as in the equation below:
\begin{equation}
	\sigma(a, b^h) =
		\mathcal{C}(a, b^h) \cdot
		\mathcal{ND}(a, b^h).
\label{eq-etri_credibility}
\end{equation}

\ssk In the end, an alternative $a$ outranks a profile $b^h$ if its
$\sigma(a, b^h)$ (the credibility threshold)  is greater than a
$\lambda$ (the threshold value) which is fixed usually in the interval $]0.5,1]$.
This can be written formally as follows:
\begin{equation}
	a \outrank b^h
	\iff
	\sigma(a, b^h) \geq \lambda.
	\label{eq-etri_outranking}
\end{equation}

\ssk Two fictive profiles $b^0$ and $b^p$ are defined and correspond to the worst and best possible performances respectively on all of the criteria.
According to \etri{}, the alternatives are assigned to a category through one of the two assignment procedures as follows.

\begin{description}
	\item[Pessimistic procedure] This procedure compares each
	alternative $a$ consecutively to $b^{p-1}, \ldots, b^1, b_0$.
	When $l$ is the first index such that $a \outrank b^l$, $a$ is
	assigned to $C^{l+1}$.

	\item[Optimistic procedure] This procedure compares each
	alternative $a$ consecutively to $b^1, b^2, \ldots, b^p$.
	When $l$ is the first index such that $b^l \outrank a$ and not
	$a \outrank b^l$, $a$ is assigned to $C^l$.
\end{description}

\ssk As explained previously, several parameters have to be determined for the \etri method. Briefly, these are, $n$ weights, $p-1$ profiles evaluated on the set of criteria, a majority threshold and at least $n$ preference, indifference and veto thresholds. If we consider different thresholds for each profile, the number of parameters increases. As the elicitation of the parameters of a MCDA model is not easy, it becomes more difficult when the number of parameters increases more.

\subsection{Elicitation of the \etri{} parameters}

Elicitation of the parameters of \etri{} is not easy. The difficulty comes from the large number of parameters, their discontinuous character (thresholds) and the interrelations between parameters (sufficient coalitions vs.\! veto effects). 

\ssk The parameters of the \etri{} method are:
\begin{itemize}
    \item the limit profiles $b^h$, $h=1, \ldots, p-1$;
    \item the criteria weights $w_i$, $i= 1, \ldots, n$;
    \item the preference, indifference and veto thresholds $p_i, q_i$ and $v_i$, $i=1, \ldots, n$;
    \item the $\lambda$ threshold, sometimes called \emph{majority threshold}.
\end{itemize}

\ssk For $n$ criteria and $p$ categories, \etri requires to elicit $n$ weights, $p-1$ profiles, one majority threshold and $n$ indifference, preference and veto thresholds. It makes $3n+np+1$ parameters to be determined in total, which can be difficult to elicit by the DM when the numbers $n$ and/or $p$ are large. If the indifference, preference and veto thresholds depend on the profile, the number of parameters grows up to $4np-3n+1$.

\ssk What is the interpretation of these parameters?

\paragraph{Limit profile $b^h$.} According to \citet[][p.~399-400]{roybouyssou1993}, the value $b_i^h$ of profile $b^h$ on criterion $i$ is a performance that defines the limit between categories $C^h$ and $C^{h+1}$. The interpretation is slightly different in the pessimistic and the optimistic rules. In the pessimistic assignment procedure, $b_i^h$ is the minimal performance to be attained by an alternative to be assigned to category $C^{h+1}$ or better, provided a similar condition is fulfilled by the alternative on a sufficient set of other criteria. 

\ssk The interpretation of $b_i^h$ is the same in the optimistic procedure, but the additional condition on other criteria is more complicated \citep[see][p.~400]{roybouyssou1993}. In the sequel, we shall focus on the pessimistic procedure, which is by far the more frequently used.

\ssk Actually, the required minimal performance is $b_i^h - q_i$. We may thus say (in the pessimistic procedure) that the minimal performance is $b_i^h$ and set $q_i=0$. 

\paragraph{Criteria weights and sufficient coalitions.} In the \electre{} methods, a criterion weight represents the power of the criterion in a set or coalition of criteria. The strength of a coalition is supposed to be represented by the sum of the weights of the criteria belonging to the coalition. A coalition is sufficiently strong if the sum of its criteria weights passes some threshold, here represented by $\lambda$.

\ssk In view of definition \eqref{eq-etri_credibility}, the degree of credibility of outranking $\sigma(a,b^h)$ is the strength of a coalition in a very particular case. If $\mathcal{C}_i (a, b^h)$ is 0 or 1 on all criteria, \ie either $a_i \geq b_i^h -q_i$ or 
$a_i \leq b_i^h - p_i$, and there is no discordance at all (\ie $\mathcal{D}_i (a, b^h) = 0$ for all $i$), then $\sigma(a,b^h) = \sum_{i:a_i \geq b_i^h -q_i} w_i$, \ie the strength of the coalitions of criteria $\{i: a_i \geq b_i^h -q_i\}$. In this special case, the interpretation of $\sigma(a, b^h)$ is clear. In the other cases, $\sigma(a, b^h)$ is a complex mixture of values comprised between 0 and 1. 

\paragraph{Preference, indifference and veto thresholds.} 

As observed before, the indifference threshold $q_i$ can be absorbed in the limit profile, in case the pessimistic rule is considered. Otherwise, its significance is a tolerance or imprecision on the evaluations within which the DM perceives no difference between an evaluation $x$ and evaluations in the interval $x-q_i, x+ q_i$. It is interpreted as a negligible difference.

\ssk The preference threshold denoted by $p_i^h$, such that $p_i^h \geq q_i^h$, represents the minimal difference between $a_i$  and $b_i^h$, from which the assertion $a \outrank b^h$ becomes partially true on criterion $i$ (see Figure~\ref{fig:partialConcDiscElIII}). Note that in general, indifference and preference thresholds may vary depending on the position on criterion scale. We shall not consider this case here.

\ssk The veto threshold $v_i$, represents the minimal difference between $a_i$ and $b_i^h$ which absolutely precludes saying that $a$ outranks $b_h$, whatever the performance of $a$ on criteria other than $i$. Between $b_i^h - p_i$ and  $b_i^h - v_i$, the discordance index increases gradually from 0 to 1. 

\ssk \citet[][p.~400]{roybouyssou1993} show that choosing veto thresholds close to the preference thresholds in conjunction with a credibility threshold close to 1, increases the conjunctive or disjunctive character of the \etri{} procedure. Axioms characterizing outranking relations based on concordance and non-discordance principles as in \electreI{} and \electreII{}  are presented by \citet{bouyssoupirlot2009,BouyssouPirlot2015AOR}.

\ssk  \electreIII and \etri require the elicitation of indifference, preference and veto thresholds in order to compute the concordance and non-discordance indices. While they think that outranking methods are well \linebreak adapted for environmental problems, \citet{rogersbruen1998} note that the use of \electre{} methods can be unreasonable if the values of the thresholds are too subjective. They propose a new approach to determine the indifference, preference and veto thresholds in a principled way which takes into account the uncertainty and human aspect.

\subsection{Direct elicitation of the \etri{} parameters}\label{ssec:directElicitEtri}

Section 6.3 in \citep{roybouyssou1993} gives general indications on how to elicit the parameters of an \etri{} model. However, the authors do not describe a strategy that avoids to question directly about the values of the parameters.

\ssk Note that “direct elicitation” does not mean questioning directly on the exact values of the parameters. In the literature \citet{PODINOVSKII1994237,RoyMousseau96}, \citet[][Section~4.4.1]{bouyssouetal2006} have argued that questioning, for example, about importance of criteria weights is bad practice. The questions to the DM should be formulated by the MCDA expert in terms of the primitives only, i.e., in terms of objects assignments into categories.

\ssk To our knowledge, no author has so far proposed a rigorous method for direct elicitation of the \etri{} parameters. This is probably due to the large number of parameters and the complexity of formulas such as \eqref{eq-etri_credibility}.


\subsection{Indirect elicitation of the \etri{} parameters}\label{ssec:indirectElEtri} 

The main difference between direct and indirect methods is that in indirect methods, the parameters of the model are inferred from a set of assignment examples. 

\ssk As direct elicitation is not easy to handle, indirect elicitation methods have been developed. Usually the parameters of the models compatible with a set of assignment examples are determined by solving a mathematical program. Assignment examples induce constraints on the parameters. 

\ssk The main problem with \etri{} is that the resulting mathematical programs are non-linear and involve binary variables. Thus, linear programming cannot be used.

\ssk \citet{MousseauSlowinski98} have been proposing the first learning algorithm of \etri{} parameters (except for vetoes). It takes a set of assignment examples and their vector of performances as input. The non-linearity of the partial concordance indices leads to a non-convex mathematical programming formulation that can be solved only for small examples. 

\ssk A number of followers have focused on the elicitation of a subset of parameters, assuming that the others were known. \citet{mousseauetal2001} determine the criteria weights and majority threshold by linear programming when limit profiles and preference and indifference thresholds are known. \citet{ngothemousseau2002} only learn the profiles on the basis of assignment examples. 

\ssk These methods do not consider veto thresholds. To avoid to determine the veto thresholds together with the other parameters \citet{diasmousseau2006} suggest to learn the veto parameters from statements of the DM, by mathematical programming. This article takes into account both learning the veto parameters individually or all of the veto parameters simultaneously. 

\ssk \citet{doumposetal2009} designed a genetic algorithm to learn all the parameters of an \etri{} model at a time. 

\ssk Two main problems may occur when trying to find an \etri{} model compatible with assignment examples: there can be no compatible model or, on the contrary, a large set of compatible models.

\ssk The former is less frequent. Yet, \citet{mousseauetal2006} have proposed tools for detecting the sources of inconsistencies in the examples which cause the absence of feasible models. 

\ssk Indeterminacy of the model is much more usual, since the set of assignment examples is often small and the number of parameters large. In such a case several, often a large variety of, models are able to reproduce the assignment examples. For dealing with such a situation, robust and stochastic methods have been suggested.

\paragraph{Robust and stochastic approaches}

A robust approach considers all model instances (of a certain type, e.g.,  \etri{} models or additive value function models) compatible with the available information, i.e., the set of assignment examples. Conclusions are drawn, after taking all compatible examples into account.

\ssk The compatible models are not explicitly determined. Rather, for each alternative of interest, one adds this alternative to the set of assignment examples, assigning it successively to each category. Whenever the augmented problem has at least a solution, this means that there is a model instance compatible with the assignment examples and the assignment of the alternative of interest. Therefore such an assignment is \emph{possible}. 

\ssk A \emph{robust conclusion} for the alternative of interest is to assign it to the set of possible categories. If there is only one possible category, then the assignment to this category is \emph{necessary}. This approach was suggested in the context of sorting using an Electre Tri model by \cite{diasmoussfig2002}. The paper presents an interactive procedure in which preferential information is gradually obtained from the DM. The information is provided directly in terms of constraints on the parameters values (e.g., an interval of possible values) or indirectly in terms of assignment examples. For objects not yet assigned, at some stages, one computes the best and worst possible assignments, from the available information. 

\ssk A more systematic robust approach is Robust Ordinal Regression (ROR) and was developed in the framework of the additive value function model (UTA) for ranking \citep{grecoetal2008} and for sorting \citep{grecoetal2010}. If all compatible models assign an object to a category (or category range), the assignment of the object to this category (or a category range) is necessary. The assignment of an object to a category (or category range) is possible if there is a suitable model assigning the object to this category (or category range). With the available information and a set of feasible model parameters, it is possible to compute, for each non-assigned object, a set of necessarily assigned categories and a set of possibly assigned categories \citep{KadzinskiTervonen2013}. If this information is used in an interactive process, it can trigger some additional preference statements by the DM.

\ssk Applying these principles to \etri{} raises difficulties because the robust approach has to solve a large number of mathematical programs, one for each augmented set of assignment examples. With \etri{}, these programs are non-linear and are thus hard to solve.  

\ssk In recent years, early ideas for a robust strategy have been methodically developed. It has been built machineries that heavily uses formulations of mathematical programming and optimization methods. It is meant to enhance interactions with the DM and support decision-making. The set of models that can be specified using preference information has been expanded. 

\ssk In the case of inputs such as imprecise individual assignment, information about relative pairwise category assignments or constraints on the cardinality of some categories and by relying on a model, possible and necessary assignments may be computed. \citet{KadzinskiCS2015SortAddValPair} is a work done in that research area. They describe how such tools can be used for sorting based on the additive value function model. The suggested modeling framework is called ROR-UTADIS. Linear or mixed integer linear programs are solved for the obtention of robust recommendations. The DM may also express his preferences relative to the values assigned to objects in some category, or on the value differences between objects assigned to different categories or to the shape of the marginals (e.g., concave).

\ssk \citet{KadzinskiCiomek21ActiveLearnOutrankSorting} exploit the same idea for making robust recommendations in outranking-based multi-criteria sorting, but not exactly for the \etri{} method. The information taken from the DM provides constraints on the outranking relation. The idea is to consider all assignments which respect the following rule [previously introduced by
\citet{RochaDias08JGlobOpt} \citet{koksalanetal2009}]: if $x$ outranks $y$, then $x$ should not be assigned to a worse category than $y$. So, in addition to be monotone \wrt dominance, the assignment rule should respect the outranking relation, a relation that contains the dominance relation. Conversely, if $x$ is assigned to a worse category than $y$, then $x$ does not outrank $y$. These constraints can define a set of feasible outranking relations but do not specify a set of sorting models. For the cases of the \electre{} and the Promethee outranking relations, these constraints can be integrated in MILP formulations. The obtained robust conclusions consider all feasible \electre{} or Promethee relations. Particularly in \electre{}, these conclusions generally do not correspond to those obtained with models such as \etri.

\ssk \citet{KadzinskiM21EnrichedPrefModel} assume that the limiting profiles are known and elicit the parameters of the \etri{-B} model. Various conclusions based on the robust or the SMAA approach are obtained. 

\ssk Monte Carlo simulation is another method which considers the whole set of model parameters compatible with the available information. When the parameters are not determined but are known to belong to some region, we can sample this region according to some probability distribution. Then we can compute the frequency with which an object is assigned to each category by the sampled models. The \emph{class acceptability index} $CAI(a,h)$ provides an estimate for the corresponding probability and is obtained as the fraction of models in the sample that assign object $a$ into category $C^h$. The first article introducing “Stochastic Multicriteria Acceptability Analysis” (SMAA) is the work of \citet{LahdelmaHS1998}. The methodology was developed and applied to various situations \citep[see][]{TervonenFigueiraJMCDA2008}, particularly, as a sorting method built on \etri{}, named SMAA-TRI \citep{tervonenEtAl_EJOR2009}.
 
\ssk Both the robust and the stochastic approaches are interesting ways of dealing with insufficient information and model indeterminacy. However, as pointed out by
\citet{BelahceneMOPS23}, in case the information is scarce, the robust conclusions are likely to be weak. This is especially true when very general models, with lots of parameters, are used. In the stochastic approach, the choice of a probability distribution on the parameters may influence the conclusions, \ie which are the most probable assignments. The choice of a particular distribution is often limited by technical constraints. 

\section{Using simpler models}\label{sec:simpleModels}

In view of the difficulties involved both in direct and indirect elicitation of the \etri{} parameters, some researchers have considered simpler models.  The characterization of the NCS model \cite{bouyssoumarchant2007a,bouyssoumarchant2007b} has opened a new research avenue. Models obtained by replacing the \electreIII{} outranking relation by \electreI{} in \etri{} have been investigated.


\subsection{\etriI{} (\mrsort{})}\label{ssec:EtriI}
Applying the same sorting rule (\eg pessimistic) as in \etri, but using the outranking relation as in \electreI{}, leads to a much more understandable model. Its parameters are more interpretable and designing a rigorous direct elicitation method seems to be more achievable. In the absence of veto, this model is known as \mrsort{} \citep{leroyetal2011,Sobrie2018}. 

\ssk The outranking relation in \electreI{} is a particular case of that in \electreIII{} (and \etri{}). The preference and the indifference thresholds on each criterion are null: $p_i = q_i = 0$. Therefore equation \eqref{eq-etri_partial_concordance} becomes:
\begin{equation}
	\mathcal{C}_i(a, b^h) =
	\begin{cases}
		0 & \text{if } a_i < b^h_i,\\
		1 & \text{if } a_i \geq b^h_i.
	\end{cases}
\label{eq-etriI_partial_concordance}
\end{equation}
The concordance degree definition \eqref{eq-etri_concordance} remains unchanged. With \electreI{}, the veto effect is no longer gradual; it is all or nothing. Therefore, the discordance index \eqref{eq-indiceDiscord} becomes:
\begin{equation}\label{eq-indiceDiscord_ElI}
	\mathcal{D}_i(a, b^h) =
	\begin{cases}
	1 & \text{if } a_i < b^h_i - v_i,\\
	0 & \text{if } a_i \geq b^h_i - v_i.
	\end{cases}
\end{equation}
The non discordance degree \eqref{eq-etri_nondiscordance} becomes
\begin{equation}
	\mathcal{ND}(a, b^h) = 	
	\begin{cases}
	1 & \text{if } a_i \geq b^h_i - v_i, \text{for all } i \in N \\
	0 & \text{if } a_i < b^h_i - v_i, \text{for some } i \in N.
	\end{cases}
\label{eq-etriI_nondiscordance}
\end{equation}
The credibility degree $\sigma(a, b^h)$ defined by equation \eqref{eq-etri_credibility} becomes: 
\begin{equation}
	\sigma(a, b^h) =
		\mathcal{C}(a, b^h) \cdot
		\mathcal{ND}(a, b^h) =
		\begin{cases}
	\mathcal{C}(a, b^h) & \text{if } \mathcal{ND}(a, b^h) = 1, \\
	0 & \text{if } \mathcal{ND}(a, b^h) = 0.
	\end{cases}
\label{eq-etriI_credibility}
\end{equation}
The definition \eqref{eq-etri_outranking} of the outranking relation $\succsim$ stays the same as well as the pessimistic and optimistic assignment procedures. 

\begin{figure}[h]
	\centering
\includegraphics{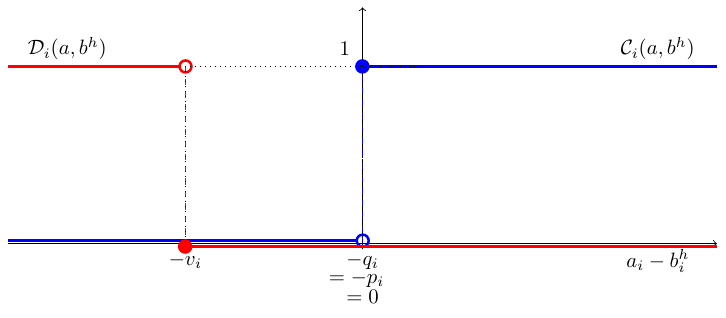}
	\caption{Partial concordance and discordance indices on the criterion $i$ in \electreI{}}
	\label{fig:prefElectreI}
\end{figure}

\ssk The partial concordance and discordance indices on criterion $i$ are represented in Figure~\ref{fig:prefElectreI}.

The pessimistic sorting rule in \etriI{} thus consists of assigning alternative $a$ to $C^h$ if 
\begin{itemize}
    \item the sum of the weights of the criteria for which $a$ is higher than or equal to the profile $b^h$ is sufficiently big, and
    \item there is no criterion on which the evaluation of $a$ is lower than the profile value minus the veto threshold, and
    \item the two above conditions are not fulfilled \wrt profile $b^{h+1}$.
\end{itemize}
These conditions are more easily interpretable than those in \etri{}.

\subsection{Elicitation of the parameters of \etriI{} (\mrsort{})}\label{ssec:elicitElI} 

In order to determine an \etriI{} model, we need to specify the following parameters :
\begin{itemize}
    \item limiting profiles, 
    \item criteria weights,
    \item a concordance threshold that tells which subsets of criteria are ``sufficiently big'',
    \item veto thresholds.
\end{itemize}
Asking questions in terms of alternatives assignments in order to determine the parameters of an \etriI{} model is still rather challenging. A further simplification assumes there are no veto thresholds. This model is known as \mrsort{} (Majority Rule Sorting). No direct elicitation method was proposed for this model. In contrast, indirect elicitation methods have been proposed and tested.

\ssk \citet{leroyetal2011} formulate the constraints on the model parameters induced by assignment examples as a MILP problem. For small sets of assignment examples, involving up to 100 examples, the problem  can be solved using MILP solvers. For larger sizes of assignment example sets, a population-based heuristic has been designed by \citet{sobrieetal2013d} \citep[see also][]{Sobrie2018}. The method and the algorithms have proved their usefulness with an application in the medical sector, namely predicting the ASA score of patients before surgery \citep{sobrieetal2016ASA}.  

\ssk \citet{OlteanuMeyer2014} have proposed a genetic algorithm for learning the parameters of a MR-Sort model with vetoes. \citet{SobrieMP17} modified their population based heuristic to elicit the parameters of MR-Sort with coalitional vetoes (\ie outranking is refused whenever an alternative is in veto \wrt a profile $b^h$ on a sufficient coalition of criteria).

\subsection{The Non Compensatory Sorting model}\label{ssec:ncs}
The model called the Non-Compensatory Sorting (NCS) model \citep{bouyssoumarchant2007a, bouyssoumarchant2007b} reflects the main idea of \etri{} while allowing for a precise characterization with fewer parameters. 

\ssk The Non Compensatory Sorting (\ncs{}) model is proposed by \citet{bouyssoumarchant2007a, bouyssoumarchant2007b} and provides a simpler and axiomatized analysis of the partitions of alternatives into categories. In this study, we work on the case of two categories, without veto.
The set of alternatives is partitioned in two categories $\calAU$, with \calA, the preferred category. 

\ssk Instead of trying to determine weights of the criteria and a concordance index, we can determine the set of sufficient coalitions which also means the minimum coalitions required to be in the concerned category. In this case, the parameters of the model become the list of the sufficient coalitions and the profile $b$. The list of sufficient coalitions can be any family \calF{} of subsets of the set of criteria that is closed by inclusion, \ie if $A \in \calF$ and $A \subseteq B$, then $B \in \calF$.  

\ssk The \ncs{} outranking rule reads as follows: 
\begin{equation}\label{eq:ncsRule}
    a \succsim b  \ \Leftrightarrow \ \{i \in N : a_i \geq b_i\} \in \calF.
\end{equation}
An alternative $a$ outranks a profile $b$ if it is at least as good as the profile on a sufficient coalition (\ie subset) of criteria. If $a$ outranks profile $b$, $a$ is assigned to the category $\calA$, which amounts to the pessimistic rule of \etri{}.

\ssk \etriI{} without veto (\ie MR-sort) is a particular case of \ncs. It corresponds to the case in which the sufficient coalitions can be represented by criteria weights and a majority threshold. More precisely, a \ncs{} model with profiles $b^h, h=1,\ldots, p$, and a family of sufficient coalitions $\calF$ is a \mrsort{} model with the same profiles if there is a set of criteria weights $w_i, i=1, \ldots, n$, and a majority threshold $\lambda$ such that
$$
A \in \calF \ \Leftrightarrow \ \sum_{i \in A} w_i \geq \lambda. 
$$
For a general NCS model, such weights and a threshold do not always exist. For instance, in the case of four criteria ($n=4$), the set of sufficient coalitions $\mathcal{F} = \{\{1,2\},\{3,4\},\{1,2,3\}, \{1,2,4\}, \{1,3,4\},$ $\{2,3,4\}, \{1,2,3,4\}\}$ cannot be represented by weights and a threshold. Therefore, the \ncs{} models with such a set of \suffco{} are not \mrsort{} models. 

\ssk Vetoes can be introduced in the NCS model \citep[as discussed and axiomatized in ][]{bouyssoumarchant2007a,bouyssoumarchant2007b}. A veto profile is a vector $v = (v_1, \ldots, v_n)$ of levels $v_i$ on each criterion $i$. These levels are below the levels of the profile $b$. If an alternative has a performance worse than the veto level, it cannot be assigned to the preferred category $\calA$ whatever its evaluations on the other criteria.

\ssk The NCS model with multiple categories $C^h, h=1, \ldots p$ uses a limit profile $b^h$ for separating category $C^h$ from $C^{h+1}$. The sets of sufficient coalitions $\calF^h$ may depend on $h$. They are nested in the sense that $\calF^h \supseteq \calF^{h+1}$ for $h=1, \ldots, p-1$. An alternative is assigned to $C^h$ if $\{i \in N : a_i \geq b_i^h\} \in \calF^h$ and $\{i \in N : a_i \geq b_i^{h+1}\} \not \in \calF^{h+1}$.    

\ssk The \ncs{} model is simple and has been axiomatized. Therefore, it is fully understood. In Chapter~\ref{chap:elicitNCS}, the model is presented in detail and methods for eliciting the parameters of this model (profiles and set of sufficient coalitions) are built on this knowledge. 

\subsection{Elicitation of the parameters of a NCS model}\label{ssec:ElicitNCS}
We are focused on a sorting model that satisfies the requirements of the non-compensatory sorting (\ncs{}) model axiomatized by \citet{bouyssoumarchant2007a, bouyssoumarchant2007b}. In this section we explain the main works suggesting elicitation methods for the parameters of a  \ncs{} model which is a generalization of \mrsort{} model.

\ssk \citet{sobrieetal2015b} compares the NCS model and the MR-Sort in terms of the additional descriptive ability. To increase the expressiveness of the model, additive weights are replaced by a capacity in order to represent the power of coalitions of criteria which is the \ncs{} model. A general capacity is intractable (because it has $2^n$ coefficients). Therefore, the paper limits itself to deal with \ncs{} models in which the sufficient coalitions can be represented by a 2-additive capacity and a threshold. Such a capacity is determined by $n+n(n-1)/2$ coefficients. Mixed Integer Programming is used in this work and a heuristic algorithm is suggested in order to learn the parameters of the model from assignment examples. Some of the \ncs{} assignment rules can be represented by the MR-Sort rules (additive weights and a threshold), but some of them cannot be represented by MR-Sort rules and require a 2-additive capacity and a threshold. It is also shown that MR-Sort models provide good approximations of the more general \ncs{} models. 

\ssk \citet{TliliBKMO22} learn the set of sufficient coalitions and the profiles of a general NCS model using a SAT formulation. This means that the problem is expressed in the form of a set of logical clauses, that are conjunctions of  Boolean variables and their negations. Determining whether the assignment examples are compatible with a \ncs{} model amounts to decide whether there is an assignment of truth values to the Boolean variables for which all clauses are simultaneously satisfied (SATisfiability problem). Efficient solvers are available for such problems. In case there is no \ncs{} model fully compatible with the examples, they use a MAXSat solver to find one that violates the least possible number of constraints. 

\ssk To our knowledge, direct elicitation of the parameters of \ncs{} models is a question that has remained untouched.

\ssk This is why, in this study, we are interested in the direct elicitation framework and our aim is to suggest efficient questioning algorithms for the sufficient coalitions of criteria and limit profiles. Several direct elicitation algorithms based on the NCS model will be constructed and studied in detail in the further chapters.

\chapter{Algorithms for direct elicitation of NCS parameters}
\chaptermark{Direct elicitation algorithms for NCS}
\label{chap:elicitNCS} 

\section{Introduction}\label{sec:introChap3}
The sorting problem consists of considering each alternative independently of the others in order to determine its category, by comparing to profiles, references or norms. In the literature, several methods like \etri have been proposed to deal with this problem. \etri assigns each alternative, object or item to one of the pre-defined and ordered categories which are defined by parameters or some typical elements.

Determining values of these preference parameters is an important requirement of the \etri method. The correctness of these parameters or norms has a key role on the sorting problem due to their direct influence on the result. In real-world decision problems, decision-makers (DMs) are facing many imperfect knowledge or judgements on the parameters which create hesitations and may imply the assignment of alternatives into a wrong category. Various elicitation procedures have been proposed in order to correctly define these parameters. Direct elicitation procedures require the DMs to provide direct answers to questions having an impact on the values of the preference parameters. The difficulty with these techniques is to understand the precise meaning of the answers given by the DMs. This is the reason why indirect elicitation techniques are more frequently used with the \etri method. However, the application of indirect elicitation procedures like the inference procedures requires the resolution of a non-linear mathematical program which is a hard task in real-world problems. Therefore, this led us to look for simple and meaningful methods to elicit all preference parameters of a sorting problem.

In this work, we rely on the results of the axiomatic analysis of the Non-compensatory Sorting (NCS) model by \citet{bouyssoumarchant2007a, bouyssoumarchant2007b}. Based on their setting and results, our central aim is to build efficient questioning algorithms that identify the sufficient coalitions and the acceptable limit values of the influential attributes while minimizing the number of questions to be asked to the decision-makers. We assume that there is no veto or discordance effects and concentrate on sorting problems with just two categories. We introduce several appropriate algorithms and compare these algorithms with each other according to the number of questions asked. Answers given by the DMs are supposed to correctly represent their preferences and not to be  contradictory.

First, we need to know how to identify the influential criteria from which the set of the sufficient coalitions is composed. Procedures for determining the category limits for each  influential criterion are needed as well as procedures for determining which coalitions of criteria are sufficient.

The suggested algorithms are thus grouped in two categories. The first category of algorithms are used for the identification of $\mathcal{F}$ (the set of the sufficient coalitions) and $\Fmin$ (the  set of minimal sufficient coalitions) or $\mathcal{G}$ (the set of the insufficient coalitions) and $\mathcal{G}_{max}$ (the set of maximal insufficient coalitions). $\Fmin$ is composed of the sufficient coalitions such that, if we remove one criterion from these coalitions, the resulting coalition belongs to the  class of insufficient coalitions \calG{}. $\mathcal{G}_{max}$ is formed of the insufficient coalitions such that, if we add any other criterion to these coalitions, the new coalition belongs to
the class of sufficient coalitions \calF{}.

The second category contains the algorithm determining the value of $b_{i}$, which represents the category limit on the $i^{th}$ criterion, in other words it is the minimal level of the $i^{th}$ criterion which contributes to assign the object to the category of satisfactory objects $\mathcal{A}$ or the maximal level of the $i^{th}$ criterion which contributes to assign the object to the category of unsatisfactory objects $\mathcal{U}$. 

We confine ourselves to analytical comparisons of these algorithms in terms of ``number of questions'' and do not carry out a thorough comparison in terms of ``speed of calculation''.

This chapter is organized as follows. First, basic definitions and some hypotheses under which we shall use the NCS model are introduced. Then, several 
algorithms for questioning about sufficient coalitions of criteria are presented and it is illustrated how they work on a few examples. Knowing the sufficient coalitions, a method is then proposed to determine the limit profile. Two apppendices describe another algorithm to elicit the sufficient coalitions and a method for efficiently finding the maximal insufficient coalitions knowing the minimal sufficient ones, and conversely.  
\section{Definitions and notation}\label{sec:defNot}
Let $N = \{1,2,...,n\}$ be the set of attributes and $X=X_{1}\times X_{2}\times\cdots \times X_{n}$ be a set of alternatives evaluated on $n$ attributes. The set $X_i$ is the set of possible values of attribute $i$. We consider that any combination of values in $X_i$ for $i\in N$ corresponds with an alternative. Furthermore alternatives are entirely characterized by their evaluation vector in $X$. We thus identify an alternative with its vector of evaluations $(x_1, \ldots, x_i, \ldots x_n)$, with $x_i \in X_i$, for $i =1, \ldots, n$. 

\ssk Let $J \subseteq N$ be a subset of attributes. The vector $(x_{J}, y_{-J})$ denotes the element $w\in X$ such that $w_{i} = x_{i}$ if $i\in J$ and $w_{i} = y_{i}$ if $i \in X \setminus J$. If $J$ is a singleton $J = \{i\}$, we simply write $(x_i, y_{-i})$ where $-i$ refers to the set of all criteria but $i$. We use $X_J$ (\resp $X_{-J}, X_{-i}$) for denoting the Cartesian product $\prod_{i \in J} X_i$ (\resp $\prod_{i \in N \setminus J} X_i$, $\prod_{j \in N \setminus \{i\}} X_j$). Hence, in $(x_J, y_{-J})$, $x_J \in X_J$ and $y_{-J} \in X_{-J}$.   

\ssk We concentrate on sorting into two categories forming a twofold partition $<\mathcal{A},\mathcal{U}>$ of the set $X$. $\mathcal{A}$ is the category of the satisfactory objects and $\mathcal{U}$ is the category of the unsatisfactory objects.

\ssk We assume that there is a preference order $\geq_i$ on the scale $X_i$ of attribute $i$, for all $i\in N$. The attributes are thus criteria. We further assume that there is an order isomorphism between $(X_i, \geq_i)$ and a subset of the reals endowed with the natural order on the reals. As a consequence, we can suppose that $X_i$ is a subset of the reals and $\geq_i$ is the order induced by the natural order on the reals.  We make this assumption from now on. The preference order on $X_i \subseteq \mathbb{R}$ is thus ``the greater the better''. The complete preference orders on the $X_i$'s induce a partial order $\geq$ on $X$.

\begin{definition}[Dominance relation $\geq$ on $X$]\label{def:dominance}  
For $x,y \in X$, we say that $x$ dominates $y$ (denoted $x \geq y$) if $x_i \geq_i y_i$, for all $i \in N$. We say that $x$ strictly dominates $y$ (denoted $x > y$) if $x$ dominates $y$ and there is $i \in N$ such that $x_i > y_i$.  
\end{definition} 
The dominance relation $\geq$ on $X$ is a partial order (\ie a reflexive, asymmetric and transitive relation). 

\ssk Let \calAU{} be an ordered bipartition of $X$. We make the following hypothesis.
\begin{hypothesis}\label{hypo:respectDom}
The ordered bipartition \calAU{} respects the dominance relation on $X$.
\end{hypothesis}
This is tantamount to assuming that, if $x\in \calA$ and $y \geq x$, then $y \in \calA$ and the other way around: if $x \in \calU$ and $x \geq y$, then $y \in \calU$.

\subsection{The \ncs{} model under some hypotheses }\label{ssec:ncsWithHypo}
According to \citet{bouyssoumarchant2007a}, a bipartition \calAU{} satisfies the \ncs{} model if the following holds. On each criterion there is a subset $\calA_i \subseteq X_i$ of ``acceptable'' levels and a family \calF{} of subsets of the set of criteria $N$ (``sufficient coalitions'') such that 
\begin{equation}\label{eq:ncsHypo1}
x \in \calA \ \Leftrightarrow \ \{i \in N: x_i \in \calA_i\} \in \calF.    
\end{equation}
Under Hypothesis \ref{hypo:respectDom}, the sets $\calA_i$ are upsets of $X_i$, \ie if $x_i \in \calA_i$ and $y_i \geq_i x_i$, then $y_i \in \calA_i$.  

\ssk In this section, we assume that $\calA_i$ is neither empty nor the whole set $X_i$. We shall deal with these cases in Section~\ref{ssec:ncsWithNonInfluentCrit}.
 
\ssk In order to introduce lower limit profiles, we need an additional hypothesis.
\begin{hypothesis}\label{hypo:lowerLimit}
The sets $\calA_i$ have a smallest element $b_i$, for all $i \in N$.
\end{hypothesis}
This hypothesis is obviously satisfied in case $X_i$ is finite.

\ssk Under this additional hypothesis, the \ncs{} assignment rule \eqref{eq:ncsHypo1} becomes: 
\begin{equation}\label{eq:ncsHypo1and2}
x \in \calA \ \Leftrightarrow \ \{i \in N: x_i \geq_i b_i  \} \in \calF.
\end{equation}
The vector $b = (b_1, \ldots, b_n) \in X$ plays the role of a lower limit profile.

\ssk Under Hypotheses~\ref{hypo:respectDom} and \ref{hypo:lowerLimit}, eliciting the parameters of an \ncs{} model with two categories amounts to identify the lower limit profile $b$ and the set of sufficient coalitions \calF. We start with the elicitation of \calF.

\subsection{Sufficiency of the coalitions}\label{ssec:sufficiencyCoal}
To identify whether a coalition $I \subseteq N$ is sufficient or not, we have to present an appropriate alternative to the DM and ask whether this alternative belongs to \calA{} or \calU. An alternative is appropriate if its evaluations are ``good'' (\ie above the profile) on the criteria in $I$ and ``bad'' (\ie below the profile) on the other criteria, in $N \setminus I := \overline{I}$ (or $-I$). Therefore, we have to identify a certainly good level and a certainly bad level on each criterion. Such levels certainly exist since we assumed that $\calA_i$ is not empty and is not all $X_i$, which implies that $\calU_i$ is not empty. 

\ssk In case the scale $X_i$ of every criterion $i$ has a biggest level $M_i$ and a smallest level $m_i$ (\wrt $\geq_i$), we are sure that the profile $b_i$ is in between : $m_i < b_i \leq M_i$. In order to know whether coalition $I \subseteq N$ belongs to \calF{}, we define the alternative $alt(I)$ with $alt(I)_i = M_i$ for $i \in I$ and $alt(I)_i = m_i$ for $i \in \overline{I}$. In words, an alternative adequate for checking whether coalition $I$ is sufficient, has maximal values on criteria in $I$ and minimal values on the complement $\overline{I}$. If the DM tells that $alt(I)$ is in \calA, then coalition $I$ belongs to \calF{}. Otherwise it is insufficient.

\ssk In case the scale $X_i$ of some criterion $i$ has no biggest level $M_i$ or no smallest level $m_i$ (\wrt $\geq_i$), we assume that there is a big value that is good and a small value that is bad. We select such values and denote them by $M_i$ (\resp $m_i$) even though there may exist bigger (\resp smaller) values in $X_i$. By choosing these values, we take some risk. If we are lucky,  the profile value $b_i$ is indeed between the chosen values : $m_i < b_i \leq M_i$. Otherwise, all values between $m_i$ and $M_i$ are either all good or all bad. In this case such a criterion will not make any difference for the assignment (see below, Section~\ref{ssec:ncsWithNonInfluentCrit}).   

\ssk The set of sufficient coalitions could thus, in principle, be elicited by presenting the alternatives $alt(I)$ for all subsets $I$ of $N$ to the DM. 
Obviously we need an efficient procedure to reduce the number of questions asked to the DM. This is one of the main topics dealt with in this thesis.

\subsection{Minimally sufficient and maximally insufficient coalitions}
The set of sufficient coalitions can be described just by listing its minimal elements. The set of all subsets of $N$ is partially ordered by inclusion. The minimal sufficient coalitions in \calF{} are such that removing any criterion from the coalition makes it insufficient. We denote by \Fmin{} the set of \emph{minimal sufficient coalitions}. 

\ssk Symmetrically, \calG{} indicates the set of insufficient coalitions, \ie $\calG = \Pn \setminus \calF$. The set of maximally insufficient coalitions is denoted by \Gmax{}. A maximally insufficient coalition is such that adding any criterion to it makes it a sufficient coalition which belongs to \calF{}.

\subsection{General \ncs{}}\label{ssec:ncsWithNonInfluentCrit}

We left aside the cases in which the set of acceptable levels $\calA_i$ is empty or all $X_i$, for some criterion $i$. In such a case, the criterion $i$ is not \emph{influential}, meaning that its value may be changed without changing the assignment of any alternative. Here is a precise definition.

\begin{definition}[Influential/Non-influential criterion]\label{def:influentCrit}
We say that a criterion $i\in N$ is influential for the bipartition $<\mathcal{A},\mathcal{U}>$, if there are $x_{i}, y_{i}\in X_{i}$ and $a_{-i}\in X_{-i}$ such that $(x_{i}, a_{-i})\in \mathcal{A}$ and $(y_{i}, a_{-i})\in \mathcal{U}$.

\ssk Criterion $i$ is non-influential if for all $x_{i}, y_{i}\in X_{i}$ and $a_{-i}\in X_{-i}$, we have that $(x_{i}, a_{-i})$ and $(y_{i}, a_{-i})$ belong both either  to $\mathcal{A}$ or to $\mathcal{U}$. 
\end{definition} 

\ssk The variation of the influential criteria values may bring a satisfactory alternative into the unsatisfactory category and conversely. According to the definition of a partition $\calAU$, $\calA$ can neither be empty nor all $N$. Therefore, we can affirm that there is at least one influential criterion in $N$. The criteria which are non-influential don't have any effect on the assignment of the alternatives and may be removed from $N$ if we know them. Unfortunately, we don't know in general which criteria are influential or not. Instead of trying to identify a priori the non-influential criteria, we shall identify the sufficient and unsufficient coalitions and, from them, see which criteria are non-influential. Therefore, we study how to identify influential/non-influential criteria from the knowledge of minimal sufficient or maximal insufficient coalitions. 

\subsection{Properties of influential/non-influential criteria}\label{ssec:propInflNonInfl} 

\begin{proposition}\label{prop:nonInflFmin}
Criterion $i$ is non-influential, if and only if, it
doesn't belong to any coalition $I$ of $\Fmin$.
\end{proposition}

\bpr Let \Mi{} denote the maximal value of $i$th criterion,
\mmi{} the minimal value of $i$th criterion.

\ssk $[\Rightarrow]$ Necessity results from the definition
of \Fmin{}. Removing a criterion from any coalition of \Fmin{} yields an insufficient coalition, while removing a non-influential criterion keeps the coalition sufficient. 

\ssk $[\Leftarrow]$
Let $I$  be any coalition from \Fmin{} and  $a= alt(I)$ the alternative defined by
\[
a_{j} = \left\{
\begin{array}{ll}
    M_{j}, & \hbox{$\forall j \in I$}\\
    m_{j}, & \hbox{$\forall j \not \in I$}\\
\end{array}
\right.
\]
Suppose that $I$ contains $i$. We have that $a = (M_i, a_{-i}) \in \calA$.   Since $I \in \Fmin$, $I \setminus \{i\} \in \calG$. Hence $(m_{i},a_{-i}) \in \calU$. This implies that  $i$ is influential, which is a contradiction. Thus $I$ does not contain $i$. \epr

By negation of Proposition~\ref{prop:nonInflFmin}, we obtain the following. 
\begin{corollary}\label{cor:inflFmin}
Criterion $i$ is influential if and only if it  belongs to at least one coalition $I$ of $\Fmin$.
\end{corollary}

\begin{corollary}\label{cor:influential}
All criteria are influential if and only if \ $\bigcup_{I\in \Fmin} I=N$.
\end{corollary}

\bpr
Using Corollary~\ref{cor:inflFmin} and Proposition~\ref{prop:nonInflFmin}, we have that $\bigcup_{I\in \Fmin} I$ is the set of influential criteria. Hence the result.
\epr

\begin{proposition}\label{prop:nonInflGmax}
Criterion $i$ is non-influential if and
only if $i$ belongs to all coalitions of \Gmax.
\end{proposition}

\bpr
$[\Rightarrow]$ Necessity results from the definition
of \Gmax. Assume there is $I\in \Gmax$ such that $i \not \in I$. Let $a = alt(I)$. Since $I \in \Gmax$, $a \in \calU$. By definition of \Gmax, adding any criterion to $I$ makes it a sufficient coalition. Hence $I \cup \{i\} \in \calF$ and $(\Mi, a_{-i}) \in \calA$. Thus $i$ is influential, which is a contradiction. Therefore $i$ belongs to all coalitions of \Gmax. 

\ssk $[\Leftarrow]$ Let $i$  such that $i\in I$ for all $I\in\Gmax$. Suppose $i$ is influential. By Proposition~\ref{prop:nonInflFmin}, there is $J \in \Fmin$ with $i\in J$. Then $J \setminus \{i\} \in\calG$. There exists $K \in \Gmax$ such that $J \setminus\{i\}\subseteq K$. We have that $i\not\in K$ otherwise, $J\subseteq K$ and $K\in \calF$. This is a contradiction with our assumption.
\epr

By negation of Proposition~\ref{prop:nonInflGmax}, we obtain the following. 
\begin{corollary}\label{cor:inflGmax}
Criterion $i$ is influential if and only if there is a coalition $I$ of $\Gmax$ such that $i \not \in I$.
\end{corollary}

\begin{corollary}\label{cor:influentGmax}
All criteria  are influential if and only if \ $\bigcap_{I\in \Gmax} I = \emptyset$.
\end{corollary}

\bpr
Using Proposition~\ref{prop:nonInflGmax}, we have that $\bigcap_{I\in \Gmax} I$ is the set of non-influential criteria. Hence the result.
\epr

\begin{proposition}\label{prop:CNSnonInfl}
The following are equivalent :
\begin{enumerate}
    \item All criteria are influential;
    \item $\bigcup_{I\in \Fmin} I= N$;
    \item $\bigcap_{I\in \Gmax} I =\emptyset$.
\end{enumerate} 
\end{proposition}


\begin{proposition}\label{prop:influentImplies}
\mbox{ }
\vspace{-0.6cm}
\[ 
i \textrm{  influential } \Rightarrow 
\left\{
\begin{array}{lll}
    \{i\}\in \mathcal{F} \\
    \textrm{or} \\
    \exists J \in\Gmax \textrm{ such that } i \in J.
\end{array}
\right.
\]
\end{proposition}

\bpr If $i$ is influential, then $\exists I \in \mathcal{F}$ such that $i\in I$. If $I=\{i\}$, the proof is done.  Otherwise, we remove $j\neq i$, $j\in I$, from I.
If $I\setminus\{j\}\in \mathcal{G}$, then, $\exists
J\in\Gmax$ such that $I\setminus \{j\}\subseteq J$. Else, $I \setminus\{j\} \in \calF$. We do $I \leftarrow I\setminus\{j\}$ and continue in a similar way until $I \setminus\{j\} \in \calG{}$ or $I \setminus\{j\} = \{i\} \in \calF$.
\epr

We may use these properties in practice. Without trying to detect in advance which criteria are influential and which are not, we may question the DM in order to determine the sufficient coalitions. We then compute \Fmin. The union of the elements in \Fmin{} is the set of influential criteria (Corollary~\ref{cor:influential}). This can also be done with \Gmax. The criteria belonging to all coalitions of \Gmax{} are non-influential. The rest are influential. 

\begin{remark}
All the results in this section remain true in case $X_i$ has no maximal and/or no minimal element. Since $\calA_i$ is neither empty nor all $X_i$, the proofs remain correct if we interpret $M_i$ (\resp $m_i$) as an element of $\calA_i$ (\resp $\calU_i$).     
\end{remark}




\subsection{The coalitions graph}\label{ssec:coalGraph}
\begin{center}
\begin{figure}[h!!!]
\includegraphics[width=130mm,height=105mm]{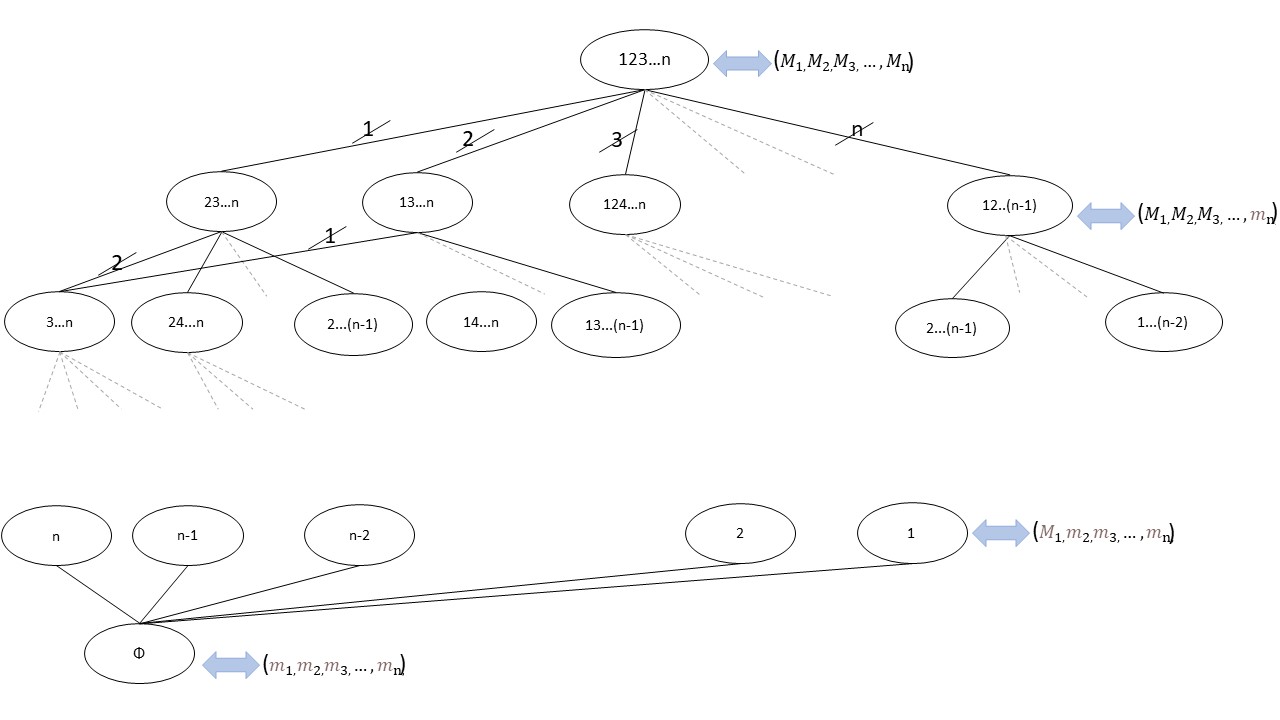}
\caption{The coalitions graph \label{fig:theGraph}}
\end{figure}
\end{center}

A useful representation for describing the algorithms for questioning about sufficient coalitions is the coalitions graph $G$. The set of nodes of graph $G$ is $\Pn$, the set of all the subsets of $N$ criteria. The edges of the graph are the inclusion relation between subsets. Figure~\ref{fig:theGraph} represents the coalitions graph. Only edges connecting immediate successors in the inclusion relation are drawn, \ie only edges linking a set of criteria and its subsets obtained by removing a single criterion.   

\ssk In the rest of this chapter, we denote coalitions by lower case letters such as $u, v, w$ (instead of capital letters $I, J, K$ as before).  

\ssk The graph $G$ is organized in levels: all coalitions in a level involve the same number of criteria. We denote by $L(k)$ the set of coalitions containing $n-k$ criteria. The set $N$ is thus the only element in $L(0)$ and $\varnothing$ is the only element in $L(n)$. For going from level $L(k)$ to level $L(k+1)$, we remove respectively one of the criteria belonging to the coalitions of level $L(k)$. The cardinality of the coalitions belonging to the same level $L(k)$ decreases by one unit when going to level $L(k+1)$. 

\ssk Any coalition $u$ is associated with an alternative $alt(u)$ having the maximal values $M_i$ for the criteria $i \in u$, and the minimal values $m_i$ for the criteria $i \not \in u$. The graph thus represents the coalitions $u$ and the alternatives $alt(u)$ used for the elicitation of sufficient and insufficient coalitions. Several questioning algorithms are proposed in the further section. These algorithms browse the whole graph, discover all elements $u$ of the graph, record them as a list and obtain the sufficient coalition sets $\mathcal{F}$ and $\Fmin$, as the outputs of the algorithms. The general purpose of this work is to find an algorithm which needs to ask a minimal number of questions efficiently.  

\section[Algorithms for the sufficient coalitions]{Algorithms for the identification of the sufficient coalitions} In this section, we present several algorithms for the identification of the sufficient and insufficient coalitions. This is the first step of our elicitation process which is based on questions asked to the DM in order to determine the sufficient coalitions and the limit profile. According to the answers given by the DM, we follow paths in the graph $G$ and stop when all nodes are labeled $S$, for sufficient, or $I$, for insufficient. We mainly develop questioning procedures based on a depth-first search (DFS) exploration strategy of the graph $G$. An alternative way based on breadth-first search is briefly described, as an appendix, in Section~\ref{sec:Breadth-First}.

\subsection{Depth-First Search (DFS)}
The principle of the depth-first search algorithm \citep[][Section~23.3]{CormenEtAl1990} is to browse the whole graph in depth, discovering all elements $u$, recording the predecessor $\Pi(u)$ of each node $u$ and backtracking when all of the edges starting from $u$ are explored. We propose three different versions of a questioning algorithm based on depth-first search.  In the first version (DFS1), a question is asked to the DM at the deepest node (the end) of each branch of the depth-first arborescence. The second version (DFS2) consists of exploring the whole graph in depth, while asking the question as soon as we reach an element for the first time. The third one (DFS3) counts the maximal length of the unexplored branch and asks the question to the midway element. With the graph structure we assumed, we obtain the sufficient coalition sets $\mathcal{F}$ and $\Fmin$ as the outputs of the first two depth-first search questioning algorithms. We obtain also the insufficient coalitions $\calG$. For a graph constructed upside-down (\ie with the empty set above and arcs representing the relation ``is included in''),  the algorithms would give us the insufficient coalition sets $\mathcal{G}$ and $\mathcal{G}_{max}$, and the sufficient coalitions $\calF$. The third questioning version gives us $F$ and $G$ as output, independently of the construction of the graph. 

A depth-first search creates an arborescence in the graph in Figure~\ref{fig:theGraph} whose set of nodes is $\Pn$, the set of all subsets of the set of criteria $N$. The set $Succ(u)$ is the set of all successors of $u$ in the graph, that is to say that $Succ(u) = \{ v: v\subseteq u \textrm{ and } |u\setminus v|=1\}$. Let $\Pi(u)$ denote the predecessor of $u$ in the search. The assignment of the alternative associated to the node $u$ is encoded as $qual(u)$.

\subsection{DFS1: Question at the end}\label{ssec:DFS1}

This is a depth-first search algorithm that starts the questioning procedure in the deepest nodes of the graph. The pseudo-code of the DFS1 algorithm, in recursive form, is given in Algorithm~\ref{algo:DFS1}. 

\begin{breakablealgorithm}
    \caption{DFS1}\label{algo:DFS1}
    \begin{algorithmic}[0]
    \For{each $u \in P(N)$}
        \State $color(u)\leftarrow$ $white$
        \State $\Pi(u)\leftarrow$ $Nil$
        \State $qual(u)\leftarrow$ ?
    \EndFor{}
    \State $\mathcal{F}\leftarrow\emptyset$
    \State $\mathcal{G}\leftarrow\emptyset$
    \State $\Fmin\leftarrow \emptyset$
    \State $u \leftarrow N$
    \If{$color(u)=white$}   \Comment{white element $u$ has just been discovered.}
        \State \Call{Depth}{$u$} 
    \EndIf
    \State \Call{Ask}{$u$} 
\State 

\Algphase{Procedure 1 - Depth}
\Procedure{Depth}{$u$}
    \State $color(u)\leftarrow gray$ 
    \For{each $v\in succ(u)$}                   \Comment{explore edge$(u,v)$}
        \If{$qual(v)=S$}                        
            \State $qual(u)=S$
        \EndIf
        \If{$color(v)=white$}
            \State $\Pi(v)\leftarrow$ $u$
            \State \Call{Depth}{$v$}
        \EndIf
    \EndFor
    \State \Call{Ask}{$u$}    
    \State $color(u)\leftarrow$ $black$ 
    \Comment{all successors of $u$ visited and labeled}
\EndProcedure  
\State

\Algphase{Procedure 2 - Ask}
\Procedure{Ask}{$u$}
    \If{$qual(u) = ?$}
        \State Ask: $alt(u)\in \calA ?$
        \If{answer = yes}
            \State $qual(u) \leftarrow S$ 
            \State $\Fmin \leftarrow\Fmin\cup\{u\}$ 
            \State $\mathcal{F}\leftarrow \mathcal{F}\cup\{u\}$
        \Else
            \State $qual(u) \leftarrow I$ 
            \State $\mathcal{G}\leftarrow \mathcal{G}\cup\{u\}$ 
        \EndIf
    \EndIf 
    \If{$qual(u) = S$} 
        \State $qual(\Pi(u)) \leftarrow S$ 
        \State $\mathcal{F} \leftarrow\mathcal{F}\cup\{\Pi(u)\}$
    \EndIf
\EndProcedure 
    \end{algorithmic}
\end{breakablealgorithm}

As explained before, this version of the depth-first search algorithm amounts to ask the question at the end of the branch on which we are situated. If the last element of the branch is a sufficient coalition, all predecessors, which contain this coalition, will be labeled as $S$ to represent their sufficiency. If the last element of the branch is an unsatisfactory coalition (labeled as $I$ to represent its insufficiency), the algorithm backtracks and looks for the other successors. An element of the graph becomes black when it is discovered and all of its successors have also been discovered and labeled. This algorithm can be preferably applied when we have the intuition or information that even small coalitions are sufficient. Our objective was to find an efficient questioning algorithm which aims to avoid a large  number of questions. According to this objective we illustrate below the performance of the algorithm in particular cases with three criteria.

\begin{center}
\begin{figure}[h!!!]
\vspace{-.5cm}
\includegraphics[width=140mm,height=9cm]{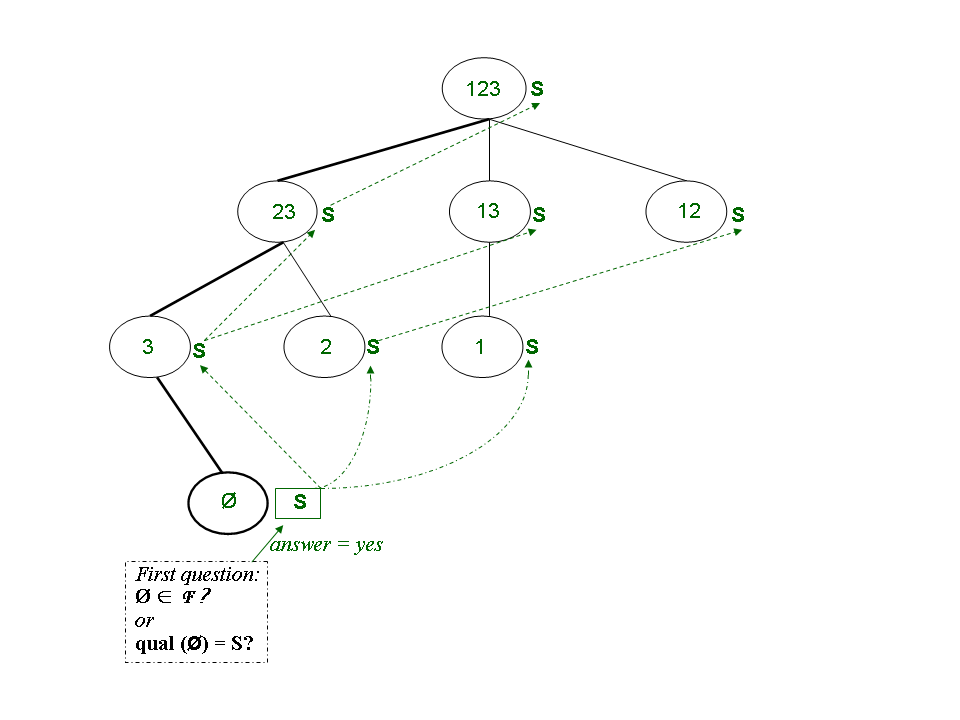}\caption{DFS1best}  \label{fig:DFS1_best}
\end{figure}
\end{center}

\paragraph{Best case} The best of the cases occurs when all elements of the graph are $S$. This situation needs just 1 question at $\varnothing$. Figure~\ref{fig:DFS1_best} illustrates the functioning of the algorithm in this case. The boldfaced edges show the initial path in depth-first search. At the end of this path, a question is asked for coalition $u = \varnothing$. The answer is ``yes, $alt(\varnothing)$ is satisfactory''. Thus $\varnothing$ is $S$. The algorithm backtracks and propagates the label $S$ to its predecessor $3$ and then $23$. When in $23$ it goes down to $2$, whose successor $\varnothing$ is $S$. Hence $qual(2) = S$. The algorithm progresses further in similar way without asking further questions. The green dashed lines with an arrow in the figure indicate how the label $S$ is propagated during the search. Note that propagation of the label $S$ of a coalition does not occur immediately. Propagation occurs only when backtracking to the predecessor or when reaching a node having the coalition among its successors. For instance, the $S$ label of  coalition $\varnothing$ is propagated to the coalitions $2$ and $1$ only when these nodes are visited by the depth-first search arborescence. 

\begin{center}
\begin{figure}[h!!!]\includegraphics[width=140mm,height=10cm]{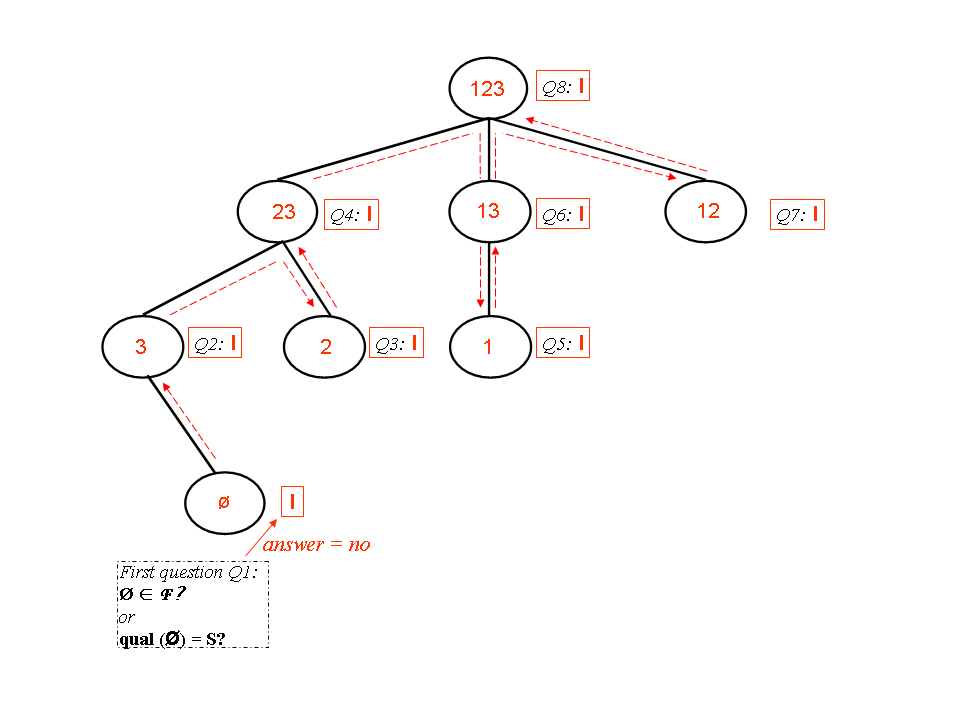}
\vspace{-2cm}\caption{DFS1worst\label{fig:DFS1_worst}}
\end{figure}
\end{center}

\paragraph{Worst case} The worst of the cases occurs when all elements of the graph are $I$. This situation needs $2^{N}$ questions to be asked, \ie all possible questions are asked.  Figure~\ref{fig:DFS1_worst} illustrates that case. The dashed lines with arrow indicate the backtracking path in the DFS arborescence. No label propagation occurs. 

\begin{center}
\begin{figure}[h!!!]
\includegraphics[width=140mm,height=10cm]{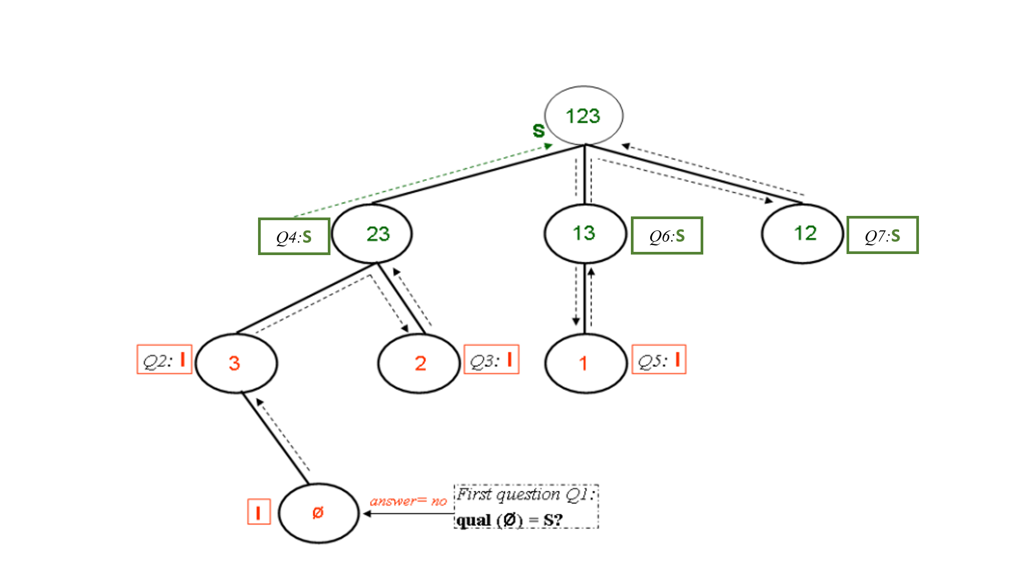}
\caption{DFS1maj\label{fig:DFS1_maj}}
\end{figure}
\end{center}

\paragraph{The case of the majority of the criteria} That is the case in which the coalitions including at least half of the criteria are sufficient. This means that: $\forall u\in P(N)$, we have $|u|> \frac{N}{2}$ if and only if  $u\in\mathcal{F}$. The algorithm asks 7 questions. Figure~\ref{fig:DFS1_maj} illustrates this case. The only label propagation occurs when the $S$ label of $23$ is transferred to $123$, when backtracking to the predecessor of $23$.

Due to the structure of the graph in Figure~\ref{fig:theGraph}, the elements having high cardinality are situated to the superior levels of the graph (levels $L(k)$ with small $k$). DFS1 deeply descends and asks questions to the elements in the lowest levels which have low cardinality. Therefore, the algorithm  DFS1 is not a good choice in the case only large coalitions are sufficient. 

\paragraph{The number of avoided questions } Each answer $I$ given for an insufficient coalition $u$ in level $L(k)$, allows us to give the label $I$ to all undiscovered subsets of $u$. Coalition $u$ in level $L(k)$ has $n-k$ elements. Therefore,
\begin{itemize}
    \item answer $I$ in node $u$ eliminates a number of questions between $|\{ v : v \subseteq u \textrm{ and } color(v) = white\}|$ and $2^{n-k}-1-|\{ v : v \subseteq u \textrm{ and } color(v)\neq black\}|$; \item answer $S$ in node $u$ eliminates: $|\{v : u \supset v \textrm{ and } qual(v)= ?\}|$ questions.
\end{itemize}
These estimations of the number of questions avoided takes into account the states of $color$ and of the labeling $qual$ of the nodes at the time the question is asked in $u$. The impact of an answer in node $u$ depends on the stage of exploration, \ie on the number of nodes already discovered and and labeled.

\subsection{DFS2: Question immediately}\label{ssec:DFS2}

The second version of the depth-first search algorithm, called ``DFS2'', immediately asks a question when it discovers an element of the graph during the deeply descending process. If the indicated element is a sufficient coalition, all predecessors which contain this coalition are labeled as $S$. If the last element of the branch is an insufficient coalition, all its successors are labeled as $I$. This algorithm begins the backtracking when it reaches the deepest element, has the answer and marks it black to say that we finished with its successors. The algorithm of DFS2 can be preferred when we have the intuition or information as the big coalitions are insufficient. 

\ssk A new variable ``Flag'' is used and permits to identify the elements in \Fmin. The pseudo-code of the DFS2 algorithm is given in Algorithm~\ref{algo:DFS2}.

\begin{breakablealgorithm}
    \caption{DFS2}\label{algo:DFS2}
\begin{algorithmic}[0]
    \For{each $u \in P(N)$}
        \State $color(u)\leftarrow$ $white$
        \State $\Pi(u)\leftarrow$ $Nil$
        \State $qual(u)\leftarrow$ ?
        \State $Flag(u)\leftarrow I$
    \EndFor
    \State $\mathcal{F}\leftarrow\emptyset$
    \State $\mathcal{G}\leftarrow\emptyset$
    \State $\Fmin\leftarrow\emptyset$
    \State $u \leftarrow N$
    \If {$color(u)=white$}  \Comment{white element $u$ has just been discovered}
            \State \Call{Depth}{$u$}
    \EndIf
\Statex

\Algphase{Procedure 3 - Depth}
\Procedure{Depth}{$u$}
    \State $ color(u)\leftarrow gray$
    \If{$\Pi(u) \neq Nil$ and $qual(\Pi(u)) = I$}
        \State $qual(u)=I$ and $\calG \gets \calG \cup \{u\}$
     \Else 
     \State \Call{Ask}{$u$}
     \EndIf
     \For{each $v\in succ(u)$}  \Comment{explore edge$(u,v)$}
        \If{$color(v)=white$}
            \State $\Pi(v)\leftarrow u$
            \State \Call{Depth}{$v$}
        \EndIf 
        \If{$qual(v)=S$} 
            \State $Flag(\Pi(v))\leftarrow$ S
        \EndIf
    \EndFor
    \State $color(u)\leftarrow$ $black$ \Comment{blacken $u$ it is finished.}
    \If{$Flag(u)=I$ and $qual(u)=S$} 
        \State $\Fmin\leftarrow\Fmin\cup\{u\}$
    \EndIf    
\EndProcedure
\State
\State

\Algphase{Procedure 4 - Ask}
\Procedure{Ask}{$u$}
    \If{$qual(u) = ?$}
        \State Ask: $alt(u)\in \calA ?$      
        \If{answer = yes} 
            \State $qual(u)\leftarrow S$ and $\mathcal{F} \leftarrow \mathcal{F}\cup\{u\}$
        \Else 
            \State $qual(u)\leftarrow I$ and $\calG \gets \calG \cup \{u\}$
        \EndIf
    \EndIf    
    \If{$qual(u) = S$} 
        \State $qual(\Pi(u)) \leftarrow S$ and $\mathcal{F}\leftarrow \mathcal{F}\cup\{\Pi(u)\}$
     \EndIf   
\EndProcedure
\end{algorithmic}    
\end{breakablealgorithm}

The performance of the algorithm is analyzed for particular situations.

\begin{center}
\begin{figure}[h!]\includegraphics[width=140mm,height=10cm]{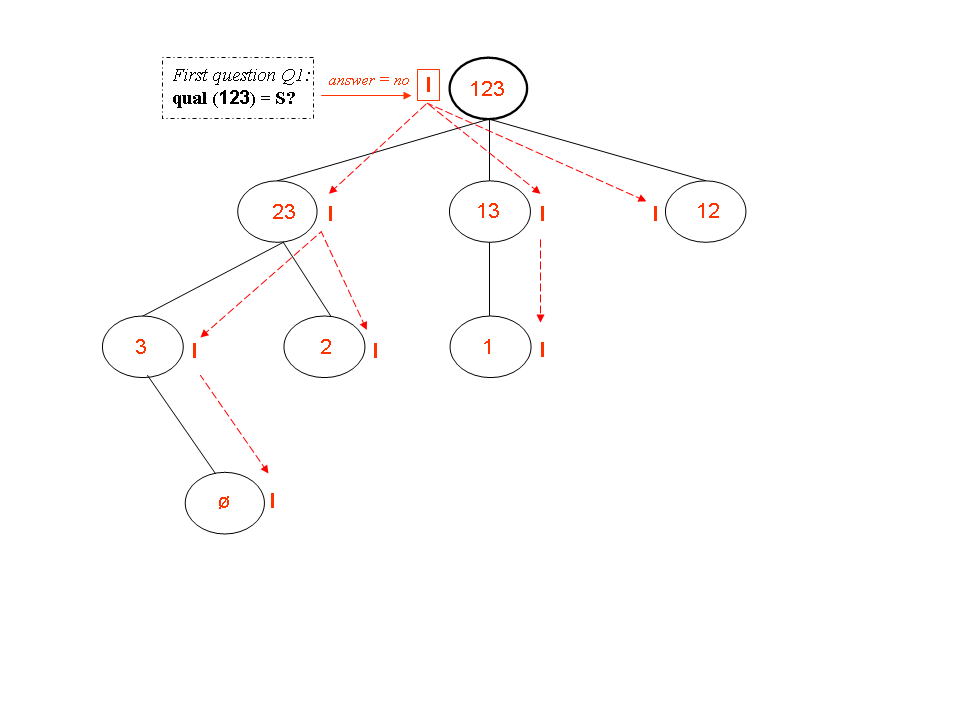}
\vspace{-3cm}
\caption{DFS2best}
\label{fig:DFS2-best}
\end{figure}
\end{center}

\paragraph{Best case } The best of the cases occurs when all elements of the graph are $I$. This situation needs just 1 question at the first level of the graph. Figure \ref{fig:DFS2-best} illustrates this case. The dashed lines with arrow represent the propagation of the $I$ label. 

\begin{center}
\begin{figure}[h!]\includegraphics[width=140mm,height=9cm]{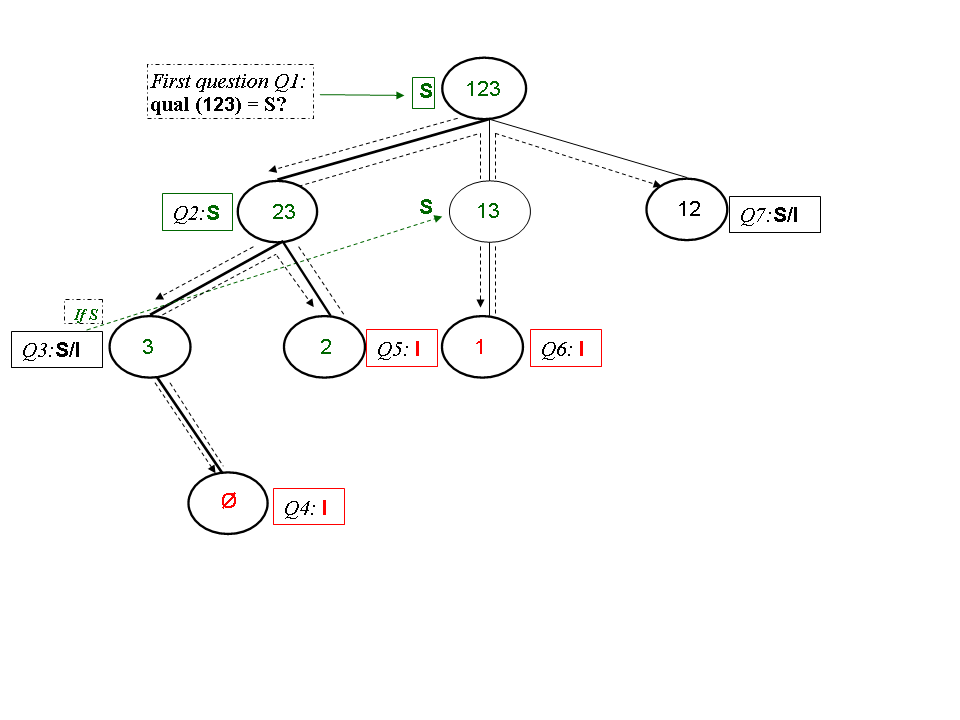}
\vspace{-3cm}
\caption{DFS2worst}
\label{fig:DFS2-worst}
\end{figure}
\end{center}

\paragraph{Worst case }  For $n\geq 3$, there is not only one specific situation which can be considered as the worst. But having sufficient coalitions at the top levels of the graph and insufficient coalitions at the last, we can obtain several graphs which require $2^{n}-1$ questions to the maximum. Figure~\ref{fig:DFS2-worst} illustrates such a case (actually, four such cases) for $n=3$. The sufficient coalitions include 123, 23, 13. The insufficient coalitions include 1, 2, $\varnothing$. The coalitions 3 and 12 may be, independently $S$ or $I$, this doesn't change the number of questions. Take coalition 3 for instance. If 3 is $S$, the propagation of this label avoids 1 question (the 13). If 3 is $I$, the question in $\varnothing$ is avoided. 

Note that there is a small improvement as compared to the worst case for DFS1. We shall prove below that this improvement is valid for all $n \geq 3$. For $n=2$ however, the worst of the cases imposes to ask $2^2 = 4$ questions. Consider the example in which, among the four elements of the graph, $\varnothing$ is insufficient and the others are considered as sufficient to belong to $\mathcal{F}$. This situation requires $2^{2}=4$ questions to ask which means that all questions will be asked.

\begin{center}
\begin{figure}[h!]\includegraphics[width=140mm,height=100mm]{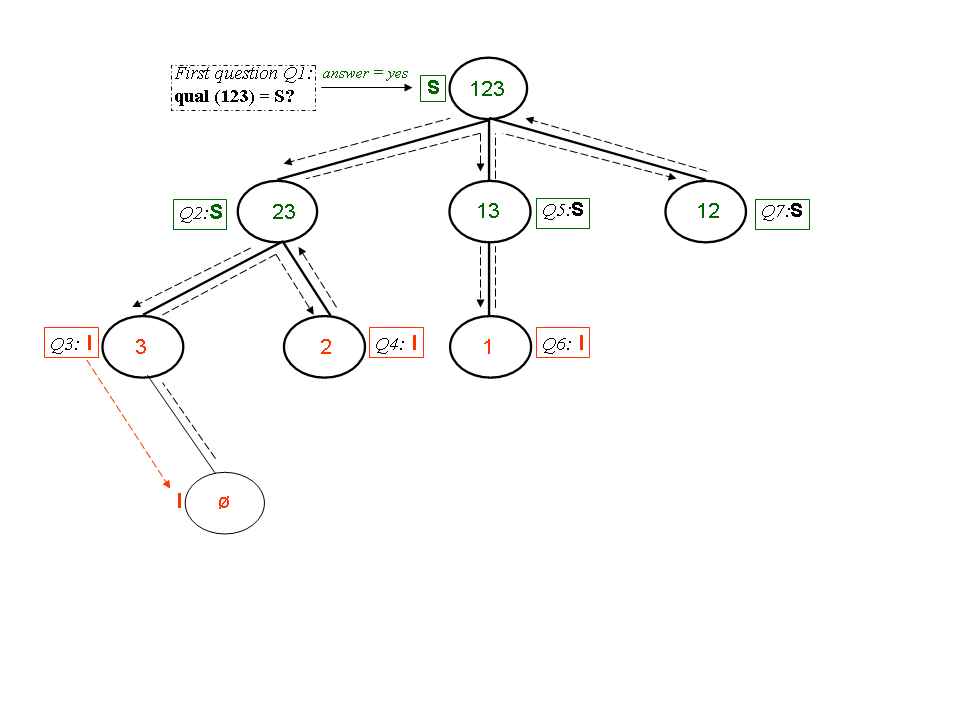}\vspace{-3cm}\caption{DFS2maj \label{fig:DFS2_maj}}
\end{figure}
\end{center}

\paragraph{The case of the majority of the criteria } The coalitions possessing the majority of the criteria are considered sufficient and questions are asked immediately, when an element is discovered. This case is illustrated on Figure~\ref{fig:DFS2_maj}.

In the case $n=3$ (Figure~\ref{fig:DFS2_maj}), beginning from the top of the graph, the first two questions are asked to elements having high cardinality. Since these are sufficient coalitions by definition, there is no propagation and we must go down the graph. With three criteria, there is no difference in the number of questions between DFS1 and DFS2 in the case of the majority of the criteria. However, for $n=4$, DFS2 asks 11 questions while DFS1 asks 15 questions. It seems that for $n \geq 4$, DFS2 performs better than DFS1 in the case of the majority of criteria.

\paragraph{The number of the avoided questions } What was said in the corresponding paragraph for DFS1 remains true. It is not obvious to count the number
of the avoided question, but, better than DFS1, we can affirm that for all $n \geq 3$, DFS2 avoids at least one question. 
\begin{proposition}\label{prop:DFS2Avoids1Question}
For $n \geq 3$ criteria, DFS2 asks at most $2^{n} -1$ questions.    
\end{proposition}
The proof of this proposition can be found in Section~\ref{sec:resultDFS2} as an appendix.

\subsection{DFS3: Question in the middle of the path}\label{ssec:DFS3}
This algorithm discovers the graph according to depth first search and questions in the middle of the branches. On each path, it numbers each unexplored and unlabeled element from 1 to $l_{max}$, the maximal length of the branch. It asks the question at the middle of the branch, \ie to the element numbered $\lceil\frac{l_{max}}{2}\rceil$. 

\ssk If the answer is $I$, all elements situated in the lower part of the branch are labeled $I$. All subsets of the concerned element will be labeled immediately without waiting to discover them through the exploration process of depth-first search. Then, we backtrack and apply the same  procedure for the unevaluated rest of the branch. 

\ssk If the answer is $S$, all elements being the supersets of the concerned element are labeled $S$. Taking the answer $S$, the depth-first search algorithm continues through the lower part of the branch and asks the next question to the unevaluated rest part of the branch.

\ssk The pseudo-code of DFS3 is sequential  Algorithm~\ref{algo:DFS3}. As before, $succ(u)$ is the set of (immediate) successors of $u$ in the graph $\Pn$. The $color$ of a node (or coalition) is ``white'' (\resp ``gray''; ``black'') when the node is non visited (\resp visited; visited and all its successors are visited). We introduce the set $VisitedSucc(u)$ (\resp $NonVisitedSucc(u)$) which is a subset of $succ(u)$ containing all visited (\resp non visited) successors of $u$. Initially, $VisitedSucc(u)$ (\resp $NonVisitedSucc(u)$) is empty (\resp is equal to $succ(u)$). Whenever the ``color'' of a node in $succ(u)$ becomes ``gray'', this node leaves $NonVisitedSucc(u)$ and enters $VisitedSucc(u)$. The union of these two sets is always $succ(u)$.  

\begin{breakablealgorithm}
    \caption{DFS3}\label{algo:DFS3}
    \begin{algorithmic}[0]
    \For{each $u \in P(N)$}
        \State $color(u)\leftarrow$ $white$ \Comment{``white'' means non visited}
        \State $\Pi(u)\leftarrow$ $Nil$
        \State $qual(u)\leftarrow$ ?
    \EndFor{}
    \State $\mathcal{F}\leftarrow\emptyset$
    \State $\mathcal{G}\leftarrow\emptyset$
    \State{}
    \State $u \leftarrow N$ \Comment{$u$ is a global variable}
    \State $color(N) \gets gray$  \Comment{``gray'' means visited}
    \While{$color(N)=gray$}   
        \State \Call{Down}{} 
    \EndWhile
\State 

\Algphase{Procedure 5 - Down}
\Procedure{Down}{}
    \State $A \gets \emptyset$ \Comment{$A$ is an array}
    \If{ $qual(u)= ?$}
           \State put $u$ in array $A$ as its first element
    \EndIf
    \While{$NonVisitedSucc(u)$ not empty}    
        \State $v \gets$ first element in $NonVisitedSucc(u)$
        \State $color(v) \gets gray$
        \State $\Pi(v) \gets u$
        \If{ $qual(v)= ?$}
           \State put $v$ in array $A$ as its next element
        \EndIf
        \State $u \gets v$
    \EndWhile
    \State \Call{AskDicho}{A}
    \State $color(u) \gets black$
    \State $u \gets \Pi(u)$
    \EndProcedure  
\State

\Algphase{Procedure 6 - AskDicho}
\Procedure{AskDicho}{$A$}
    \While{$\text{Card}(A)>0$}
        \For{$i\gets 1, \text{card}(A)$}
            \State{$K(A(i))\gets i$}    \Comment{$K(w)$ is the level of $w$ in the current subbranch}
        \EndFor
        \State $midA \gets \lceil \frac{\text{Card}(A)}{2} \rceil $
        \State $w \gets A(midA)$            \Comment{the $midA$-th element in $A$}
        \State Ask: $alt(w) \in \calA ?$
        \If{answer is ``yes''}
            \State $qual(w) \gets S$ and $\calF \gets \calF \cup \{w\}$
            \For{ all $x$ with $K(x) < midA$}
                \State delete $x$ from $A$; 
            \EndFor
            \For{all $x$ superset of $w$}
                \State $qual(x) \gets S$ and $\calF \gets \calF \cup \{x\}$
            \EndFor
        \Else
            \State $qual(w) \gets I$ and $\calG \gets \calG \cup \{w\}$
            \For{ all $x$ with $K(x) > midA$}
                \State delete $x$ from $A$; 
            \EndFor
            \For{all $x$ subset of $w$}
                \State $qual(x) \gets I$ and $\calG \gets \calG \cup \{x\}$
            \EndFor
        \EndIf
        \State delete $w$ from $A$
    \EndWhile
\EndProcedure 
    \end{algorithmic}
\end{breakablealgorithm}

\begin{center}
\begin{figure}[h!!!]
\centering
\includegraphics[width=120mm]{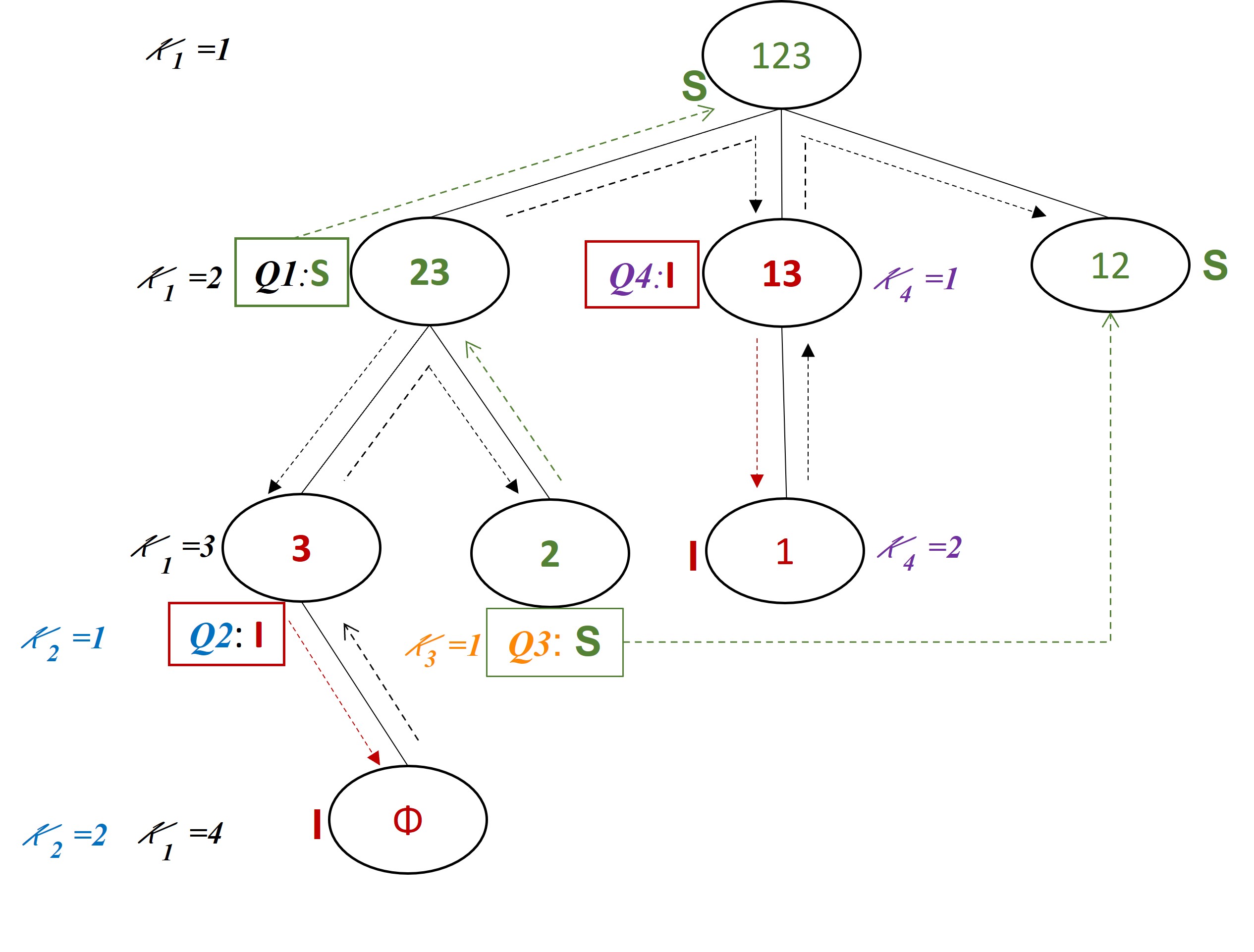} \caption{DFS3 in case the sole minimal sufficient coalition is the singleton $2$   \label{fig:DFS3}}
\end{figure}
\end{center}

To fix the ideas, consider the example in Figure~\ref{fig:DFS3}. 
In this example, the set of minimal coalitions $\Fmin = \{2\}$. We start by numbering the elements of the first path 123-23-3-$\varnothing$. These are numbered $K_1=1$ to $K_1=4$, respectively. The middle element has number $K_1=2$. We ask the question : is $23$ sufficient or not ? Here $23$ is $S$. Then we propagate to its superset $123$ in the first branch and label it as $S$. Then, we number the elements in the rest 3-$\varnothing$ of the first branch, without including $23$ which is now labeled, \ie $K_2=1$ for $3$, $K_2=2$ for $\varnothing$. The middle element of the new branch with two elements is the first one, which is $3$. After questioning $3$, $3$ becomes $I$. Hence its subset $\varnothing$ which is in the same branch becomes $I$ without asking a question. 
We backtrack towards $23$ and number the elements of the branch 23-2, without numbering 23, which is already labeled. Element 2 is numbered $K_3=1$. The question is for the unique unlabeled element $K_3=1$, \ie $2$, which, after questioning, is labeled as $S$. We propagate the label $S$ to all unlabeled supersets of $2$, here only $12$. We then, backtrack to $123$ and number elements in the second unexplored branch starting from 123. The unlabeled part of the branch 13-1 is numbered as respectively $K_4= 1$ for $13$, $K_4=2$ for $1$. The question is for $13$ and takes $I$ as an answer. Hence its subset $1$ becomes $I$ without asking a question. This branch is explored and labeled, and we backtrack to $123$. $12$ coming from $123$ is still not explored but it is already labeled $S$. As all elements of the graph have been explored and labeled, we found that $\Fmin = \{2\}$ using $4$ questions, hence avoiding $4$ questions. 

\begin{center}
\begin{figure}[h!]
\centering
\includegraphics[width=120mm]{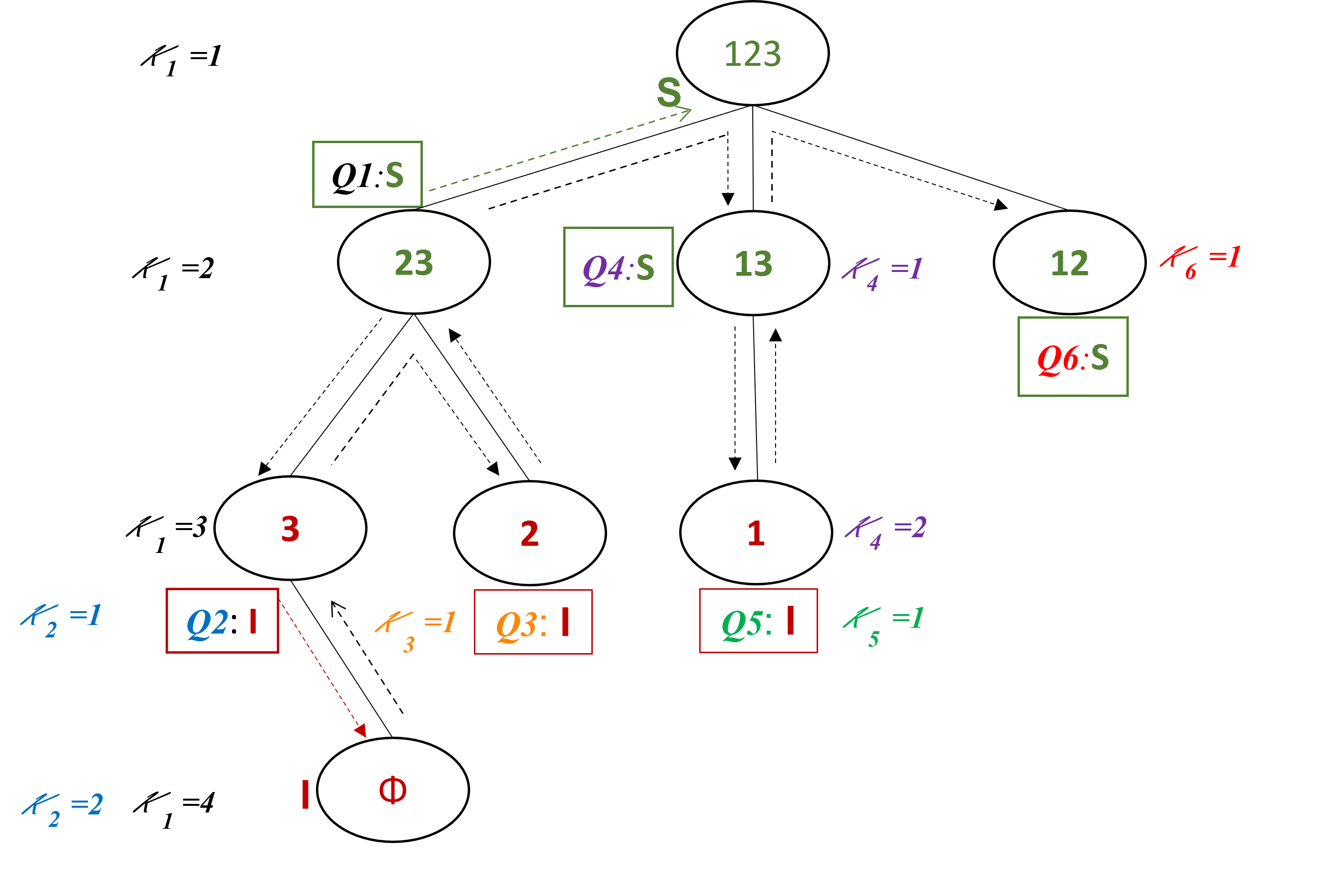} \caption{DFS3 Majority of the criteria \label{fig:DFS3MAJ}}
\end{figure}
\end{center}

\ssk Let us consider now the case of the majority of the criteria, as with DFS1 and DFS2, which is illustrated in Figure~\ref{fig:DFS3MAJ}. In this example, the set of minimal coalitions $\Fmin = \{23, 13, 12\}$. We start by numbering the elements of the first path 123-23-3-$\varnothing$. These are numbered $K_1=1$ to $K_1=4$. The middle element has number $K_1=2$. We ask the question : is $23$ sufficient or not ? Here $23$ is $S$. Then we propagate to its superset $123$ and label as $S$. 
Then, we number the elements in the rest of the first branch, without including $23$,  \ie $K_2=1$ for $3$, $K_2= 2$ for $\varnothing$. The middle element is situated at level $2/2=1$, which is $3$. The question is thus asked to $3$ and the answer is $I$. Hence $\varnothing$ is $I$. 
We backtrack towards $23$ and number the second branch starting from $23$ containing $2$. As we don't give numbers to labeled elements, $K_3=1$ for the only undiscovered element $2$. The third question is $K_3=1$ for $2$, which its answer is $I$. We backtrack to $123$ and number the unexplored branch starting from $123$. The unexplored branch 13-1 is numbered by respectively $K_4= 1, 2$. We ask in the middle for $K_4=1$ and get that $13$ is $S$. Then we propagate, but $123$ was already labeled as $S$. As there is an unexplored element below the branch, we continue descending to $1$. The fourth answer is $I$ and there is no unexplored elements below, so we backtrack to the top. We are now on the third path coming from $123$. The only unlabeled element in this branch is $12$; it is numbered $K_6=1$ and the sixth question is asked to $12$. The answer is $S$ and all elements of the graph have been labeled. We found that \Fmin $= \{23, 13, 12\}$ using $6$ questions, hence avoiding 2 questions which are 123 and $\varnothing$.



\ssk We shall describe the  implementation of the DFS3 algorithm in Chapter~\ref{chap:questioner}, and see that the best case requires $3$ questions (for instance, when $\Fmin =\{1\}$). The worst case requires $6$ questions (for instance, when $\Fmin = \{3\}$). The performance of this algorithm in terms of the number of questions and the average number of questions asked will be studied in depth there.   


\ssk With DFS3, all questions are asked in the middle of the undiscovered part of a branch. Each answer allows us to avoid the questions of about half of the undiscovered elements situated at the upper or lower part of the mid-point element. Currently this is our best algorithm which avoids a maximal number of questions. The resulting software is named ``the main program'' and detailed in Chapter~\ref{chap:questioner}. The simulation module ``questioner'' which executes ``the main program'' for all possible sets of sufficient coalitions is also explained in Chapter~\ref{chap:questioner}.  


\section{Determining the limit profile values}\label{sec:limitProfile} 
After having determined the set of the minimum sufficient coalition \Fmin and the set of maximum the insufficient coalition \Gmax, we may determine which criteria are non-influential (using Propositions~\ref{prop:nonInflFmin} or \ref{prop:nonInflGmax}). For each influential criterion $i$, we must determine the value of $b_i$ which represents the category limit. We assume that $X_{i}$ is a numerical interval and that the maximal and the minimal values are respectively $M_i$ and $m_i$. Thus $b_i \in [m_i, M_i]$. First, we need to decide, for each criterion $i \in N$, an acceptable precision threshold $\delta_{i}$. 

\ssk The main idea is the following.  If an (efficient) criterion belongs to coalition $I \in \Fmin$, the alternative $alt(I)$ (which has value $M_j$ if criterion $j$ belongs to $I$ and, else, has value $m_j$) is in \calA{}. If we change its value $M_i$ into $m_i$, it no longer belongs to $\calA$ because the coalition of criteria for which the alternative is at least as good as the limit profile, is now $I \setminus \{i\}$, which is an insufficient coalition. By decreasing the value of $alt(I)$ on criterion $i$, we shall come to the lowest one for which the alternative is still in \calA{}, which is $b_i$, the value of the profile on criterion $i$.  We search for $b_i$ by dichotomy. We first ask the DM if substituting the value $M_i$ of $alt(I)$ on criterion $i$ into the average of $m_i$ and $M_i$ is a satisfactory alternative.  According to the answer we  proceed with the upper or the lower half part of the value range $[m_i, M_i]$. If the answer is ``yes'', we keep the lower part, otherwise the upper part. The dichotomy continues until obtaining a value range of length at most $\delta(i)$. The last value obtained before reaching the precision threshold is identified as the value of $b_i$, the category limit for the $i^{th}$ criterion. 

\ssk The pseudo-code of the corresponding procedure is in Algorithm~\ref{algo:bi}.  The input of the algorithm consists of the label of criterion $i$, a coalition $I \in \Fmin$ containing $i$, the maximal and minimal values on all criteria scales, and the precision threshold $\delta_i$. 

\ssk The algorithm must be applied for each influential criterion. By Corollary~\ref{cor:inflFmin}, for each influential criterion $i$ there is a coalition in \Fmin{} which contains it. Therefore, the procedure can be applied. 

\begin{breakablealgorithm}
    \caption{Limit profile value $b_i$\label{algo:bi}}
\begin{algorithmic}[0]
    \Procedure{Dicho}{$i, I, m_i, M_i, \delta_i$}
        \Repeat    
            \State $x_{i} \gets \frac{M_{i} + m_{i}}{2}$
            \State Let $x = (x_1, \ldots, x_i, \ldots, x_n)$, with $x_j = M_j$ for $j \in I \setminus \{i\}$ 
            \State and $x_j = m_j$ for $j \in N \setminus I$  
            \State Ask: is alternative $x \in \calA$ ?       
            \If{answer = yes}
                \State $M_{i} \gets x_{i}$
            \Else
                \State $m_{i} \gets x_{i}$           
            \EndIf
            \State $b_i \gets M_i$
        \Until {$M_{i}-m_{i} \leq \delta_i$}        
    \EndProcedure 
\end{algorithmic}    
\end{breakablealgorithm}

\begin{center}
\begin{figure}[h!!!]
\centering
\includegraphics[width=130mm]{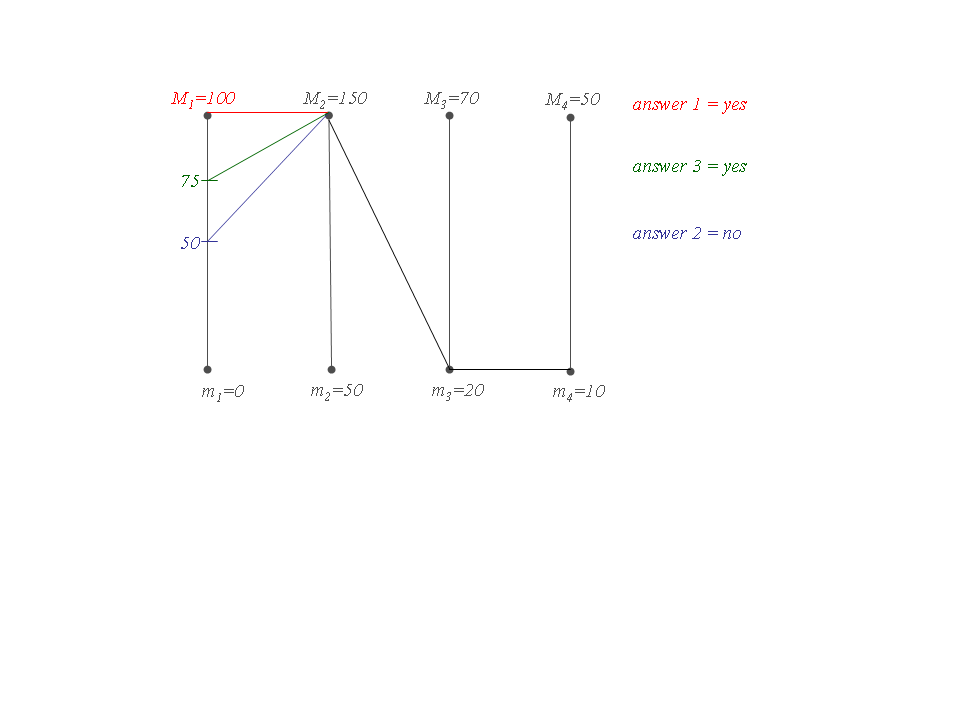}\vspace{-4cm}\caption{Determination of the limit value for the 1st criterion} \label{fig:illustrProcbi}
\end{figure}
\end{center}

\clearpage 
We illustrate this procedure on an example. Assume that we have four criteria with their value ranges $[m_i, M_i]$ represented in Figure~\ref{fig:illustrProcbi}. Let $\Fmin=\{12, 134\}$, and the precision threshold on criterion 1 be $\delta_1=25$. The profile value $b_1$ can be elicited using either coalition in \Fmin{}. Let us use $12$. The alternative $(100, 150, 20, 10)$ belongs to the satisfactory category $\calA{}$ since it has the maximal values on the criteria of the sufficient coalition 12 (and minimal values on 34). We know that alternative $(0, 150, 20, 10)$ belongs to \calU{}.  
The first question is asked for $(50,150,20,10)$, which evaluation on criterion 1 is the middle of the range. Assume that the DM finds this alternative unsatisfactory thus answering ``no'' to the question. This means that $50 < b_1$. Therefore we keep the upper part of the value range and do another dichotomy. Assume that the answer to the second question ``Is $(75,150,20,10)$ satisfactory ?'' is ``yes''. We conclude that $50 < b_1 \leq 75$. The imprecision threshold of $25$ is reached hence the algorithm stops while assigning the value $75$ to $b_1$. In case the answer to the second question had been ``no'', then the value retained for $b_1$ would be $100$ instead of $75$.  If a better precision was desired, the algorithm could continue further. The limit profile value $b_2$ for the second criterion can be determined in a similar way, using the same coalition $12$. For $b_3$ and $b_4$, the coalition $134$ is to be used.

\ssk In practice, the imprecision about the limit profile value $b_i$ can be dealt with in different ways. This question is briefly addressed in Chapter~\ref{chap:conclusion}, Section~\ref{ssec:hesitation}.

\section{Chapter conclusion}\label{sec:concluChap3}

In this chapter we explored possible ways of eliciting the parameters of a NCS model with two categories and no veto. Several algorithms for direct elicitation of the NCS model parameters are presented. The parameters that have to be determined are the sufficient coalitions of criteria and the profiles. The main difficulty is the elicitation of the sufficient coalitions. 

\ssk We analyzed different approaches based on an exploration of the coalitions graph by means of depth-first search (DFS). Three questioning strategies using the DFS arborescence were studied. The most promising seems to be DFS3 which questions in the middle of the paths. Asking the questions in the middle of the graph's branches, gives us the possibility to avoid about half of the questions in the considered branch. In Chapter~\ref{chap:questioner}, we describe the implementation of DFS3 and assess its performance. 

\ssk In the present chapter, we also described a dichotomic procedure for eliciting the limit profile. We didn't investigate the case where there are more than two categories, hence more than one profile. In the same way, we didn't address the question of determining the veto profile(s). These questions will be briefly addressed in Chapter~\ref{chap:recul}.

\ssk In appendix to the present chapter, we describe another type of exploration algorithm that uses the breadth-first search approach. We didn't pursue investigations in that direction, because we followed the idea of dichotomy that is implemented in DFS3. Another appendix (Section~\ref{sec:Kuntzmann}) deals with the question of efficiently computing $\Gmax$ knowing $\Fmin$ and conversely, which is not so simple but has been solved in the context of Boolean functions.    

\clearpage

\section{Appendix : Breadth-First Search Algorithms}\label{sec:Breadth-First}
The Breadth-First Search (BFS) algorithm explores all elements of each level of the graph in width and passes to the elements of a lower level when all elements of the superior levels are explored. The elements of the graph are explored horizontally and step by step. The output of the algorithm are the insufficient coalitions $\mathcal{G}$ and $\mathcal{G}_{max}$ with the coalition graph as it is represented in Figure~\ref{fig:theGraph}. Were the graph constructed upside-down, the algorithm would output $\mathcal{F}$ and $\Fmin$. The pseudo-code of the BFS algorithm is given below.

\begin{algorithm}
    \caption{BFS: Breadth-First Search}\label{algo:BFS}
    \begin{algorithmic}[0]
    \For{each $u\in P(N)$}
        \State $color(u)\leftarrow$ $white$
        \State $\Pi(u)\leftarrow$ $Nil$
        \State $qual(u)\leftarrow$ $S$      \Comment{the assumption}
    \EndFor
    \State $\mathcal{G}_{max}\leftarrow\emptyset$
    \State $u \leftarrow N$
    \State $Q\leftarrow \{u\}$               \Comment{the queue has been constituted by $u$}
    \While{$Q\neq \varnothing$} 
        \State $u\leftarrow$ head($Q$)
        \State Ask: $alt(u)\in \calU ?$
        \If {answer=yes}
            \State $qual(u)\leftarrow I$ 
            \State $\mathcal{G}_{max} \leftarrow \mathcal{G}_{max}\cup\{u\}$
        \Else
            \For {each $v\in succ(u)$ } \Comment{explore edge$(u,v)$}
                \If {$color(v)=white$}
                    \State $color(v)\leftarrow gray$
                    \State $\Pi(v)\leftarrow$ $u$
                    \If {$qual(u)=I$}  
                        \State $qual(v)=I$
                    \Else 
                        \State {Ask: $alt(v)\in \calU ?$}
                        \If {answer=yes and $\left(v \nsubseteq w \text{ for all } w \in \Gmax\right),$}
                            \State $\mathcal{G}_{max} \leftarrow \mathcal{G}_{max}\cup\{v\}$
                        \EndIf
                    \EndIf
                    \State $Enqueue(Q,v)$
                \EndIf
            \EndFor
        \EndIf
        \State $Dequeue(Q)$
        \State $color(u)\leftarrow black$ \Comment{blacken $u$ it is finished.}
    \EndWhile
\end{algorithmic}
\end{algorithm}

\paragraph{The best of the cases } The best case occurs when all elements of the graph are
$I$. Only 1 question is asked at the first level of the graph. It is the same situation with the graph of DFS2. This algorithm should be preferred when we have the intuition or information that even the big coalitions are insufficient and the elements situated in the higher levels of the
graph are labeled as ``$I$''.

\begin{center}
\begin{figure}[!h]
\centering
\includegraphics[width=80mm]{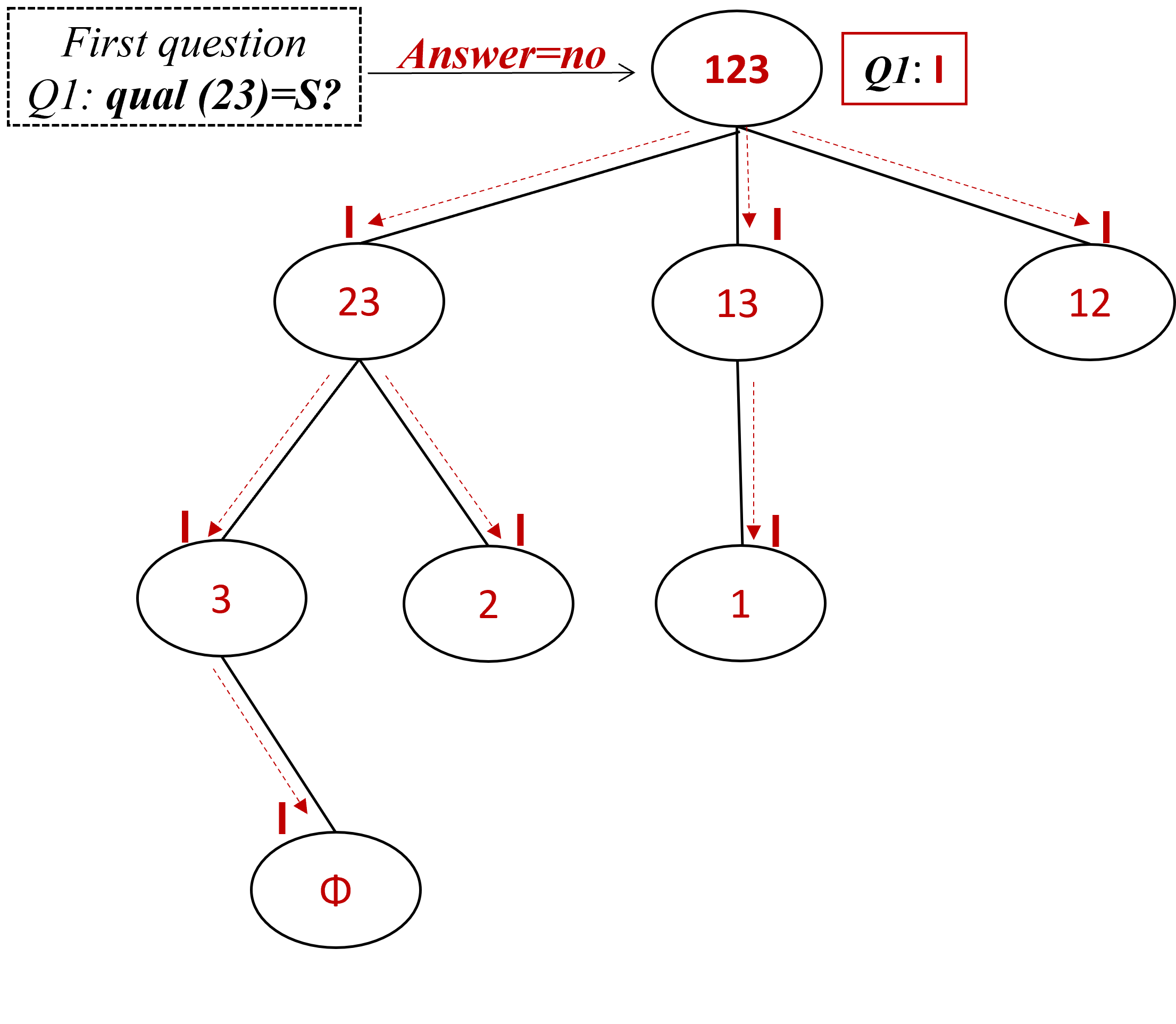}\caption{The best case of the BFS algorithm }
\label{fig:BFSbest}
\end{figure}
\end{center}

\clearpage 

\begin{center}
\begin{figure}[h!]
\centering
\includegraphics[width=140mm]{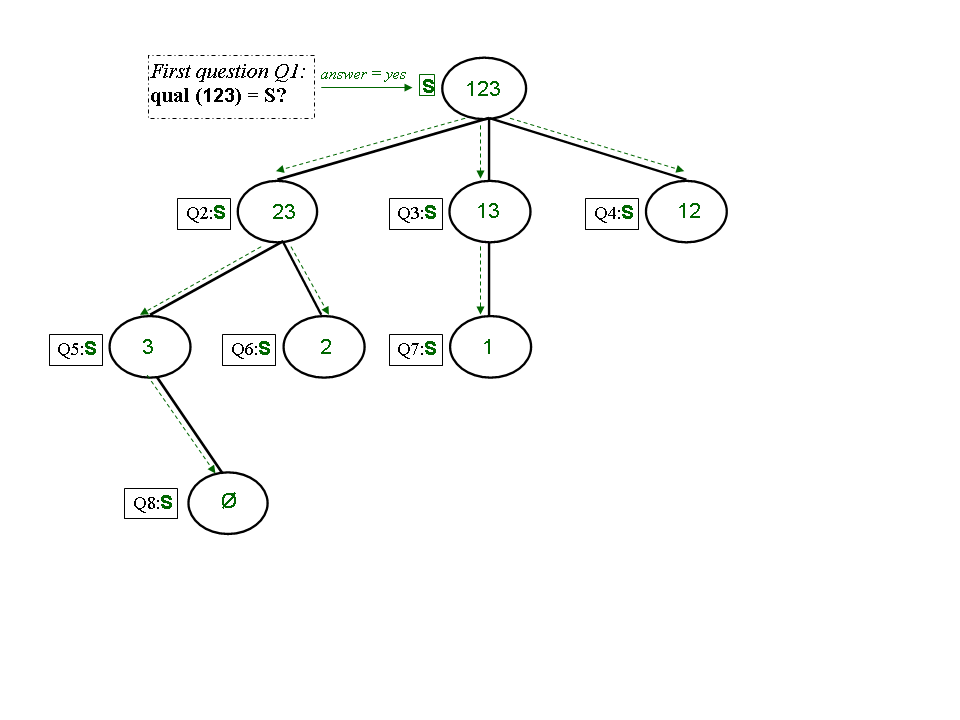}\vspace{-2cm}\caption{The worst case of the BFS Algorithm}
\label{fig:BFSworst}
\end{figure}
\end{center}

\paragraph{The worst of the cases } An illustration of the BFS algorithm is Figure~\ref{fig:BFSworst}. The worst case occurs when all elements of the graph
are $S$. It needs to question all of the elements of the graph, which means $2^{N}$ questions to ask.\\

\begin{center}
\begin{figure}[!h]
\centering
\includegraphics[width=140mm]{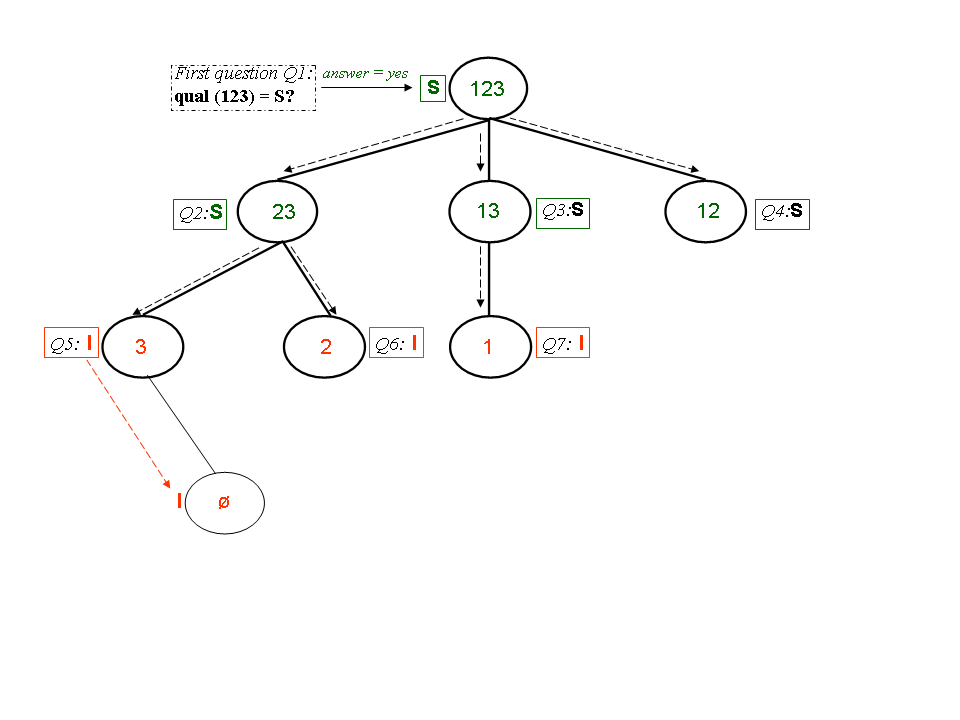}\vspace{-2cm}\caption{The BFS algorithm in the case of the majority of the criteria}
\label{fig:BFSmaj}
\end{figure}
\end{center}

\clearpage
\paragraph{The case of the majority of the criteria } The
breadth-first search algorithm can be preferred for such a
situation. The elements of the same level have a unique answer and
they are quickly discovered in width. It is even more effective for
an elevated number of criteria. An illustration of the BFS algorithm In the case of the majority of the criteria is shown in Figure~\ref{fig:BFSmaj}.

\clearpage

\section{Appendix: DFS2 avoids at least one question}\label{sec:resultDFS2}
This section is devoted to the proof of Proposition~\ref{prop:DFS2Avoids1Question}. Actually we establish a formula for the lower bound on the number of avoided questions as a function of the first time an answer ``no'' is obtained when asking questions on the first (and longest) branch of the DFS arborescence. We  prove that this bound is strictly positive for $n\geq 3$. 

\begin{lemma}\label{le:DFS2AvoidedQuest} Assume that $N$ is $S$ and $\varnothing$ is $I$. Let $f(r)$ be the function counting the avoided questions coming from a first answer $I$ at the $r$th question ($2\leq r\leq n+1$) in the first branch of the DFS arborescence (thus $S$ is the answer at the $(r-1)$th question) while using the DFS2 algorithm. We have : 
\begin{equation}\label{eq:f(r)DFS2}
f(r)=(2^{n+1-r}-1)+(2^{r-2}-(r-1)). 
\end{equation} 
This number is positive as soon as $n \geq 3$. If $n=2$, $f(r) =0$ if and only if $r=3$. 
\end{lemma}

\begin{center}
\begin{figure}[h!!!]
\centering
\includegraphics[width=120mm]{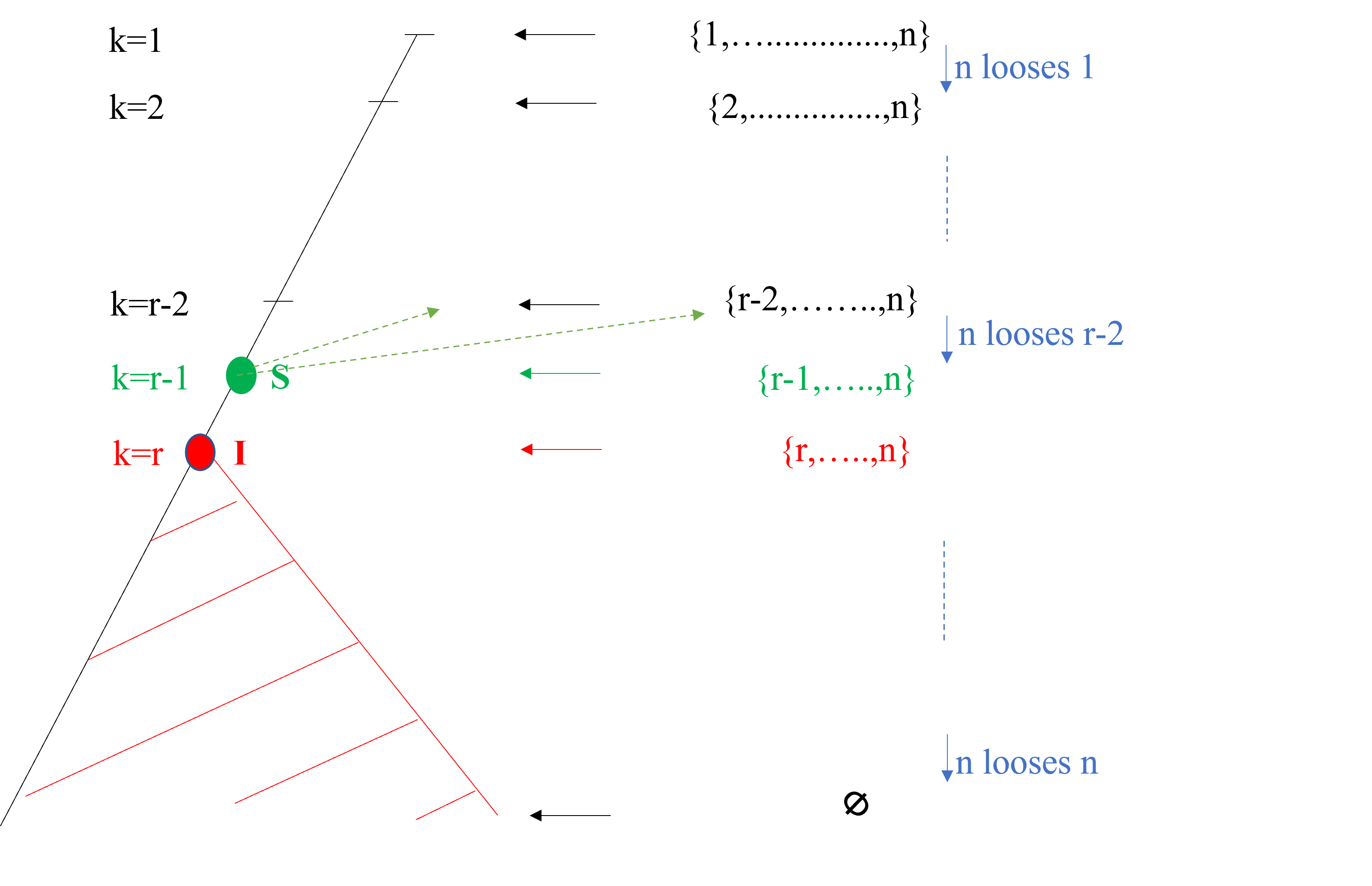}\caption{The process of applying DFS2 \label{fig:DFS2proof}}
\end{figure}
\end{center}

\bpr 
The process of applying DFS2 is represented in Figure~\ref{fig:DFS2proof}. 

With DFS2, questions are asked on each node of the first branch of the DFS arborescence until the first $I$ node is met. We may assume without loss of generality that the first $r-1$ answers are ``yes'', corresponding to the $r-1$ initial nodes in the first branch being sufficient coalitions; the $r$th node is the first $I$ node (with $2 \leq r \leq n+1$, since $N$ is $S$ and $\varnothing$ is $I$).  In this situation, the function $f(r)$ counts the avoided questions coming from the propagation of $S$ and $I$. 

\ssk The coalition in which the $r$th question is asked contains $n-r+1$ criteria. Since this coalition is insufficient, we know that all its subsets are insufficient without asking further questions. Therefore, the number of avoided questions due to the propagation of this $I$ label is $2^{n-r+1}-1$. 

\ssk The $S$ coalition obtained at the $(r-1)$th question involves $n-r+2$ criteria. The number of coalitions that contain this specific coalition $S$ is $2^{r-2}$. We should subtract $r-1$ which is the number of questions already asked and yielding the answer ``yes''. There are thus $2^{r-2}-(r-1)$ questions which are avoided due to the propagation of $S$. 

\ssk Formula~\eqref{eq:f(r)DFS2} is thus established. It remains to prove that $f(r)$ is strictly positive whenever $n\geq 3$. The first term in \eqref{eq:f(r)DFS2} represents the number of avoided questions due to the propagation of insufficiency downwards the graph. It is strictly positive as soon as $r < n+1$. If $r=n+1$, this term is null (because $I$ is obtained when asking whether $\varnothing$ is sufficient or not). The second term counts the avoided questions due to the propagation of sufficiency upwards the graph. If $r=2$, this term is null (because there is no propagation of sufficiency). As soon as $r \geq 3$, there is some propagation of sufficiency except for $n=2$ hence the number is strictly positive unless $n=2$. Consequently, $f(r) > 0$ unless $n=2$. We already know that the number of avoided questions can be zero in case $n=2$ and $r=3$. If $n=2=r$, $f(2)=1$.        
\epr  

\vspace{-.5cm}
\paragraph{Remark } It is possible to extend the expression of the number of avoided questions to the case $N$ is $I$ and to the case $\varnothing$ is $S$.     If coalition $N$ is insufficient then $r=1$ and all coalitions are insufficient. In this case,  the number of avoided questions is $2^n -1$, since all coalitions are insufficient.  If coalition $\varnothing$ is sufficient, then $r-1=n+1$, \ie $r=n+2$. All coalitions are sufficient and the number of avoided questions is $2^n-n-1$. We may thus extend the domain of $f(r)$ as follows:
\begin{equation*}
f(r)=
\left\{
\begin{array}{rl}
   2^n -1 & \textrm{for } r=1 \\
  2^{n+1-r}-1+2^{r-2}-(r-1)   &  \textrm{for } 2\leq r \leq n+1\\
    2^n - n - 1  & \textrm{for } r=n+2.
\end{array}    
\right.
\end{equation*}


\section{Appendix: Maximal insufficient vs minimal sufficient coalitions}\label{sec:Kuntzmann}
\sectionmark{Appendix: Max insufficient vs min sufficient}

How can we efficiently compute the set of maximal insufficient coalitions \Gmax{} when we know the set of sufficient coalitions $\mathcal{F}$ or the set of minimal sufficient coalitions \Fmin{} ? We initially found how to do that in an old book by \citet{Kuntzmann1965}. More recently we discovered a clearer and henceforth classical presentation in \citet{CramaHammer2011}'s book. This presentation leans on the theory of Boolean functions (which shall also be used in Chapter~\ref{chapter-Hansel}). 

\ssk The set of sufficient coalitions $\mathcal{F}$ in $2^n$ can be described as the inverse image of 1 by a monotone (or positive) Boolean function in $n$ Boolean variables $x_1, \ldots, x_n$. Such a function is monotone (positive) if $f(y_1, \ldots, y_n) \geq f(x_1, \ldots, x_n)$ whenever $y_i \geq x_i$ for all $i=1, \ldots, n$ and $x_i, y_i \in \{0,1\}$. Let $f: 2^n \rightarrow \{0,1\}$ be defined by 
\begin{equation}\label{eq:booleanF}
    f(x_1, \ldots, x_n) = 
    \left\{
    \begin{array}{ll}
        1 & \mathrm{if }\  (x_1, \ldots, x_n) \in \mathcal{F} \\
        0 & \mathrm{otherwise,\ \ie \ if }\ (x_1, \ldots, x_n) \in \mathcal{G} .
    \end{array}
    \right.
\end{equation}
This Boolean function is monotone (or positive) since $f(x_1, \ldots, x_n) = 1$ iff $(x_1, \ldots, x_n) \in \mathcal{F}$ and $(y_1, \ldots, y_n) \geq (x_1, \ldots, x_n) \Rightarrow (y_1, \ldots, y_n) \in \mathcal{F}$ iff $f(y_1, \ldots, y_n) =1$.

\ssk Any monotone Boolean function can be represented by a Boolean formula in disjunctive normal form (DNF) in which no variable negation appears \citep[][Theorem~1.21]{CramaHammer2011}.

\ssk Consider the following example for $n=3$. Let $\mathcal{F} = \{1, 23, 12, 13, 123\}$ with $\Fmin{} = \{1, 23\}$. Thus, $\mathcal{G} = \{2, 3, \varnothing\}$ with $\Gmax = \{2,3\}$. In Boolean form, we have $\mathcal{F} = \{100, 011, 110, 101, 111\}$ and $\Fmin{} = \{100, 011\}$.  Also, $\mathcal{G} = \{010, 001, 000\}$ with $\Gmax = \{010,001\}$. The Boolean function corresponding to $\mathcal{F}$ can be described by the disjunction $x_1 \vee x_2x_3$, where $x_2x_3$ is shorthand for $x_2 \wedge x_3$. Indeed, all Boolean vectors in $\mathcal{F}$ make the logical formula $x_1 \vee x_2x_3$ true, while all Boolean vectors in $\mathcal{G}$ make the formula false. Note that the DNF for $\mathcal{F}$ exactly corresponds to \Fmin{}. The clauses $x_1$ and $x_2x_3$ form a minimal description of $f$. 

\ssk We now define the \emph{dual} of $f$, which is related to $\mathcal{G}$ and \Gmax{}. The dual $f^d$ of $f$ is defined as
\begin{equation}
    f^d(x_1, \ldots, x_n)= f'(x'_1, \ldots, x'_n), 
\end{equation}
where $x'_i$ is the negation of the Boolean variable $x_i$ (\ie $x'_i = 1$ iff $x_i =0$) and $f' = 1 $ iff $f = 0$. 

\ssk Consider for instance the dual of the function $f$ associated with $\mathcal{F}$. 
We have:
\begin{equation*}
    f^d(x_1, x_2, x_3) = f'(x'_1, x'_2, x'_3) = 
    \left\{
    \begin{array}{ll}
        1 & \mathrm{if }\ f(x'_1, x'_2, x'_3) = 0  \\
        0 & \mathrm{otherwise}.
    \end{array}
    \right.
\end{equation*}
Hence $f^d(x_1, x_2, x_3) =1$ whenever $(x'_1, x'_2, x'_3) \in \mathcal{G}$. This happens when  $(x_1, x_2, x_3) \in \mathcal{G}^c$, the complement of $\mathcal{G}$, \ie $\mathcal{G}^c = \{101, 110, 111\} = \{13, 12, 123\}$.

\ssk The Boolean function $f^d$ is monotone (\ie positive). It describes the set $\calG^c$, an upset of $2^n$. It thus has a description in DNF in which the clauses correspond to the elements of $\Gmax^c$, the complements of the elements of \Gmax{}. 

\ssk The standard algorithm for computing the DNF of $f^d$ from that of $f$ works by sequentially applying the distributivity property of the $\vee$ and $\wedge$ operators. It is therefore known as the \emph{SD-Dualization algorithm}. It has been discovered and rediscovered by many authors, including Kuntzmann \citep[see][p~187, for historical details]{CramaHammer2011}. 

\ssk To compute $f^d$, we apply complementation to the DNF of $f$. Let us do that using the example $f(x_1, x_2, x_3) = x_1 \vee x_2x_3$. 
We have $f^d(x_1,x_2,x_3) = f'(x'_1, x'_2, x'_3) = (x'_1 \vee x'_2x'_3)' = x_1 \wedge (x_2 \vee x_3)$. To obtain the DNF of $f^d$, we apply distributivity (and reduction):
$$
f^d(x_1,x_2,x_3) = x_1x_2 \vee x_1x_3.
$$
The DNF is immediately obtained (without requiring reduction) and yields $\Gmax^c$. The set $\Gmax^c = \{110, 101\}$, which is the set of minimal solutions of $f^d$. By complementation, we get $\Gmax = \{001, 010\} = \{3, 2\}$. 

\ssk In order to illustrate \emph{sequential} distributivity and \emph{reduction}, we need a more complex example, with at least 3 elements in \Fmin{}. Let $n=3$ and $\Fmin = \{12, 13, 23\}$. The corresponding Boolean function is \\ $f(x_1,x_2,x_3) = x_1x_2 \vee x_1x_3 \vee x_2x_3$. The dual $f^d(x_1,x_2,x_3) = (x'_1x'_2 \vee x'_1x'_3 \vee x'_2x'_3)' = (x_1 \vee x_2) \wedge (x_1 \vee x_3) \wedge (x_2 \vee x_3)$. We start by applying distributivity to the first two terms:
$$
(x_1 \vee x_2) \wedge (x_1 \vee x_3) = x_1x_1 \vee x_1x_3 \vee x_1x_2 \vee x_2x_3.  
$$
Reduction here consists in omitting the terms $x_1x_2$ and $x_1x_3$ which imply $x_1$  (if $x_1x_2$ is satisfied then $x_1$ is satisfied; alternatively, the coalitions 12 and 13 contain the coalition 1). We then proceed with applying  distributivity involving the third term:
\begin{align*}
& (x_1 \vee x_2x_3) \wedge (x_2\vee  x_3)  \\
= & x_1x_2 \vee x_1x_3 \vee x_2x_3 \vee x_2x_3 \\
= & x_1x_2 \vee x_1x_3 \vee x_2x_3.
\end{align*}
This expression is reduced. The minimal solutions of $f^d = 1$ are $\{110, 101,\\ 
011\}$. Thus $\Gmax^c = \{12, 13, 23\}$ and $\Gmax = \{3, 2, 1\}$.

\chapter{A program for eliciting the sufficient coalitions of criteria}
\chaptermark{The questioner program}
\label{chap:questioner}

In this chapter, we describe the implementation of the DFS3 algorithm introduced in the previous chapter. \textit{The main program}, based on the DFS3 algorithm principles, explores the set of coalitions following the depth-first search algorithm. In order to determine which coalitions are sufficient or insufficient, the questioning procedure consists to ask questions ``in the middle'' of the branches containing unlabeled elements of the depth-first search arborescence. It is also possible to include in the main program, some additional informations coming from the DM about the relative importance of criteria, if needed. First, the DM has to enter the number of criteria of the decision problem. Then, the main program asks to the DM if ``Are there any known importance?'' which means is there is a preference/importance order between criteria. If no, the user enters ``n''. Otherwise, the user enters ``y'' and the program asks to enter the available importance information between criteria. For instance, the user enters ``1>2>3>4'' in the case of four criteria. If no further information  exists, the user enters ``0'' to finish adding information and the program continues with the next step. The next step consists in exploring the coalition graph in depth-first search and questioning the DM in the middle of the concerned branch. This can be seen as a dichotomy of non-explored branches of the coalition graph. 

\ssk The program can be used in two manners:
\begin{itemize}
    \item as an interactive questioning tool referred to as \emph{the main program}. According to the DFS3 algorithm, the main program uses a direct questioning procedure and asks to the DM interactively, if the current coalition of criteria is sufficient or not;
    \item as a simulation tool referred to as \textit{questioner}. With the set of all possible minimal sufficient coalitions generated previously and for every \Fmin{}, the simulation module \textit{questioner} runs. The main program generates the sequence of questions and answers are given according to the concerned \Fmin{}. This constitutes the elicitation process in case a DM had such a set of sufficient coalitions in mind. Finally, the simulation module \textit{questioner} returns the number of questions needed for each \Fmin and computes the average number of questions needed for a specified number of criteria.  
\end{itemize}

\section{Notation and Definitions}\label{sec:notationChap4}
For a given number of criteria $n$, the set of criteria is $N=\{1,2,..n\}$ and the set of the coalitions of criteria $\Pn$ is the set of subsets of $N$. As $\Pn$ is partially ordered by $\subseteq$, the inclusion relation on subsets, it is a poset (partially ordered set), we can define its chains and antichains as:

\ssk A \textbf{\emph{chain}} of a poset $(\Pn,\subseteq)$, is a set $\{B_{1},...,B_{k}\}$ of subsets such that $\forall i,j \in \{1,..,k\}$, $B_{i}\subseteq B_{j}$ or $B_{j}\subseteq B_{i}$.

\ssk An \textbf{\emph{antichain}} of a poset $(\Pn,\subseteq)$, is a set $\{B_{1},...,B_{k}\}$ of subsets such that $\forall i\neq j; i,j\in \{1,..,k\}$, $B_{i}\nsubseteq B_{j}$ and $B_{j}\nsubseteq B_{i}$.

\ssk We note that, \Fmin, the set of minimally sufficient coalitions of criteria, is an antichain of the set of subsets of the set of criteria, partially ordered by inclusion. 

\ssk The possible $\mathcal{F}_{min}$ are the antichains of ($\Pn,\subseteq$). The number of antichains in ($\Pn,\subseteq$) forms the integer sequence of \emph{Dedekind numbers} $D_n$ \citep[][Sequence A003182]{OEIS}, which grows very fast. 

\ssk For using the simulation tool \textit{questioner}, we shall need to list all possible sets of minimal coalitions \Fmin{} of $n$ criteria, \ie all antichains of ($\Pn,\subseteq$). This will be possible for values of $n$ up to $n=6$.  

\section{The Main Program}\label{sec:mainProgram}

The main program is based on a questioning procedure which combines depth-first search algorithm with dichotomy described in Section~\ref{ssec:DFS3}. It is an implementation of an interactive questioning process. First, the number of criteria $n$ has to be entered, then all subsets of $n$ are stored in lists as parent or children ordered by the inclusion relation. If they exist, all importance relations between coalitions of criteria are entered as input and the program starts. For a given $n$, the coalition graph is constructed and then the exploration step starts. Questions are asked in the middle of the explored but not labeled branches. According to the answer given by the DM, all possible propagations have to be done, which results in  elements of the graph (coalitions of criteria) being labeled as sufficient or insufficient. The program continues to explore the graph by the rest of the non-labeled coalitions. The main program stops when all elements of the graph are explored and labeled.

\ssk This program which is based on an interactive questioning algorithm has been implemented in Java. The main program generates a sequence of questions that have to be answered by the user. The output of the main program is the list of all questions and the set of \Fmin{} extracted from the given answers.

\subsection{An example for \texorpdfstring{$n=2$}{n=2}}\label{ssec:examplen2} 
\begin{figure}[h!!!]
\centering
\includegraphics[width=70mm] {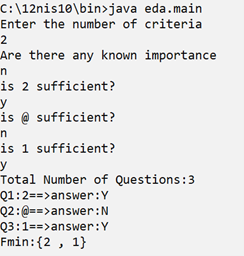}\caption{An output of the main algorithm for $n=2$ and $\Fmin{} = \{1,2\}$ \label{fig:main for n=2}}
\end{figure}

A screenshot of the main program's output for a simple example of $n=2$ without any additional information in terms of an importance order between criteria, is illustrated in Figure~\ref{fig:main for n=2} in case $\Fmin{} =\{1, 2\}$, \ie the case in which each single criterion is a sufficient coalition. The program starts with the number of criteria given by the DM. Then the graph of $\Pn, \subseteq$ for $n=2$ is constructed.  According to the depth-first search algorithm, the first question is asked in the middle of the first discovered branch. The first branch is $12-2-\varnothing$ and $2$ is the element situated in the middle of this branch. The answer to the question ``is 2 sufficient?'' is ``yes (y)'' and the program labels ``2'' and all coalitions containing ``2'', \ie $\{12\}$, with the label ``y''. The program continues to follow the depth-first search within the unexplored elements of the branch (here $\varnothing$). As the only remaining element of the branch is $\varnothing$, the second question is ``is $\varnothing$ sufficient ?''. The given answer is ``no (n)'' and $\varnothing$ is labeled ``n''. As there is no subset of $\varnothing$, we do not propagate. Then the next question is asked to the only unexplored element $1$. The given answer is ``yes'' and there is no other unexplored element of the graph. Thus, the program stops, lists all the questions and the given answers. Thus, according to the answers, the set $\Fmin{} = \{1,2\}$ is extracted and shown.

\clearpage

\subsection{Example for \texorpdfstring{$n=3$}{n=3}}\label{ssec:Examplen3}
\begin{figure}[h!!!]
\centering
\includegraphics[width=70mm]
{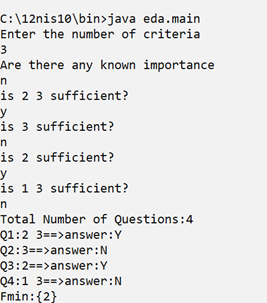}\caption{Output of the main algorithm for $n=3$ and $\Fmin{} = \{2\}$ \label{fig:main for n=3 Fmin=2}}
\end{figure}

We consider the case of $\Fmin{} = \{2\}$ as in Section~\ref{ssec:DFS3}, Figure~\ref{fig:DFS3}.
For $n=3$, the graph contains 8 coalitions of criteria which are: $123,23,13,\\ 12,3,2,1,\varnothing$. According to the DFS3 algorithm, the first discovered branch starting from the top to the end is $123-23-3-\varnothing$ and the first question is asked for $23$ which is situated in the middle of the concerned branch. In the case of $\Fmin{}=\{2\}$, without any importance order between criteria, the output of the main program is shown in Figure~\ref{fig:main for n=3 Fmin=2}. Since the answer given to ``is $23$ sufficient?'' is ``yes (y)'',  $123$ is also labeled as ``y'' and exluded from the list of unexplored elements of the graph. The unexplored part of the first branch is now $3,\varnothing$ and the question to come is for $3$. The answer is ``no (n)'' and as $\varnothing$ is the subset of $3$, both are labeled as ``n''. After finishing the exploration of the first branch in depth the program is going up to $23$ and down to the unexplored branch with the only unexplored element which is $2$. The answer to ``is $2$ sufficient ?'' is ``yes'', hence the program labels with ``y'' the element $2$ and all unexplored elements which contain $2$, more specifically $12$ for this example ($23$ and $123$ were already labeled). The next question is asked in the middle of the unexplored elements in the branch 13-1, \ie for $13$. The answer is ``n'' which thus labels $13$ and its unlabeled subset $1$. The element of the last branch $12$ was already labeled.  Therefore, the questioning process terminates.  All the four questions are listed with their answers. Finally, the set \Fmin{} is written as the output. The program output is shown in Figure~\ref{fig:main for n=3 Fmin=2}.

\begin{figure}[h!]
\centering
\includegraphics[width=130mm]{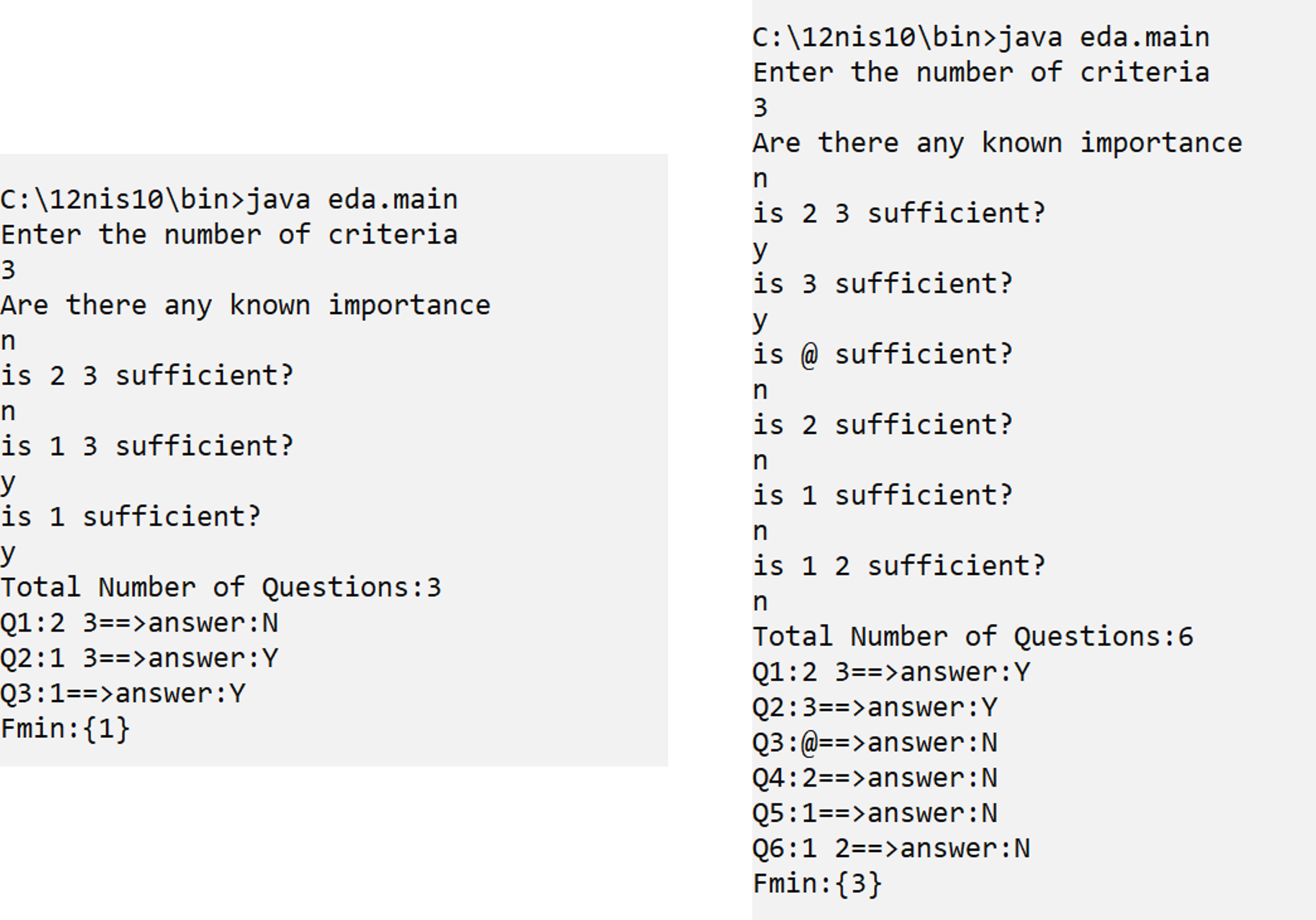}\caption{A view of the best (left) and worst (right) cases for $n=3$\label{fig:bestworst for n=3}}
\end{figure}

\ssk A view of the worst and best cases for $n=3$ is shown in Figure~\ref{fig:bestworst for n=3}. The best case is for $\Fmin = \{1\}$ and needs 3 questions to explore and label all coalitions of the graph. The worst case is for $\Fmin = \{3\}$ and needs $6$ questions. These results are obtained by running the simulation module \textit{questioner} on all instances of \Fmin{} for $n=3$ (see Section \ref{ssec:questioner}).

\section{The simulation module  \textit{questioner}}\label{ssec:questioner}
In order to compute the average number of questions for each $n$, a simulation module called \textit{questioner} is added to the main program. A list of the possible \Fmin for a given $n$ is taken from a code written in Python. The \textit{questioner} program runs this Python code, breaks up its output list, takes the characters and puts them into buffer. Each row is read by ``bufferreader'' and the coding of $\Fmin$ in the Python program is transformed into the type of coding that we use in the ``main program''. For instance, for $n=2$, an antichain listed as $11$ in Python, becomes $12$ in the main algorithm; an antichain listed as $10,01$ in Python, becomes $1,2$ in the main algorithm. For each possible \Fmin{} located in a row of the extracted list, \textit{questioner} begins the depth-first search algorithm with dichotomy (DFS3). The answers are all known since \Fmin{} is known and given by \textit{questioner} for each step of the algorithm functioning as a decision maker. The total number of questions is stored for each \Fmin and their arithmetic average is displayed at the end. 

\begin{figure}[h!]
\centering
\includegraphics[width=130mm]{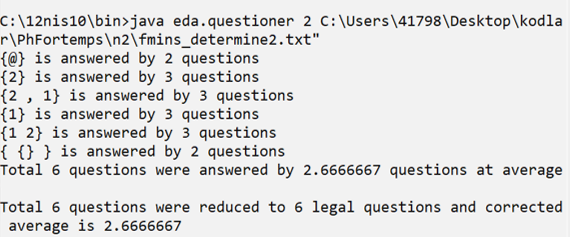}\caption{Output of questioner for n=2 \label{fig:questioner for n=2}}
\end{figure}

\ssk Figure~\ref{fig:questioner for n=2} shows the output of the simulation module \textit{questioner} for $n=2$. 
The set of all antichains of $n=2$ are computed in Python and transferred to a .txt document. The simulation module \textit{questioner} runs the main program, counts the number of questions for each possible $\Fmin$ and gives $2.6666667$ as the average number of questions. 

\ssk Under the hypothesis that all antichains are equally likely,  a lower bound on the expected number of questions is $\log_{2}\mathcal{D}_{n}$, which corresponds to $2.5850$ for $n=2$. This bound is the \emph{informational bound}, \ie the minimal number of questions to find an object in a set of objects by using dichotomy. For more on this bound, see Chapter~\ref{chap-average}, Section~\ref{sec:4crit}.

\section{Simulation results for \texorpdfstring{$n=2$}{n=2} to \texorpdfstring{$n=6$}{n=6}}\label{sec:simulationN2toN6}
The simulation module \textit{questioner} runs the main program for each possible \Fmin and gives us the average number of questions for the whole set of the possible \Fmin for $n$ the number of criteria entered by the user. 

\ssk The complete output of \textit{questioner} for $n=3$ and $n=4$ is listed in Section~\ref{sec:appendix:resultsQuestioner_n3_4}, as an appendix. Depending on the size of the antichain the running time for each is less than half a second in most of the cases. 

\ssk For $n=3$, the \Fmin that require the least number of questions (3 questions) are $\varnothing$, $1$ and $\{\}$. The maximal number of questions is 6 for $\Fmin = \{3\}$ or  $\Fmin = \{3, 12\}$ or $\Fmin = \{13, 23\}$ or $\Fmin = \{23, 13, 12\}$. Note that there are no obvious symmetries that could facilitate the execution of \textit{questioner}. For instance, the number of questions for \Fmin = $\{1\}$ is 3, while it is  6 for \Fmin = $\{3\}$. The number of questions depends on the labeling of the criteria.   

\ssk For $n=4$, the minimal number of questions is 3 for $\Fmin = \varnothing$ (and no other \Fmin). The maximal number of questions is 11 for  $\{34, 24, 14\}$ and for $\{34, 24, 14, 123\}$. Again, no obvious symmetry can be observed: for instance,  $\{14, 13, 12\}$  requires 8 questions.

\ssk The results for $n=5$ can be obtained in 2 hours approximatively. For $n=6$, a need for a server-based computer has appeared and the program ran during 24 days. Memory constraints do not allow to list here the number of questions required by the $7581$ possible $\Fmin$ for ${n=5}$ and $7828354$ for ${n=6}$. The computed average numbers of questions are ${15.294948}$ for ${n=5}$ and ${27.94281}$ for ${n=6}$. The number of possible $\Fmin$ can be reduced if some additional information can be taken from the decision-maker. With the available additional information, the average number of questions will be also decrease. 

The number of antichains $D_n$, \ie the number of $\Fmin{}$, grows very rapidly. The largest known value is $D_9$. Already for $n=7$, it is impossible to list all antichains since  $D_7 = 2414682040998 $.  That is why, while the main program works for any value of $n$, the simulation module \textit{questioner} is limited to $n=6$. 

\begin{figure}[h!]
\centering
\includegraphics[width=130mm]{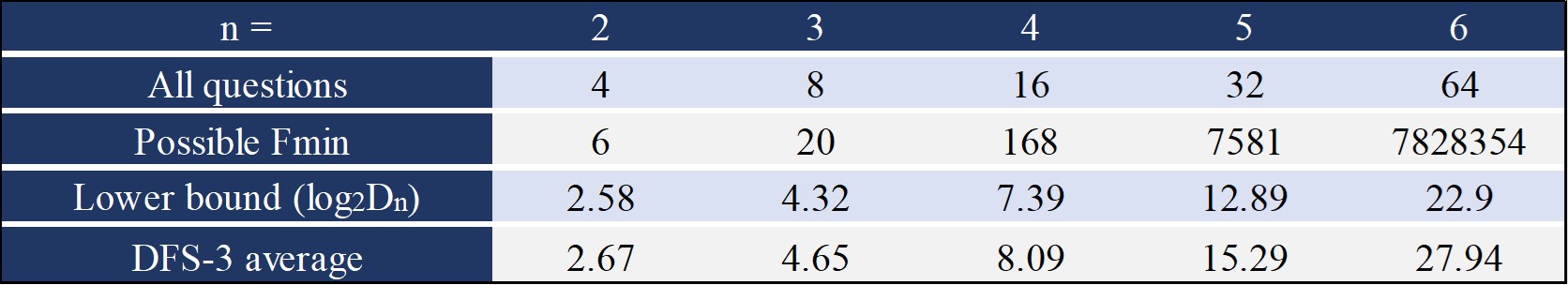}\caption{Comparison of DFS3 results}
\label{fig:ComparisonDFS3}
\end{figure}

\ssk Figure~\ref{fig:ComparisonDFS3}
shows a comparison of DFS3 results for $n=\{2,3,4,5,6\}$. We can easily observe that, the gap between the DFS3 results and the related lower bounds increases with the number criteria. Starting with this observation, a question that arises is ``Is it possible to do better on average than DFS3?''. 

Looking for effective answers to this question, we suggest different strategies and solutions in Chapter~\ref{chapter-Hansel} and Chapter~\ref{chap-average}. The implementation of an algorithm based on ``Hansel chains'' is suggested in Chapter~\ref{chapter-Hansel} and the results are compared to DFS3. In Chapter~\ref{chap-average}, the optimal expected values are computed manually and explained in detail for $n=2,3,4$ criteria.

\section{Main algorithm with additional information}\label{sec:additionalInfo}

In certain cases, additional information may be available and facilitate the elicitation of sufficient coalitions. 

\ssk Assume that an importance relation $\unrhd$ is known on the set of coalitions. We suppose that this relation, interpreted as ``at least as important than'', is reflexive and transitive. We also assume that it respects the inclusion relation on coalitions, \ie $A \supseteq B$ implies $A \unrhd B$. Finally, we suppose that this relation is monotone \wrt the categorization of coalitions as sufficient or insufficient, \ie 
\begin{equation}\label{eq:importanceRel}   
\begin{array}{lcl}
    A \unrhd B \textrm{ and } A \textrm{ is } I & \Rightarrow & B \textrm{ is } I,\\
   A \unrhd B \textrm{ and } B \textrm{ is } S & \Rightarrow & A \textrm{ is } S.
\end{array}
\end{equation}
Let us denote the symmetric part of $\unrhd$ as $\triangleq$. This is an equivalence relation. We assume that it is limited to identical pairs, \ie 
$$
\textrm{If } A \neq B \quad \Rightarrow \quad   Not(A \triangleq B).
$$

For example, consider the particular case in which the individual criteria can be ordered by importance and assume that the relation on coalitions of criteria respects this order. More precisely, if $i \in A$, $j \not\in A$ and $\{j\} \unrhd \{i\}$, then $A \setminus \{i\} \cup \{j\} \unrhd A$. Note this implies conversely the following: if $i \in A$, $j \not\in A$ and $\{i\} \unrhd \{j\}$, then $A \unrhd A \setminus \{i\} \cup \{j\}$.  

\begin{figure}[h!]
\centering
\includegraphics[width=135mm]{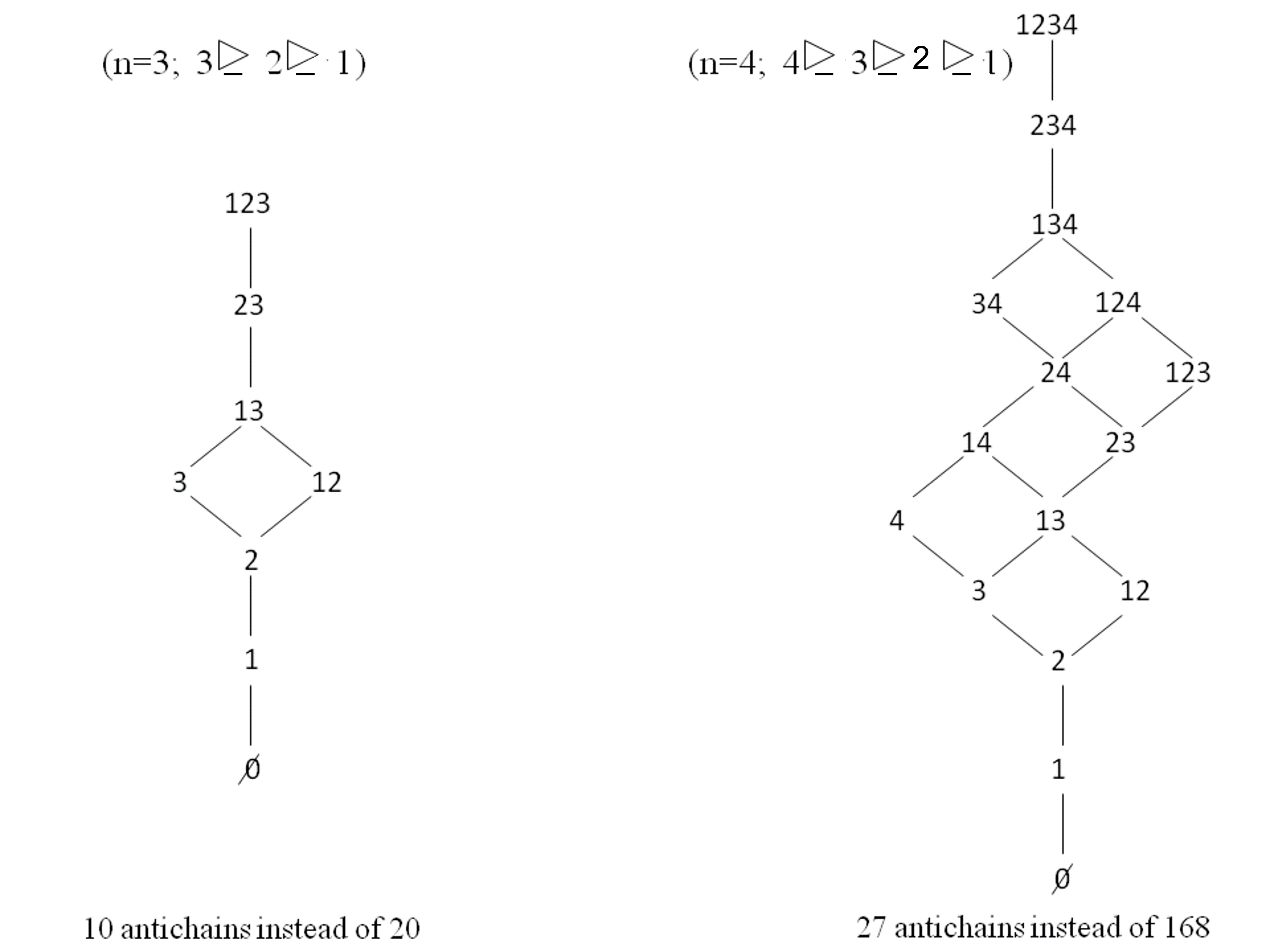}\caption{Transformation of the graph structure when an importance order on the criteria is known. On the left, $n=3$; on the right, $n=4$
\label{fig:additionalInfoGraph}}
\end{figure}

\ssk When an order between criteria is known, this information changes the structure of the graph which is constructed via lists of coalitions. According to the given information new parent-child relations are built. This new structure is shown by a Hasse diagram which is a simple picture of a finite partially ordered set, forming a drawing of the covering relation of the partial order. In other words, only relations between direct successors are represented. For instance, the right part of Figure~\ref{fig:additionalInfoGraph} shows the case of four criteria with $4\unrhd 3\unrhd 2\unrhd 1$,  an additional information given by the decision maker (for simplicity, we omit the brackets for coalitions involving a single criterion).

\ssk In this case, $4$ becomes the parent of $3$, $3$ the parent of $2$, $2$ the parent of $1$ and the graph of $n=4$ is transformed into the represented Hasse diagram. 
The set of possible \Fmin{} in this situation is reduced from 168 to 27. 

\subsection{The main algorithm with additional information}
We may apply the algorithm DFS3 in this case. The main algorithm uses the standard graph for the questioning procedure, but the Hasse diagram is used for the propagations according to the given answers. 

\ssk For example, as listed previously for $n=4$, $\Fmin=\{34, 24, 14\}$ requires 11 questions with the main algorithm when no additional information is available. Using the given information on the importance ordering of the criteria, $\Fmin$ is reduced to $\{14\}$ and requires 6 questions instead of 11. The functioning of the main algorithm for $n=4$, with the additional information $4\unrhd 3\unrhd 2\unrhd 1$ is displayed below and shows the question/answer process for the main algorithm in this case. 

\label{verb:dialogue}
\begin{verbatim}
C:\12nis10\bin>java eda.main
Enter the number of criteria
4
Are there any known importance
y
Enter Importance like: '1>2>3>4' or '2 3>1 3' or '0' to end
4>3>2>1
Enter Importance like: '1>2>3>4' or '2 3>1 3' or '0' to end
0
is 3 4 sufficient?
y
is 4 sufficient?
n
is 2 4 sufficient?
y
is 2 3 sufficient?
n
is 1 4 sufficient?
y
is 1 2 3 sufficient?
n
Total Number of Questions:6
Q1:3 4==>answer:Y
Q2:4==>answer:N
Q3:2 4==>answer:Y
Q4:2 3==>answer:N
Q5:1 4==>answer:Y
Q6:1 2 3==>answer:N
Fmin:{1 4}
\end{verbatim}

\begin{remark}\label{rem:DFS3withAdditInfo}
The DFS3 algorithm could be applied directly to the Hasse diagram when additional information is available. It is likely that this would reduce the number of questions since the algorithm would typically question in the middle of longer chains. We didn't implement such an algorithm. 
\end{remark}

\ssk The \textit{questioner} module simulates for a reduced set of antichains in the presence of additional information. A serious decrease in the number of antichains may be witnessed, when additional information exists. This is especially meaningful for high values of $n$. 

\begin{figure}[h!]
\centering
\includegraphics[width=133mm]{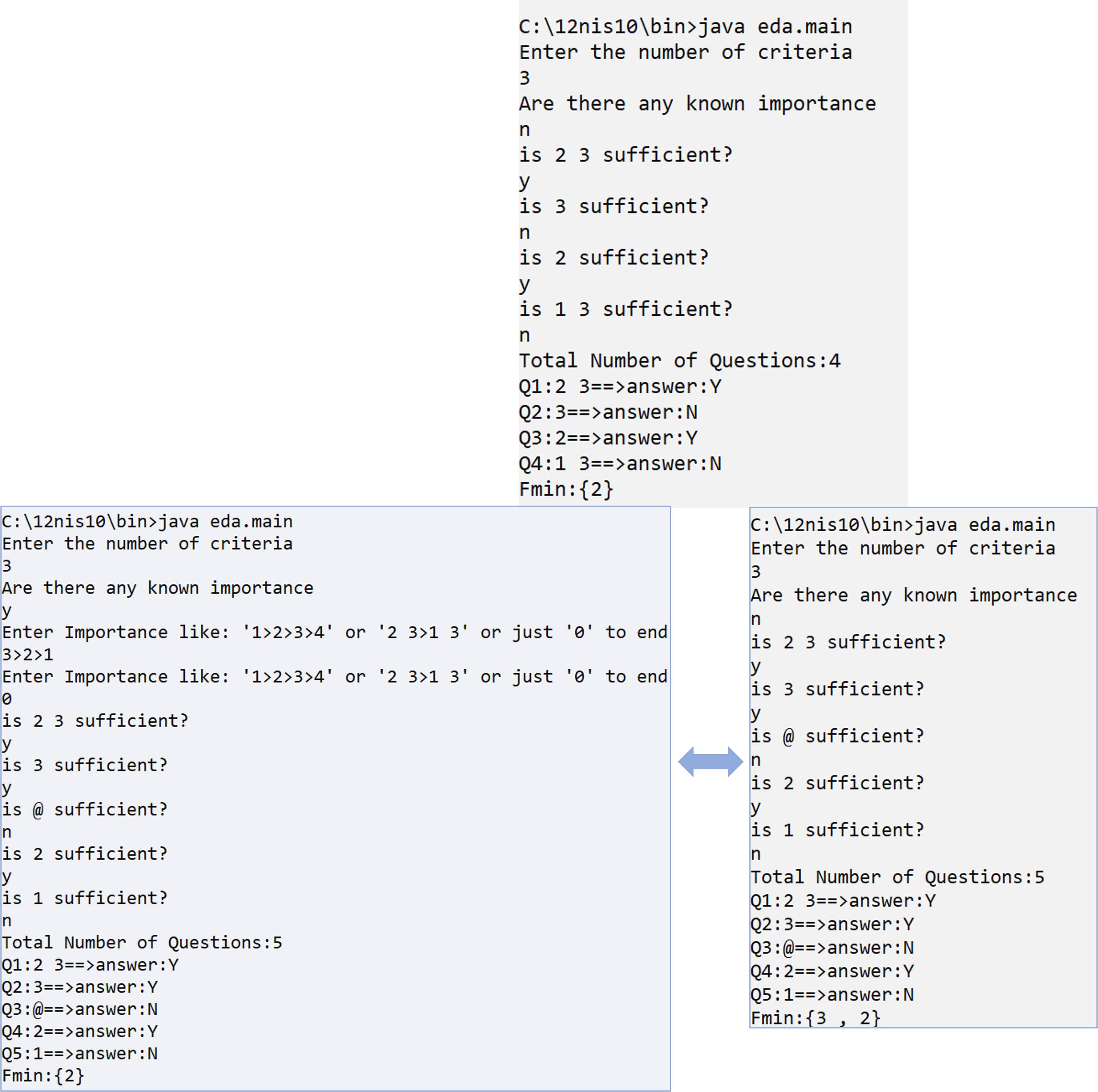}\caption{Conversion of \Fmin{} for a given information\label{fig:ConversionaddInfon=3}}
\end{figure}

\ssk Also, the number of questions for a given \Fmin is reduced or remains the same. Note however an apparently paradoxical effect. Figure~\ref{fig:ConversionaddInfon=3} gives a comparative example of the questioning procedure for $n=3$ and $\Fmin=\{2\}$. The first output is without any information and $\Fmin=\{2\}$ requires $4$ questions. When the additional information $3 \unrhd 2 \unrhd 1$ is given, the same example takes $5$ questions. Despite a given additional information, this increase in the number of questions may be misleading while making a comparison. Actually, $\Fmin=\{2\}$ with $3\unrhd 2\unrhd 1$ is equivalent to $\Fmin=\{2,3\}$ without information. The latter requires $5$ questions as shown in the lower right part of Figure~\ref{fig:ConversionaddInfon=3}. Thus, the right comparison should be done by taking into account the transformations of $\Fmin$ in the presence of additional information.

\paragraph{Type of additional information allowed for } The main program and ``questioner'' not only accept additional information in the guise of an order on individual criteria. They also accept information saying that a coalition is more important than another (see for instance the prompted examples in the dialogue on page \pageref{verb:dialogue}). 
The user may introduce as many additional information of this type as wanted. It should be noted that information such as ``$23\unrhd 13$'' is understood as implying for instance that  ``$234\unrhd 134$''.

\subsection{Using \textit{questioner} with additional information}
The simulation module \textit{questioner} can also integrate additional information given as an importance order between criteria or betwwen coalitions. Due to the exponential increase of the number of antichains forming the integer sequence of Dedekind numbers (A00372), the computations take long time for high values of $n$. As already mentioned, for $n=6$ simulations take 24 days. This time span can make the algorithm useless, in a decision making process. Here the integration of a given additional information to the simulation module becomes vital. In the case of $n=5$, \textit{questioner} runs for $7581$ antichains in two hours. When additional information like $5\unrhd 4\unrhd 3\unrhd 2\unrhd 1$ is given, the $7581$ antichains are reduced to $118$ and the simulation lasts for $30$ minutes. 

\paragraph{Results of \textit{questioner} with additional information }
It is also possible to enter additional information if it exists. The type of this additional information can be a preference order between criteria. This order can be given as between coalitions of criteria or between every single criterion. For instance if criterion 2 is more important than criterion 1, we can add this information to the simulation module as ``2 >1''. Likewise, if the coalition of criteria 12 is worse than the criterion 3, this information can be added as  ``3> 12''.

\ssk If we run \textit{questioner} for two criteria with an additional information such that ``2>1'', we obtain the following: 

\begin{verbatim}
C:\12nis10\bin>java eda.questioner 2 "C:\Users\41798\Desktop
\kodlar\PhFortemps\n2\fmins_determine2.txt" "2>1"
{@} is answered by 2 questions
{2} is answered by 3 questions
{2 , 1} is reduced to {1} and was answered by 3 questions
{1} is answered by 3 questions
{1 2} is answered by 2 questions
{ {} } is answered by 2 questions
Total 6 questions were answered by 2.5 questions at average
Total 6 questions were reduced to 5 legal questions and 
corrected average is 2.4
\end{verbatim}

We observe that with the additional information, the number of sets of minimal coalitions to be elicited decreases to $5$ from $6$. The average number of questions for all possible \Fmin becomes $2.4$ from $2.6666667$. The last sentence of the output which is ``Total 6 questions were reduced to 5 legal questions and corrected average is 2.4'' indicates the reduction of the number of antichains and as its result the decreased average number of questions. With the additional information, the set of possible \Fmin becomes smaller, thus, \textit{questioner} runs less times and the average number of questions is smaller.
 
\ssk When we run \textit{questioner} for three criteria without any additional information, we obtain the following: 

\begin{verbatim}
C:\12nis10\bin>java eda.questioner 3 "C:\Users\41798\Desktop
\kodlar\PhFortemps\n3\fmins_determine3.txt"
{@} is answered by 3 questions
{3} is answered by 6 questions
{3 , 2} is answered by 5 questions
{3 , 2 , 1} is answered by 5 questions
{3 , 1} is answered by 5 questions
{3 , 1 2} is answered by 6 questions
{2} is answered by 4 questions
{2 , 1} is answered by 5 questions
{2 , 1 3} is answered by 5 questions
{2 3} is answered by 5 questions
{2 3 , 1} is answered by 5 questions
{2 3 , 1 3} is answered by 6 questions
{2 3 , 1 3 , 1 2} is answered by 6 questions
{2 3 , 1 2} is answered by 5 questions
{1} is answered by 3 questions
{1 3} is answered by 4 questions
{1 3 , 1 2} is answered by 4 questions
{1 2} is answered by 4 questions
{1 2 3} is answered by 4 questions
{ {} } is answered by 3 questions
Total 20 questions were answered by 4.65 questions at average
Total 20 questions were reduced to 20 legal questions and 
corrected average is 4.65
\end{verbatim}
For a decision problem with three criteria, there are $20$ possible \Fmin, \ie antichains. The simulation module \textit{questioner}, runs the main program for every \Fmin among the $20$ possible ones. Without any additional information, \ie preference order between criteria, the average number of questions is $4.65$.

\ssk Running \textit{questioner} for three criteria with an additional information, \linebreak ``3>2>1'' gives the following as output: 
\begin{verbatim}
C:\12nis10\bin>java eda.questioner 3 "C:\Users\41798\Desktop\
kodlar\PhFortemps\n3\fmins_determine3.txt" "3>2>1"
{@} is answered by 3 questions
{3} is answered by 5 questions
{3 , 2} is reduced to {2} and was answered by 5 questions
{3 , 2 , 1} is reduced to {1} and was answered by 
5 questions
{3 , 1} is reduced to {1} and was answered by 5 questions
{3 , 1 2} is answered by 5 questions
{2} is answered by 5 questions
{2 , 1} is reduced to {1} and was answered by 5 questions
{2 , 1 3} is reduced to {2} and was answered by 5 questions
{2 3} is answered by 3 questions
{2 3 , 1} is reduced to {1} and was answered by 5 questions
{2 3 , 1 3} is reduced to {1 3} and was answered by 
4 questions
{2 3 , 1 3 , 1 2} is reduced to {1 2} and was answered by 
4 questions
{2 3 , 1 2} is reduced to {1 2} and was answered by 
4 questions
{1} is answered by 5 questions
{1 3} is answered by 4 questions
{1 3 , 1 2} is reduced to {1 2} and was answered by 
4 questions
{1 2} is answered by 4 questions
{1 2 3} is answered by 2 questions
{ {} } is answered by 2 questions
Total 20 questions were answered by 4.2 questions at average
Total 20 questions were reduced to 10 legal questions and 
corrected average is 3.8
\end{verbatim}

\ssk An additional information such that ``3>2>1'', transforms the set of possible \Fmin and the reduced set of possible \Fmin contains $10$ elements instead of $20$. Thus, the simulation module \textit{questioner} runs the questioning procedure $10$ times and the average number of questions becomes $3.8$ instead of $4.2$.

\ssk In addition to the first information of ``3>2>1'', if the decision-maker provides another additional information like ``3>12'', the following results are obtained: 

\begin{verbatim}
C:\12nis10\bin>java eda.questioner 3 "C:\Users\41798\Desktop
\kodlar\PhFortemps\n3\fmins_determine3.txt" "3>2>1" "3>1 2"
{@} is answered by 3 questions
{3} is answered by 5 questions
{3 , 2} is reduced to {2} and was answered by 5 questions
{3 , 2 , 1} is reduced to {1} and was answered by 5 questions
{3 , 1} is reduced to {1} and was answered by 5 questions
{3 , 1 2} is reduced to {1 2} and was answered by 5 questions
{2} is answered by 5 questions
{2 , 1} is reduced to {1} and was answered by 5 questions
{2 , 1 3} is reduced to {2} and was answered by 5 questions
{2 3} is answered by 3 questions
{2 3 , 1} is reduced to {1} and was answered by 5 questions
{2 3 , 1 3} is reduced to {1 3} and was answered by 
3 questions
{2 3 , 1 3 , 1 2} is reduced to {1 2} and was answered by 
5 questions
{2 3 , 1 2} is reduced to {1 2} and was answered by 
5 questions
{1} is answered by 5 questions
{1 3} is answered by 3 questions
{1 3 , 1 2} is reduced to {1 2} and was answered by 
5 questions
{1 2} is answered by 5 questions
{1 2 3} is answered by 2 questions
{ {} } is answered by 2 questions
Total 20 questions were answered by 4.3 questions at average
Total 20 questions were reduced to 9 legal questions and 
corrected average is 3.6666667
\end{verbatim}

The second additional information ``3>1 2'', means that criterion 3 is preferred to the coalition of criteria 12. Having this supplementary information, we observe that the set of \Fmin reduces to $9$ and the average number of questions becomes $3.6666667$ instead of $3.8$.

\section{Chapter conclusion}\label{sec:concluChap4}

In this chapter, we presented the implementation of the most promising questioning algorithm, the DFS3 and its simulation tool \textit{questioner}. It is possible to integrate additional information to the main program and its simulation module. The results with and withour additional information are compared and discussed. The additional information is an importance order between criteria. With this information given by the DM, the propagation and labelization processes increase and it is possible to obtain the results faster.

\ssk The evaluation of the effectiveness of such algorithms is still a matter that has to be addressed. One way would be to try to have an algorithm that asks as few questions as possible.  What is the minimal number of questions? If we knew which coalitions are sufficient and which are insufficient, \ie \calF{} and \calG{}, we need at least to ask $|\Fmin| + |\Gmax|$ questions to verify that \calF{} (\resp \calG{}) is the set of sufficient (\resp insufficient) coalitions. 

\ssk For instance, suppose that all coalitions of at least $k+1$ criteria are sufficient and those having less than $k+1$ criteria are insufficient. Then the minimal number of questions to ask is  $|\Fmin| + |\Gmax| = \binom{n}{k+1} + \binom{n}{k}$. 
So, one way of assessing algorithms is to count how many more questions are asked  than the minimal number of questions. This number depends on $\Fmin$ and $\Gmax$. In Chapter~\ref{chapter-Hansel}, we shall study an algorithm that is optimal in the worst case, \ie  minimizing the number of questions in the worst case.


\clearpage
\section{Appendix: Number of questions when \texorpdfstring{$n=3, 4$}{n=3 or 4}} \label{sec:appendix:resultsQuestioner_n3_4}

\subsection{The case \texorpdfstring{$n=3$}{n=3}} 
\emph{$\{\varnothing\}$ is answered by 3 questions\\
$\{3\}$ is answered by 6 questions\\
$\{3 , 2\}$ is answered by 5 questions\\
$\{3 , 2 , 1\}$ is answered by 5 questions\\
$\{3 , 1\}$ is answered by 5 questions\\
$\{3 , 1 2\}$ is answered by 6 questions\\
$\{2\}$ is answered by 4 questions\\
$\{2 , 1\}$ is answered by 5 questions\\
$\{2 , 1 3\}$ is answered by 5 questions\\
$\{2 3\}$ is answered by 5 questions\\
$\{2 3 , 1\}$ is answered by 5 questions\\
$\{2 3 , 1 3\}$ is answered by 6 questions\\
$\{2 3 , 1 3 , 1 2\}$ is answered by 6 questions\\
$\{2 3 , 1 2\}$ is answered by 5 questions\\
$\{1\}$ is answered by 3 questions\\
$\{1 3\}$ is answered by 4 questions\\
$\{1 3 , 1 2\}$ is answered by 4 questions\\
$\{1 2\}$ is answered by 4 questions\\
$\{1 2 3\}$ is answered by 4 questions\\
$\{ \}$ is answered by 3 questions\\
Total 20 possible \Fmin{} were answered by \textbf{4.65} questions on average.}

\subsection{The case \texorpdfstring{$n=4$}{n=4}}
\emph{$\{\varnothing\}$ is answered by 3 questions\\
$\{4\}$ is answered by 10 questions\\
$\{4 , 3\}$ is answered by 7 questions\\
$\{4 , 3 , 2\}$ is answered by 6 questions\\
$\{4 , 3 , 2 , 1\}$ is answered by 6 questions\\
$\{4 , 3 , 1\}$ is answered by 6 questions\\
$\{4 , 3 , 1 2\}$ is answered by 7 questions\\
$\{4 , 2\}$ is answered by 7 questions\\
$\{4 , 2 , 1\}$ is answered by 6 questions\\
$\{4 , 2 , 1 3\}$ is answered by 7 questions\\
$\{4 , 2 3\}$ is answered by 9 questions\\
$\{4 , 2 3 , 1\}$ is answered by 7 questions\\
$\{4 , 2 3 , 1 3\}$ is answered by 9 questions\\
$\{4 , 2 3 , 1 3 , 1 2\}$ is answered by 9 questions\\
$\{4 , 2 3 , 1 2\}$ is answered by 9 questions\\
$\{4 , 1\}$ is answered by 7 questions\\
$\{4 , 1 3\}$ is answered by 9 questions\\
$\{4 , 1 3 , 1 2\}$ is answered by 9 questions\\
$\{4 , 1 2\}$ is answered by 9 questions\\
$\{4 , 1 2 3\}$ is answered by 10 questions\\
$\{3\}$ is answered by 6 questions\\
$\{3 , 2\}$ is answered by 6 questions\\
$\{3 , 2 , 1\}$ is answered by 7 questions\\
$\{3 , 2 , 1 4\}$ is answered by 7 questions\\
$\{3 , 2 4\}$ is answered by 7 questions\\
$\{3 , 2 4 , 1\}$ is answered by 7 questions\\
$\{3 , 2 4 , 1 4\}$ is answered by 8 questions\\
$\{3 , 2 4 , 1 4 , 1 2\}$ is answered by 8 questions\\
$\{3 , 2 4 , 1 2\}$ is answered by 7 questions\\
$\{3 , 1\}$ is answered by 6 questions\\
$\{3 , 1 4\}$ is answered by 7 questions\\
$\{3 , 1 4 , 1 2\}$ is answered by 7 questions\\
$\{3 , 1 2\}$ is answered by 7 questions\\
$\{3 , 1 2 4\}$ is answered by 7 questions\\
$\{3 4\}$ is answered by 9 questions\\
$\{3 4 , 2\}$ is answered by 7 questions\\
$\{3 4 , 2 , 1\}$ is answered by 7 questions\\
$\{3 4 , 2 , 1 4\}$ is answered by 8 questions\\
$\{3 4 , 2 , 1 4 , 1 3\}$ is answered by 8 questions\\
$\{3 4 , 2 , 1 3\}$ is answered by 7 questions\\
$\{3 4 , 2 4\}$ is answered by 10 questions\\
$\{3 4 , 2 4 , 2 3\}$ is answered by 9 questions\\
$\{3 4 , 2 4 , 2 3 , 1\}$ is answered by 8 questions\\
$\{3 4 , 2 4 , 2 3 , 1 4\}$ is answered by 10 questions\\
$\{3 4 , 2 4 , 2 3 , 1 4 , 1 3\}$ is answered by 10 questions\\
$\{3 4 , 2 4 , 2 3 , 1 4 , 1 3 , 1 2\}$ is answered by 10 questions\\
$\{3 4 , 2 4 , 2 3 , 1 4 , 1 2\}$ is answered by 10 questions\\
$\{3 4 , 2 4 , 2 3 , 1 3\}$ is answered by 9 questions\\
$\{3 4 , 2 4 , 2 3 , 1 3 , 1 2\}$ is answered by 9 questions\\
$\{3 4 , 2 4 , 2 3 , 1 2\}$ is answered by 9 questions\\
$\{3 4 , 2 4 , 1\}$ is answered by 8 questions\\
$\{3 4 , 2 4 , 1 4\}$ is answered by 11 questions\\
$\{3 4 , 2 4 , 1 4 , 1 3\}$ is answered by 10 questions\\
$\{3 4 , 2 4 , 1 4 , 1 3 , 1 2\}$ is answered by 10 questions\\
$\{3 4 , 2 4 , 1 4 , 1 2\}$ is answered by 10 questions\\
$\{3 4 , 2 4 , 1 4 , 1 2 3\}$ is answered by 11 questions\\
$\{3 4 , 2 4 , 1 3\}$ is answered by 9 questions\\
$\{3 4 , 2 4 , 1 3 , 1 2\}$ is answered by 9 questions\\
$\{3 4 , 2 4 , 1 2\}$ is answered by 9 questions\\
$\{3 4 , 2 4 , 1 2 3\}$ is answered by 10 questions\\
$\{3 4 , 2 3\}$ is answered by 8 questions\\
$\{3 4 , 2 3 , 1\}$ is answered by 7 questions\\
$\{3 4 , 2 3 , 1 4\}$ is answered by 9 questions\\
$\{3 4 , 2 3 , 1 4 , 1 3\}$ is answered by 9 questions\\
$\{3 4 , 2 3 , 1 4 , 1 3 , 1 2\}$ is answered by 9 questions\\
$\{3 4 , 2 3 , 1 4 , 1 2\}$ is answered by 9 questions\\
$\{3 4 , 2 3 , 1 3\}$ is answered by 8 questions\\
$\{3 4 , 2 3 , 1 3 , 1 2\}$ is answered by 9 questions\\
$\{3 4 , 2 3 , 1 3 , 1 2 4\}$ is answered by 9 questions\\
$\{3 4 , 2 3 , 1 2\}$ is answered by 9 questions\\
$\{3 4 , 2 3 , 1 2 4\}$ is answered by 9 questions\\
$\{3 4 , 1\}$ is answered by 7 questions\\
$\{3 4 , 1 4\}$ is answered by 10 questions\\
$\{3 4 , 1 4 , 1 3\}$ is answered by 9 questions\\
$\{3 4 , 1 4 , 1 3 , 1 2\}$ is answered by 9 questions\\
$\{3 4 , 1 4 , 1 2\}$ is answered by 9 questions\\
$\{3 4 , 1 4 , 1 2 3\}$ is answered by 10 questions\\
$\{3 4 , 1 3\}$ is answered by 8 questions\\
$\{3 4 , 1 3 , 1 2\}$ is answered by 9 questions\\
$\{3 4 , 1 3 , 1 2 4\}$ is answered by 9 questions\\
$\{3 4 , 1 2\}$ is answered by 9 questions\\
$\{3 4 , 1 2 4\}$ is answered by 10 questions\\
$\{3 4 , 1 2 4 , 1 2 3\}$ is answered by 10 questions\\
$\{3 4 , 1 2 3\}$ is answered by 9 questions\\
$\{2\}$ is answered by 5 questions\\
$\{2 , 1\}$ is answered by 5 questions\\
$\{2 , 1 4\}$ is answered by 6 questions\\
$\{2 , 1 4 , 1 3\}$ is answered by 6 questions\\
$\{2 , 1 3\}$ is answered by 6 questions\\
$\{2 , 1 3 4\}$ is answered by 6 questions\\
$\{2 4\}$ is answered by 8 questions\\
$\{2 4 , 2 3\}$ is answered by 7 questions\\
$\{2 4 , 2 3 , 1\}$ is answered by 6 questions\\
$\{2 4 , 2 3 , 1 4\}$ is answered by 8 questions\\
$\{2 4 , 2 3 , 1 4 , 1 3\}$ is answered by 8 questions\\
$\{2 4 , 2 3 , 1 4 , 1 3 , 1 2\}$ is answered by 8 questions\\
$\{2 4 , 2 3 , 1 4 , 1 2\}$ is answered by 8 questions\\
$\{2 4 , 2 3 , 1 3\}$ is answered by 8 questions\\
$\{2 4 , 2 3 , 1 3 , 1 2\}$ is answered by 8 questions\\
$\{2 4 , 2 3 , 1 3 4\}$ is answered by 8 questions\\
$\{2 4 , 2 3 , 1 3 4 , 1 2\}$ is answered by 8 questions\\
$\{2 4 , 2 3 , 1 2\}$ is answered by 7 questions\\
$\{2 4 , 1\}$ is answered by 6 questions\\
$\{2 4 , 1 4\}$ is answered by 9 questions\\
$\{2 4 , 1 4 , 1 3\}$ is answered by 8 questions\\
$\{2 4 , 1 4 , 1 3 , 1 2\}$ is answered by 8 questions\\
$\{2 4 , 1 4 , 1 2\}$ is answered by 8 questions\\
$\{2 4 , 1 4 , 1 2 3\}$ is answered by 9 questions\\
$\{2 4 , 1 3\}$ is answered by 8 questions\\
$\{2 4 , 1 3 , 1 2\}$ is answered by 8 questions\\
$\{2 4 , 1 3 4\}$ is answered by 9 questions\\
$\{2 4 , 1 3 4 , 1 2\}$ is answered by 8 questions\\
$\{2 4 , 1 3 4 , 1 2 3\}$ is answered by 9 questions\\
$\{2 4 , 1 2\}$ is answered by 7 questions\\
$\{2 4 , 1 2 3\}$ is answered by 8 questions\\
$\{2 3\}$ is answered by 7 questions\\
$\{2 3 , 1\}$ is answered by 6 questions\\
$\{2 3 , 1 4\}$ is answered by 8 questions\\
$\{2 3 , 1 4 , 1 3\}$ is answered by 8 questions\\
$\{2 3 , 1 4 , 1 3 , 1 2\}$ is answered by 8 questions\\
$\{2 3 , 1 4 , 1 2\}$ is answered by 8 questions\\
$\{2 3 , 1 3\}$ is answered by 8 questions\\
$\{2 3 , 1 3 , 1 2\}$ is answered by 9 questions\\
$\{2 3 , 1 3 , 1 2 4\}$ is answered by 9 questions\\
$\{2 3 , 1 3 4\}$ is answered by 8 questions\\
$\{2 3 , 1 3 4 , 1 2\}$ is answered by 9 questions\\
$\{2 3 , 1 3 4 , 1 2 4\}$ is answered by 9 questions\\
$\{2 3 , 1 2\}$ is answered by 8 questions\\
$\{2 3 , 1 2 4\}$ is answered by 8 questions\\
$\{2 3 4\}$ is answered by 8 questions\\
$\{2 3 4 , 1\}$ is answered by 6 questions\\
$\{2 3 4 , 1 4\}$ is answered by 9 questions\\
$\{2 3 4 , 1 4 , 1 3\}$ is answered by 8 questions\\
$\{2 3 4 , 1 4 , 1 3 , 1 2\}$ is answered by 8 questions\\
$\{2 3 4 , 1 4 , 1 2\}$ is answered by 8 questions\\
$\{2 3 4 , 1 4 , 1 2 3\}$ is answered by 9 questions\\
$\{2 3 4 , 1 3\}$ is answered by 8 questions\\
$\{2 3 4 , 1 3 , 1 2\}$ is answered by 9 questions\\
$\{2 3 4 , 1 3 , 1 2 4\}$ is answered by 9 questions\\
$\{2 3 4 , 1 3 4\}$ is answered by 9 questions\\
$\{2 3 4 , 1 3 4 , 1 2\}$ is answered by 9 questions\\
$\{2 3 4 , 1 3 4 , 1 2 4\}$ is answered by 10 questions\\
$\{2 3 4 , 1 3 4 , 1 2 4 , 1 2 3\}$ is answered by 10 questions\\
$\{2 3 4 , 1 3 4 , 1 2 3\}$ is answered by 9 questions\\
$\{2 3 4 , 1 2\}$ is answered by 8 questions\\
$\{2 3 4 , 1 2 4\}$ is answered by 9 questions\\
$\{2 3 4 , 1 2 4 , 1 2 3\}$ is answered by 9 questions\\
$\{2 3 4 , 1 2 3\}$ is answered by 8 questions\\
$\{1\}$ is answered by 6 questions\\
$\{1 4\}$ is answered by 9 questions\\
$\{1 4 , 1 3\}$ is answered by 8 questions\\
$\{1 4 , 1 3 , 1 2\}$ is answered by 8 questions\\
$\{1 4 , 1 2\}$ is answered by 8 questions\\
$\{1 4 , 1 2 3\}$ is answered by 9 questions\\
$\{1 3\}$ is answered by 7 questions\\
$\{1 3 , 1 2\}$ is answered by 8 questions\\
$\{1 3 , 1 2 4\}$ is answered by 8 questions\\
$\{1 3 4\}$ is answered by 8 questions\\
$\{1 3 4 , 1 2\}$ is answered by 8 questions\\
$\{1 3 4 , 1 2 4\}$ is answered by 9 questions\\
$\{1 3 4 , 1 2 4 , 1 2 3\}$ is answered by 9 questions\\
$\{1 3 4 , 1 2 3\}$ is answered by 8 questions\\
$\{1 2\}$ is answered by 6 questions\\
$\{1 2 4\}$ is answered by 7 questions\\
$\{1 2 4 , 1 2 3\}$ is answered by 7 questions\\
$\{1 2 3\}$ is answered by 7 questions\\
$\{1 2 3 4\}$ is answered by 7 questions\\
$\{ {} \}$ is answered by 6 questions\\
Total 168 possible \Fmin{} were answered by \textbf{8.089286} questions on average.}\\\\

%

\chapter{Worst case optimal algorithm}
\chaptermark{Worst case optimal algorithm}
\label{chapter-Hansel}

\section{Introduction}\label{sec:intro}
As previously seen, the DFS3 questioning algorithm which starts questioning in the middle of each chain of the DFS arborescence is not proven optimal and it is not, as we shall see. There are at least two ways of being optimal: minimizing the number of questions in the worst case or on average. Optimal questioning procedures in the worst case have been investigated but in a different context, namely, the elicitation of monotone Boolean functions. 

\ssk A Boolean function is a function from $\{0,1\}^n$ into $\{0,1\}$ ($n$ a positive integer). Such a function is monotone if it respects the natural partial order $\geq$ on $\{0,1\}^n$ defined as follows: $x = (x_1, \ldots, x_n) \geq y = (y_1, \ldots, y_n)$, with $x_i, y_i \in \{0,1\}$, if $x_i \geq y_i$, for all $i =1, \ldots, n$. A monotone Boolean function $f$ satisfies $f(x) \geq f(y)$ whenever  $x\geq y$. The set $\{0,1\}^n$ is in bijection with the set $\Pn$ of subsets of $N= \{1, \ldots, n\}$. The natural order on $\{0,1\}^n$ corresponds to set inclusion $\supseteq$. A monotone Boolean function corresponds to a partition of the subsets of $N$ into sufficient and insufficient coalitions. The sufficient (\resp insufficient)  coalitions correspond to the Boolean vectors $x$ such that $f(x) =1$  (\resp $f(x) =0$). The monotonicity of the Boolean function corresponds to the stability by inclusion of the sufficient and insufficient coalitions. 
Therefore, identifying the set of sufficient coalitions in our problem is equivalent to eliciting a monotone Boolean function. 

\ssk 
\citet{Korobkov1965} proved that the number of questions of the best algorithm for eliciting a monotone Boolean function of $n$ variables, in the worst case, is comprised between two bounds, the lower bound being: 
\begin{equation}\label{eq:lowerBound}
\varphi(n) = \binom{n}{\lfloor n/2 \rfloor} + \binom{n}{\lfloor n/2 \rfloor +1}. 
\end{equation}

\begin{figure}[h!!!]\label{fig:Hansel}
    \centering
  \includegraphics [width=8cm]{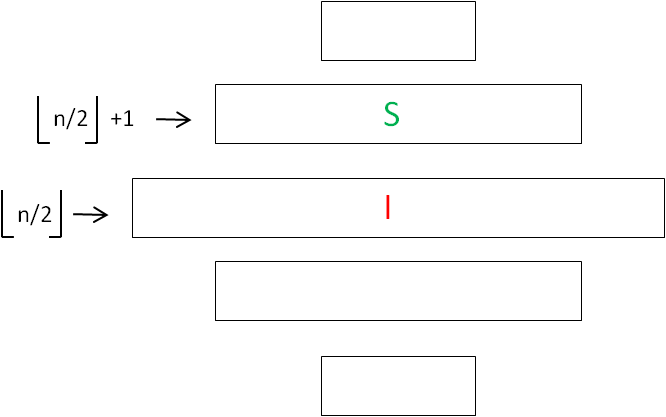}
  \caption{A situation in which the number of questions is at least\\ \texorpdfstring{$\varphi(n) = {n \choose {\lfloor n/2 \rfloor}} + {n \choose {{\lfloor n/2 \rfloor}} +1}$}{phi(n)=binom(n)(n/2)+binom(n)(n/2)+1} \ssk \label{fig:proofLowerBound}}
\end{figure}

\ssk It is easy to show that it is indeed the minimal number of questions in a special case. Determining the set of minimal coalitions always requires to identify all elements in \Fmin{} and all elements in \Gmax{} (the latter proving that the sufficient coalitions in \Fmin are indeed minimal). Hence the number of questions is at least $|\Fmin| + |\Gmax|$.  The largest antichain in $B_n$ has  $\binom{n}{\lfloor n/2 \rfloor}$ 
elements (Sperner's theorem, see, \eg \citet[][p.~119]{CasparLeclercMonjardetBook}). It is the number of subsets of $N$ of cardinal $\lfloor n/2 \rfloor$. If we take this set as \Gmax{} and the set of subsets of cardinal $\lfloor n/2 \rfloor +1$ as \Fmin{}, the result follows. See Figure~\ref{fig:proofLowerBound} for an illustration.

\ssk \citet{Hansel1966} gave a constructive proof that  this lower bound is exact, \ie one can manage to never use more than this number of questions and there is (at least) one case that requires this number of questions. \citeauthor{Hansel1966}'s proof leads to an algorithm that is thus optimal in the worst case. Other authors \citep{kovalerchuketal1996,torvikEtAl2002,torvikEtAl2003} have worked on this topic in the framework of monotone Boolean functions.  \citet{kovalerchuketal1996} implement Hansel's algorithm and test it on some examples. In view of an application to breast cancer detection, the authors structure the 11 criteria in a two-level hierarchy. This allows to aggregate subsets of criteria at the lower level, using Boolean functions, and then aggregate these Boolean functions at the main level. This reduces the number of criteria involved in the Boolean functions. \citet{torvikEtAl2002} investigate the problem of minimizing the \emph{average} number of questions needed for eliciting a monotone Boolean function. They compare several algorithms \wrt this objective. \citet{torvikEtAl2003} extend their investigation of good algorithms on average to the case of \emph{nested} monotone Boolean functions, which corresponds to sorting into more than 2 categories. We shall come back to the work of \citet{torvikEtAl2002,torvikEtAl2003} in Chapter~\ref{chap-average}. 

\ssk  Based on the principle that the Hansel chains solve the problem optimally in the worst case, we had the idea to modify the depth-first search algorithm to obtain the Hansel chains. This is our main original contribution in this chapter. In addition, we implemented Hansel's algorithm (DFSHansel) as a variant of depth-first search and we tested it exhaustively for up to 6 criteria. We compare the average number of questions used in DFS3 and DFSHansel.

\section{Hansel's result}\label{sec:HanselResult}
We recall that a chain in a partially ordered set (such as $B_n= \{0,1\}^n$) is a totally ordered subset of the set.
Let us define the \emph{length} of a chain in $B_n$ as the number of elements of $B_n$ in the chain.  

\ssk The Lemma in \citep{Hansel1966} states the following. The $n$-dimensional unit cube $B_n = \{0,1\}^n$ provided with its usual order structure, can be covered by a set of disjoint chains which enjoy the following properties: 
\begin{itemize}
 \item The number of the chains of length $n-2p+1$ is $\binom{n}{p}-\binom{n}{p-1}$ with $\lfloor n/2 \rfloor$$\geq{p}\geq{0}$;
  \item Given three elements $a_1 < a_2 < a_3$, forming a connected chain (\ie two consecutive elements differ on exactly one dimension) belonging to the same chain of length $n-2p+1$, the relative complement of $a_2$ for the interval $\lfloor a_1,a_3\rfloor$ belongs to a chain of length $n-2p-1$.
\end{itemize}
We start with illustrating the first property. We shall come back to the second and explain the expression ``relative complement'' later. 

\subsection{The construction of Hansel chains}\label{ssec:ConstructionHansel}

To fix the ideas, the Hansel chains for $n=1, 2, 3$ are as follows:
\begin{itemize}
    \item For $n=1$, we have only one chain in $B_1$: $\{1,0\}$;
    \item For $n=2$, $p$ takes the values $0,1$. The chain for $p=0$ has $n-2p+1= 3$ elements~: $\{(1,1), (0,1), (0,0)\}$. The chain for $p=1$ has one element: $\{(1,0)\}$.
\end{itemize} 
\ssk \noindent In the sequel, we shall simply write $x_1x_2 \ldots x_n$ for designating the vector $(x_1, x_2, \ldots, x_n) \in B_{n}$. So, for instance, we write $\{11, 01, 00\}$ instead of \\ 
$\{(1,1), (0,1), (0,0)\}$. 

\begin{itemize}
    \item For $n=3$, $p$ takes the values 0,1. The chain for $p=0$ has $n-2p+1= 4$ elements~: $\{111, 011, 001, 000\}$. The two chains for $p=1$ have 2 elements : $\{110, 010\}$ and $\{101,100\}$.
\end{itemize}

\citeauthor{Hansel1966} builds up the chains recursively, using the chains in $B_{n-1}$ to construct those in $B_n$. 
For constructing the Hansel chains of $B_n$, we start from those of $B_{n-1}$. We create two copies of $B_{n-1}$, denoted $B^0_{n-1}$ (\resp $B^1_{n-1}$) by adding a dimension: we add 0 (\resp 1) as initial coordinate of the vectors in $B^0_{n-1}$ (\resp $B^1_{n-1}$). The Hansel chains of    $B^0_{n-1}$ are completed by the copy in $B^1_{n-1}$ of their greatest element. The greatest element of each Hansel chain in $B^1_{n-1}$ is pruned. This process yields the Hansel chains of $B_n$. 

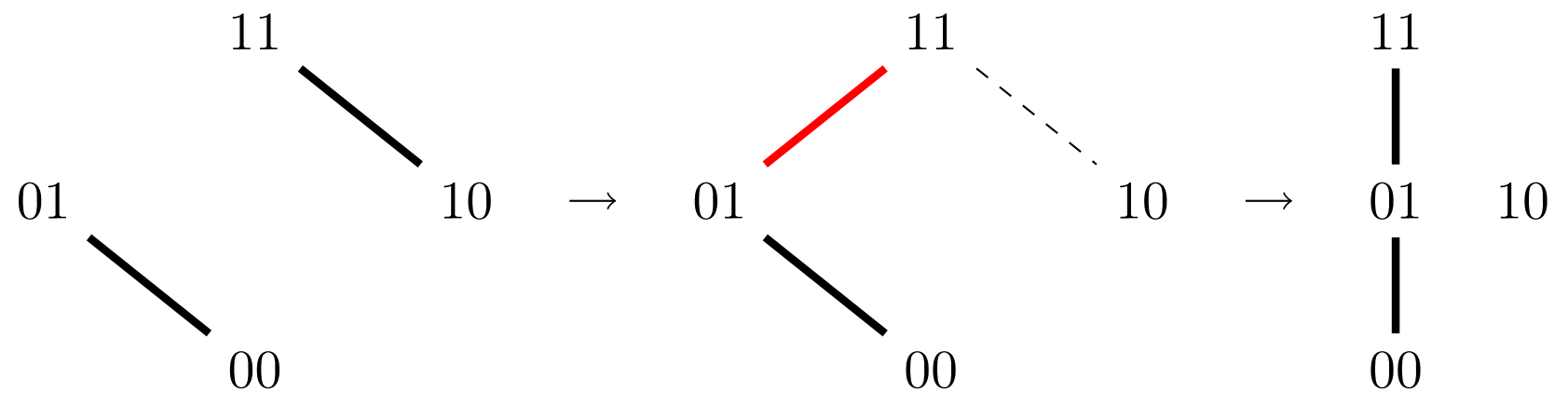
\begin{figure}[h!!!]
\begin{center}
\begin{tikzpicture}[scale=.6]
\tikzstyle{mynodestyle} = [ font={\huge\bfseries}, shape=circle, minimum size=0.5cm, circular drop shadow, text=black, very thick, draw=black!55, top color=white,bottom color=green!80, text width=0.5cm, align=center]
\tikzstyle{myedgestyle} = [ultra thick]
\node (v23) at (-8,2.5) {11};
\node (v25) at (-10.5,0.5) {01};
\node (v27) at (-5.5,0.5) {10};
\node (v28) at (-8,-1.5) {00};
\draw [myedgestyle] (v23) edge (v27);
\draw [myedgestyle] (v25) edge (v28);
\node (v99) at (-4.5,.5) {};
\node (v910) at (-3.5,.5) {};
\draw [->] (v99) edge (v910);
\node (v3) at (0,2.5) {11};
\node (v5) at (-2.5,0.5) {01};
\node (v7) at (2.5,0.5) {10};
\node (v8) at (0,-1.5) {00};
\draw [dashed] (v3) edge (v7);
\draw [myedgestyle] (v8) edge (v5);
\draw [ultra thick,red] (v3) edge (v5);
\node (v9) at (3.5,.5) {};
\node (v10) at (4.5,.5) {};
\draw [->] (v9) edge (v10);
\node (v13) at (5.5,0.5) {01};
\node (v12) at (5.5,2.5) {11};
\node (v14) at (5.5,-1.5) {00};
\node (v16) at (7,0.5) {10};
\draw [ultra thick] (v12) edge (v13);
\draw [ultra thick] (v13) edge (v14);
\end{tikzpicture}
\caption{Hansel's construction for $n=2$ starting from $n=1$ \label{fig:twohansel}}
\end{center}
\end{figure}

\begin{figure}[h!!]
\begin{center}
\begin{tikzpicture}[scale=.6]
\tikzstyle{mynodestyle} = [ font={\huge\bfseries}, shape=circle, minimum size=0.5cm, circular drop shadow, text=black, very thick, draw=black!55, top color=white,bottom color=green!80, text width=0.5cm, align=center]
\tikzstyle{myedgestyle} = [ultra thick]
\node (v21) at (-8,4.5) {111};
\node (v22) at (-10.5,2.5) {011};
\node (v23) at (-8,2.5) {101};
\node (v24) at (-5.5,2.5) {110};
\node (v25) at (-10.5,0.5) {001};
\node (v26) at (-8,0.5) {010};
\node (v27) at (-5.5,0.5) {100};
\node (v28) at (-8,-1.5) {000};
\draw [myedgestyle] (v22) edge (v25);
\draw [myedgestyle] (v23) edge (v27);
\draw [myedgestyle] (v25) edge (v28);
\draw [myedgestyle] (v23) edge (v21);
\node (v99) at (-4.5,1.5) {};
\node (v910) at (-3.5,1.5) {};
\draw [->] (v99) edge (v910);
\node (v1) at (0,4.5) {111};
\node (v2) at (-2.5,2.5) {011};
\node (v3) at (0,2.5) {101};
\node (v4) at (2.5,2.5) {110};
\node (v5) at (-2.5,0.5) {001};
\node (v6) at (0,0.5) {010};
\node (v7) at (2.5,0.5) {100};
\node (v8) at (0,-1.5) {000};
\draw [dashed] (v1) edge (v3);
\draw [myedgestyle] (v2) edge (v5);
\draw [myedgestyle] (v3) edge (v7);
\draw [myedgestyle] (v5) edge (v8);
\draw [ultra thick,red] (v2) edge (v1);
\draw [ultra thick, red] (v6) edge (v4);
\node (v9) at (3.5,1.5) {};
\node (v10) at (4.5,1.5) {};
\draw [->] (v9) edge (v10);
\node (v13) at (5.5,0.5) {001};
\node (v12) at (5.5,2.5) {011};
\node (v11) at (5.5,4.5) {111};
\node (v14) at (5.5,-1.5) {000};
\node (v15) at (7,2.5) {110};
\node (v16) at (7,0.5) {010};
\node (v17) at (8.5,2.5) {101};
\node (v18) at (8.5,0.5) {100};
\draw [ultra thick] (v11) edge (v12);
\draw [ultra thick] (v12) edge (v13);
\draw [ultra thick] (v13) edge (v14);
\draw [ultra thick] (v15) edge (v16);
\draw [ultra thick] (v17) edge (v18);
\end{tikzpicture}
\caption{Hansel's construction for $n=3$ starting from $n=2$ \label{fig:threehansel}}
\end{center}
\end{figure}
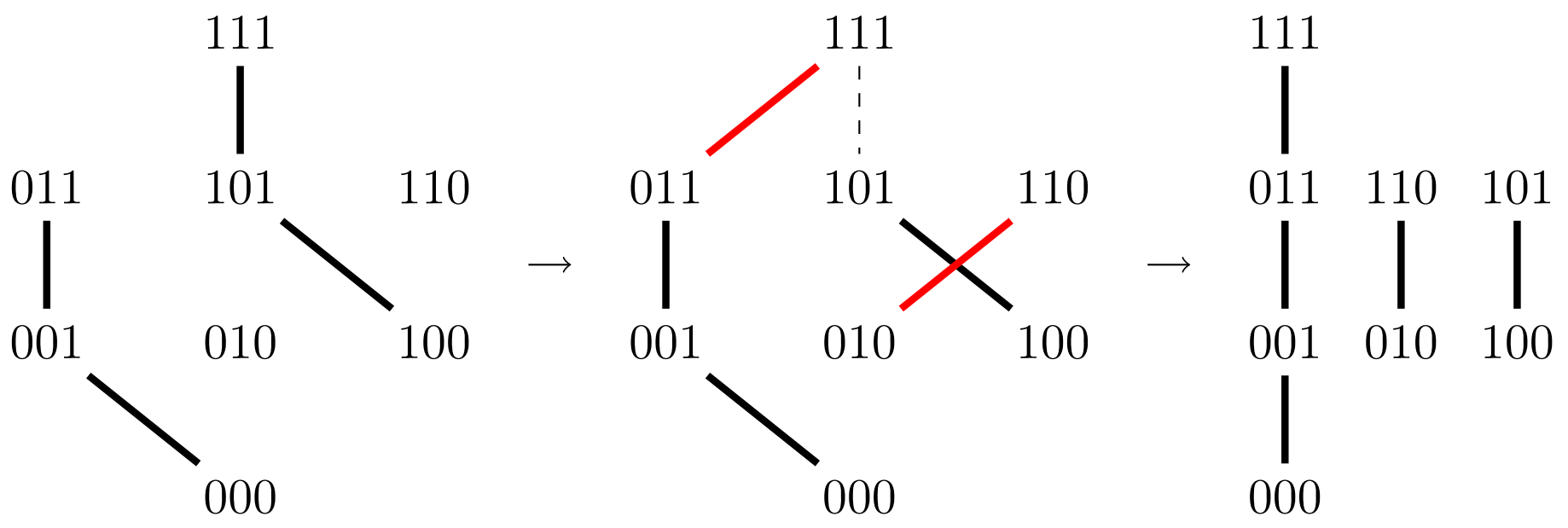

\ssk This process is illustrated in Figures~\ref{fig:twohansel} and~\ref{fig:threehansel}. On the left part of Figure~\ref{fig:twohansel}, we see the two copies $B_1^0$ and $B^1_{1}$. In the middle of this figure,  11 is added to the chain $\{01, 00\}$ in order to create the new Hansel chain $\{11, 01, 00\}$. The element 11 is removed from the chain $\{11, 10\}$ of $B_1^1$ yielding the chain $\{10\}$.

\ssk Similarly, in Figure~\ref{fig:threehansel},  we use the two copies $B_2^0$ and $B_2^{1}$. After that, the element $111$ is added to the chain $\{011, 001, 000\}$ in order to create the new Hansel chain $\{111, 011, 001, 000\}$. The element 111 is removed from the chain $\{111, 101, 100\}$ of $B_2^1$ yielding the chain $\{101, 100\}$. In the same way, the chain $\{010\}$ of $B_2^0$ is completed by the element $\{110\}$ from $B_2^1$.

\begin{figure}[h!!!]
    \centering
    \includegraphics[height=7cm]{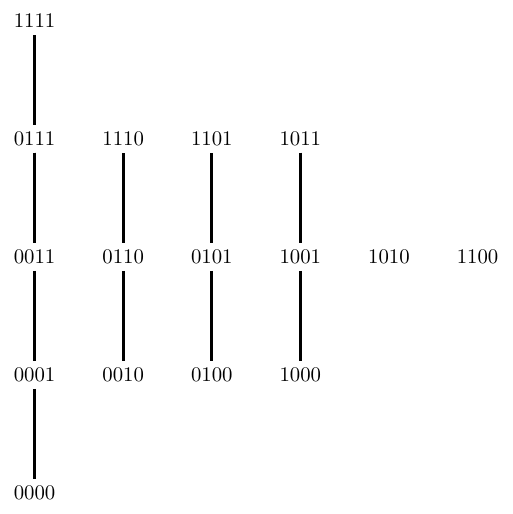}
    \caption{Hansel chains for $n=4$     \label{fig:fourHansel}}
\end{figure}

\ssk Figure \ref{fig:fourHansel} shows the Hansel chains for $n=4$. It is easy to see how these chains are obtained from two copies \Bnull[3] and \Bone[3] of the Hansel chains of $B_3$. As announced in the Lemma, there are 2 chains of length 1 ($p=2$), 3 chains of length 3 ($p=1$) and 1 chain of length 5 ($p=0$).  

\ssk In general, as stated in \citep{Hansel1966}'s Lemma, the Hansel chains provide a decomposition of the $n$-dimensional cube in disjoint chains :
\begin{itemize}
  \item 1 chain of length $n+1$;
  \item ${n \choose 1} = n$ chains of length $\geq n-1$;
  \item ${n \choose 2} = n (n-1)/2$ chains of length $\geq n-3$;
  \item $n \choose p$ chains of length $n-2p +1$, for $0 \leq p \leq \lfloor n/2 \rfloor$; 
  \item ${n \choose {\lfloor n/2 \rfloor}}$ chains of length $\geq n-2{\lfloor n/2 \rfloor}+1 = 1 $ or 2.
\end{itemize}




\subsection{Hansel's second property}\label{ssec:secondHansel}
The second property in Hansel's Lemma is crucial for the proof of Hansel's bound on the number of questions, because it implies that asking at most two questions in each chain will be enough.

\ssk Let's recall and explain this property:
\begin{quote}
    Given three elements $a_1 < a_2 < a_3$, forming a connected chain (\ie two consecutive elements differ on exactly one dimension) belonging to the same Hansel chain of length $n-2p+1$, the relative complement of $a_2$ for the interval $\lfloor a_1,a_3\rfloor$ belongs to a chain of length $n-2p-1$.
\end{quote}

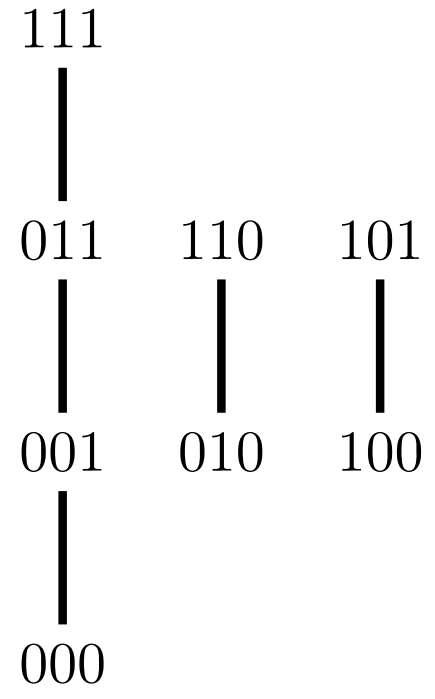
\begin{figure}[!h]
\begin{center}
\begin{tikzpicture}[scale=.7]
\tikzstyle{mynodestyle} = [ font={\huge\bfseries}, shape=circle, minimum size=0.5cm, circular drop shadow, text=black, very thick, draw=black!55, top color=white,bottom color=green!80, text width=0.5cm, align=center]
\tikzstyle{myedgestyle} = [ultra thick]
\node (v13) at (5.5,0.5) {001};
\node (v12) at (5.5,2.5) {011};
\node (v11) at (5.5,4.5) {111};
\node (v14) at (5.5,-1.5) {000};
\node (v15) at (7,2.5) {110};
\node (v16) at (7,0.5) {010};
\node (v17) at (8.5,2.5) {101};
\node (v18) at (8.5,0.5) {100};
\draw [ultra thick] (v11) edge (v12);
\draw [ultra thick] (v12) edge (v13);
\draw [ultra thick] (v13) edge (v14);
\draw [ultra thick] (v15) edge (v16);
\draw [ultra thick] (v17) edge (v18);
\end{tikzpicture}
\caption{Hansel's chains for $n=3$ \label{fig:Hansel3}}
\end{center}
\end{figure}

Take for instance the case $n=3$ (Figure~\ref{fig:Hansel3}). 
The property applies only here to the longest chain  ($p=0$) of length $n-2p+1= 4$: $\{111,011,001,000\}$. There are only two connected subchains $a_1 < a_2 < a_3$ in this chain, namely $\{011,00,000\}$ and  $\{111,011, 001\}$.  The relative complement of the middle element $a_2$ in $\{011,001,000\}$ is $010$. The two extreme elements $a_1, a_3$ differ in two coordinates; the middle element $a_2$ differs from the extremes on one coordinate; the relative complement of $a_2$ differs from the extremes on the other coordinate. With respect to the subchain $\{011,001,000\}$, the relative complement of the middle element $010$ belongs to the leftmost chain of two elements in the figure. Considering the other subchain $\{111,011,001\}$, the relative complement of the middle element is $101$; it  belongs to the rightmost chain of two elements in the figure. 

\begin{figure}[h!!!]\label{fig:exHansel}
\begin{center}
\includegraphics[width=4cm]{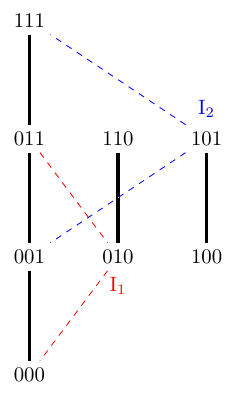}
\caption{Propagation in Hansel chains}
\label{fig:propagationHanseln=3}
\end{center}
\end{figure}

\ssk This has an important consequence. If we know the status $I$ or $S$ of the elements in the chain of length 2, we also know the status of one extreme element in all subchains of consecutive elements in the chain of length 4. For example, consider Figure~\ref{fig:propagationHanseln=3} .

\ssk If, for instance,  $101$ and $010$ are insufficient, we know that the smallest element $001$ in the subchain $\{111,011,001\}$ is insufficient as well as the smallest element $000$ in the subchain $\{011,001,000\}$. Therefore  we need only two questions to determine the status of $111$  and $011$ in the long chain. 

\ssk This reasoning is general. Whenever we know the status of the elements of the smallest chains we need at most two questions to determine the status of all elements in the longer chains. In the example with $n=3$, no more than 6 questions are needed, whatever \Fmin{}.  

\ssk In general, in chains of length one (if there are), one question is needed. In all longer chains at most two questions are needed. This simple observation allows to establish that the lower bound on the number of questions in the worst case is feasible.
The minimum number of questions $\varphi(n)$ of the best algorithm for determining an increasing arbitrary Boolean function or, equivalently, determining the set of minimal coalitions,  $f(x_1,...,x_n)$ is
$$
\varphi(n) = {n \choose {\lfloor n/2 \rfloor}} + {n \choose {\lfloor n/2 \rfloor +1}}
$$

For $n=3$, the number of question needed is ${3 \choose 1} + {3 \choose 2} = 2 {3 \choose 1} = 2 \times 3 = 6$.

\ssk For $n=4$, it is ${4 \choose 2} + {4 \choose 3} =  6 + 4 = 10$. 

\ssk For $n=5$, it is ${5 \choose 2} + {5 \choose 3} = 2 {5 \choose 2} = 2 \times 10 = 20$. 

\ssk For $n=6$, it becomes $35$.  

\ssk For $n=7$, the minimum number of question of the best algorithm is $70$ in the worst case.

\msk Hansel's constructive proof that the lower bound is feasible, suggests an optimal algorithm in the worst case.
After the obtention of the Hansel chains, 


\begin{itemize}
    \item Start questioning with the shorter chains;
    \item Propagate what is known for chains of length $n-2p-1$ to chains of length $n-2p+1$;
    \item Ask the remaining two questions (at most) for each chain of length $n-2p+1$.
\end{itemize}

\section{A new algorithm for generating Hansel chains}\label{sec:algorithmDFSHansel}
An optimal questioning strategy (in the worst case) relying on Hansel's proof, requires to build recursively the decomposition in Hansel chains beforehand. We want to avoid that preliminary step and generate the decomposition by means of a modified DFS. In this section, we show how to do this modification and then how to apply an optimal questioning strategy using the arborescence obtained in the modified DFS.

\ssk Let $B_n$ denote the set of all binary vectors of dimension $n$. 
For $x$ and $y$ in $B_n$, we say that $x \supseteq y$ if $x_i \geq y_i$, for all $i \in \ensn$. The set $(B_n$, endowed with relation $\supseteq)$, is a partially ordered set.  
A \textit{successor} of $x$ in that order is any vector $y$ which differs from $x$ on a single coordinate $i \in N =\ensn$; on this coordinate $x_i=1$ while $y_i =0$. The set of all successors of $x$ is denoted by $\vsucc$. For all $y$ in $\vsucc$, there is no $z$ in $B_n$ such that $x \supset z \supset y$. The set $\vsucc$ has as many elements as $x$ has components equal to 1. In all cases $|\vsucc| \leq n$.

\ssk A different order, a complete one, on $B_n$ is the lexicographic order $\succ_{lex}$. We have $x \suclex y$ if, for some $i \in N$, $x_i > y_i$ and $x_j = y_j$, for all $j<i$. 
We call \suclex-maximal (\resp \suclex-minimal) the maximal (\resp minimal element of a set \wrt the relation \suclex. Sometimes we will use ``largest'' (\resp ``smallest'') in the same sense.  

\ssk $B_n$ can be partitioned in $n+1$ classes, which we call \textit{levels}. Level $L(p)$ is the set of elements of $B_n$ which exactly have $p$ components equal to 0 and consequently $n-p$ components equal to 1. We have $|L(p)| = {n \choose p}$. All successors of an element of $L(p)$ are in $L(p+1)$.

\ssk In  Algorithm \ref{algo:hanselChains}, any element $x \in B_n$ receives the status \textit{visited} as soon as it has been added to a chain.

\ssk Variables used in the algorithm are:
\begin{itemize}
\item $n$ is the number of criteria given by user;
\item $numChains = {n \choose {\lfloor n/2 \rfloor}}$ is the number of elements in the largest level;
\item $numLevel$ is $n + 1$;
\item $HC(m)$ is the $m$th Hansel chain.
\end{itemize}

\ssk We prove in Section~\ref{sec:proofAlgoHansel} that Algorithm~\ref{algo:hanselChains} indeed generates the Hansel chains.

\ssk We start by giving an informal description of the algorithm. It works as follows. Level $L(0)$ contains a single element $(1, \ldots, 1)$. Chain $HC(1)$ starts with this element. We shall call it the \emph{head} of the chain (\ie its largest element \wrt $\supseteq$). The second element of $HC(1)$ is the successor of the head that is minimal in lexicographic order (\suclex-minimal) in $L(1)$, \ie $(0,1, \ldots, 1)$. The next element is always chosen as the \suclex-minimal successor of the previous element. This process thus generates the chain $HC(1)$ ending with the element $(0, \ldots, 0)$. Its element in $L(p)$ is $(0, \ldots, 0, 1, \ldots, 1)$ with the first $p$ coordinates equal to 0 and the last $n-p$ equal to 1. Each element included in a chain is marked as ``visited''. 

\ssk Chain $HC(2)$ starts with the \suclex-maximal element in $L(1)$. The next step selects its \suclex-minimal non-visited successor in $L(2)$. The process goes on always selecting the \suclex-minimal non-visited successor of the last element included in the chain. This generates $HC(2)$, a chain of length $n-1$ ending in $L(n-1)$. For the next chains, we always select their heads as the \suclex-maximal non-visited element in $L(1)$ and proceed with choosing iteratively the \suclex-minimal non-visited successor in the next levels. We repeat this until there is no element left in $L(1)$. 

\ssk  We then start chains with the \suclex-maximal non-visited element in the next level and complete the chains as before. The last series of chains starts from level $L(\lfloor n/2 \rfloor)$. These chains have 1 element if $n$ is even and 2 elements if $n$ is odd.  

\ssk A formal description of Algorithm~\ref{algo:hanselChains} follows.


\begin{breakablealgorithm}
  \caption{Construction of Hansel Chains}\label{algo:hanselChains}
  \begin{algorithmic}[0]
    \For{$m$ = 0 to ($numChains$ - 1)}
    \For{$p$ = 0 to ($numLevels$ - 1)}
    \If {There is an element in $L(p)$}
    \parState{Extract \suclex-max element in level $L(p)$ \\ \phantom{.} \hfill and add at the end of chain $HC(m)$}
    \State $k \gets p $
    \While {There is an element in $L(k+1)$}
    \parLoop { Over remaining elements in $L(k+1)$ \\ \phantom{.} \hfill starting from the \suclex-min element}
    \parIf{ Current element of $L(k+1)$ is\\ \phantom{.} \hfill subset of last element in $HC(m)$}
    \parState {Extract current element of $L(k+1)$\\ \phantom{.} \hfill and add at the end of chain $HC(m)$}
    \State Break the loop
    \EndparIf
    \EndparLoop
    \State $k \gets k + 1$
    \If{ $k+1 \geq numLevel$}
    \State Break the While loop
    \EndIf
    \EndWhile
    \State Break the For loop $p$
    \EndIf
    \State $p \gets p+1$
    \EndFor
    \State $m \gets m+1$
    \EndFor
  \end{algorithmic}
\end{breakablealgorithm}

\begin{figure}[h!!!]
    \centering
  \includegraphics[width=7cm]{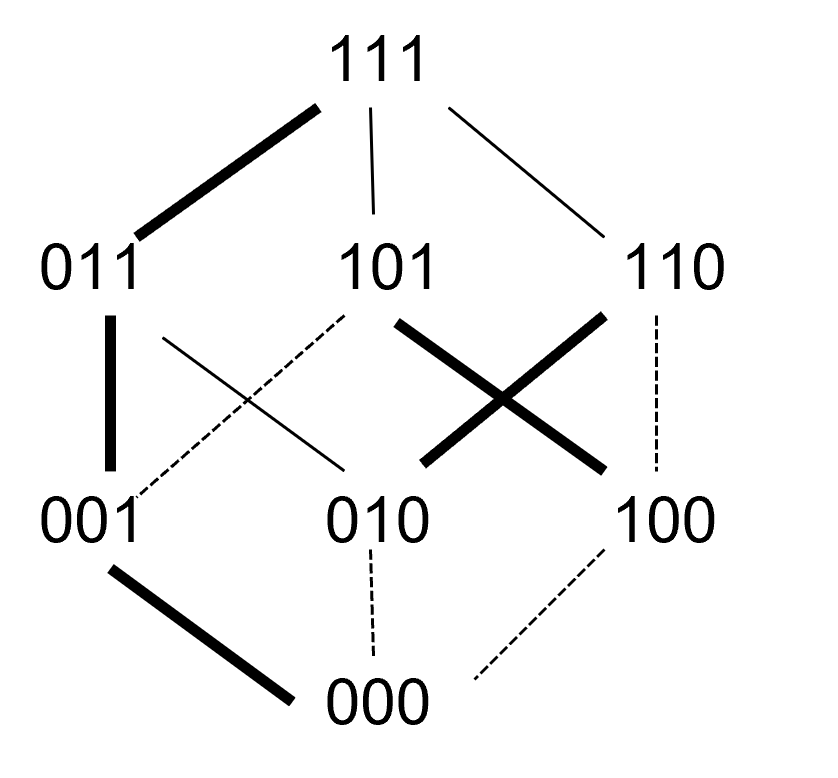}
  \caption{Cube $n=3$\label{fig:Cuben=3}}
  \end{figure}

\begin{figure}[h!!!]
    \centering
  \includegraphics[width=12cm]{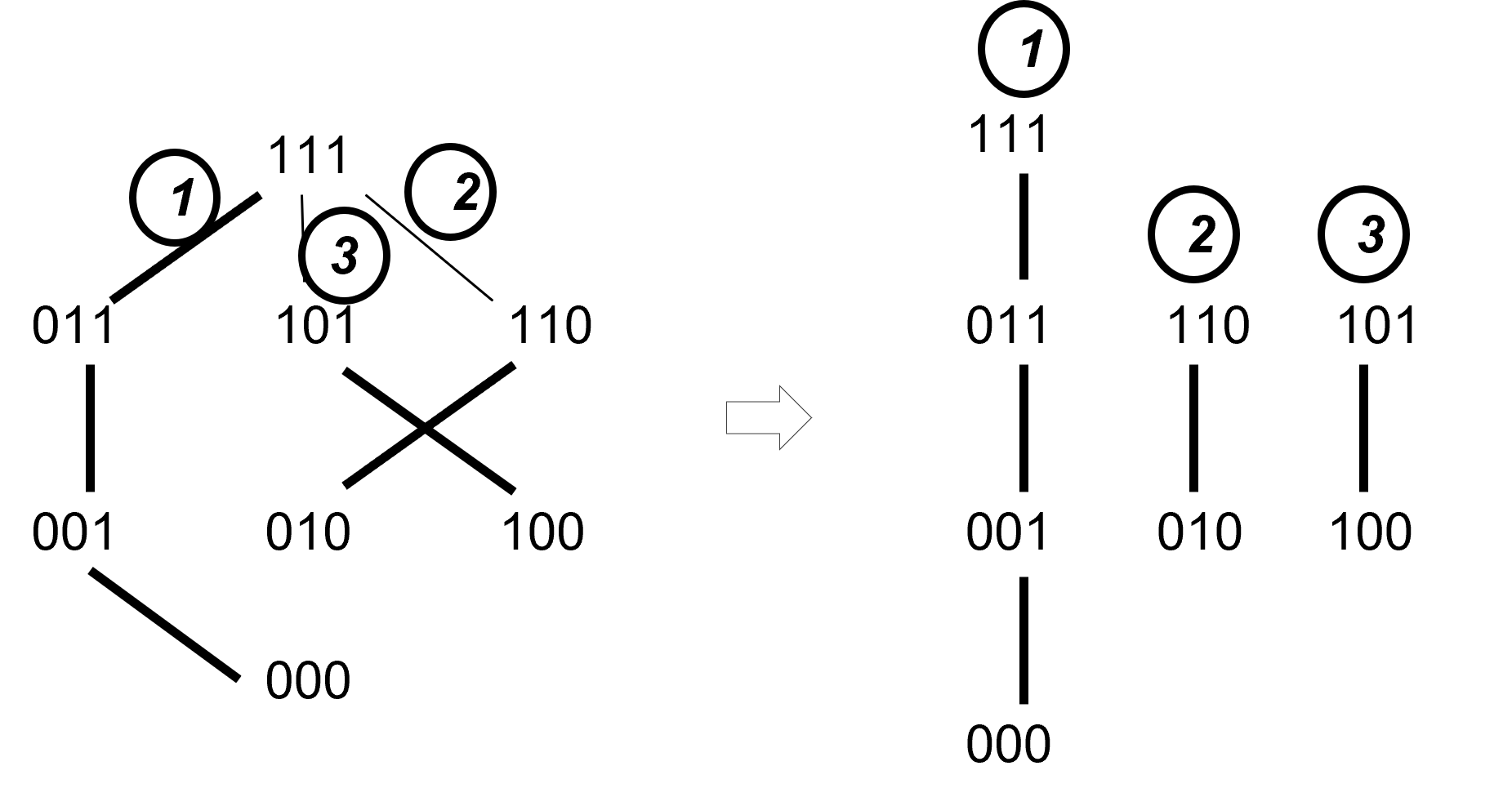}
  \caption{Modified DFS (left) yielding the Hansel chains (right) for $n=3$. The circled numbers label the Hansel chains $HC(m)$, $m=1, 2, 3$ \label{fig:DFSHansel}}
\end{figure}

\ssk For $n=2$, it is easily checked that the algorithm returns the chains $HC(0) = (11, 01, 00)$ and $HC(1) = (10)$. For $n=3$, the first chain $HC(1)$ starts with $111$ thus successively includes $011$, $001$ and  $000$. The head of chain $HC(2)$ is the \suclex-maximal non-visited element in $L(1)$, \ie  $110$. The chain proceeds with the \suclex-minimal non-visited successor of $110$ in $L(2)$, \ie $010$, which is the second and last element of this chain. Chain $HC(3)$ has its head in $101$ and finishes with $001$ (see Figures~\ref{fig:Cuben=3} and~\ref{fig:DFSHansel}).

\paragraph{Remark.} Algorithm~\ref{algo:hanselChains} can be viewed as a modified DFS. The first chain is the longest one as in ordinary DFS. Then we backtrack to the level $L(1)$ instead of to the first node from which it is possible to make a descent. In $L(1)$, we select the \emph{largest} non visited element (in lexicographic order) and go down as deep as possible by always choosing the \emph{smallest} non visited successor. Then we backtrack again to $L(1)$ and so on until all elements in $L(1)$ have been used. Then, we select the largest non-visited element in $L(2)$ (if there are), and proceed similarly. The process stops as soon as all elements have been included. The resulting contains all Hansel chains plus dummy edges linking the top element to the top element of each Hansel chain. This is illustrated in Figures~\ref{fig:Cuben=3} and~\ref{fig:DFSHansel} for the case $n=3$.

\begin{figure}[h!!!]
    \centering
  \includegraphics[width=9.5cm]{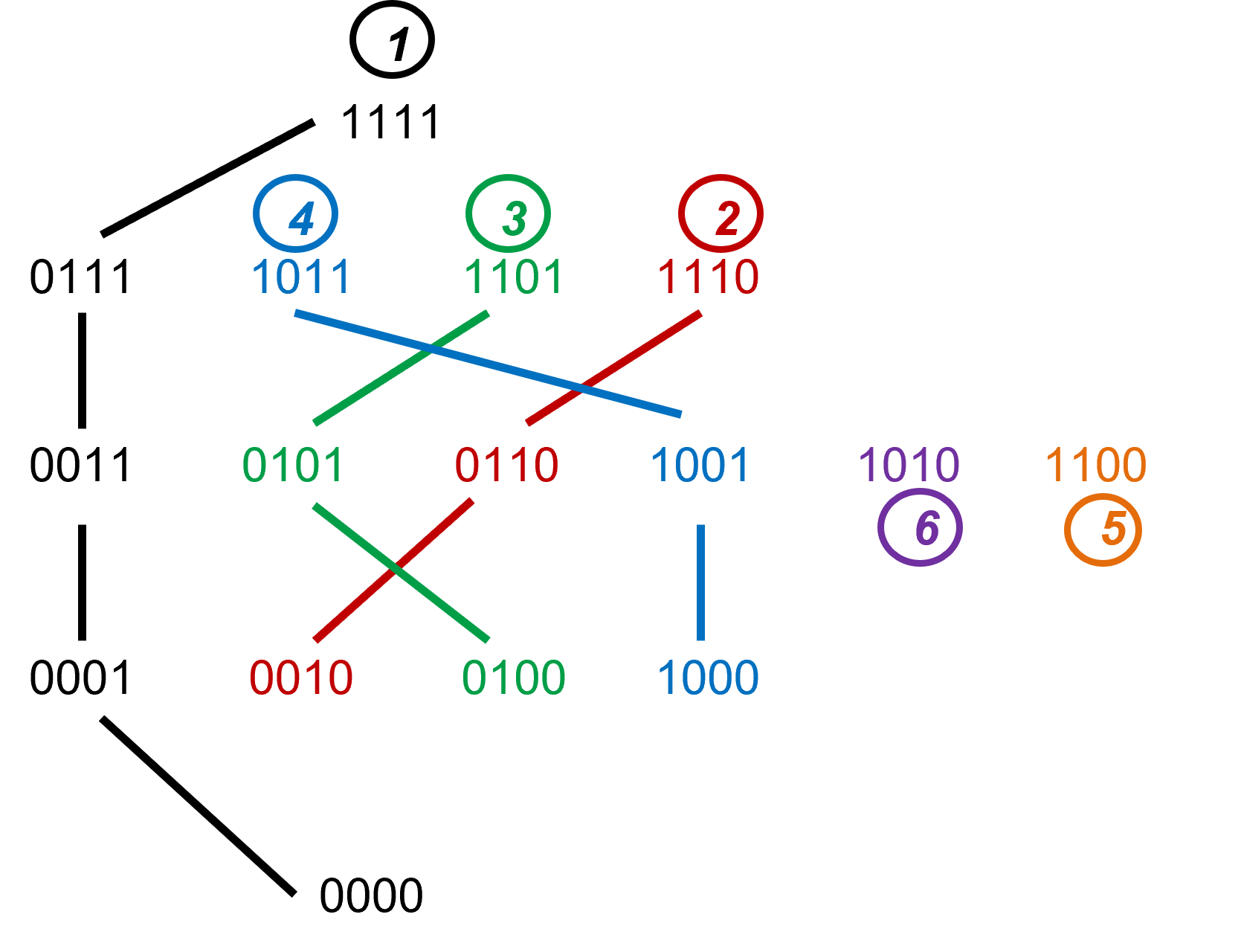}
  \caption{Modified DFS yielding the Hansel chains for $n=4$\label{fig:DFSHanseln=4}. The circled numbers label the Hansel chains $HC(m)$, $m=1, \ldots, 6$}
\end{figure}

\begin{figure}[h!!!]
    \centering
  \includegraphics[width=132mm]{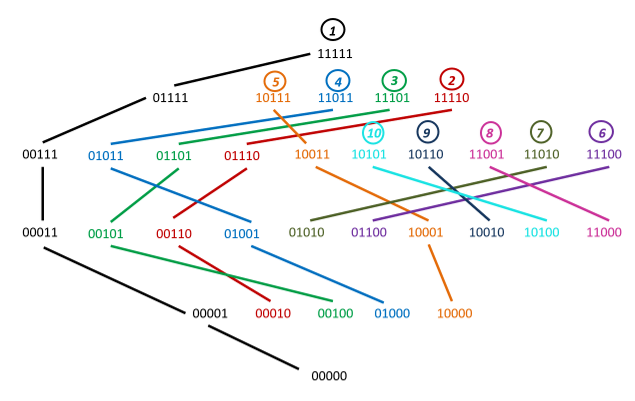}
  \caption{Modified DFS yielding the Hansel chains for $n=5$\label{fig:DFSHanseln=5}. The circled numbers label the Hansel chains $HC(m)$, $m=1, \ldots, 10$}
\end{figure}

\ssk The Hansel chains for $n=4$ and $n=5$ are shown in Figure~\ref{fig:DFSHanseln=4} and~\ref{fig:DFSHanseln=5}, respectively.

\clearpage

\section{Proof of Algorithm~\ref{algo:hanselChains}}\label{sec:proofAlgoHansel}

This section aims to prove that Algorithm~\ref{algo:hanselChains} generates the Hansel chains of $B_n$ for all $n$. We first study the way Hansel chains are produced by the original Hansel's construction. Then we use these properties to prove our algorithm. 

\subsection{Properties of Hansel chains}\label{ssec:propHanselChains}

We analyze how many Hansel chains cross each level $L(p)$ of $B_n$ or start (\ie have their head) in $L(p)$. We also distinguish between those that are augmented Hansel chains of $B_{n-1}^0$ and those that are truncated Hansel chains of $B_{n-1}^1$. In this section, we use ``HC'' as a shorthand  for ``Hansel chain''. The levels of $B_n$ are simply denoted by $L(p)$. For the levels of $B_{n-1}$, we write $L_{n-1}(p)$.    

\ssk We recall that there are $n \choose p$ HCs of length $n-2p+1$. This implies that these chains start from level $L(p)$ or before that level (we go through levels starting from $L(0)$). The number of chains starting precisely from $L(p)$ hence is ${n\choose p} - {n\choose {p-1} }$. 

\paragraph{Level $L(0)$ }  For all $n$, the longest chain starts from $(1 \ldots 1)$. It is an augmented chain of $B_{n-1}^0$ of length $n-2p +1 = n+1$.

\paragraph{Level $L(1)$ } There are $n$ elements in $L(1)$. Only one element has 0 as its first coordinate. This element, $01 \ldots 1$ is visited by the longest chain, starting from $L(0)$. The other $n-1$ elements are the starting points of $n-1$ HCs. One of them is the longest truncated chain of $B_{n-1}^1$ thus of length $n-1$; the head of this chain is vector $(101 \ldots 1)$. The other $n-2$ HCs are augmented chains of $B_{n-1}^0$. There are indeed ${{n -1} \choose 1} -1 $ HCs in $B_{n-1}$ with length $n-2$.   

\paragraph{General case $L(p)$ for $p\geq 2$ and $n \geq 4$} Table~\ref{ta:levelLp} describes the number of HCs of the two types that cross or start in level $L(p)$. This description is established below.  There are $n \choose p$ elements in level $L(p)$. Among them ${n-1} \choose {p-1} $ have 0 as their first coordinate;  ${n-1} \choose {p} $ have 1 as first coordinate. The number of HCs starting before level $L(p)$ is $n \choose {p-1}$; their length is at least $n-2(p-1)+1 = n-2p+3$. Among them, ${n-1} \choose {p-1}$ are augmented HCs from $B_{n-1}$ of length at least $n-2p+2$. There are ${n-1} \choose {p-1}$ such HCs. This means that all of the elements of $L(p)$ with first coordinate equal to 0 belong to augmented chains of $B_{n-1}$ starting in previous levels. The only unvisited elements hence have their first coordinate equal to 1. The number of truncated chains of $B_{n-1}^1$ starting before $L(p)$ is the number of HCs in $B_{n-1}$ of length at least $n-2p +4$; these thus have their first element in level $L_{n-1}(p-2)$ of $B_{n-1}^1$; there are ${n-1} \choose {p-2}$ such chains. Hence, among the ${n-1} \choose p$ elements with 1 as first coordinate in $L(p)$, ${n-1} \choose {p-2}$ belong to truncated chains of $B_{n-1}^1$ starting above $L(p)$. 

\ssk How many new HCs start in $L(p)$? The number of HCs of $B_{n-1}$ starting precisely in $L_{n-1}(p)$ is ${{n-1}\choose {p}} - {{n-1}\choose{p-1}}$. Therefore, this is the number of augmented chains from $B_{n-1}^0$ starting in $L(p)$. The rest, \ie the ${{n- 1}\choose{p-1}} - {{n-1}\choose{p-2}}$ remaining elements are the starting points of truncated HCs of $B_{n-1}^1$.

\renewcommand{\arraystretch}{1.5}

\begin{table}[h!!!]
    \centering
    \resizebox{\columnwidth}{!}{%
    \begin{tabular}{|c|c|c|c|} 
\hline
    $(0 \ldots)$     & \multicolumn{3}{|c|}{$(1 \ldots )$} \\
    ${n-1}\choose {p-1}$     & \multicolumn{3}{|c|}{${n-1}\choose{p}$} \\
\cline{2-4}
    & ${n-1}\choose{p-2}$ & ${{n-1}\choose{p-1}} - {{n-1}\choose{p-2}}$ & ${{n-1}\choose{p}} - {{n-1}\choose{p-1}}$ \\
augmented    & truncated & truncated & augmented\\
started before $L(p)$    & started before $L(p)$& started in $L(p)$& started in $L(p)$ \\
\hline
    \end{tabular}
    }
    \caption{Partition of the elements of $L(p)$ (for $p \geq 2$ and $n\geq 4$) according to whether they belong to an augmented or truncated Hansel chain starting before $L(p)$ or precisely in $L(p)$. The binomial coefficients indicate the number of elements of each kind in $L(p)$}
    \label{ta:levelLp}
\end{table}

\ssk Which are the elements in $L(p)$ (for $p\geq 2$ and $n\geq 4$) from which the new chains start ? We establish the following result.

\begin{proposition}\label{prop:HCmaxLexico}
In each level $L(p)$, the augmented HCs from $B_{n-1}^0$ starting from this level are the ${{n-1}\choose{p}} - {{n-1}\choose{p-1}}$ \suclex-maximal elements in $L(p)$.
\end{proposition}

\bpr
We can see from Figures~\ref{fig:twohansel}, \ref{fig:threehansel} and \ref{fig:fourHansel} that the property is true for $n=2, 3, 4$. The proof then proceeds by induction. Assuming that the property is true up to $n-1$, we prove that it is also true for $n$. 

\ssk We know from what precedes that the initial elements of new augmented chains have their first coordinate equal to 1, \ie they belong to $B_{n-1}^1 \cap L(p)$. The elements of $B_{n-1}^1 \cap L(p)$ are in one-to-one correspondence with those of level $L_{n-1}(p)$ in $B_{n-1}$; the former just differ from the latter by an additional first coordinate equal to 1. Similarly, the elements of $B_{n-1}^0 \cap L(p+1)$ are in one-to-one correspondence with those of level $L_{n-1}(p)$ in $B_{n-1}$; the former just differ from the latter by an additional first coordinate equal to 0. The \suclex\ order on $L_{n-1}(p)$ is not changed by the adjunction of an initial coordinate equal to 1 or to 0. Therefore, the ${{n-1}\choose{p}} - {{n-1}\choose{p-1}}$ \suclex-maximal elements in $B_{n-1}$ correspond to ${{n-1}\choose{p}} - {{n-1}\choose{p-1}}$ \suclex-maximal elements in $B_{n-1}^1 \cap L(p)$ and in $B_{n-1}^0 \cap L(p+1)$. By the induction hypothesis, the HCs of $B_{n-1}$ starting from level $L_{n-1}(p)$ start from the ${{n-1}\choose{p}} - {{n-1}\choose{p-1}}$ \suclex-maximal elements of $L_{n-1}(p)$. The corresponding chains in $B_{n-1}^0$ hence start from the ${{n-1}\choose{p}} - {{n-1}\choose{p-1}}$ \suclex-maximal elements from $B_{n-1}^0 \cap L(p+1)$ and are augmented by the ${{n-1}\choose{p}} - {{n-1}\choose{p-1}}$ \suclex-maximal elements in $B_{n-1}^1 \cap L(p)$.
\epr

\paragraph{Illustration for $n=5$ }  The construction of HCs for $n=5$ is represented in Table~\ref{ta:HCn5}. The copies in $B_{n-1}^0$ (\resp $B_{n-1}^1$) of the six HCs of $B_{n-1}$ are displayed in the six leftmost (\resp rightmost) columns of the table. The six HCs of $B_{n-1}^0$ are augmented by the six boldface elements heading the six rightmost columns.  The rest of the six rightmost columns, without the boldface elements are the truncated HCs from $B_{n-1}^1$.       

\begin{table}[h!!!]
    \centering
\resizebox{\columnwidth}{!}{%
     \begin{tabular}{|c|c|c|c|c|c|c|c|c|c|c|c|}
\hline
        & & & & & & \textbf{11111} & & & & &  \\
\hline
\textit{01111}   & & & & & & 10111 & \textbf{11110} & \textbf{11101} & \textbf{11011} & &  \\
\hline        
00111   & \textit{01110} & \textit{01101} & \textit{01011} & & & 10011 & 10110 & 10101 & 11001 & \textbf{11010} & \textbf{11100} \\
\hline        
00011   & 00110 & 00101 & 01001 & \textit{01010} & \textit{01100} & 10001 & 10010 & 10100 & 11000 & &  \\
\hline        
00001   & 00010 & 00100 & 01000 & & & 10000 & & & & &  \\
\hline        
00000   & & & & & & & & & & &  \\
\hline         
    \end{tabular}
    }
    \caption{Construction of Hansel chains in $B_5$. The six leftmost (\resp rightmost) columns represent $B_{n-1}^0$ (\resp $B_{n-1}^1$). The rows represent the levels $L(p)$, $p=0, \ldots, 5$. The elements in bold are the head of augmented chains of $B_{n-1}^0$; the elements in italics are the heads of the chains in $B_{n-1}^0$ which are augmented by the elements in bold. The truncated chains appear in the columns headed by boldfaced elements.}
    \label{ta:HCn5}
\end{table}

We may check that the binomial coefficient in the four columns of Table~\ref{ta:levelLp} indeed yield the correct numbers of elements of the different categories in the case of $B_5$. For instance, for level $L(2)$,  
\begin{itemize}
    \item the number of augmented HCs started before $L(2)$ is ${{n-1}\choose{p-1}} = {4\choose 1} = 4 $;
    \item the number of truncated HCs started before $L(2)$ is ${{n-1}\choose{p-2}} = {4\choose 0} = 1 $;
    \item the number of truncated HCs started in $L(2)$ is ${{n-1}\choose{p-1}} - {{n-1}\choose{p-2}} = 4 -1 =3 $;
    \item the number of augmented HCs started in $L(2)$ is ${{n-1}\choose{p}} - {{n-1}\choose{p-1}} = 6 -4 =2 $.
\end{itemize}
We also observe that the heads of the augmented HCs are the \suclex-maximal elements in each level.  

\ssk We may also check that applying Algorithm~\ref{algo:hanselChains} to $B_5$ yields the HCs. In each level $L(p)$, the algorithm starts building the chains by considering the \suclex-maximal elements. Therefore, it starts by building all the augmented HCs starting in $L(p)$. It is then left with the first elements of the truncated chains starting in $L(p)$, which are the only non-visited elements of $L(p)$. An important observation is the following. The order in which the algorithm is applied to the ${{n-1}\choose{p}} - {{n-1}\choose{p-1}}$ \suclex-maximal elements of $L(p)$ is irrelevant. The algorithm generates the  correct HCs also when the ${{n-1}\choose{p}} - {{n-1}\choose{p-1}}$ \suclex-maximal elements are not considered in \suclex\ decreasing order. A similar remark holds for the group of the next ${{n-1}\choose{p-1}} - {{n-1}\choose{p-2}}$  elements (in $\suclex$ order) that are the heads of the truncated chains starting in $L(p)$. The important thing is that  the algorithm generates the set of extended chains starting in level $L(p)$ before generating the truncated ones. 

\subsection{Main result}\label{ssec:theoremAlgoChains}
We are now ready to prove that, for all $n$, Algorithm~\ref{algo:hanselChains}  actually constructs the Hansel chains of $B_n$ directly, and not recursively, as in the original work by \citet{Hansel1966}. This is the goal of the next theorem.

\begin{theorem}\label{th:algo1}
  Algorithm \ref{algo:hanselChains} returns the Hansel chains covering $B_n$, for all $n$.
\end{theorem}



\bpr
The proof is by recurrence. We verified that the theorem is true for $n=2$ to 5. We assume that it is true up to $n-1$ and prove it is true for $n$. 

\ssk When using Algorithm~\ref{algo:hanselChains}, the chains are generated sequentially  level after level. So, in level $L(0)$ there is only one element $1 \ldots 1$. By using the algorithm, we generate a chain starting from $1\ldots 1$. The next element in the chain is obtained by transforming the leftmost 1 into 0 in the previous element of the chain. It ends up with element $0 \ldots 0$. This chain is the longest Hansel chain HC(1). The algorithm proceeds with level $L(1)$. There is only one element of $B_{n-1}^0$ in $L(1)$; it is already visited by chain HC(1). The $n-1$ remaining elements belong to $B_{n-1}^1$. The $n-2$ \suclex-maximal elements among them are the initial elements of the augmented chains starting in $L(1)$ (by Proposition~\ref{prop:HCmaxLexico}). Applying the algorithm to these elements amounts to replace their initial 1 coordinate by 0. We are thus in level $L(2)$ of $B_{n-1}^0$. By the recurrence hypothesis, applying the algorithm to each of these elements yields Hansel chains of $B_{n-1}$ with an additional initial element equal to 0. The remaining element of $L(1)$ is $101 \ldots 1$. Applying the algorithm to this element does not yield $001 \ldots 1$ because the latter is already visited by HC(1). The next element of the chain will keep the 1 in initial position. This element is indeed the head of a truncated HC of $B_{n-1}^1$. By the recurrence hypothesis, applying the algorithm to this element while the augmented chains already have been generated produces the first truncated Hansel chain of $B_{n-1}^1$. After having visited all elements in $L(0)$ and $L(1)$, we   generated the $n$ first augmented and the first truncated HCs of $B_n$.

\ssk The proof proceeds with a second recurrence hypothesis. Assuming that the chains generated by the algorithm when going through levels $L(0)$ to $L(p-1)$ are the $n \choose {p-1}$ HCs starting from these levels, we prove that those obtained when going through level $L(p)$ are the ${{n}\choose {p}} - {{n} \choose {p-1}}$ HCs starting from $L(p)$.  

\ssk When it comes to $L(p)$, the algorithm finds that all elements of $B_{n-1}^0 \cap L(p)$ were visited by the augmented HCs started before $L(p)$. Another ${n-1}\choose{p-2}$ elements of $B_{n-1}^1 \cap L(p)$ were visited by truncated HCs started before $L(p)$. The ${{n-1}\choose {p}} - {{n-1} \choose {p-1}}$ \suclex-maximal elements of $L(p)$ are non-visited (because they are the head of the augmented HCs starting from $L(p)$). The algorithm first takes each of them in turn. For each of them, its \suclex-minimal successor is obtained by transforming its initial 1 coordinate into 0. This element  is non-visited because it is the second element of an augmented HC that was not yet started. This element is thus added to the chain. We are now in $B_{n-1}^0$, \ie a copy of $B_{n-1}$. By the first recurrence hypothesis, we know that applying the algorithm to this element generates the HC of $B_{n-1}$, prefixed by an initial 0 coordinate,  starting from this element. The chains generated by the algorithm from the ${{n-1}\choose {p}} - {{n-1} \choose {p-1}}$ \suclex-maximal elements of $L(p)$ are all the augmented HCs starting from $L(p)$. At this stage, the ${{n-1}\choose {p}} - {{n-1} \choose {p-2}}$ remaining non-visited elements in $L(p)$ are the second elements of the  HCs of length $n-1 - 2(p-1) +1 = n- 2p +2$ of $B_{n-1}^1$ from which the heads is removed, \ie the truncated chains of length $n-2p+1$ starting in $L(p)$. Using again the first recurrence hypothesis, when applying the algorithm to these elements (in any order), we obtain the rest of the truncated HCs starting from $L(p)$.

\epr

\section{Modified DFS1 Algorithm: DFSHansel}\label{sec:DFSHansel}

An optimal questioning algorithm in the worst case is obtained as follows:
\begin{itemize}
     \item Generate the Hansel chains by means of Algorithm~\ref{algo:hanselChains} and connect the top element of each chain to the top element of $B_n$ by a dummy edge;
    \item Run through each branch of the arborescence, starting with the shortest ones and ask questions as in DFS1; propagate the consequences of each answer. 
\end{itemize}
Note that, instead of asking questions as in DFS1, we could do as in DFS2 or choose another questioning option such as first questioning in the middle of each Hansel chain, starting with the shortest ones. The optimality in the worst case is guaranteed anyway.

\ssk There are (at least) two possible ways of propagating:
\begin{itemize}
    \item full propagation: draw all consequences of each answer;
    \item limited propagation: propagate only the consequences to the chains of length ``+2'', \ie the chains that have two elements more than the current chain.
\end{itemize}
Due to Hansel's Lemma, the limited propagation guarantees that the number of questions will not exceed the lower bound. The full propagation may reduce the number of questions (except in the worst case) at the cost of additional computational effort. 

\ssk Algorithm \ref{alg:DFSHansel} implements an optimal (in the worst case) questioning strategy  and  updates the status (visited or not) of chains elements according to user response, using full propagation.


\algloop{Continue}
\algloop{Input}

\begin{algorithm}
  \caption{DFSHansel}\label{alg:DFSHansel}
  \begin{algorithmic}
    \State $numQuestions \gets 0$
    \For{$p$ = ($numChains$ - 1) to 0}
    \For{$k$ = (Number of elements in $HC(i)$ - 1) to 0}
    \If{$k^{th}$ element of $HC(i)$ (named as $X$) is visited}
    \Continue
    \EndIf
    \State Ask user if current element $X$ is marked as $S$ (i.e. Sufficient) or $I$ (i.e. Insufficient)
    \State Update this element's status and mark as visited.
    \State $numQuestions \gets numQuestions + 1$
    \Loop{ Over all elements of chain (named as $Y$)}
    \If{Input is $S$}
    \If{$Y$ is superset of $X$}
    \State Mark the element $Y$ as visited and $S$
    \EndIf
    \ElsIf{Input is $I$}
    \If{$Y$ is subset of $X$}
    \State Mark the element $Y$ as visited and $I$
    \EndIf
    \EndIf
    \EndLoop
    \State $k \gets k-1$
    \EndFor
    \State $p \gets p-1$
    \EndFor
    \State{$F_{min} \gets$ Elements which are marked as $S$ and cannot be defined as subset of each other.}
    \State{Number of questions asked $\gets numQuestions$}
  \end{algorithmic}
\end{algorithm}



\clearpage
\paragraph{Illustration of DFSHansel}\label{ssec:DFSHanselIllustr}

We give some examples of the functioning of DFSHansel algorithm in different cases. Full propagation is done for each situation. 

\begin{figure}[h!!]
\begin{center}
\includegraphics[width=8cm]{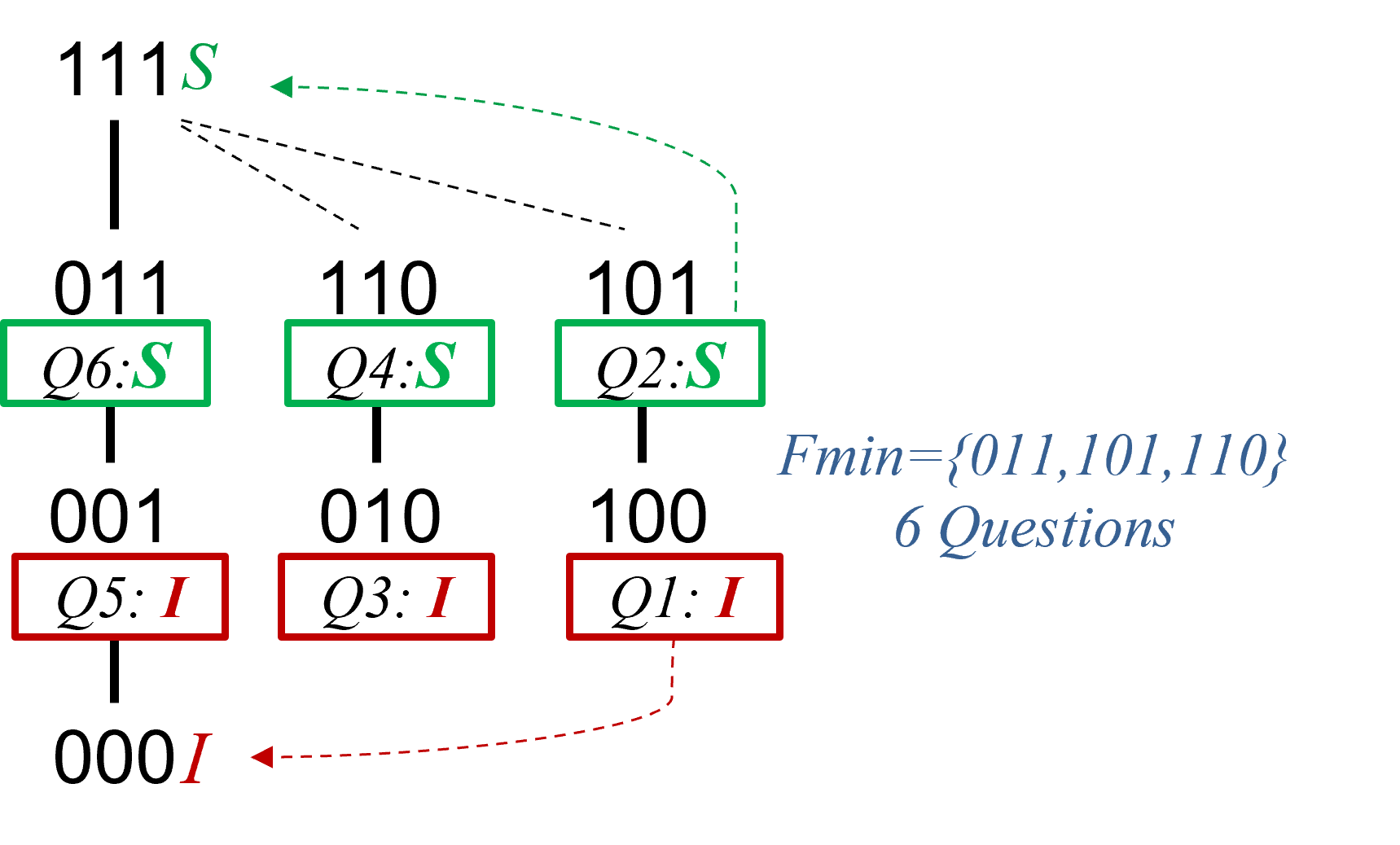}
\vspace{0.5cm}
\caption{DFSHansel $n=3$ Majority. Dashed lines terminating by an arrow indicate propagation results}
\label{fig:DFSHanseln=3MAJ}
\end{center}
\end{figure}

\paragraph{Case $n=3$ and \Fmin = $\{maj\}$}
If we require that sufficient coalitions involve at least a majority of criteria, \ie two, we are in the worst case. Both the minimal and the maximal propagation lead to 6 questions. Note that DFS3 also requires 6 questions in this case, while DFS1 and DFS2 require 7 questions. This situation is shown in Figure~\ref{fig:DFSHanseln=3MAJ}

\begin{figure}[h!!!]
\begin{center}
\includegraphics[width=10cm]{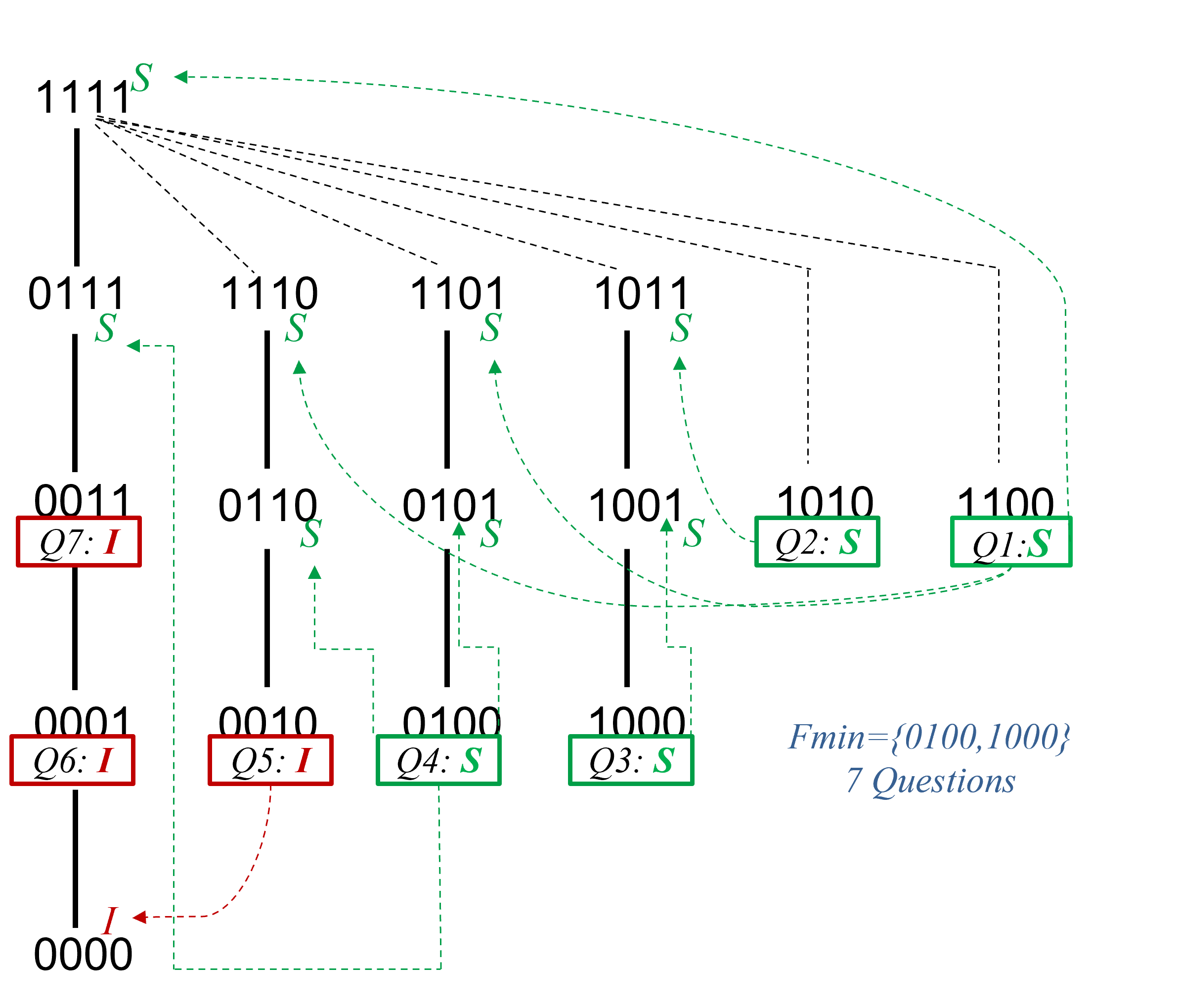}
\vspace{0.5cm}
\caption{DFSHansel for $n=4$, $\Fmin = \{1,2\}$. Dashed lines terminating by an arrow indicate propagation results \label{fig:DFSHanseln=4Ex1}}
\end{center}
\end{figure}

\clearpage
\paragraph{Case $n=4$ and \Fmin $=\{1,2\}$}
In the case of four criteria with $\Fmin=\{1,2\}$, \ie  $\Fmin{} = \{1000,0100\}$, $7$ questions are asked via DFSHansel. The mechanism of the algorithm for the given \Fmin is illustrated in Figure~\ref{fig:DFSHanseln=4Ex1}. With DFS3, the same \Fmin requires $5$ questions which is a better performance compared to DFSHansel. 

\begin{figure}[h!!!]
\begin{center}
\includegraphics[width=10cm]{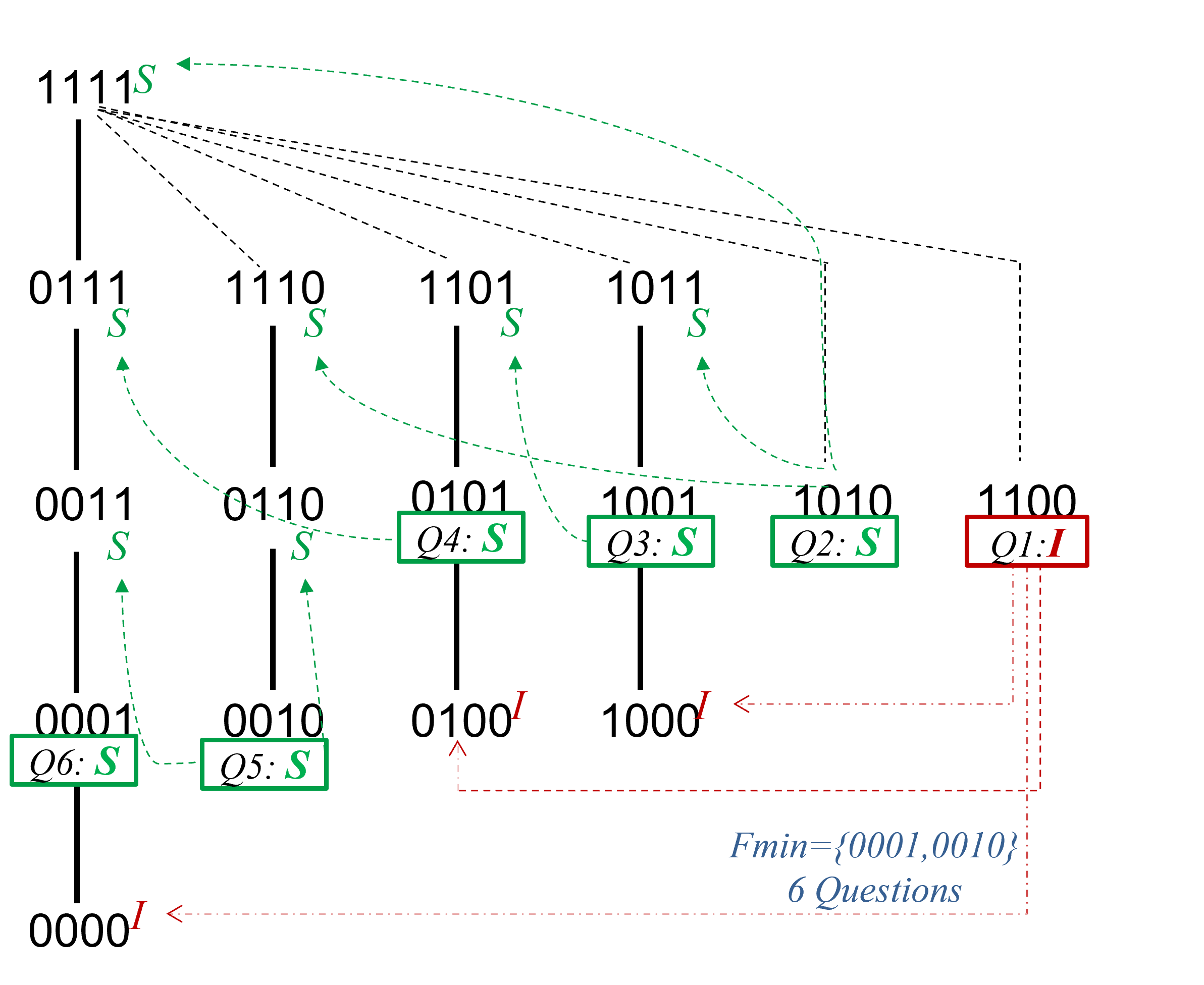}
\vspace{0.5cm}
\caption{DFSHansel for $n=4$ and $\Fmin=\{3,4\}$. Dashed lines terminating by an arrow indicate propagation results}
\label{fig:DFSHanseln=4Ex2}
\end{center}
\end{figure}

\clearpage
\paragraph{Case $n=4$ and \Fmin$=\{3,4\}$ } In the case of \Fmin$=\{3,4\}$, \ie $\{0010,0001\}$, $6$ questions are asked with DFSHansel. Figure~\ref{fig:DFSHanseln=4Ex2} shows the questioning procedure. DFS3 asks $7$ questions for this \Fmin. In this case, DFSHansel is more efficient.

\ssk Even if DFS3 performs better in some cases,  like $n=4$ and $\Fmin= \{1,2\}$, there is no guarantee of not exceeding the minimum number of questions in the worst case. For instance if $\Fmin =\{14, 24, 34\}$, DFS3 requires 11 questions while the bound is 10 questions for $n=4$. This bound is always guaranteed by DFSHansel. In order to make an objective comparison of the algorithms, it will be interesting to calculate the average number of questions for all possible \Fmin with DFSHansel (see Section~\ref{ssec:DFSHanselOnAverage}).


\subsection{The implementation of DFSHansel }\label{ssec:ImplementationDFSHansel}
An implementation of the DFSHansel algorithm is done in Python in order to construct the Hansel chains for a given number of criteria and then to run the questioning procedure. 

\ssk The construction of the Hansel chains is done by  Algorithm~\ref{algo:hanselChains}, for a value of $n$ entered by the user. After the construction of the Hansel chains, the program starts asking questions from the bottom element of the shortest Hansel Chain. It asks the user if the current coalition is sufficient or not. If it is labeled as ``sufficient (S)'', then the program labelizes $S$ all coalitions that are supersets of the current coalition. If it is labeled as ``insufficient (I)'', then the program labelizes $I$ all coalitions that are subsets of the current coalition, and moves on to the next coalition that is unlabeled. When moving to the next coalition, it again picks the shortest Hansel Chain with an unlabeled coalition and starts asking questions from the element (coalition) situated in the bottom of the shortest chain. When all elements are marked, \ie there are no more questions to ask, the program determines \Fmin{}. 

\ssk As mentioned in the previous chapters, \Fmin{} should be chosen from among sufficient coalitions that cannot be represented as subsets of each other, and all sufficient coalitions should be a superset of at least one of the selected coalitions.

\begin{figure}[h!!!]
\begin{center}
\includegraphics[width=135mm]{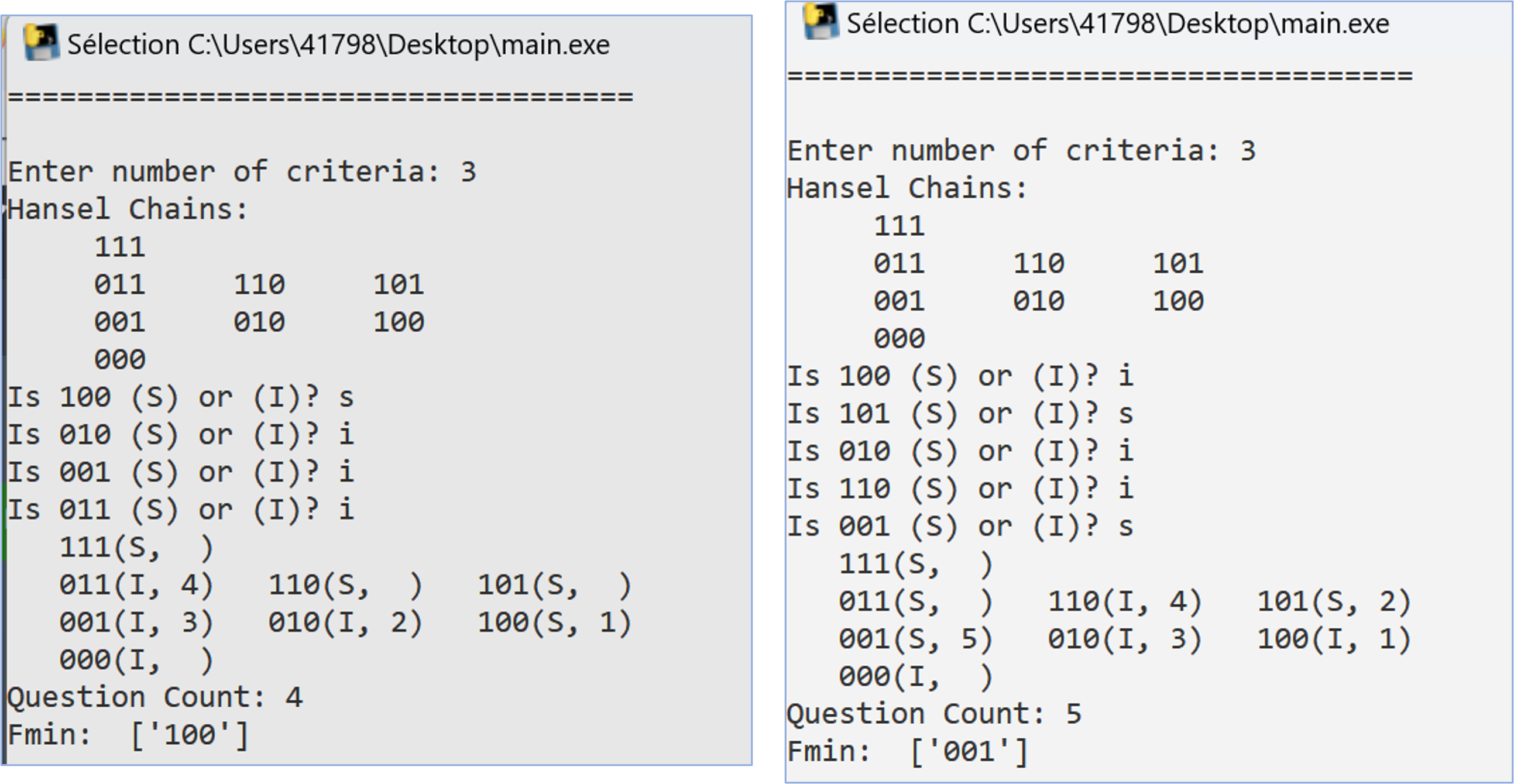}
\vspace{0.5cm}
\caption{DFSHansel for $n=3$, $\Fmin=\{1\}$ (left) and $\Fmin=\{3\}$(right)}
\label{fig:DFSHansel_outputn3}
\end{center}
\end{figure}

\paragraph{An example of an output of DFSHansel}
Figure~\ref{fig:DFSHansel_outputn3} illustrates the outputs of DFSHansel for $n=3$, in the cases of $\Fmin=\{1\}$, \ie  $\{100\}$, and $\Fmin=\{3\}$, \ie  $\{001\}$, analyzed in the previous chapter. First, the number of criteria $n$ is entered by the user. Then the program displays all Hansel chains for this given $n$ and the questioning procedure starts. As the DFSHansel questioning algorithm consists to start from the smallest chains, for $n=3$ the first question is for 100, which is the element situated at the bottom of the smallest chain. This coalition is the single criterion $1$. The type of the program question is ``Is 100 (S) or (I)?'' and the answer has to be given as Sufficient(S) or Insufficient(I). On the left side, we can see the questioning sequence for $\Fmin=\{1\}$ and on the right the questioning sequence for $\Fmin=\{3\}$. Four questions have to be asked for the obtention of $\Fmin=\{1\}$. This \Fmin{} was obtained in 3 questions by DFS3 and was an example of the best cases. 

\ssk With DFSHansel, $\Fmin=\{3\}$ is obtained through 5 questions. This \Fmin was also obtained in 5 questions with DFS3 and was an example of a worst case. We note that, for $\Fmin=\{1\}$, the performance of DFS3 is better than the performance of DFSHansel (3 questions vs. 4 questions). DFSHansel and DFS3 ask both 5 questions and have the same performance for $\Fmin=\{3\}$.

\paragraph{Existence of additional information } 
When the DM can give additional information in terms of an importance order between criteria, for instance, the DFSHansel algorithm could take into account this information when propagating the consequences of the given answers. This was not implemented but it could be done without difficulty. 

\ssk We observed in Section~\ref{sec:additionalInfo}, Remark~\ref{rem:DFS3withAdditInfo}, that the principle of the DFS3 algorithm can be applied without change to the poset representing the importance relation on the coalitions of criteria, \ie the inclusion relation enriched by the importance order information (it was not implemented, however). In contrast, there seems to be nothing similar to Hansel's theory for general posets. 

\ssk In other words, we don't know how to modify Hansel chains when the Hasse diagram of the importance relation between criteria coalitions is richer than the inclusion relation.  We thus emphasize that DFSHansel does not make the DFS3 principle obsolete. When importance information between criteria coalitions exists, an implementation of the DFS3 principle for the corresponding Hasse diagram could be efficient and useful. 

\begin{figure}[h!!!]
\begin{center}
\includegraphics[width=132mm]{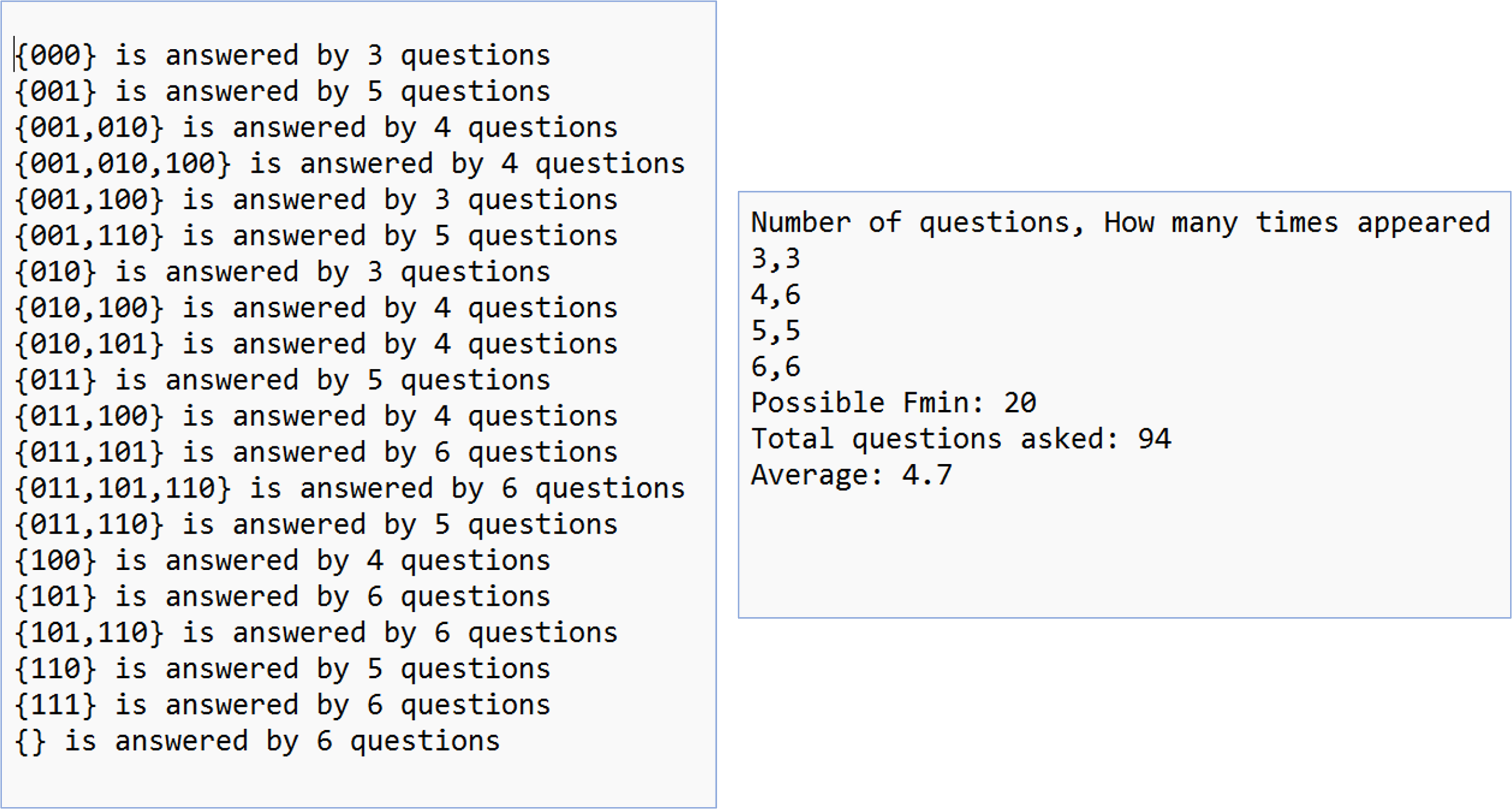}
\caption{The outputs of H-questioner for $n=3$}
\label{fig:Hquestioner_outputn3}
\end{center}
\end{figure}

\paragraph{The simulation module of DFSHansel: H-questioner} 
In order to compare the performance of DFSHansel, a simulation module  \textit{H-questioner} is implemented. For a given number of criteria, it takes the existing list of the possible \Fmin{} and executes the program of DFSHansel for each \Fmin{}. The results are given in two lists. The first list consists of the number of questions asked for every \Fmin{} and the second list gives the distribution and the average number of questions. Figure~\ref{fig:Hquestioner_outputn3} shows these two outputs for $n=3$.

\subsection{Performance of DFSHansel on average}\label{ssec:DFSHanselOnAverage} 
After the implementation of DFSHansel with a full propagation strategy, we wanted to compare the average number of questions for different numbers of criteria. Lists of possible \Fmin for $n=2, 3, 4, 5, 6$ were previously generated. The simulation module H-questioner is similar to the ``questioner'' module using DFS3 but has been implemented for DFSHansel algorithm. For a given $n$, ``H-questioner'' takes the corresponding list of possible \Fmin and answers the questions of the main DFSHansel program according to each \Fmin set.

\begin{figure}[h!!!]
\begin{center}
\includegraphics[width=132mm]{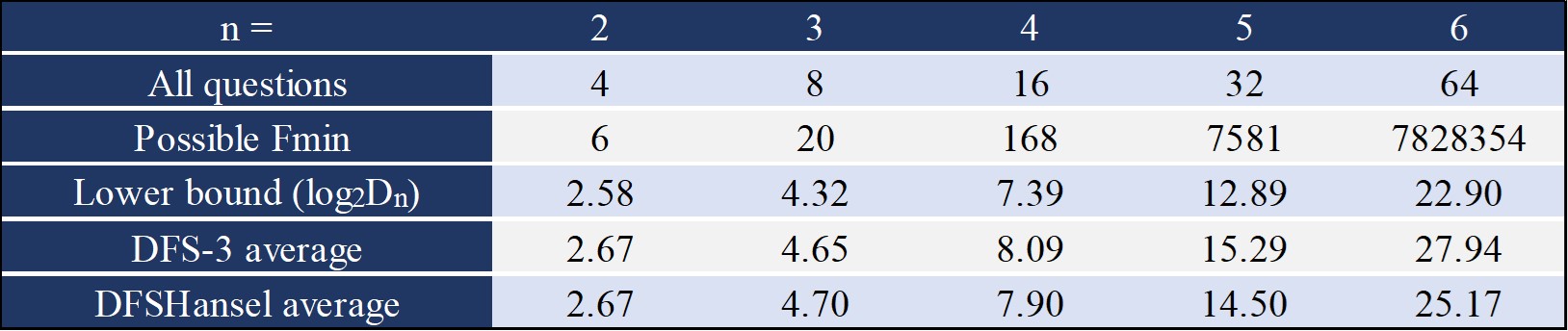}
\caption{Average number of questions by DFSHansel. Comparison with DFS3}
\label{fig:comparisonDFSHansel}
\end{center}
\end{figure}

\ssk The comparison of ``H-questioner'' and ``DFS3-questioner''  is illustrated in Figure~\ref{fig:comparisonDFSHansel}. According to the results of H-questioner, we observe that DFSHansel performs better than DFS3 on average. DFS3 is only slightly better for $n=3$ on average. 
The difference in favor of DFSHansel seems to grow with $n$, at least for $3 < n \leq 6$. This difference however is not very large.


\ssk In order to analyze this difference more in detail, we reproduce in Tables~\ref{ta:CompaNumbQuestn=3} and~\ref{ta:CompaNumbQuestn=4}, the distribution of the number of questions posed by DFS3 and DFSHansel for $n=3$ and $n=4$.

\begin{table}[h!]
\renewcommand{\arraystretch}{1}
    \centering
    \begin{tabular}{crr}
    \hline
      $\#$ of Questions   & DFS3 & DFSHansel \\
    \hline  
         3 & 3 & 3 \\
         4 & 5 & 6 \\
         5 & 8 & 5 \\
         6 & 4 & 6 \\
\hline
Total & 93 & 94 \\
\hline
    \end{tabular}
    \caption{Frequency of the number of questions by DFS3 and DFSHansel for $n=3$ criteria}
    \label{ta:CompaNumbQuestn=3}
\end{table}
We observe that for $n=3$, DFS3 uses one question less than DFSHansel in total. DFS3 asks 6 questions 4 times, when DFSHansel asks 6 times. Such an advantage fades out for $n\geq 4$.

\begin{table}[h!!!]
\renewcommand{\arraystretch}{1}
    \centering
    \begin{tabular}{crr}
    \hline
      $\#$ of Questions   & DFS3 & DFSHansel \\
    \hline  
         3 & 1 & - \\
         4 & - & - \\
         5 & 2 & 4 \\
         6 & 18 & 16 \\
         7 & 29 & 43 \\
         8 & 48 & 52 \\
         9 & 51 & 35 \\
         10 & 17 & 18 \\
         11 & 2 & - \\
\hline
Total & 1359 & 1328 \\
\hline
    \end{tabular}
    \caption{Frequency of the number of questions by DFS3 and DFSHansel for $n=4$ criteria}
    \label{ta:CompaNumbQuestn=4}
\end{table}

\ssk For $n=4$, we observe that DFS3 uses slightly more questions than DFSHansel (31 less on a total of 1328 for DFSHansel). From $n=4$ on, the distribution of the number of questions is more concentrated for DFSHansel. The maximum number of questions asked by DFS3 passes the lower bound (11 instead of 10 for $n=4$). For $n=5$, DFS3 asks up to 22 questions (while the bound is 20). Note that the maximal number of questions asked by DFSHansel is equal to the bound \eqref{eq:lowerBound}. 

\section{Chapter conclusion}\label{sec:concluChap5}
In this chapter, we implemented and tested an algorithm based on Hansel chains. This algorithm guarantees that the lower bound on the number of questions is never exceeded. Our contribution \wrt the existing literature is to place Hansel's algorithm in the framework of Depth-First Search. An in-depth study of Hansel chains was required in order to design an algorithm that generates the Hansel chains as a variant of the depth-first search arborescence.

\ssk Using the DFSHansel questioning algorithm as a simulation tool, we were able to compute the average number of questoins required by DFSHansel up to 6 criteria. Comparing the results with those obtained with DFS3, we may conclude the DFSHansel uses slightly less questions on average than DFS3 for $n=4,5, 6$ criteria and never uses more questions than the lower bound. However, the idea of DFS3 remains relevant because it is a general procedure that can be adapted to the case where additional information is available on the relative importance of the criteria.  



\chapter{Efficient strategy on average }
\label{chap-average}

\section{Introduction}
In order to obtain, from the decision-maker, all the necessary information to specify our model, we are interested in assessing efficient/optimal questioning procedures. Focusing on the elicitation of \Fmin, the set of minimally sufficient coalitions of criteria in a Non Compensatory Sorting (NCS) model, our aim is to establish elicitation algorithms minimizing the number of questions.

\ssk In Chapter~\ref{chapter-Hansel}, we developed an algorithm that minimizes the number of questions to elicit the sufficient coalitions in the worst case. In this chapter we are interested in minimizing the \emph{average} number of questions. Then, we will compare the results obtained with the algorithm DFSHansel to the minimal number of questions on average.

\ssk Discovering the unknown set of sufficient coalitions by asking questions to the DM can be seen as a decision process under uncertainty. One has to decide which question should be asked in order to find as quickly as possible which coalitions are sufficient. The answer to a question eliminates certain \Fmin, which are no longer possible given the current information (\ie the answers to the questions asked sofar). 

\ssk Assuming an a priori uniform probability distribution on the minimal sufficient coalitions (antichains of \Pn, the set of subsets of the set $N$ of the criteria), we try to find a sequence of questions that identifies the right \Fmin{} as quickly as possible on average. The next question depends on the answer to the previous ones. We thus aim at establishing an optimal questioning strategy, \ie a strategy with which the \emph{expected} number of questions is minimal. This can be done, in principle, through representing the problem in a decision tree \citep[see, \eg][]{raiffa1968decision}. 

In this section we compute optimal strategies for the cases of 2, 3 and 4 criteria. Since the number of possible \Fmin{} increases extremely rapidly with the number $n$ of criteria (the Dedekind numbers sequence \citep[][sequence A000372]{OEIS}) it cannot be expected to do that even for $n$ slightly larger than 4. However, knowing the minimal expected number of questions for up to 4 criteria allows to assess the performance of the DFS3 and the DFSHansel algorithms regarding the average number of questions they ask.

\section{Decision tree}\label{sec:decisionTree}
The process of questioning the DM for eliciting the set of sufficient coalitions or the set of minimal sufficient coalitions can be represented  as a decision tree. There are two types of nodes, decision nodes and chance nodes. For instance, in the case of two criteria, the set of possible sets of minimal coalitions, \Fmin, is  $\{\varnothing,\{1\},\{2\},\{1,2\},\{12\},\{\}\}$. 

\begin{center}
\begin{figure}[h!!!]\includegraphics[width=130mm]{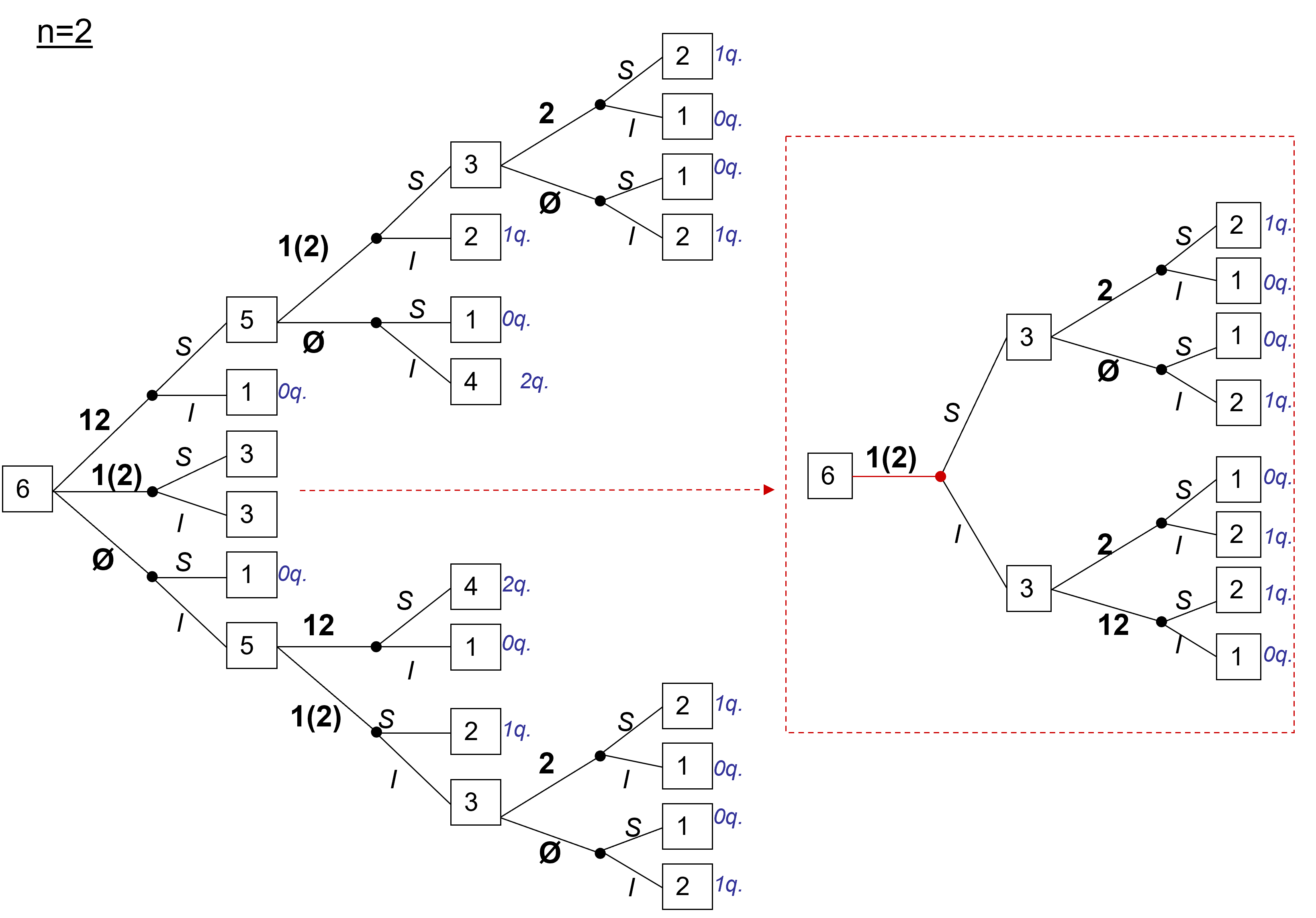}\caption{Decision tree in the case of two criteria. Possible questionning paths. \label{fig:decisionTree2crit}}
\end{figure}
\end{center}

\ssk As a reminder, $\Fmin = \{\varnothing\}$ corresponds to the case in which $\varnothing$ is a sufficient coalition, hence all coalitions are sufficient.  In contrast, $\Fmin= \{\}$ means that not any coalition is sufficient. At each step, a question of the type ``Is coalition $A$ sufficient or insufficient ?'' (expressed in terms of an alternative having the best evaluations on the set of criteria $A$ and the worst on the rest of the criteria) is asked to the DM. This is a decision node. The answer of the DM is either ``Sufficient ($S$)'' or ``Insufficient ($I$)''; this is a node of chance. The whole set of possible questioning processes are represented on Figure~\ref{fig:decisionTree2crit}. Square nodes represent decision nodes (one decides which question is to be asked). Bullet nodes (black filled circles) represent chance nodes (the analyst doesn't know whether the DM will answer $S$ or $I$). 

There are only four possible questions since the possible coalitions are $12$, $1$, $2$ and $\varnothing$ (we drop the brackets, for simplicity). To start, one can ask whether any of them is sufficient. The process starts in the lefthand square node in  Figure~\ref{fig:decisionTree2crit}. Three branches emanate from this node, corresponding to the four possible questions. Since questionning processes starting by asking whether $1$  is sufficient or whether $2$ is sufficient are completely symmetric, we represent them in the same branch. The number (6) in the initial square node represents the number of possible sets of minimal coalitions at this stage.

\ssk For instance, let us explain the upper branch of the tree. The process starts with asking whether $12$ is $S$ or $I$. If the answer is $S$, the number in the second square node is 5, because only one of the six possible \Fmin is eliminated, namely, $\{\}$ (because all other families of sufficient coalitions are contained by $12$). If the answer is $I$, the number in the second square is 1, the only possible \Fmin is $\{\}$. In this case, there is no further question, the process stops there. 

\ssk Let us continue on the path 12S. There are three possible questions. If we ask about $1$ (or $2$) and the answer is $S$, the next square node contains 3 possible \Fmin : $\{1\},\{1,2\},\{\varnothing\} $. If the answer is $I$, the next square node contains 2 possible \Fmin :$\{2\},\{12\}$. One question remains to be asked: is $2$ sufficient ? (not represented). Continuing in the upper branch 12S-1S, we ask whether $2$ is sufficient or whether $\varnothing$ is sufficient. If we ask about $2$, and if the answer is $S$, we arrive to the 4th square node of the upper branch which contains 2 possible \Fmin : $\{\varnothing\},\{1,2\}$. Asking whether $\varnothing$ is sufficient terminates the elicitation (not represented). If the answer is $I$ when asking about 2, there is only one possible \Fmin, which is $\{1\}$.

\ssk The rest of the decision tree can be explained similarly. 

\ssk In particular, we observe a more complex symmetry between upper and lower branches in Figure~\ref{fig:decisionTree2crit}. For example, the number of possible \Fmin in branch 12S-2I (\resp 12S-2S) is equal to the number of possible \Fmin in branch $\varnothing$I-1S (\resp $\varnothing$I-1I). The possible antichains knowing 12S-2I (\resp $\varnothing$I-1S) are 1 or 12 (\resp 1 or $\{1,2\}$). The corresponding sets of sufficient (S) and insufficient coalitions (I) are displayed in Table~\ref{ta:symmetryNeq2}. 

\begin{table}[h!!!]
\begin{center}
\begin{tabular}{ccc|ccc}
\hline
\multicolumn{3}{c|}{  $\varnothing$I-1S}    & \multicolumn{3}{c}{12S-2I}  \\
\hline      
\Fmin  & S & I & \Fmin & S & I \\
\hline 
1  & 1, 12 & $\emptyset$, 2 & 1 & 1, 12 & $\emptyset$, 2 \\
  1, 2 & 1, 2, 12 & $\emptyset$ & 12 & 12 & $\emptyset$, 1, 2 \\
\hline
\end{tabular}  
\end{center}
    \caption{Example of symmetry between upper and lower branches}
    \label{ta:symmetryNeq2}
\end{table}

The correspondance is as follows. For each antichain in the upper branch, the sets of sufficient (\resp insufficient) coalitions are the complements of the insufficient (\resp sufficient) coalitions corresponding to an antichain in the lower branch, and conversely. For instance, for the antichain 12 in the 
upper branch 12S-2I, the set of sufficient coalitions is just 12, hence the insufficient coalition in the corresponding antichain $\{1, 2\}$ in the branch $\varnothing$I-1S is the complement of 12, \ie $\varnothing$. 

Let us be more formal. Let $A_1 \R_1 - A_2 \R_2 - \ldots - A_k \R_k$ denote a path in the graph, with $A_i$, a coalition, and $\R_i$, the status of this coalition, \ie $\R_i = S$ or $I$, for $i=1, \ldots, k$. Given a path, the status $\R $  of all other coalitions is either determined or remains unknown. In particular, if coalition $A$ contains $A_i$ and $\R_i=S$, then $\R(A) = S$. If coalition $A$ is contained in $A_i$ and $\R_i = I$, then $\R(A) = I$. Otherwise, $\R(A)$ is undetermined, which we note $\R(A) = ?$. Thus, for a given path, we have a status function defined for all $A \in 2^n$ and taking one out of the three values $S, I $ or $?$.  When considering a path, we assume that the path is not contradictory, \ie assigning the status $\R_i$ to $A_i$ given $A_1\R_1 - A_2 \R_2 - \ldots - A_{i-1}\R_{i-1}$ is a possible assignment. Let $\overline{A\,}$ be the complement of $A$ in $2^n$ and $\overline{\R\,}(A)$ be the opposite assignment, \ie  $\overline{\R\,}(A) =S$ (\resp $I, ?$) iff ${\R}(A)=I$ (\resp $S,?$).

\begin{proposition}[Upper-Lower symmetry]\label{prop:symmetryUpperLower}
The status of $\R(A)$ of $A \in 2^n$ given the path $A_1 \R_1 - \ldots - A_k\R_k$ is $S$ (\resp $I, ?$) iff the status of $\overline{A\,}$ given $\overline{A_1} \, \overline{\R_1} - \ldots - \overline{A_k} \, \overline{\R_k}$ is $I$ (\resp $S, ?$).   
\end{proposition}

\bpr
Suppose $\R(A)=S$ given $A_1 \R_1 - \ldots - A_k\R_k$. This implies that there is $i\in \{1,\ldots, k\}$ with $A \supseteq A_i$ and $\R(A_i)= \R_i = S$. Therefore, $\overline{A\,} \subseteq \overline{A_i}$ and the assignment of $\overline{A_i}$ given $\overline{A_1} \, \overline{\R_1} - \ldots - \overline{A_k} \, \overline{\R_k}$ is $\overline{\R_i}=I$.  Therefore, the status of $\overline{A\,}$ is $I$. 
One proves similarly that $\R(A)=I$ given $A_1 \R_1 - \ldots - A_k\R_k$ implies that the status of $\overline{A\,}$ is $S$. Consequently, the status of $A$ is $?$ iff the status of $\overline{A\,}$ is $?$.
\epr

In the sequel we refer to this sort of symmetry as to \emph{Upper-Lower symmetry}. We shall also say that the paths $A_1 \R_1 - A_2 \R_2 - \ldots - A_k \R_k$ and  $\overline{A_1} \, \overline{\R_1} - \ldots - \overline{A_k} \, \overline{\R_k}$ are \emph{dual} of one another. Such paths will also be referred to as \emph{dual patterns}. 

\section{Expected number of questions}\label{sec:expectedNumQuest}
If we assume a probability distribution on the set of possible \Fmin, we can use the decision tree to compute the average number of questions. This computation corresponds to using decision trees for deciding under uncertainty (see \eg \citet{raiffa1968decision}). 

\begin{center}
\begin{figure}[h!]\includegraphics[width=130mm]{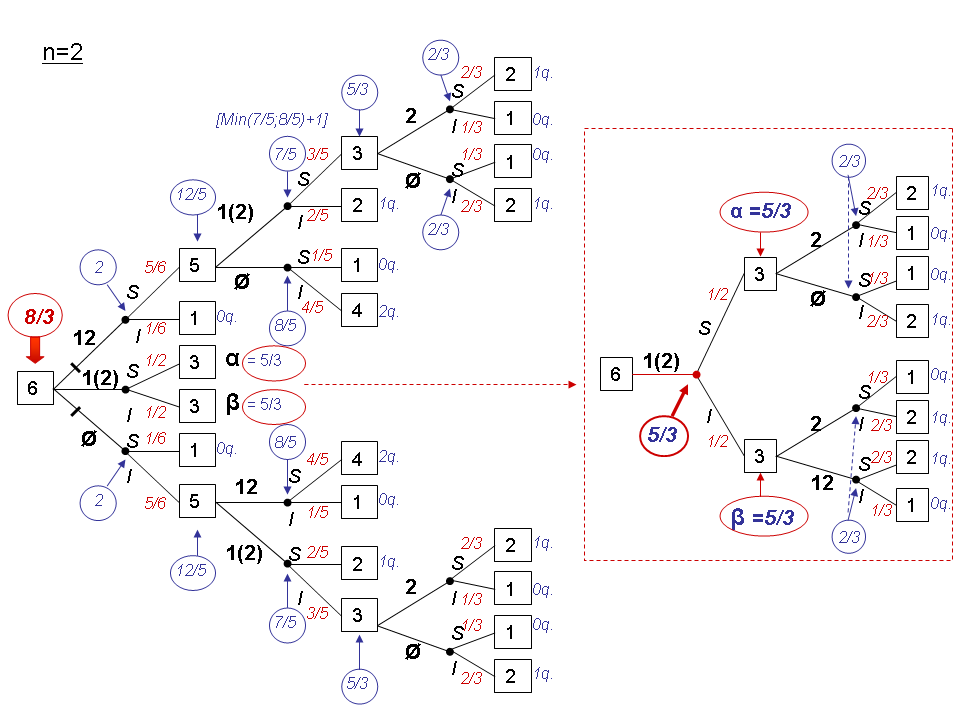}\caption{Evaluated decision tree in the case of two criteria. The probability distribution on the sets of minimal coalitions is uniform. \label{fig:decisionTree2critEvaluated}}
\end{figure}
\end{center}

Assume the uniform distribution on the set of all possible \Fmin. In the case of two criteria, each of the 6 possible \Fmin{} is supposed to have probability $1/6$. This allows us to evaluate the tree in Figure~\ref{fig:decisionTree2crit}, yielding the evaluated tree represented in Figure~\ref{fig:decisionTree2critEvaluated}. 

Our goal is to determine the optimal path(s) in the decision tree, \ie the path(s) on which the expected number of questions is minimal. To do that, we consider each path of the graph in Figure~\ref{fig:decisionTree2crit} from right to left.

\paragraph{Chance nodes.} We compute the probabilities of each answer in the chance nodes. These are determined by the number of remaining possible \Fmin{} depending on the given answer. Knowing the expected number of remaining questions after an $S$ or an $I$answer, we compute the average of these numbers weighted by the probability of each answer. This yields the expected number of questions in the chance node. For instance, at the rightmost chance node in the upper branch of the graph (12S-1S-2), we know that 12 and 1 are sufficient. Therefore the remaining possible \Fmin{} are: $\{1\}, \{1,2\}, \{\varnothing\}$. If 2 is sufficient then, we are left with two possible \Fmin{}, which are $\{1,2\}, \{\varnothing\}$. The probability of this answer is thus 2/3 and 1 more question is to be asked (Is $\varnothing$ sufficient ?). If $2$ is insufficient, we are left with one possible \Fmin{}, which is $\{1\}$. The probability of this answer is thus 1/3 and there is no other question to be asked. The expected number of questions in chance node 2 is thus $2/3*1+1/3*0 = 2/3$. The probabilities of each answer in chance nodes are written on the edge $S$ and $I$ from the node. The chance node itself is evaluated by the average number of remaining questions in all paths to the right of the node. 

\paragraph{Decision nodes.} Assuming that we know the average number of questions in all chance nodes following immediately a given decision node, we know which question yields the minimal expected number of questions. We thus choose this question, we discard the others and we evaluate the decision node by the minimal expected number of questions of the following chance nodes plus one (since we have to ask the question realising the minimal expected value). For instance, consider the decision node at the end of the subpath (12S1S) in Figure~\ref{fig:decisionTree2crit}. There are two possible questions : $2$ or $\varnothing$. If we ask $2$, we compute the expected number of questions, which is 2/3. If we ask $\varnothing$, the expected number of questions is also 2/3 (the situation is symmetric). So, here, any of the two branches is optimal. The decision node receives the evaluation $2/3 + 1 =5/3$. 

\ssk Let us take another example. Consider the decision node at the end of the subpath (12S). The possible questions are $1$, $2$ or $\varnothing$. We know that asking in $1$ is equivalent to asking in $2$. Chance nodes $1$ and $2$ lead to an expected number of questions equal to 7/5 (computed by the usual rule for chance nodes, see above). The expected number of questions in chance node $\varnothing$ is 8/5. Hence we choose the minimal value and ask $1$ (or $2$) and cut the path going through edge $\varnothing$. The evaluation of decision node (12S) is $7/5 +1 = 12/5$.  

\ssk Doing this for all nodes, working backwards from right to left, yields the evaluated decision tree in Figure~\ref{fig:decisionTree2critEvaluated}.

\begin{center}
\begin{figure}[h!]\includegraphics[width=130mm]{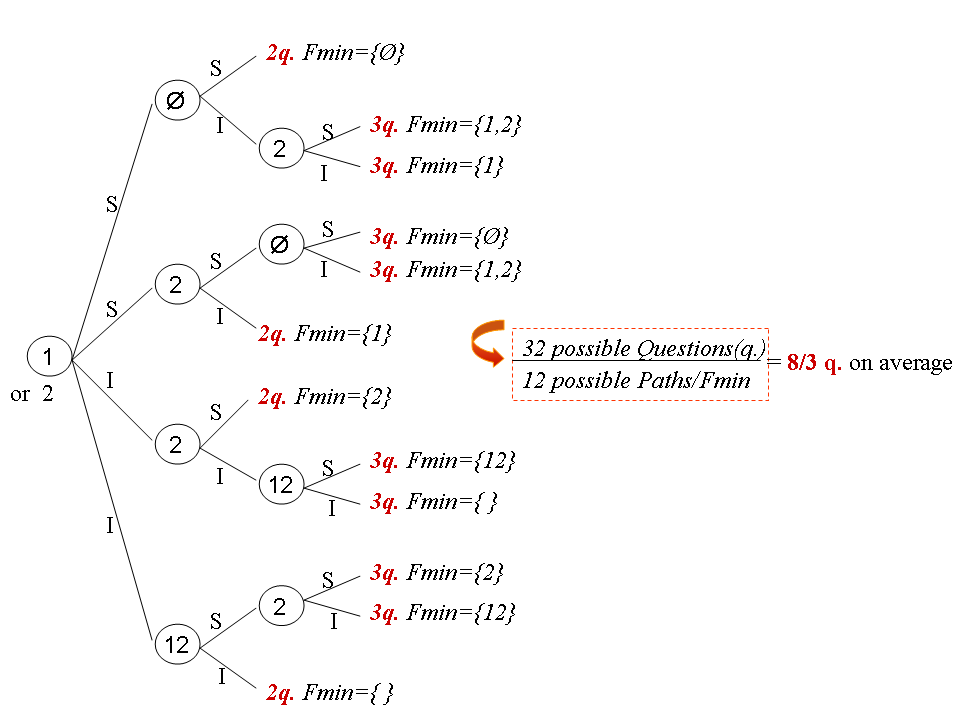}\caption{Optimal questioning policies for $n=2$\label{fig:optimalPaths2crit}}
\end{figure}
\end{center}

\ssk The resulting optimal decision paths for the case of 2 criteria are represented in Figure~\ref{fig:optimalPaths2crit}. 
The figure actually represents two optimal questioning policies. The initial question is in $1$ (or equivalently, in $2$). Whatever the answer (S or I), one may then ask in $2$. Alternatively, one may ask in $\varnothing$ if $1$ is S or in $12$ if $1$ is I. The final questions follow the pattern shown in the figure. Whatever the policy followed, the total number of questions used for identifying the 6 \Fmin{} is 16. Hence the average number of questions in each policy is $16/6 = 8/3$. Note that the number of questions needed to reach a given \Fmin{} may depend on the chosen optimal policy. For instance, $\Fmin = \{1\}$ is reached after 2 questions with one policy and 3 questions with the other.    

\section{The case of three criteria}\label{sec:Expected3crit}
\begin{center}
\begin{figure}[h!]\includegraphics[width=155mm]{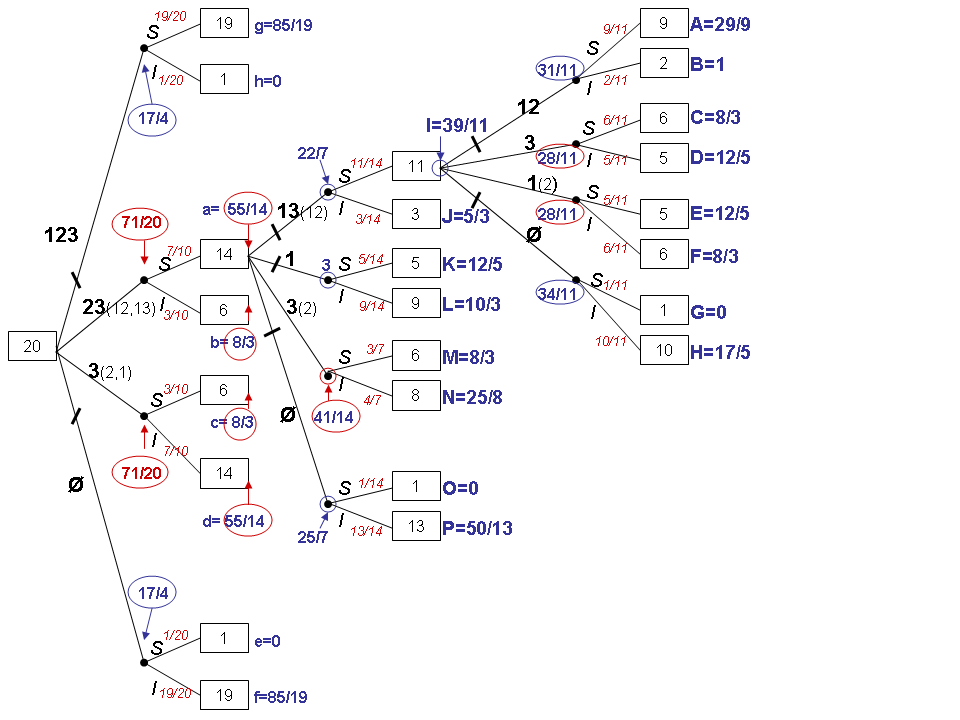}\caption{The case of  three criteria}
\label{fig:decisionTree3critEvaluated}
\end{figure}
\end{center}

Figure~\ref{fig:decisionTree3critEvaluated} shows a general view of the case of the three criteria. As the graph is very big, it was necessary to present it in several sub-graphs illustrated in appendix to this chapter (Section~\ref{app:figures3n}). 

For $n=3$ criteria, there are 8 possible questions since the possible coalitions are $123$, $23$, $13$, $12$, $3$, $2$, $1$ and $\varnothing$. To start, one can ask whether any of them is sufficient. The process starts in the lefthand square node in Figure~\ref{fig:decisionTree3critEvaluated}. Four branches emanate from this node, corresponding to the eight possible questions. Since questionning processes starting by asking whether $23$ is sufficient or whether $12$ or $13$ is sufficient are completely symmetric, we represent them in the same branch. A similar symmetry exists between $3$, $2$ and $1$. Thus, they are represented by the same branch.  
 
An upper-lower symmetry (see Proposition~\ref{prop:symmetryUpperLower}) also exists between branches 23 and 1 (or 13 and 2 or 12 and 3). Thanks to Proposition~\ref{prop:symmetryUpperLower}, we only need to develop the upper branches (or the lower branches but not both). The optimal strategy in the branch starting with 1 will be symmetric to the optimal strategy in the branch starting with 23.  The number (20) in the initial square node represents the number of possible sets of minimal coalitions at this stage. In the case of three criteria, the set of possible sets of minimal coalitions \Fmin, is 20. 
 
The set of possible \Fmin{} is listed as follows:
\begin{center}
\small{$\{\varnothing\},\{1\},\{2\},\{3\},\{1,2\},\{1,3\},\{2,3\},\{1,2,3\},\{23\},\{13\},\{12\}, \{23,13\},$ \\ $\{23,12\},\{13,12\},\{23,13,12\},\{23,1\},\{2,13\},\{3,12\},\{123\},\{\}\}$.}
\end{center}

As illustrated in Figure~\ref{fig:decisionTree3critEvaluated}, we have to choose one of the 4 possible branches for starting. Our main idea for establishing the whole graph for $n=3$, was to find some symmetries and to check the values of the sub-graphs showing patterns already evaluated in the case of $n=2$.

\ssk For instance, let us explain the branch starting with asking whether $23$ is $S$ or $I$. If the answer is $S$, the number in the second square node is 14, because the set of possible \Fmin{} knowing that $23$ is $S$ are:
\begin{center}   
\vspace{-1mm}
$\{\varnothing,\{2\},\{3\},\{1,2\},\{1,3\},\{2,3\},\{1,2,3\},$\\
$\{23\},\{23,13\},\{23,12\},\{23,13,12\},\{23,1\},\{2,13\},\{3,12\}.$
\vspace{-1mm}
\end{center}
Indeed, this answer eliminates the following 6 possible \Fmin{}: \\
$\{1\},\{13\},$ $\{12\},\{13,12\},\{123\},\{\}$. These 6 sets of minimal coalitions are the possible \Fmin{} if $23$ is $I$. Let's continue on the branch (23S). We have $13$, $12$, $1$, $2$, $3$ and $\varnothing$ as remaining questions. $13$ and $12$ can be represented in the same branch because of their symmetry. Similarly, $2$ and $3$ are symmetric and are represented by the same branch. We need a branch $1$ because it has a special position \wrt $23$. 

\ssk Let us continue by (23S-13S). The number in the third square node is 11, because the possible \Fmin{} knowing that $23$ and $13$ are $S$ are: 
\begin{center}
$\varnothing,\{3\},\{1,2\},\{1,3\},\{2,3\},\{1,2,3\},\{23,13\},\{23,13,12\},$\\ $\{23,1\},\{2,13\},\{3,12\}$.
\end{center}

If we take the branch (23S-13I), the number in the third square node is 3, because the set of possible \Fmin{} knowing that $23$ is $S$ and $13$ is $I$ are:
$\{\{2\},\{23\},\{23,12\}\}$. 

\ssk Consider the decision node in (23S-13S). There are 5 remaining possible questions: $12$, $3$, $1$, $2$, $\varnothing$. Here $1$ and $2$ can be represented by the same branch because they are symmetric \wrt the current knowledge, \ie $23$ and $13$ are $S$. We explore now the consequences of the question: ``is $3$ sufficient ? ''.   

\ssk Let us take the branch (23S-13S-3S), the number in the fourth square node is 6, because the set of possible sets of minimal coalitions, knowing that $23$,$13$ and $3$ are $S$ are: $\{\varnothing,$$\{3\},\{1,3\},\{2,3\},\{1,2,3\},\{3,12\}\}$. The other branch (23S-13S-3I), contains 5 possible sets of minimal coalitions which are: 
$\{1,2\},\{23,13\},\{23,13,12\},$ $\{23,1\},\{2,13\}$.

\begin{center}
\begin{figure}[h!]\includegraphics[width=135mm]{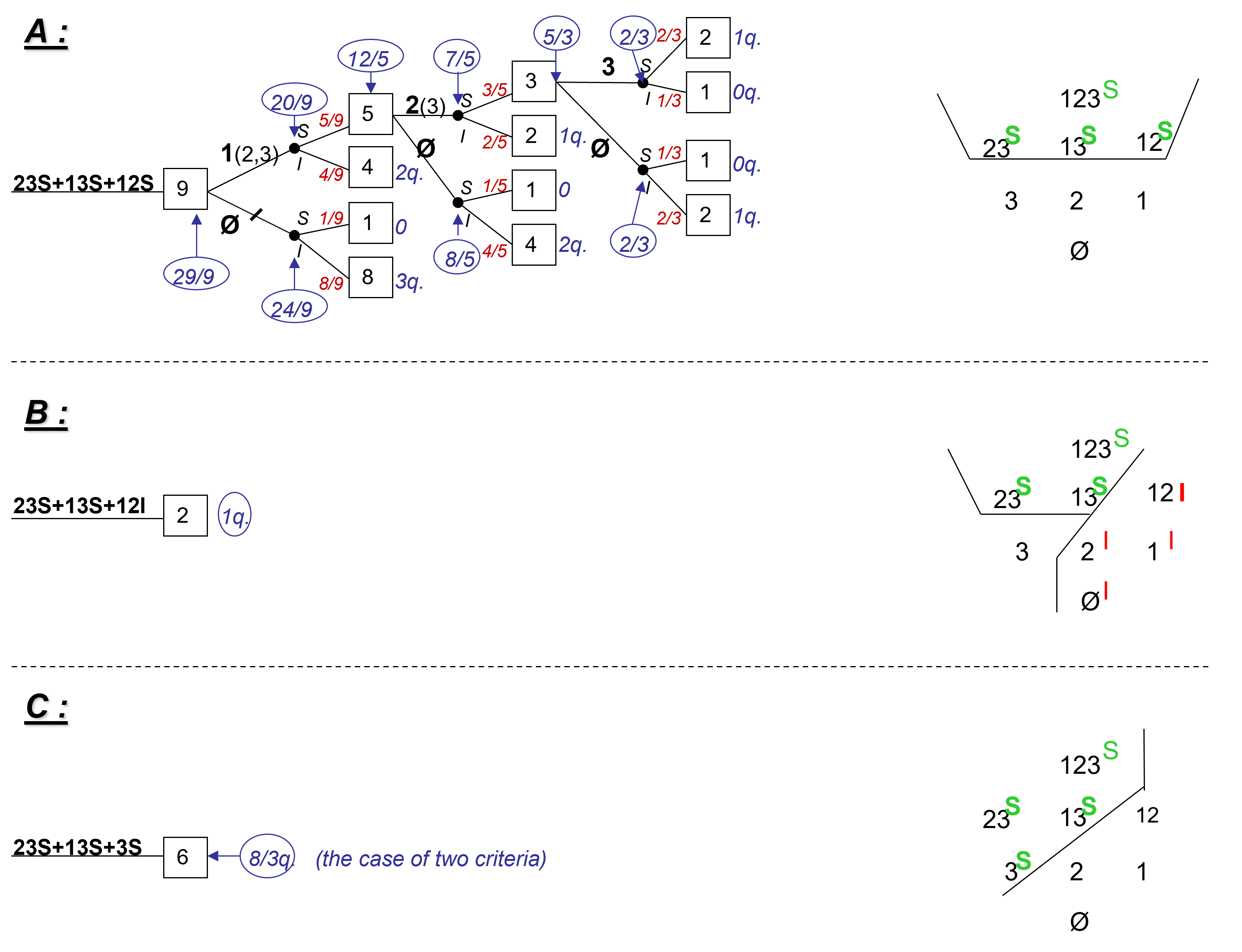}\caption{n=3, ABC}
\label{fig:decisionTree3critABC}
\end{figure}
\end{center}

\ssk Shown in  Figure~\ref{fig:decisionTree3critEvaluated}, following the path (23S-13S-3S), we arrive to  situation C represented in the bottom part of Figure~\ref{fig:decisionTree3critABC}. The graph to the right shows all coalitions separated by a polygonal line. Above the line, the coalitions known as sufficient. Below, the status of the coalitions is unknown. We see, that the situation of the unknown coalitions is exactly the case of $n=2$ criteria. We can thus use its expected number of questions of 8/3 for our computations. Hence $C=8/3$ in Figure~\ref{fig:decisionTree3critEvaluated}.

\ssk The two other cases A and B illustrated in Figure~\ref{fig:decisionTree3critABC} correspond with the situations at the end of the branches (23S-13S-12S) or (23S-13S-12I). The case B is simple. If $12$ is $I$, so are $1$, $2$ and $\varnothing$. The only unknown coalition status is that of $3$. This is illustrated in the graph on the right of the middle part of Figure~\ref{fig:decisionTree3critABC}. In the left part above the separating line are coalitions known as sufficient. To the right of the separating line are the coalitions known as insufficient. On the lower left part, the only coalition $3$ whose status is unknown.
Therefore there is only one question to be asked in this case: is $3$ sufficient or insufficient ? Hence $B=1$ in Figure~\ref{fig:decisionTree3critABC}.

\ssk In the decision node (23S-13S-12S), the situation is called A. It is represented in the upper part of Figure~\ref{fig:decisionTree3critABC}. We know that all coalitions involving two or three criteria are sufficient. We don't know whether any singleton or $\varnothing$ is sufficient (see right part of the Figure). Therefore, the whole questioning procedure is analyzed in the left part of the figure, leading to an expected number of questions of 29/9. Hence $A=29/9$.   

\ssk All other branches are evaluated in a similar way. The detailed assessments are described in figures labelled D to P and b to h. These are represented in Figure~\ref{fig:decisionTree3critDEF}, Figure~\ref{fig:decisionTree3critGHJ}, Figure~\ref{fig:decisionTree3critKLM}, Figure~\ref{fig:decisionTree3critNOP}, Figure~\ref{fig:decisionTree3critbcd}, Figure~\ref{fig:decisionTree3critRS}, Figure~\ref{fig:decisionTree3critefgh}.

\ssk Consider the decision node at the end of the path (23S-13S) in Figure~\ref{fig:decisionTree3critEvaluated}. The possible questions are 12, 3, 1(or 2) and $\varnothing$. We know that 1 and 2 are equivalent. Chance node 3 leads to an expected number of questions equal to 28/11. The value of C can be seen in Figure~\ref{fig:decisionTree3critABC} and is taken directly from the case of two criteria. The value of D in Figure~\ref{fig:decisionTree3critDEF} is computed by the usual rule for chance nodes. The expected number of questions in chance node 3 comes from the computation of $8/3*6/11+12/5*5/11=28/11$. This computation is made for all other chance nodes. Hence we choose the minimal value and ask 3 (or 1 or 2 on the other branch) and cut the path going through $12$ and $\varnothing$. The evaluation of decision node (23S-13S) is $28/11 +1 = 39/11$.  The value of the chance node (23S-13) is $22/7$ and computed by $39/11*11/14+5/3*3/14$. Thus, the expected number of questions at the decision node 23S is the minimum value coming from the 4 edges. The value computed for (23S-13) is not the smallest one and will not be choosed. As seen in Figure~\ref{fig:decisionTree3critEvaluated}, for this stage, the minimal value of expected questions is for the edge 3. On the other hand, we obtain also the same value of $28/11$ at the chance node coming from the path 23S-13S-1(or 2). Thus, taking the branch 3 or 1(or 2) are the same for this step. The average number of questions at the decision node I: 23S-13S is $28/11+1=39/11$. The decision node J, represented by the path 23S-13I, has $5/3$ questions. As seen in Figure~\ref{fig:decisionTree3critGHJ}, the case of J shows the same pattern of $n=2$ with 1I and $\varnothing$I. Thus, we can use the value computed in the case of two criteria, following the paths $\varnothing$I-1I or 1I. Both of these paths are conducting to the same average number of questions $5/3$ which is also been used as the value of the sub-graph J.
By backtracking from right to left of the graph, we arrive to the chance node (23S-13). At this node, the average number of questions is calculated as: $39/11*11/14+5/3*3/14=22/7$.    

\ssk Doing these calculations for all nodes, working backwards from right to left, we obtain all expected values of the decision tree shown in Figure~\ref{fig:decisionTree3critEvaluated}. As the average number of questions computed for the chance node (23S-3) is $41/14$, we choose this path which needs the minimum number of questions compared to the paths (23S-13), (23S-1) and (23S-$\varnothing$). Consequently, at the decision node 23S, the average number of questions is $41/14+1=55/14=a$. For (23I) we are in the sub-graph b illustrated in Figure~\ref{fig:decisionTree3critbcd}. We observe that the pattern of b is the same of the pattern of a whole graph for two criteria. Thus, we can use the average number of questions for $n=2$ and $b=8/3$. 

\ssk At the chance node of (23), we compute $a*7/10+b*3/10=71/20$ as the average number of questions. This is also the case of (12) and (13), due to the symmetry. Figure~\ref{fig:decisionTree3critEvaluated} shows that, the chance node of (3) has $71/20$ questions. This case is the same for (2) and (1). The two other starting branches (123) and ($\varnothing$) have $17/4$ questions. Thus, the minimal average number of questions is $71/20+1=91/20$ by choosing the branches (23) (or (12),(13)) and (3) (or (2),(1)).

\begin{center}
\begin{figure}[h!]\includegraphics[width=130mm]{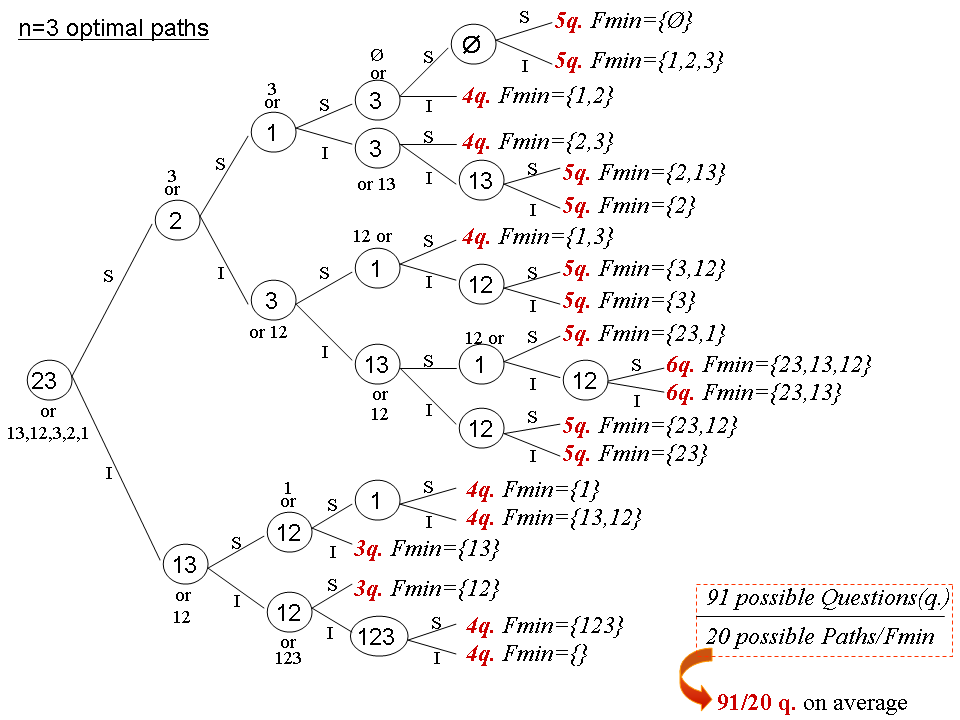}\caption{Optimal questioning paths  for $n=3$}
\label{fig:optimalPaths3crit}
\end{figure}
\end{center}

\ssk For the case of three criteria, an example of an optimal path of the decision tree is 23S-2S-1S-3S-$\varnothing$S. Figure~\ref{fig:optimalPaths3crit} shows an optimal questioning policy found by evaluating the tree. The minimum number of questions is $91/20  \simeq 4.5$ on average. It can be interesting to see what happens in an ``extreme'' case in which all coalitions are sufficient, \ie for \Fmin$=\{\varnothing\}$. The suggested path, with the optimal policy on average, is 23S-2S-1S-3S-$\varnothing$S; it involves 5 questions. Ideally, if we suspect in advance that \Fmin{} is $\varnothing$, we have only to ask one question to confirm this.  In case \Fmin $=\{23,13,12\}$, the optimal policy on average requires 6 questions. If we suspect that this is \Fmin{} and we want to confirm it, the number of questions is also 6. 

The symmetries between some elements of the graphs exist also for $n=3$. These symmetries come from the structure of the graph and the distribution among the supersets and subsets of the considered elements.
Several patterns of $n=2$ can be identified in C, D, E, J, K, L, M, N, P. Due to the similarities between these patterns of $n=2$, found in the case of $n=3$, we used the values of $n=2$ previously calculated. Consider, for instance, Case C in Figure~\ref{fig:decisionTree3critABC}. 
Knowing 23S-13S-3S, we propagate this information and get 123S. Four sets remain to be labelled: $\varnothing$, 1, 2, 12. We know how to question optimally in this case since we previously solved the case $n=2$. The expected number of questions is $8/3$. We refer to such a shortcut in the computations as \emph{dimensional reduction}. 

\ssk Another, less straightforward example occurs in Case~J in Figure~\ref{fig:decisionTree3critGHJ}. We know 23S and 13I. We can label 123S and 1I, 3I, and $\varnothing$I. It remains to label 2 and 12, which is a subcase of $n=2$. We find the same pattern in Figure~\ref{fig:decisionTree2critEvaluated} as pattern $\beta$. The expected number of questions for $\beta$ is $5/3$. Note that this case can also be viewed as the case $n=1$ with 2 (\resp 12) playing the role of $\varnothing$ (\resp 1). 

\ssk A further example is Case~K in Figure~\ref{fig:decisionTree3critKLM}. The unlabeled sets are $\varnothing$, 2, 3. This corresponds to a subcase of $n=2$ while knowing 12S. The role of 3 is played by 1. This case is in Figure~\ref{fig:decisionTree3critEvaluated} at node 12S with expected number of questions $12/5$.                                         

\section{Simplifying tools}\label{sec:simplTools}

Computing the average number of questions in case $n \geq 4$ becomes more difficult. We have the three following  tools at our disposal to simplify the computations. We illustrate them on the case $n=4$.
\begin{itemize}
    \item \textbf{Symmetry}: from the first node, we need only 5 branches corresponding to the 5 possible sizes of the coalitions: 0, 1, 2, 3, 4. For instance, for size 2 (\ie coalitions consisting of a pair of criteria), starting by asking whether 12 is sufficient is equivalent to asking whether 34 (or any coalition of two criteria) is sufficient. 
    \item \textbf{Upper-lower symmetry}: due to Proposition~\ref{prop:symmetryUpperLower}, it is sufficient to develop the branches corresponding to coalitions of size 4, 3 and 2. The branches related to coalitions of size 0 and 1 are obtained by complementation from the cases of size 4 and 3 respectively. 
    \item \textbf{Dimension reduction}: as we did in the case $n=3$, we may try to identify patterns already calculated in the smaller dimension cases ($n= 2, 3$).   
\end{itemize}

This is not sufficient however. We introduce an additional ingredient, which is \emph{Branch \& Bound} \citep[][Chap.~7]{Wolsey1998}.

\paragraph{Informational Bound and Branch \& Bound strategy}

For each branch, in a chance node, we know the number, say $M$, of possible antichains given the answers to the questions raised so far. We may thus give a lower bound on the number of questions needed to determine the right antichain in the set of the $M$ possible ones. The number of questions is at least $\log_2(M)$. This is the \emph{informational bound} (denoted IB in the sequel). This bound is attained when each question-answer divides the number of possible antichains by 2. Attaining this bound is often not possible in our case, since there may not exist question that divide exactly in two the set of possible antichains. 

Assume a decision node, from which two branches are issued. If one branch has been completely explored and the minimal number of questions is known, and in the other branch, the \emph{informational bound} is greater than the minimal number of questions in the other branch, then the second branch need not be explored. 


\section{The case of four criteria}\label{sec:4crit}
We will show how the average number of questions can be calculated in case of $4$ criteria. Since, for $n=4$, the decision tree is too big, we have to use the simplifications mentioned in the previous section. Because of the figure complexity, it will not be possible to represent the whole graph of $n= 4$ and the computations will be shown on sub-parts of the decision tree. 

\subsection{Tree exploration policy}\label{ssec:n4ExplorationPolicy}
In order to simplify the computations, all possible symmetries or patterns already known should be identified on the tree. The \emph{informational bound} (IB) is also an indicator which can be useful to eliminate some branches of the tree and has to be computed. Several patterns of $n=2$ and $n=3$ can be identified in $n=4$. For these patterns of $n=2$ and $n=3$, found in the case of $n=4$, we use the values of the expected number of questions  previously computed for $n=2$ and $n=3$. 

\begin{figure}[h!!!]
\begin{center}
\includegraphics[width=115mm]{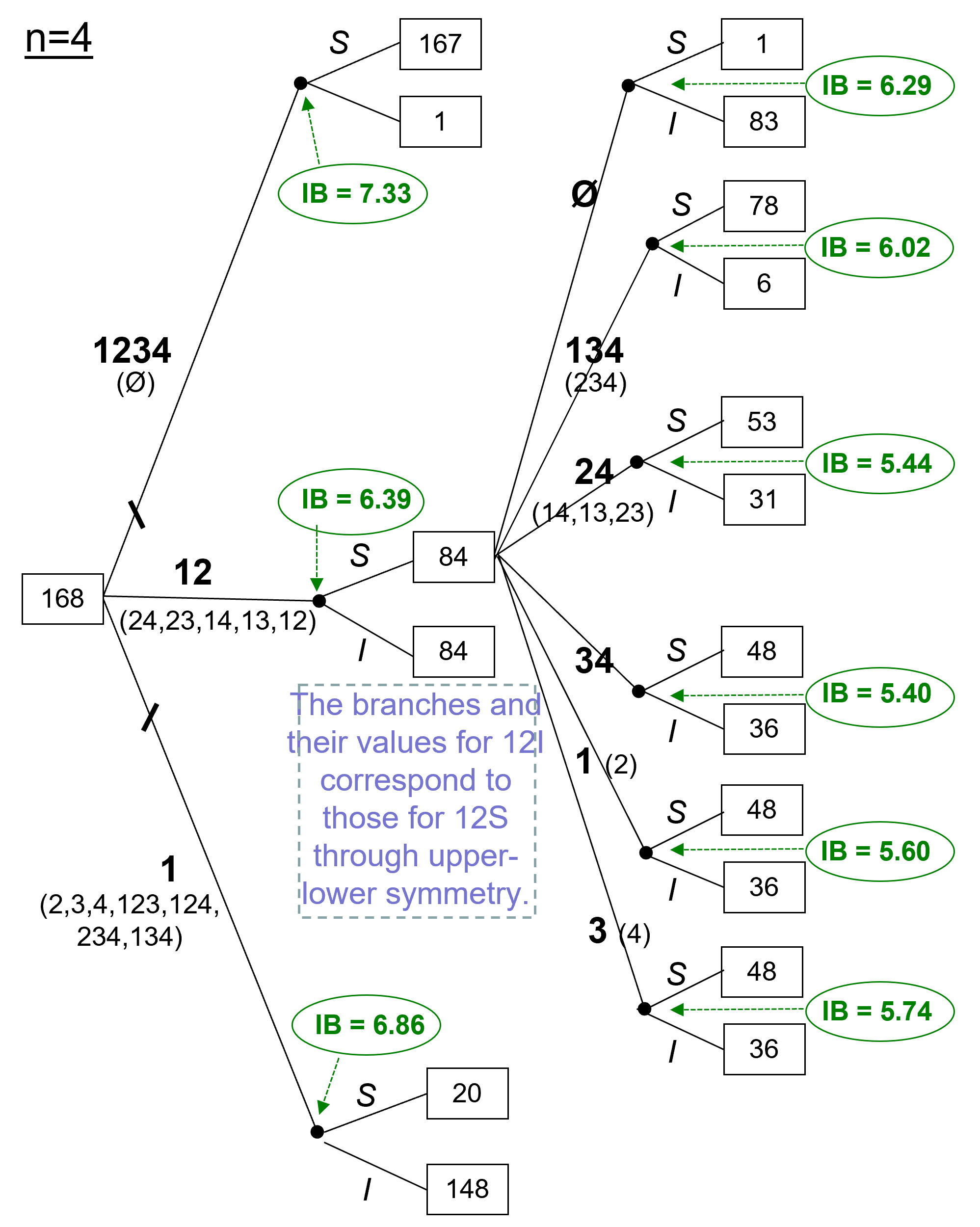}
\vspace{0.5cm}
\caption{Informational bounds for n=4}
\label{fig:Expectedn=4_1}
\end{center}
\end{figure}

\ssk In the case of $4$ criteria, the number of possible \Fmin (antichains) is $168$. As explained before, this is the value of the increasing sequence known as ``Dedekind Numbers'' for $n=4$. The decision tree starts with a decision node containing this number of antichains. Using the property of symmetry, we  observe that $\varnothing$ and $1234$ are upper-lower symmetric and can be represented by a single branch (see Figure~\ref{fig:Expectedn=4_1}). On the other hand, $12, 24, 23, 14$ and $13$ are all symmetric and can be represented by the second branch. Finally, $1, 2, 3, 4, 123, 124, 234$ and $134$ also are symmetric or upper-lower symmetric and represented by a third branch. Before deciding which branch to start within, we compute the \emph{informational bound} (IB) for every possible branch. Our policy will be to explore the tree starting with the most promising branch, \ie that with the lowest IB. The most promising decision  consists to ask about $12$ (or $24, 23, 14, 13, 12$). If the answer of $12$ is S, the number of remaining \Fmin is $84$. This is also the value of possible \Fmin when $12$ is I. This branch has the most homogenous repartition of the set of possible \Fmin. Thus, its \emph{informational bound} is minimal. Indeed, it is 
$$
\frac{1}{168} (84(\log_2 84)+ 84(\log_2 84)) \simeq 6.86.
$$
Hence the average number of questions needed in each case is at least $6.86$. 

\ssk The path of decision and chance nodes leading to a node is used as an identifier of this node. If \textit{path} ends in a chance node, we add S (\resp I) for denoting the path obtained if the answer in chance node ending \textit{path} is S (\resp I). Let $\Fmin(\textit{path})$, be the set of $\Fmin$ remaining at the end of \textit{path}.   
Let $b(\textit{path})$ denote the IB for the node at the end of \textit{path}. For instance $b(12)$ is the IB computed above ($\approx 6.86$). In general,

\begin{equation}\label{eq:}
\begin{split}
b(\textit{path})= \frac{1}{\# \Fmin(\textit{path})} [\# \Fmin(\text{\textit{path}+S}) \log_2(\# \Fmin(\text{\textit{path}+S})) \\ 
 + \ \# \Fmin(\text{\textit{path}+I}) \log_2(\# \Fmin(\text{\textit{path}+I}))].    
\end{split}    
\end{equation}

The branch $\varnothing$ (and $1234$) has the least homogenous distribution of antichains (167, 1) and the value of the IB is $7.34$. The second branch corresponds to the coalition  $12$ (or $24, 23, 14, 13, 12$). It has the smallest IB, i.e.,  $6.39$. The third branch representing $1, 2, 3, 4, 123,$ $ 124, 234$ and $134$ as the coalitions of criteria, has a distribution of (148, 20) and it corresponds to an IB of $6.86$. 

We thus select the most promising branch and continue with $12$.  
We observe that $12$ (or $24, 23, 14, 13, 12$) is situated in the middle of the graph and the graph is symmetric with respect to the level of $12$. Thus, we obtain the same number of questions on average when $12S$ or $12I$ and it is sufficient to evaluate just one of these paths because of upper-lower symmetry (we shall turn back to this later). When we arrive to the decision node of $12S$, we have 6 possible branches which are: $\varnothing$, $134$ (or $234$), $24$ (or $13, 14, 23$), $34$, $1$ (or $2$) and $3$ (or $4$). The distributions of the antichains for each possible path are respectively (1, 83), (78, 6),  (53, 31), (48, 36), (20, 64) and (14, 70). The corresponding IBs are $6.29, 6.02, 5.44, 5.40, 5.60$ and $5.74$. Figure~\ref{fig:Expectedn=4_1} shows the possible paths and their IBs.

\begin{figure}[h!!!]
\begin{center}
\includegraphics[width=16.5cm]{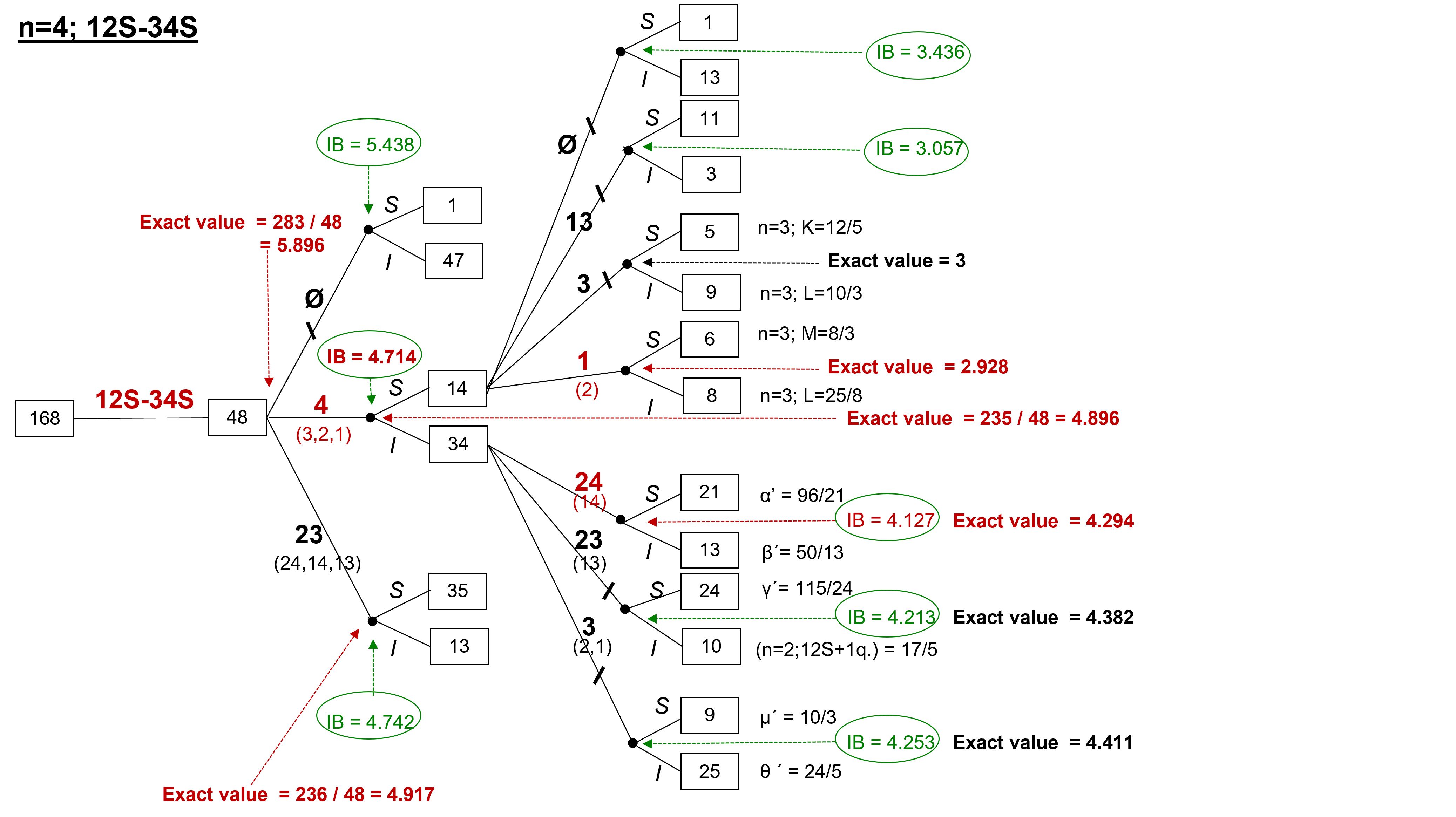}
\vspace{0.5cm}
\caption{Expected number of questions for n=4 for 12S-34S}
\label{fig:Expectedn=4-12S34S}
\end{center}
\end{figure}

The IB values are lower (\resp higher) when the distribution of the antichains is more balanced (\resp less balanced). Node $34$ ending path 12S-34 has (48, 36) as a distribution of the antichains and the least IB of $5.40$. If we consider the path 12S-34S at the chance node 12S-34, $48$ possible antichains remain. The decision node situated at 12S-34S, suggests three different branches that are $23$ (or $24$, $14$,$13$), $\varnothing$ and $4$ (or $3,2,1$). The path 12S-34I contains $36$ possible antichains, but we will continue to explain the computations via the branch 12S-34S. The IBs calculated by the distribution of the antichains are respectively $4.74, 5.43$ and $4.71$ for $23$ (or $24$, $14$,$13$), $\varnothing$ and $4$ (or $3,2,1$), see Figure~\ref{fig:Expectedn=4-12S34S}. Let us continue on the branch having the smallest IB, thus we come to the chance node of 12S-34S-4. If $4$ is S, $14$ antichains remain. If not, $34$ antichains remain. When $4$ is S, four possible branches arise. These are $\varnothing$, $13$ (or $23$), $3$ and $1$ (or $2$). Chance nodes IBs for each of them are respectively $3.43, 3.05, 2.86$ and $2.92$. Figure~\ref{fig:Expectedn=4-12S34S} also shows the exact values that have been computed in the usual way (yet using the simplifying tools described in previous section).  

\subsection{Tree evaluation}\label{ssec:n4TreeEval}
Figure~\ref{fig:Expectedn=4_2} shows all exact values computed for all possible paths. The branches with worse IB scores or worse exact values are pruned. Detailed computations for our starting path 12S-34S are illustrated in Figure~\ref{fig:Expectedn=4-12S34S}. 

\ssk At the decision node 12S-34S-4S (see Figure~\ref{fig:Expectedn=4-12S34S}), we identify some patterns that we have already computed. The branches that have the most homogenous distribution of antichains are $1$ (or $2$) and $3$ with (6-8) and (5-9) antichains, respectively. For 12S-34S-4S-3S we identify the pattern of Case~K that we found for $n=3$ (see  Figure~\ref{fig:decisionTree3critKLM} in Appendix~\ref{app:figures3n}). Thus, we can directly use the value $12/5$ already known as an exact expected value. For 12S-34S-4S-3I we identify the pattern of the Case~L that we found also in Figure~\ref{fig:decisionTree3critKLM}. The exact value of L is $10/3$. Thus, we compute the exact expected value at the chance node  12S-34S-4S-3 as $12/5*5/14+10/3*9/14=3$. In the same way, we identify two already known patterns at 12S-34S-4S-1S and  12S-34S-4S-1I which are respectively Case~M in Figure~\ref{fig:decisionTree3critKLM} and Case~N  in Figure~\ref{fig:decisionTree3critNOP}. By using the exact expected values of M$=8/3$ and N$=25/8$ we obtain $8/3*6/14+25/8*8/14= 41/14 \approx 2.92$ at the chance node of 12S-34S-4S-1. As the IBs of $\varnothing$ and $13$ (or $23$) are respectively $3.46$ and $3.05$, we can directly eliminate these branches without computing the exact values. We can also eliminate the path 12S-34S-4S-3 whose exact expected value is 3 (for this path, the IB was not low enough to eliminate the branch without computing the exact expected value). 

\ssk In node 12S-34S-4I (see Figure~\ref{fig:Expectedn=4-12S34S}), there are 34 antichains and three possible branches, which are 24 (or 14), 23 or (13) and 3 (or 2 or 1). New patterns $\alpha', \beta', \gamma', \mu', \theta' $ are identified; they are shown in Figures~\ref{fig:Expectedn=4-alphaprime}, \ref{fig:ExpectedN=4_betaprime}, \ref{fig:Expectedn=4-gammaprime} and \ref{fig:Expectedn=4-muthetaprime} in Appendix~\ref{app:figures4n}. The exact computed values of the branches are displayed in Figure~\ref{fig:Expectedn=4-12S34S}. The best branch is 12S-34S-4I-24, with an exact expected value of $96/21*31/34 + 50/13*13/34 = 146/34 \approx 4.294$. 

\ssk The expected value in node 12S-34S-4 is thus $(41/14 + 1)*14/48 + (146/34 + 1)*34/48  = 235/48 \approx 4.896$. In node 12S-34S, We may eliminate the branch $\varnothing$ since its IB (=5.438) is larger than the exact value in node $4$. Branch 23 cannot be eliminated by IB. We computed its exact expected value (not detailed here), which is $236/48$, slightly more than for branch 4. The optimal question in 12S-34S is thus $4$. Therefore the exact expected value in node 12S-34S is $235/48 + 1 = 283/48 \approx 5.896$.

\begin{figure}[h!!!]
\begin{center}
\includegraphics[width=11cm]{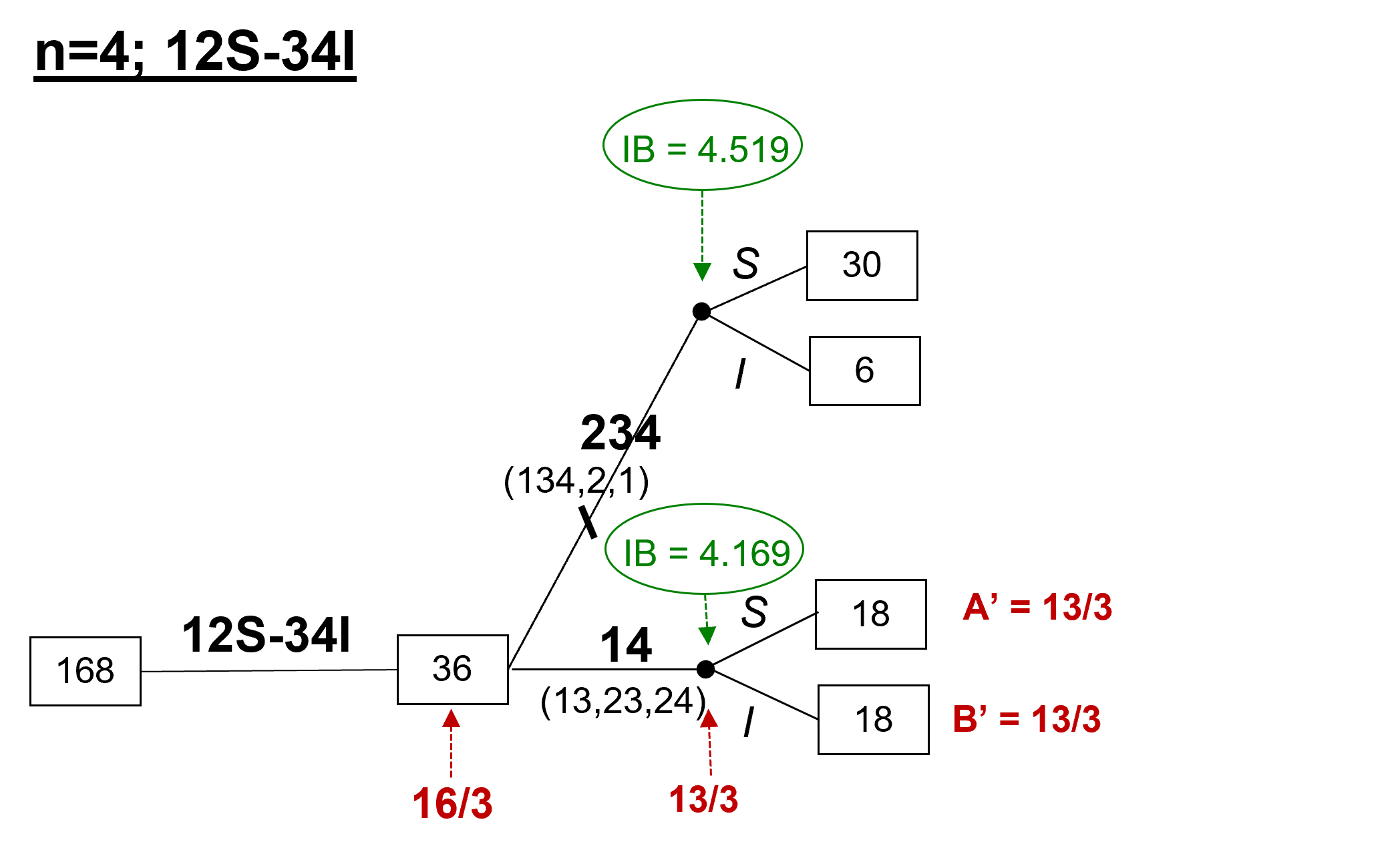}
\vspace{0.5cm}
\caption{Expected values for n=4, 12S-34I}
\label{fig:Expectedn=4_12S34I}
\end{center}
\end{figure}

\begin{figure}[h!!!]
\begin{center}
\includegraphics[width=135mm]{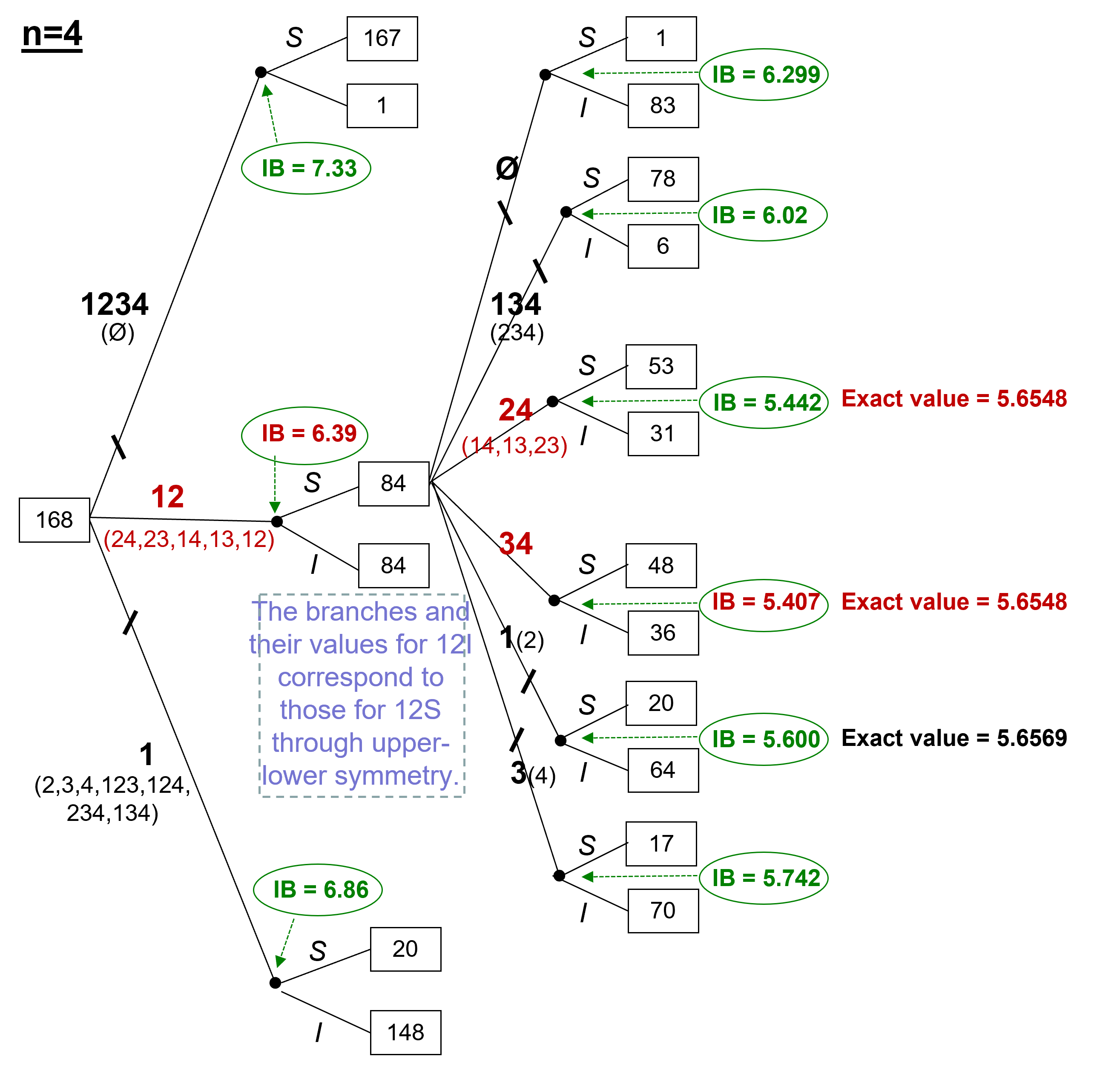}
\vspace{0.5cm}
\caption{Expected Number of Questions for n=4}
\label{fig:Expectedn=4_2}
\end{center}
\end{figure}

\ssk On the other hand, the path 12S-34I, illustrated in Figure~\ref{fig:Expectedn=4_12S34I}, contains $36$ antichains. At this decision node, we have 2 branches to explore. 
The first one is $234$ (or $134$) with a distribution (30-6) of the antichains. Asking whether $1$ (or $2$) is sufficient gives rise to the same distribution of antichains (\ie (6-30)) because there is an upper-lower symmetry between the case 12S-34I-234I (\resp -234S, -134I, -134S) and the case 34I-12S-1S (\resp -1I, -2S, 2I).   We may thus treat these cases in the same branch as $234$ (or $134$). 

\ssk The second branch is $14$ (or $13, 23$ and $24$) with a distribution (18-18) of the antichains. The IB of the chance node situated at 12S-34I-234 of the (30-6) distribution is equal to $4.52$. The branch of $14$ (or $13, 23$ and $24$) has a well-balanced distribution of antichains (18-18); the number of the antichains is divided by $2$ for $14$S or $14$I. Thus, we start to compute its exact expected value. 

\ssk The decision node A' of the path 12S-34I-14S, contains $18$ antichains and the possible branches are $234$ (or $2$), $1$ (or $13$) and $24$ (or $23$). Figure~\ref{fig:Expectedn=4_12S34IA'} in Appendix~\ref{app:figures4n} shows the exploration of this case. The IBs of $234$ (or $2$) and $1$ (or $13$) are respectively $3.51$ and $3.25$. As $24$ (or $23$) has the lower IB ($=3.16$) because of its balanced distribution, we compute its exact expected value.
The pattern identified in the path of 12S-34I-14S-24S is the same as Case~L of Figure~\ref{fig:decisionTree3critKLM}. Thus, we can use the exact expected value, which is $10/3$, found in the Case~L for $n=3$. 

\ssk The path 12S-34I-14S-24I has also the pattern of Case~L illustrated in  Figure~\ref{fig:decisionTree3critKLM}. Its exact expected value is again $10/3$. Thus, the exact expected value at the chance node 12S-34I-14S-24 is $10/3=3.33$ and the branch $234$ (or $2$) is eliminated because its IB is $3.51$, higher than $3.33$. We don't need to compute the rest of this branch. But the IB of $1$ (or $13$) is $3.25$, thus we have to compute its exact expected value. 

\ssk At the chance node 12S-34I-14S-1, the distribution of the antichains is (6-12). For 12S-34I-14S-1S, we identify the pattern of $n=2$ and its expected value is $8/3$. For 12S-34I-14S-1I, we observe a pattern of $n=2$ plus an independent element. Thus, the expected value is $8/3+1=11/3$. The average expected value at the chance node 12S-34I-14S-1 is $10/3$. Consequently, the paths that start by asking 24, 23, 1 or 13 are equivalent. The expected value for the decision node 12S-34I-14S is $10/3+1=13/3$.

\ssk The chance node B' of the 12S-34I-14I is illustrated in Figure~\ref{fig:Expectedn=4_12S34IB'}. 
We observe that, there is an upper-lower symmetry between A' and B'. Actually 12S-34I-14I, \ie A' is upper-lower symmetric with 34I-12S-23S, which is equivalent (by symmetry) with 12S-34I-14S, \ie B'. Therefore, the expected value is the same, \ie $13/3$. Consequently, the expected value of the decision node of 12S-34I is $13/3+1=16/3$. 

\ssk The expected value at the chance node 12S-34 is $283/48*48/84+16/3*36/84=475/84 \approx 5.6548$. Using this exact value, we can eliminate all branches coming from 12S except 24 (or 13, 14, 23) and 1 (or 2) (see Figure~\ref{fig:Expectedn=4_2}). We evaluated branch 12S-24, which turns out to be equal to branch 12S-34. Branch 12S-1 is slightly worse (exact value = 5.6569)  Thus, the expected value at the decision node 12S is $6.6548$.

\ssk For 12S-34I (12I-34S is the same by symmetry)-14S (or I)-24 (or 23,1,13), we ask $6.6548$ questions and it is another optimum path. 

\ssk On the other hand, 12I gives the same number of questions as 12S, because of the upper lower symmetry.

\ssk Thus, the optimal number of questions on average at the chance node 12 is $6.6548$. By adding the first question 12, we obtain $1286/168 = 7.6548$ as the optimal number of questions on average for $n=4$.

\subsection{An optimal questioning tree}\label{ssec:n4OptimalTree}
\begin{center}
\begin{figure}[h!]\includegraphics[width=125mm]{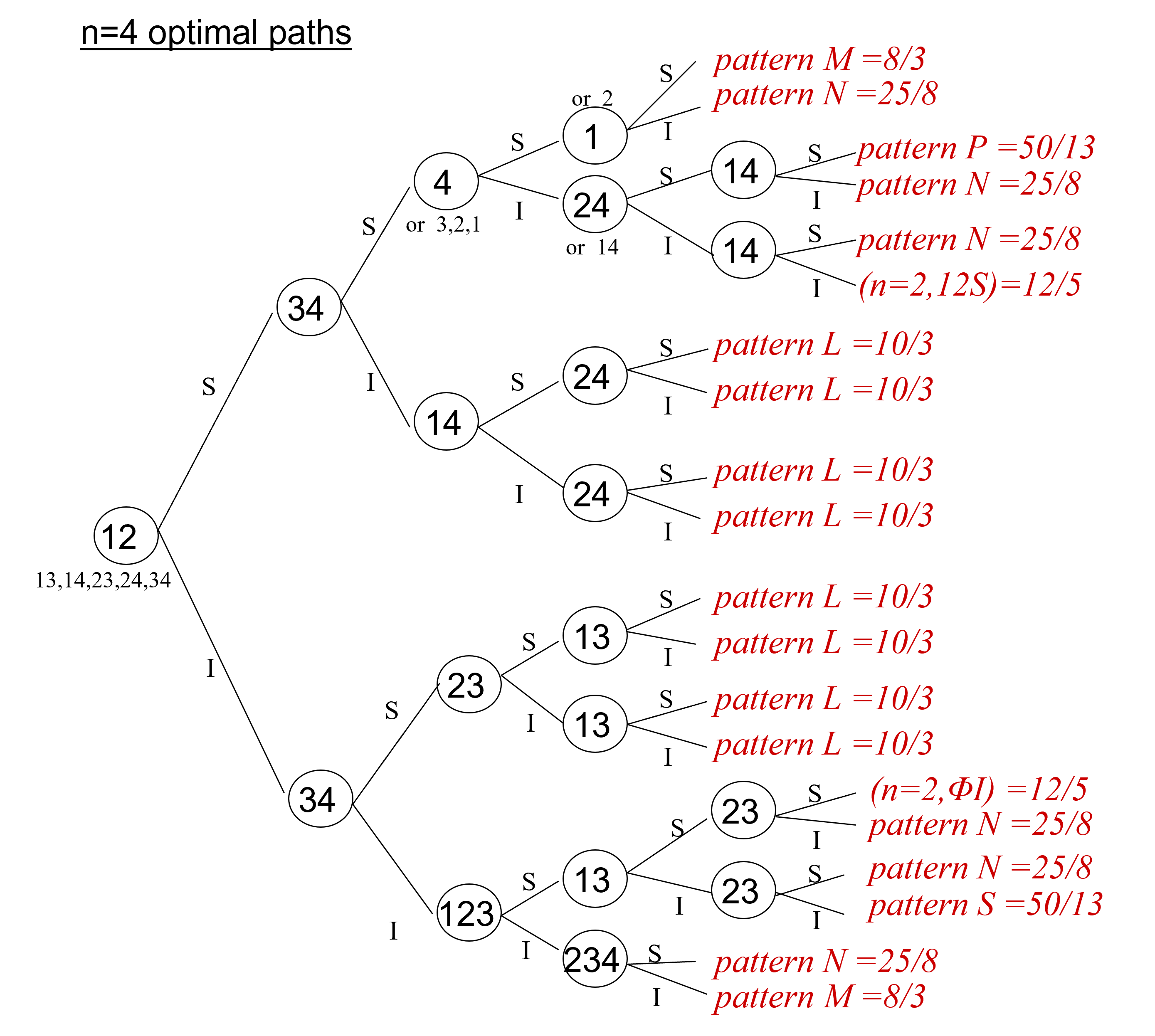}\caption{Optimal questioning tree for $n=4$}
\label{fig:optimalPaths4crit}
\end{figure}
\end{center}

From Figure~\ref{fig:Expectedn=4_2}, we can derive an optimal questioning tree (note there are several such optimal trees). An example of an optimal questioning tree is represented in Figure~\ref{fig:optimalPaths4crit}. 

\ssk The paths are represented up to 4 or 5 questions.  12S-34I-14S-24S is one of the optimal paths that needs 4 questions to ask before finding the pattern L. Of course the followed path depends on the answers to each question. In the present case whatever the answer to the first question (``Is 12 sufficient?''), the second question is asked in 34. Then, if we arrived in 34 with 12S and the answer in 34 is S, we ask in 4, else we ask in 14 and so on.

\ssk The tree is not detailed up to the determination of a particular \Fmin. With path 12S-34I-14S-24S, we end up with pattern L. Thus, the rest of the questioning process, leading to a particular \Fmin, is obtained by following an optimal questioning path of one of the following patterns from $n=3$ or $n=2$: L, M, N, P, S, ($n=2$, with 12S) or ($n=2$, with $\varnothing$I). Patterns L, M et N are self upper-lower symmetric or self-dual. When we take the duals of each one, we obtain again the same pattern. For instance, the dual of the pattern L ($n=3$, with 23S, 1I) is ($n=3$, with 1I, 23S). Pattern P is not self-dual. Its dual is pattern S and conversely. The same is true for patterns ($n=2$, with 12S) and ($n=2$, with $\varnothing$I). The optimal questioning trees of dual patterns are dual. Hence the optimal number of questions is the same. The optimal questioning trees for patterns L, M, N, P and ($n=3$, with 23S, 1I) are displayed in Appendix~\ref{app:n4} (Figures~\ref{fig:ExpectedL_nbq}, \ref{fig:ExpectedM_nbq}, \ref{fig:ExpectedN_nbq}, \ref{fig:ExpectedP_nbq} and \ref{fig:ExpectedN2_12S_nbq}).
  

\ssk The optimal questioning tree is not unique. Another example is shown in Figure~\ref{fig:anotherOptPath4crit} in Appendix~\ref{app:otherOptimalN4}. It starts with 12S-24 (or 12I-13). The branch 12I-13 obtains by symmetric complementation of the upper branch 12S-24. The operation is less straightforward than for the tree in Figure~\ref{fig:optimalPaths4crit}. In the latter, complementing the first two questions in the branch 12S-34 yields 34I-12, which are entirely symmetric. This is not true for the tree starting with 12S-24, which yields 34I-13. We first create the lower branch of the tree by taking the complementary coalitions at each question, then we relabel the criteria in order that the lower branch starts with 12. The relabeling used to obtain the lower branch in Figure~\ref{fig:anotherOptPath4crit} is $1 \leftrightarrow 3$ and $4 \leftrightarrow 2$. 

\ssk Following this questioning tree also results in an average number of questions of $1286/168$. 

\section{Comparison of questioning algorithms }\label{sec:compaExpectedNumberQ} 

\ssk The optimal expected number of questions has been computed manually for $n= 2, 3, 4$. Doing this for $n\geq 5$ is impossible by hand.  

\begin{figure}[h!!!]
\begin{center}
\includegraphics[width=13cm]{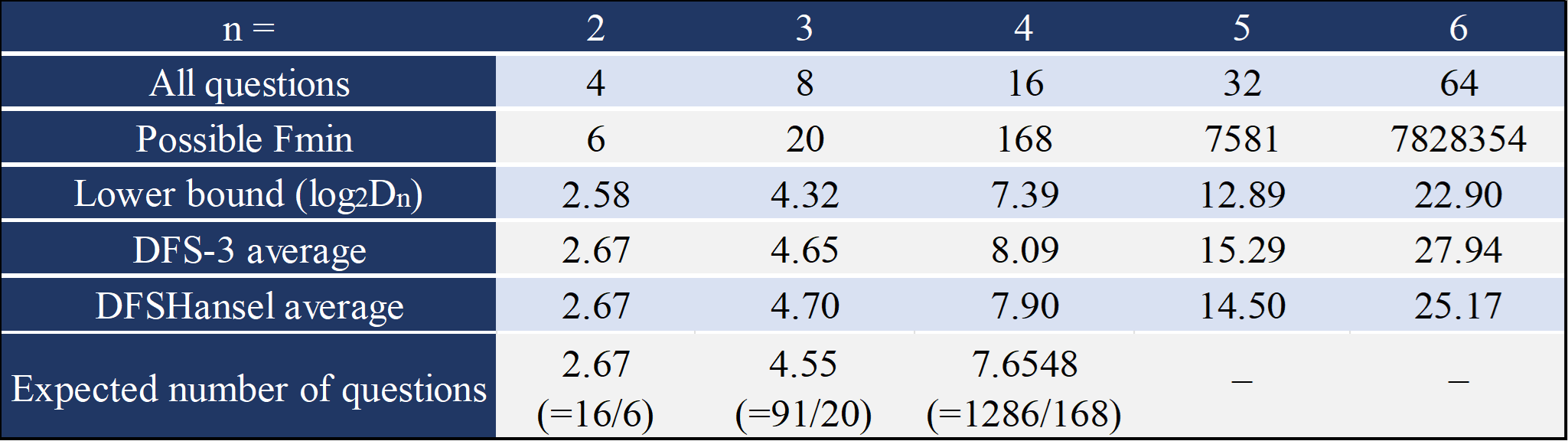}
\vspace{0.5cm}
\caption{Comparison of Expected Number of Questions}
\label{fig:comparisonExpected}
\end{center}
\end{figure}

\ssk Figure~\ref{fig:comparisonExpected} shows the values of the optimal expected number of questions for $n=2,3,4$ compared to the expected number of questions obtained by simulation using the questioning algorithms DFS3 and DFSHansel. 

\ssk The average expected number of questions for $n=4$ is $7.6548$. This value is found as $8.09$ with DFS3 and $7.90$ with DFSHansel. We observe that, both of the suggested algorithms results, DFS-3 and DFSHansel, are not far from the optimal expected number of questions. The gap between DFSHansel and an optimal questioning policy on average is less than a quarter of a question for $n=4$.

\ssk Using Figure~\ref{fig:optimalPaths4crit} complemented by the optimal questioning trees in appendix \ref{app:n4}, enables to compute the distribution of the number of questions for the optimal questioning tree. The results are given in Table~\ref{ta:distribNQuestn4Average}.
\begin{table}[h!!!]
\renewcommand{\arraystretch}{1}
    \centering
    \resizebox{\columnwidth}{!}{%
    \begin{tabular}{lcccccccr}
    \hline
      Patterns (p.)  &  & L & M & N & N & P or S & n=2, &  \\
      &&&&&&& 12S or $\varnothing$I & \\
     \hline 
     \# \Fmin in p. & & 9 & 6 & 8 & 8 & 13 & 5 & \\
     \# q. before p. & & 4 & 4 & 4 & 5 & 5 & 5  &\\
     \hline
     & \# q. &&&&&&& Distr. \# q.\\
     \hline
     & 6 & 1 & 2 & 1 &&&& 14\\
     & 7 & 4 & 4 & 5 & 1 & & 3 & 60\\
     & 8 & 4 &  & 2 & 5 & 4 & 2 & 68\\
     & 9 & & & & 2 & 7 & & 22 \\
     & 10 & & & & & 2 & & 4\\
     \hline
     \# q. for 1 p. & & 66 & 40 & 57 & 65 & 115 &  37  & \\
     \# p. & & 8 & 2 & 2 & 4 & 2 & 2 & \\     
     \# q. for \# p. & & 528 & 80 & 114 & 260  & 230 & 74 & =1286\\
     \hline
    \end{tabular}
    }
    \caption{Distribution of number of questions (\# q.) for an optimal  questioning tree for $n=4$}
    \label{ta:distribNQuestn4Average}
\end{table}

The first line in  Table~\ref{ta:distribNQuestn4Average} lists the patterns on the right of Figure~\ref{fig:optimalPaths4crit}. The total number of each of these patterns is in the second line. The third line displays the number of questions asked before coming to the concerned pattern. Note that pattern N is special:  for two (\resp four) occurences of pattern N, the number of questions for reaching the pattern is 4 (\resp 5). For each of the observed numbers of questions, from 6 to 10,  lines 5 to 9 display the frequency of this number of questions for each \Fmin belonging to each pattern. For pattern L, for instance, 30 questions are needed to determine the 9 \Fmin belonging to the pattern. Before reaching the pattern, 4 questions are required for each of the \Fmin in the pattern. In total, $66 = 30 + 9 \times 4$ questions are needed to determine the 9 \Fmin in that pattern. In line 10, we see the total number of questions asked in each pattern. Line 12 is obtained by multiplying the line 10 and the line 11. The sum of the numbers in line 12 is the total number of questions in the optimal questioning tree for all possible \Fmin (\ie 1286). The distribution of the number of questions, from 6 to 10, can be seen in the rightmost column. 

\ssk Table~\ref{tab:compaHanselAverage} shows the distributions of the number of  questions using the optimal questioning tree and DFSHansel (see Chapter~\ref{chapter-Hansel}, Section~\ref{ssec:DFSHanselOnAverage}) for $n=4$. The range of the number of questions is narrower with the optimal tree. The maximal number of  questions (=10) is used only 4 times with the optimal tree while it is used 18 times with DFSHansel.  

\begin{table}[h!!!]
\renewcommand{\arraystretch}{1}
    \centering
    \begin{tabular}{ccc}
        \# q. &  DFSHansel & Optimal tree \\
        \hline
         5 & 4 & \\
         6 & 16 & 14 \\
         7 & 43 & 60 \\
         8 & 52 & 68 \\
         9 & 35 & 22 \\
         10 & 18 & 4 \\
         \hline
         Total & 1328 & 1268 \\
         \hline       
    \end{tabular}
    \caption{Comparing the distribution of the number of questions for DFSHansel and an optimal questioning tree}
    \label{tab:compaHanselAverage}
\end{table}

\paragraph{Question 1 }  For $n=4$, the optimal questioning tree is also optimal in the worst case (as is Hansel's algorithm). Is it the case for all $n$?

\ssk Note that the distribution of the number of questions is exactly the same in both optimal questioning trees represented in Figures~\ref{fig:optimalPaths4crit} and \ref{fig:anotherOptPath4crit}. 

\paragraph{Question 2 } Is the distribution of the number of questions identical in all optimal questioning trees?
 
\section{A heuristic algorithm}\label{sec:heuristic} 
An important observation made during the selection of optimal paths with the computation of the expected number of questions, was that the branches that make a ``homogeneous'' distribution of \emph{possible} $\mathcal{F}_{min}$, are chosen as optimal paths. ``Homogenous distribution'' means the most possible balanced distribution of \emph{possible} $\mathcal{F}_{min}$, \ie dividing the set of \emph{possible} $\mathcal{F}_{min}$ in two subsets as equally as possible. A second observation was that the optimal paths usually ask questions for coalitions that are in the middle of the graph (in terms of size), and the difference between the numbers of unlabeled subsets and supersets of the coalition, is minimal. Instead of trying to design an algorithm that implements an optimal policy on average (which is a difficult task), a simple heuristic, easy to use, may be a reasonable alternative. Such an heuristic is especially helpful for higher values of $n$, when computation by hand becomes impossible. The functioning of the heuristic is as follows:
\begin{itemize}
   \item Construct the graph of coalitions for a given $n$
   \item For each coalition count the number $\sigma^+$ (\resp $\sigma^-$) of unlabeled supersets (\resp subsets) of the coalition;
   \item Find the coalition(s) for which the $\min(\sigma^+, \sigma^-)$ is maximal and, among these those for which $\max(\sigma^+, \sigma^-)$ is maximal (if several coalitions fulfill these conditions, choose one arbitrarily);
   \item Ask a question about this coalition, propagate the obtained label and iterate the process until all coalitions are labeled.  
 \end{itemize}

\begin{center}
\begin{figure}[h!]\includegraphics[width=130mm]{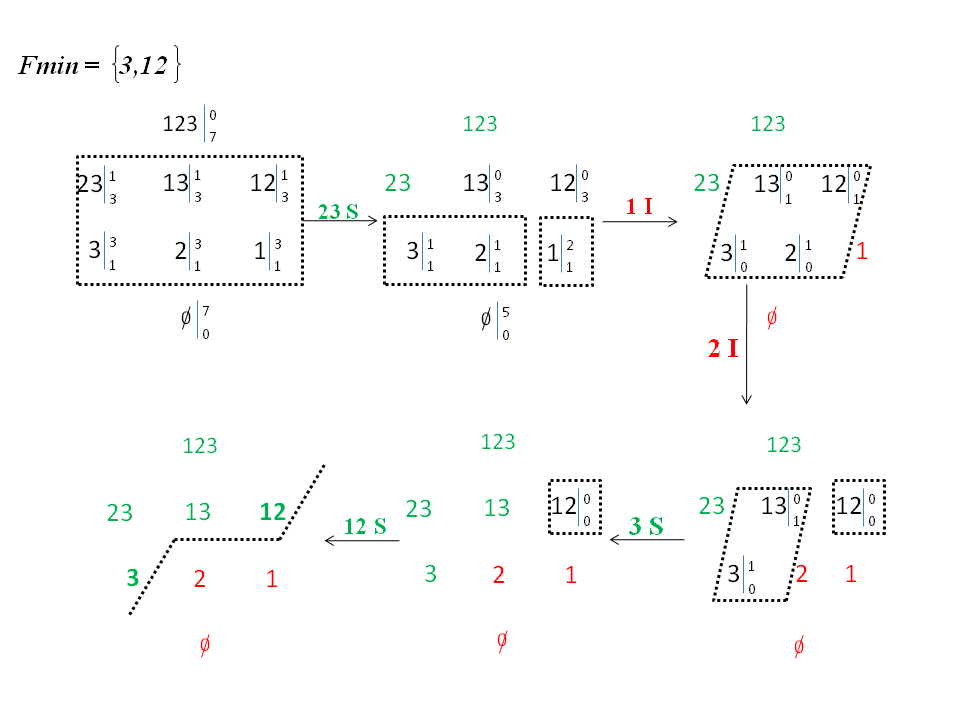}\caption{An alternative path of the suggested heuristic \label{fig:heuristicPathEx2}}
\end{figure}
\end{center}

\begin{center}
\begin{figure}[h!]\includegraphics[width=135mm]{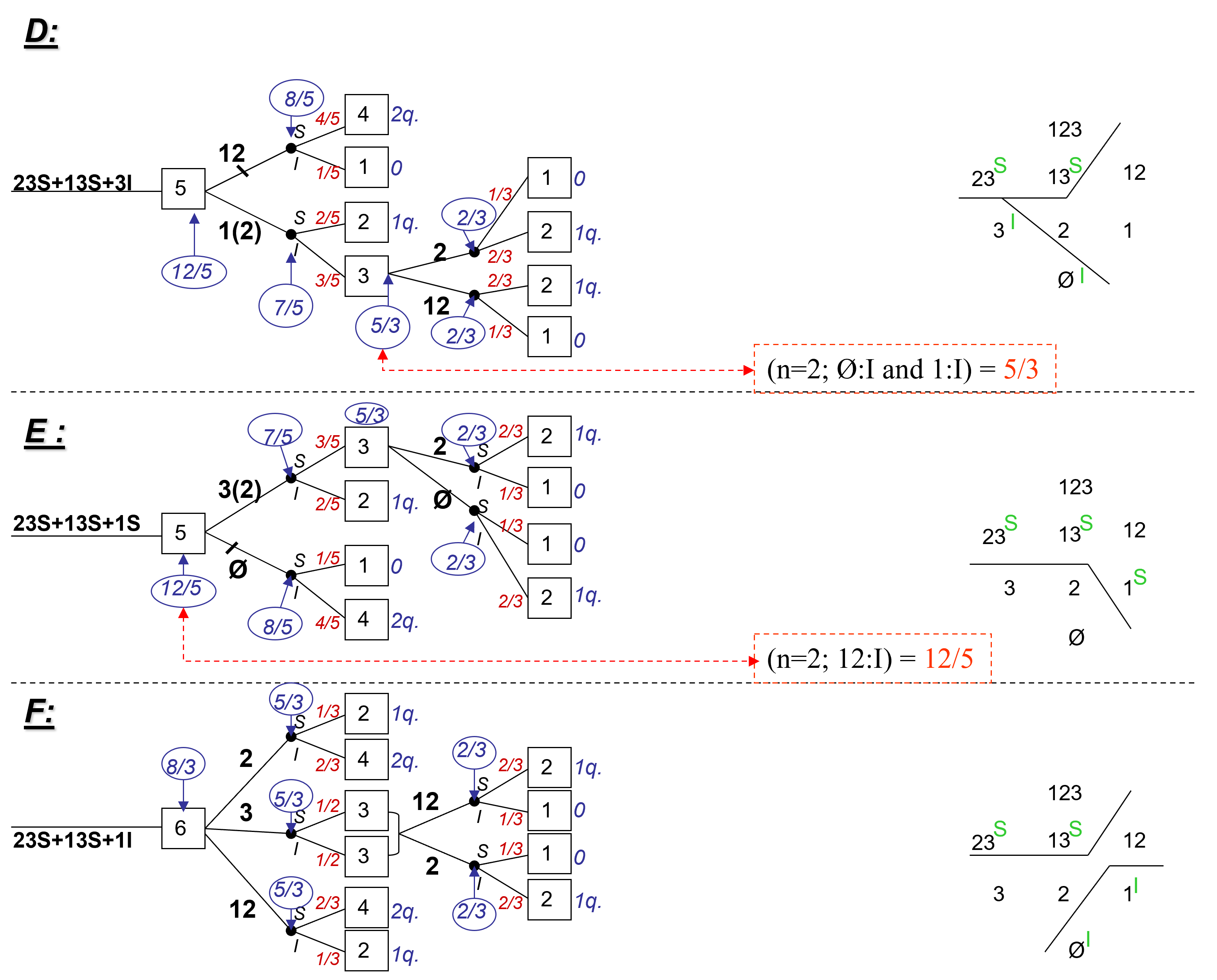}\caption{n=3, DEF}
\label{fig:decisionTree3critDEF}
\end{figure}
\end{center}

\begin{center}
\begin{figure}[h!]\includegraphics[width=135mm]{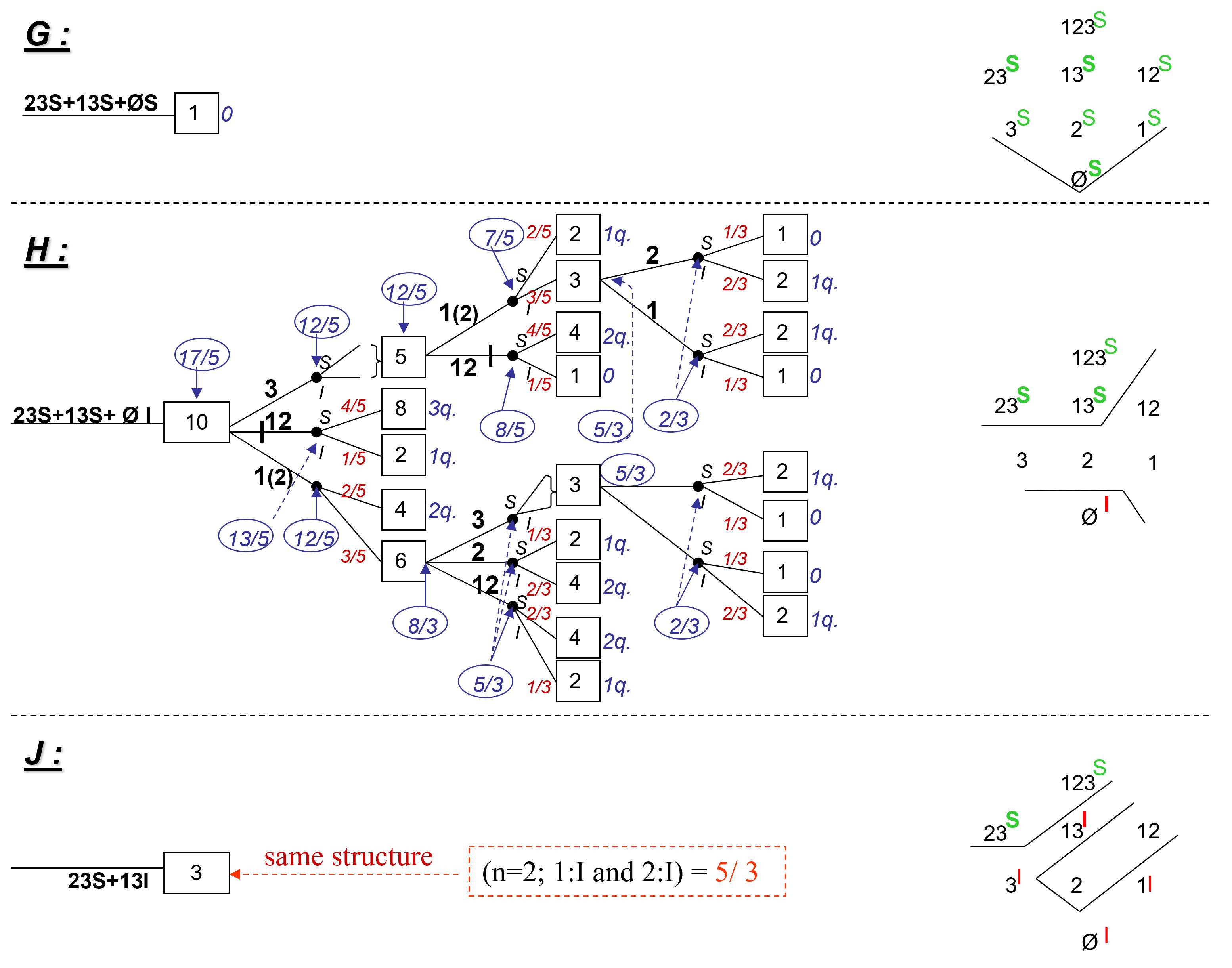}\caption{n=3, GHJ}
\label{fig:decisionTree3critGHJ}
\end{figure}
\end{center}

Figure~\ref{fig:heuristicPathEx2} shows a possible path suggested by the heuristic for $\mathcal{F}_{min}=\{3 , 1 2\}$. For each unlabeled coalition the numbers in exponent (\resp in index) is $\sigma^+$ (\resp $\sigma^-$). At the first step on may arbitrarily choose to question one of the 6 coalitions in the dashed rectangle. We choose to question in $23$. The answer is $S$, which labels $23$ and $123$. At the next step, the heuristic only points to coalition $1$. The answer to the question in $1$ is $I$, which labels $1$ and $\varnothing$. The process continues similarly and ends up after 5 questions. Note that this is also the number of questions asked when following the optimal policy on average represented in Figure~\ref{fig:optimalPaths3crit}. Using DFS3 would require 6 questions.  

\ssk For $n=4$, we observed, for some particular \Fmin{}, that the suggested heuristic gives acceptable results. For example $\mathcal{F}_{min}=\{34, 24, 14\}$ needs 11 questions with the main algorithm and 7 questions with $34S-12I-24S-23I-4I-14S-13I$ as the suggested path of the heuristic. On the other hand, $\mathcal{F}_{min}=\{\varnothing\}$ which requires 3 questions trough the main algorithm, is obtained by 5 questions following the path: $34S-12S-4S-3S-\varnothing S$.

As can be seen from the example, the heuristic favors the choice of a coalition with balanced, or homogeneous, distribution of unlabeled supersets and subsets. Actually the question is asked for the coalition maximizing the minimal gain (in terms of number of propagated labels). For coalitions tied on this criterion, the ones maximizing the maximal gain are preferred. A dichotomic principle is again essential here. 

\citet{torvikEtAl2002} have implemented and tested an idea of a heuristic algorithm similar to the one proposed above. They question a coalition that minimizes the difference $|K_1 - K_0|$, which is exactly the difference $\sigma^+ - \sigma^-$. They test their heuristic exhaustively up to $n=5$ and by sampling in the possible $\Fmin{}$ from $n=6$ to $n=11$. Their conclusion is that the heuristic performs slightly better on average than Hansel's algorithm. \citet{torvikEtAl2002} also proved that their heuristic is not optimal on average. In general, using an optimal decision tree leads to a smaller expected number of questions than the heuristic.

\section{Chapter conclusion and further research} \label{sec:concluChap6}

In this chapter, we focused on minimizing the average number of questions to ask. A decision tree corresponding to the number of criteria contains all possible paths that the DM can choose and allows to compute the expected number of questions for the considered paths. Optimal questioning paths can be suggested through the computations of the expected number of questions. The computations through the decision tree become difficult with the increase of the number of criteria. 

\ssk 
In order to deal with (somewhat) larger numbers of criteria, some simplifying tools can be used, such as existing symmetries, the upper-lower symmetries and the identification of some patterns that we have already known for smaller dimension cases. We can also compute the Informational Bounds and use the the Branch $\&$ Bound strategy to simplify the task.

\ssk  The results obtained by this computation method are done for \linebreak $n=\{2, \ldots ,4\}$ and compared in Fig~\ref{fig:comparisonExpected} with the results of other suggested algorithms: DFS3 and DFSHansel evaluated on average.

\ssk As a consequence of the study, we made a meaningful remark concerning the choice of the optimal paths which consists in choosing the most ``homogeneous distribution'' of \textit{possible} \Fmin{}. Starting from this observation, we introduce a simple heuristic that can be a facilitator, especially when the number of criteria increases.

\paragraph{Further research issue. } The proposed heuristic refines the one proposed earlier by \citet{torvikEtAl2003}. It could be interesting to visualise and compare the performance of our heuristic to DFSHansel and to the optimal expected questioning tree. This can be done by programming the heuristic and by simulating for different values of $n$. 

\clearpage

\section{Appendix: Figures in case \texorpdfstring{$n=3$}{n=3}}\label{app:figures3n}  
\vspace{2cm}

\begin{center}
\begin{figure}[h!]\includegraphics[width=135mm]{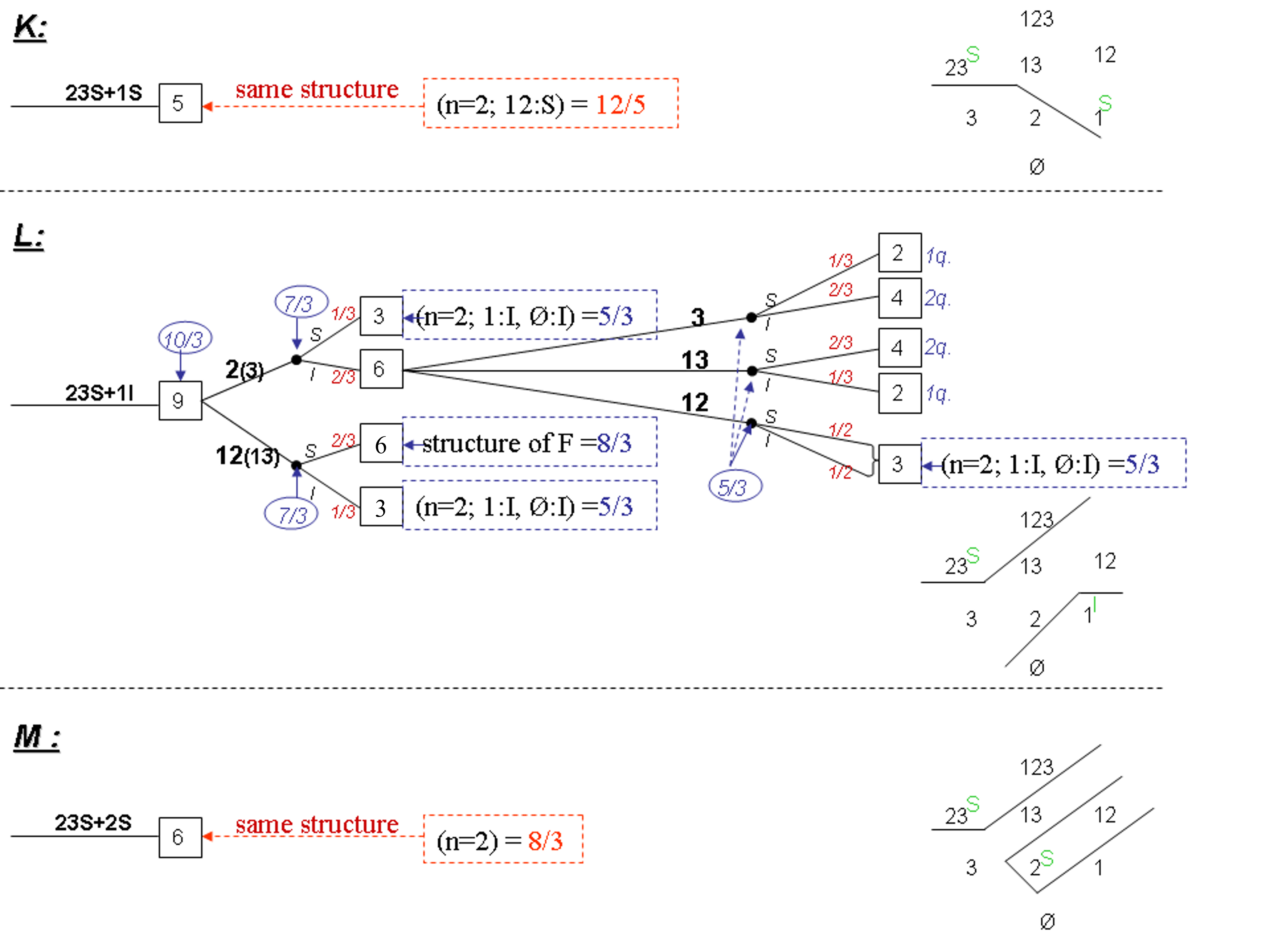}\caption{n=3, KLM}
\label{fig:decisionTree3critKLM}
\end{figure}
\end{center}

\begin{center}
\begin{figure}[h!]\includegraphics[width=135mm]{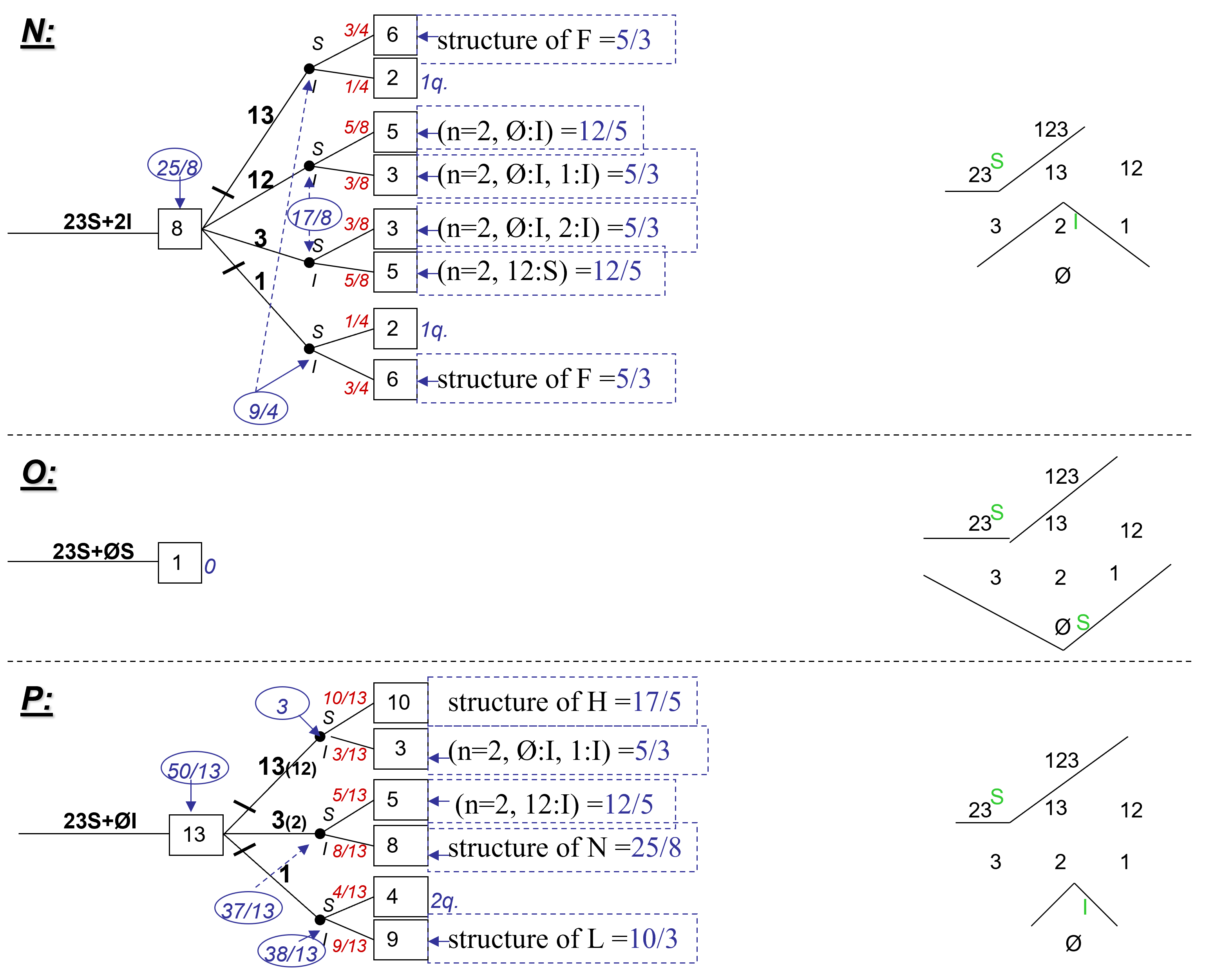}\caption{n=3, NOP}
\label{fig:decisionTree3critNOP}
\end{figure}
\end{center}

\begin{center}
\begin{figure}[h!]\includegraphics[width=135mm]{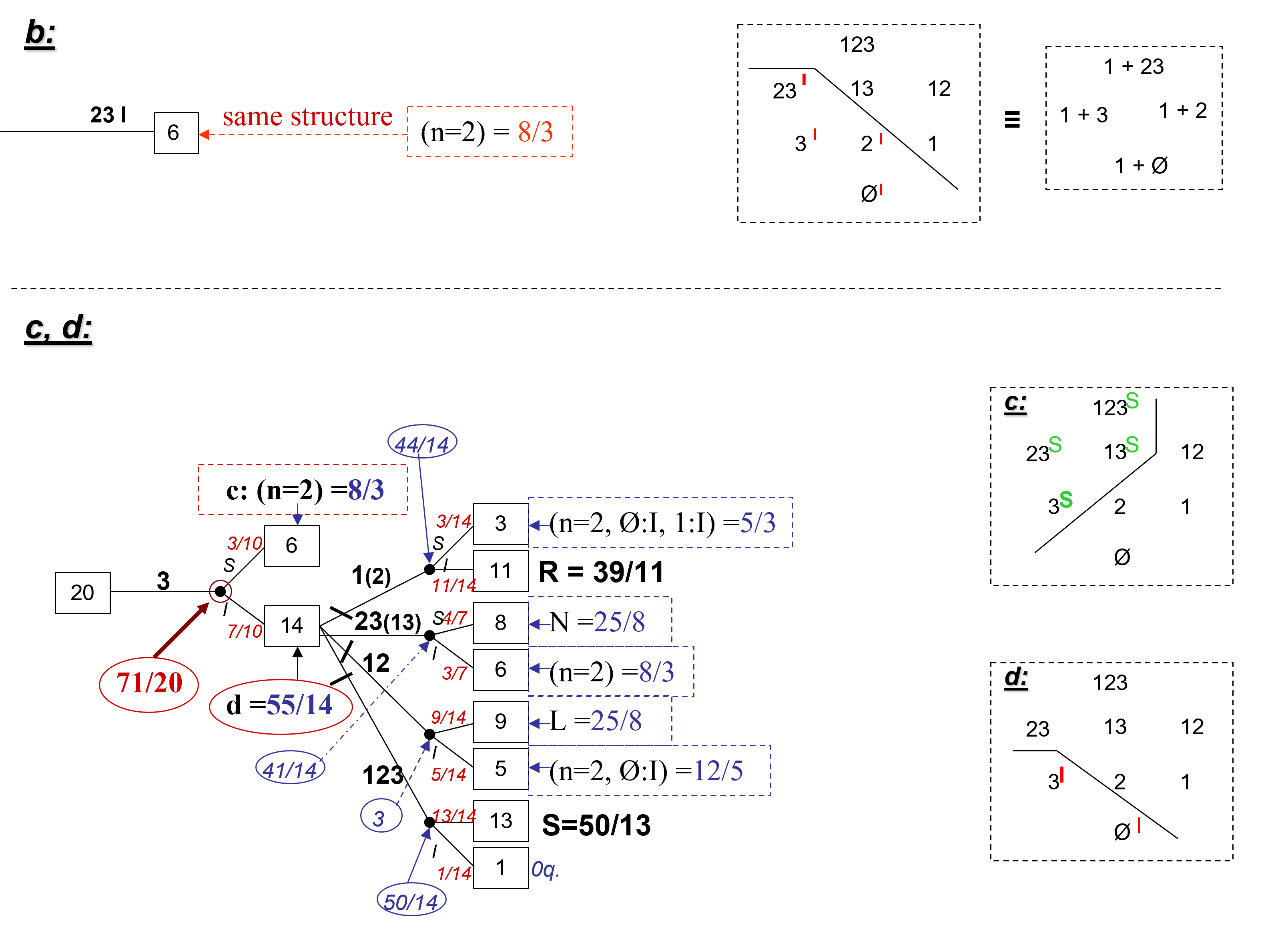}\caption{$n=3$, bcd}
\label{fig:decisionTree3critbcd}
\end{figure}
\end{center}

\begin{center}
\begin{figure}[h!]\includegraphics[width=135mm]{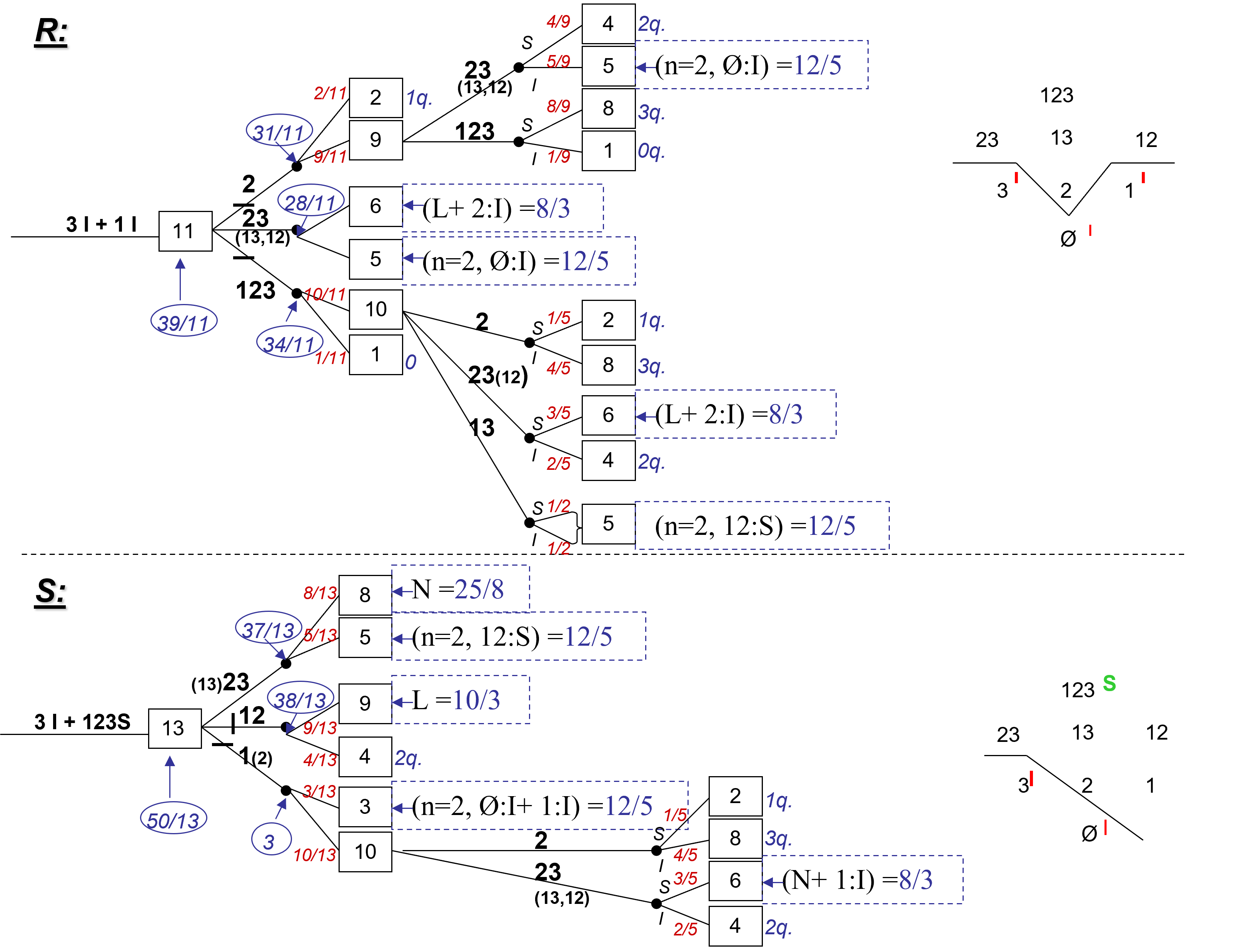}\caption{$n=3$, RS}
\label{fig:decisionTree3critRS}
\end{figure}
\end{center}

\begin{center}
\begin{figure}[h!]\includegraphics[width=135mm]{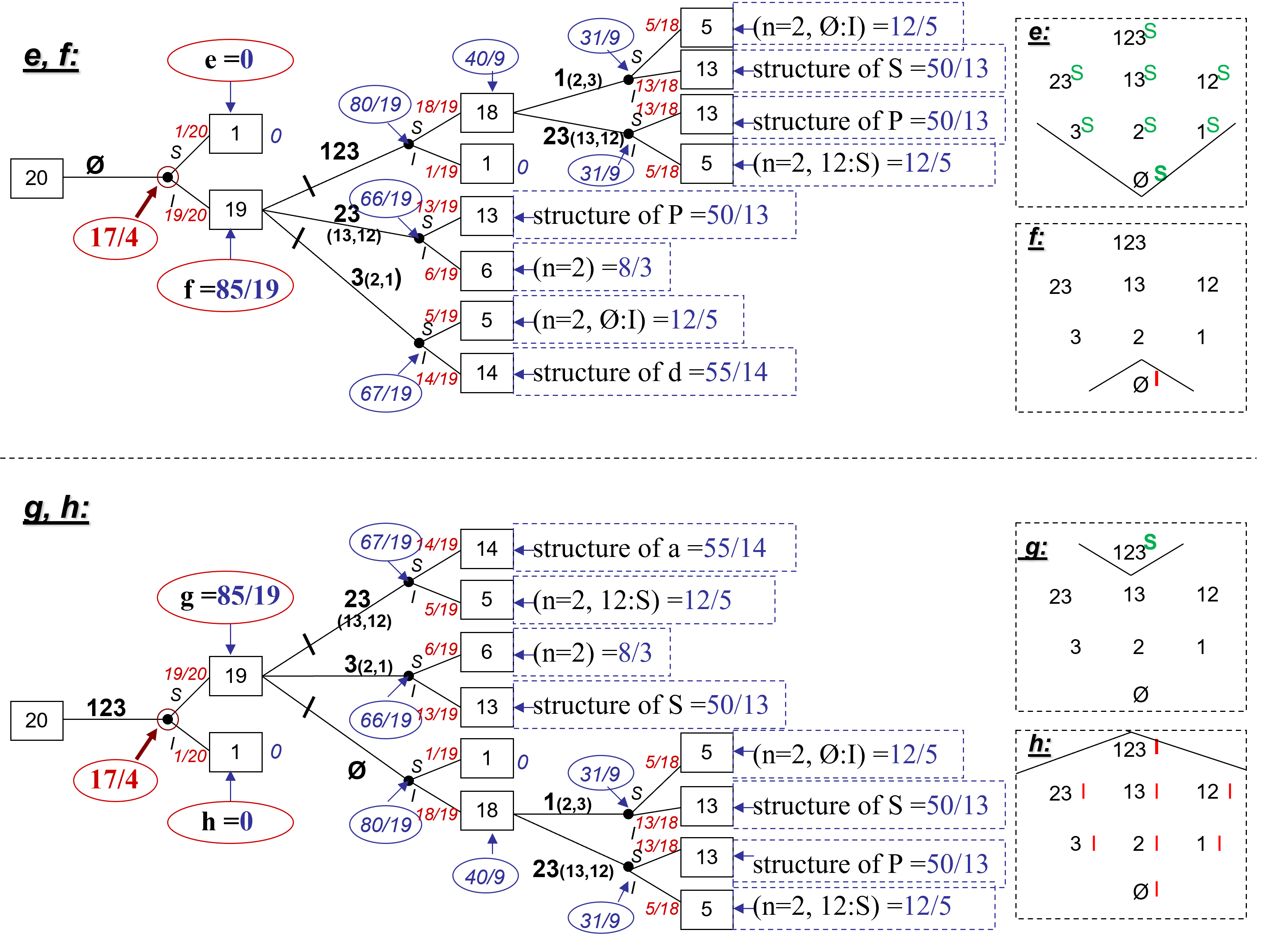}\caption{n=3, efgh}
\label{fig:decisionTree3critefgh}
\end{figure}
\end{center}

\clearpage

\section{Appendix: Figures in case \texorpdfstring{$n=4$}{n=4}}\label{app:figures4n}  
\vspace{2cm}

\begin{figure}[h!!!]
\begin{center}
\includegraphics[width=135mm]{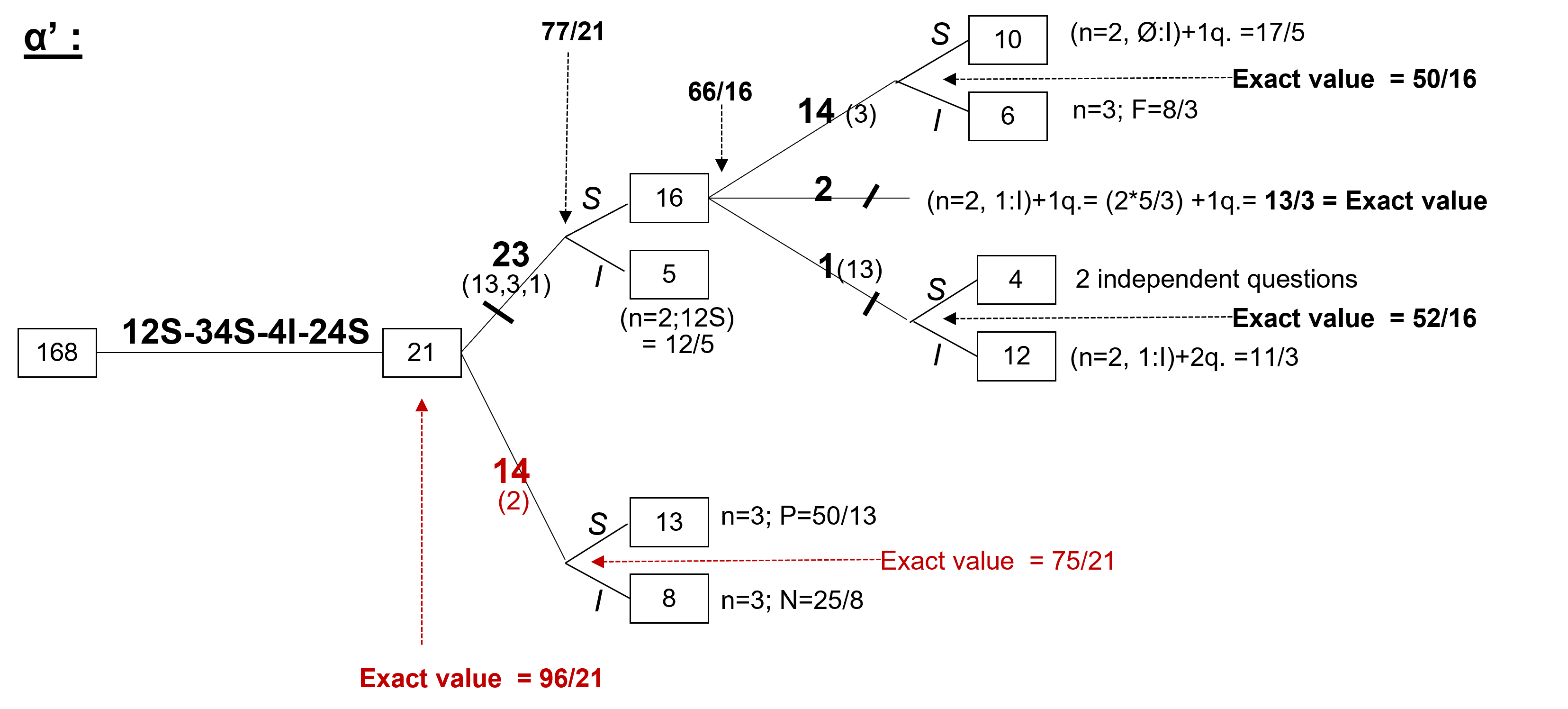}
\vspace{0.5cm}
\caption{Expected Number of Questions for n=4, Pattern $\alpha'$}
\label{fig:Expectedn=4-alphaprime}
\end{center}
\end{figure}

\begin{figure}[h!!!]
\begin{center}
\includegraphics[width=130mm]{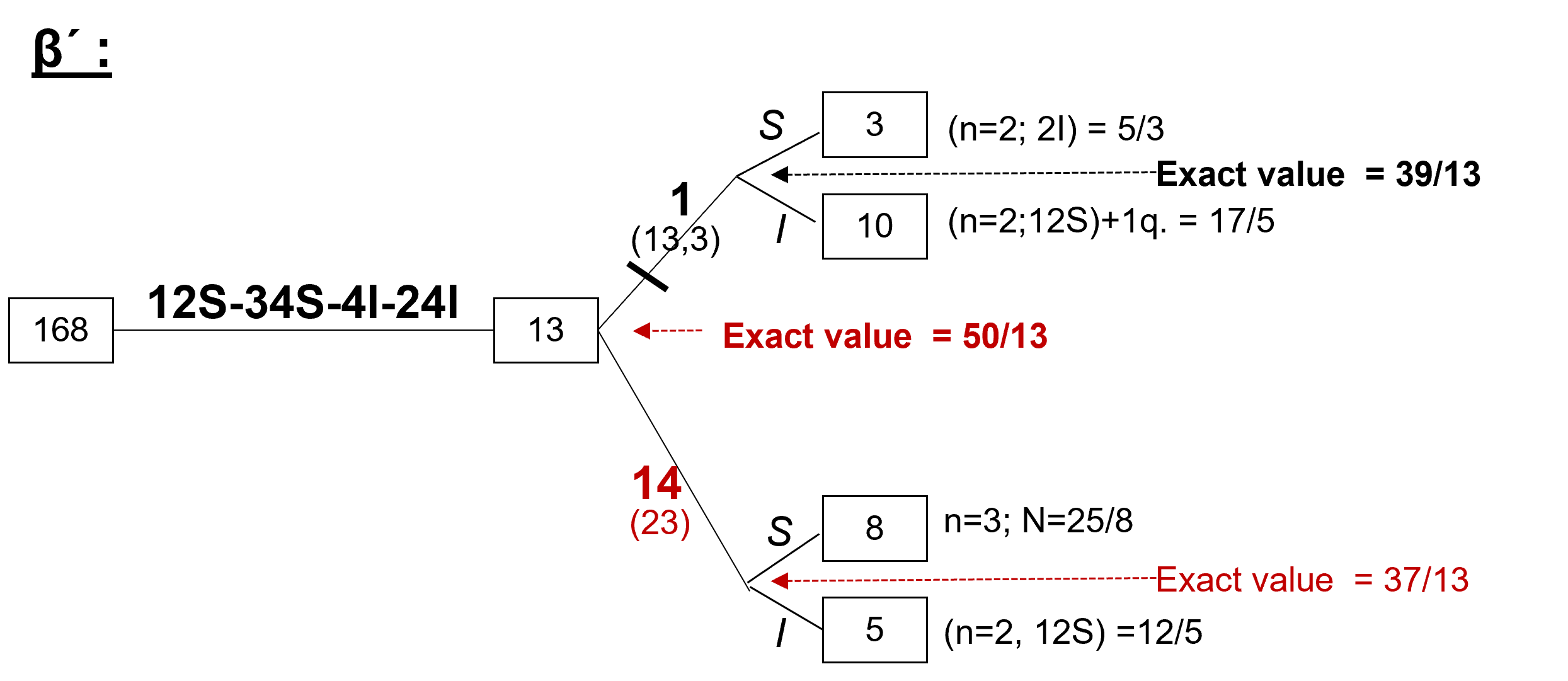}
\vspace{0.5cm}
\caption{Expected Number of Questions for n=4, Pattern $\beta'$}
\label{fig:ExpectedN=4_betaprime}
\end{center}
\end{figure}

\begin{figure}[h!!!]
\begin{center}
\includegraphics[width=135mm]{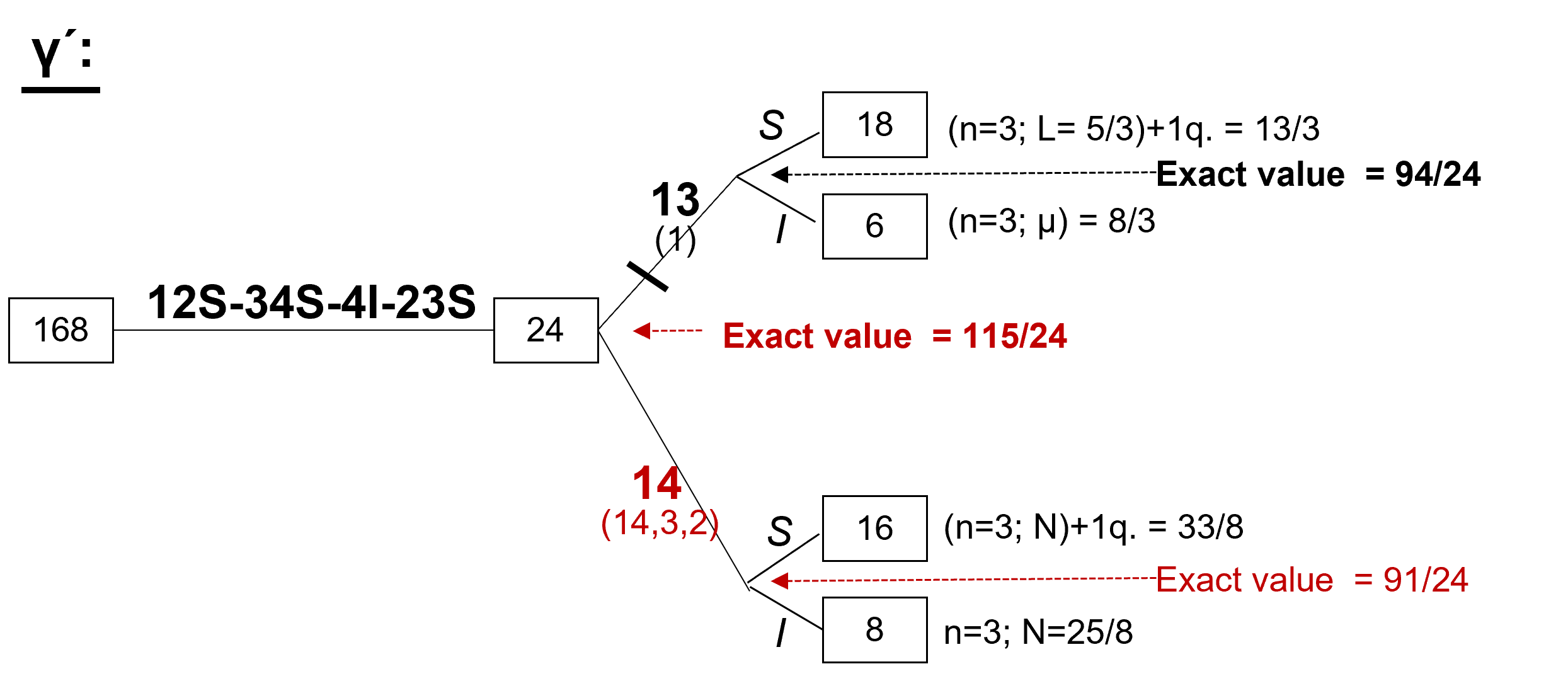}
\vspace{0.5cm}
\caption{Expected Number of Questions for n=4, Pattern $\gamma'$}
\label{fig:Expectedn=4-gammaprime}
\end{center}
\end{figure}

\begin{figure}[h!!!]
\begin{center}
\includegraphics[width=14cm]{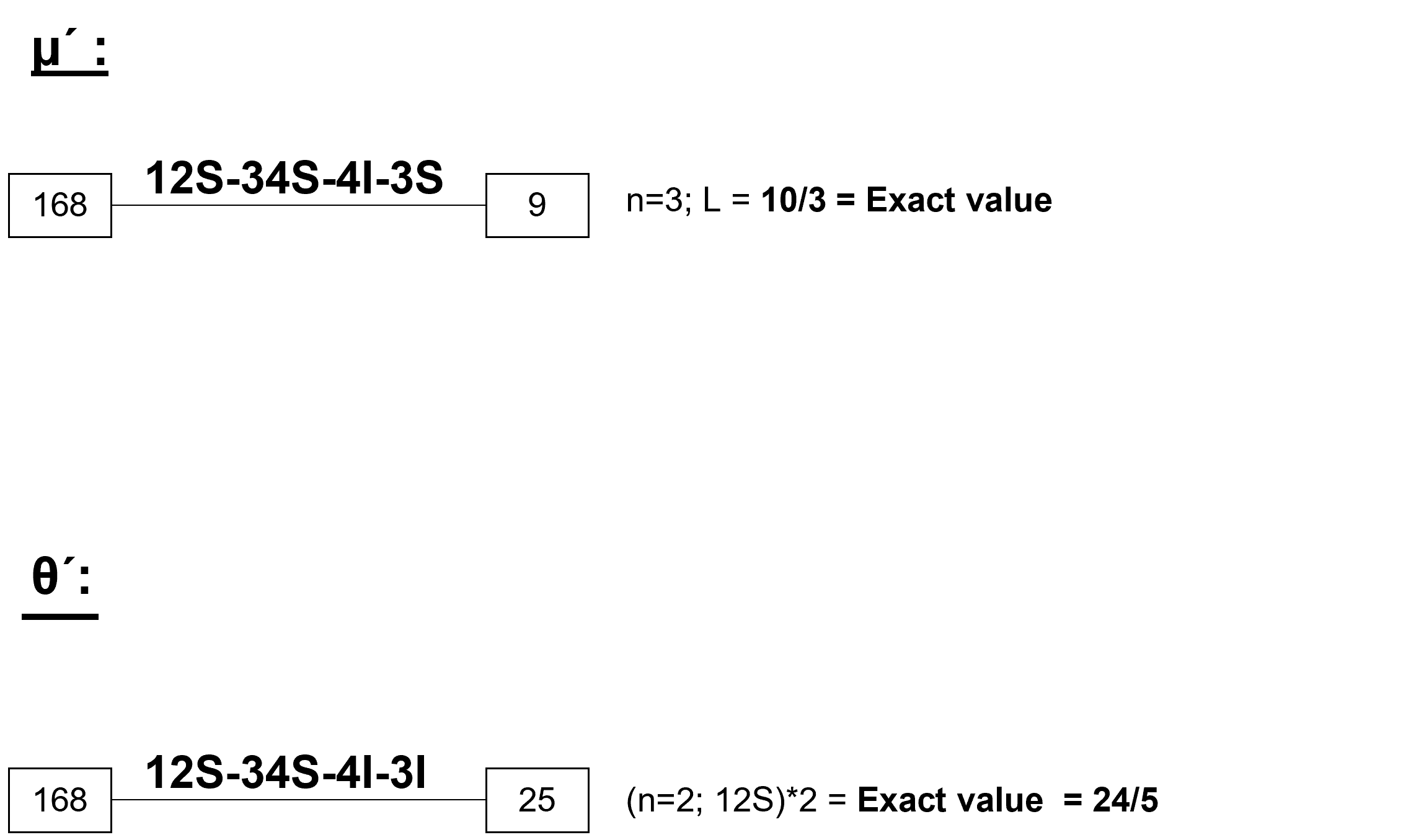}
\vspace{0.5cm}
\caption{Expected Number of Questions for n=4 $\mu$', Pattern $\theta'$}
\label{fig:Expectedn=4-muthetaprime}
\end{center}
\end{figure}

\clearpage

\begin{figure}[h!!!]
\begin{center}
\includegraphics[width=130mm]{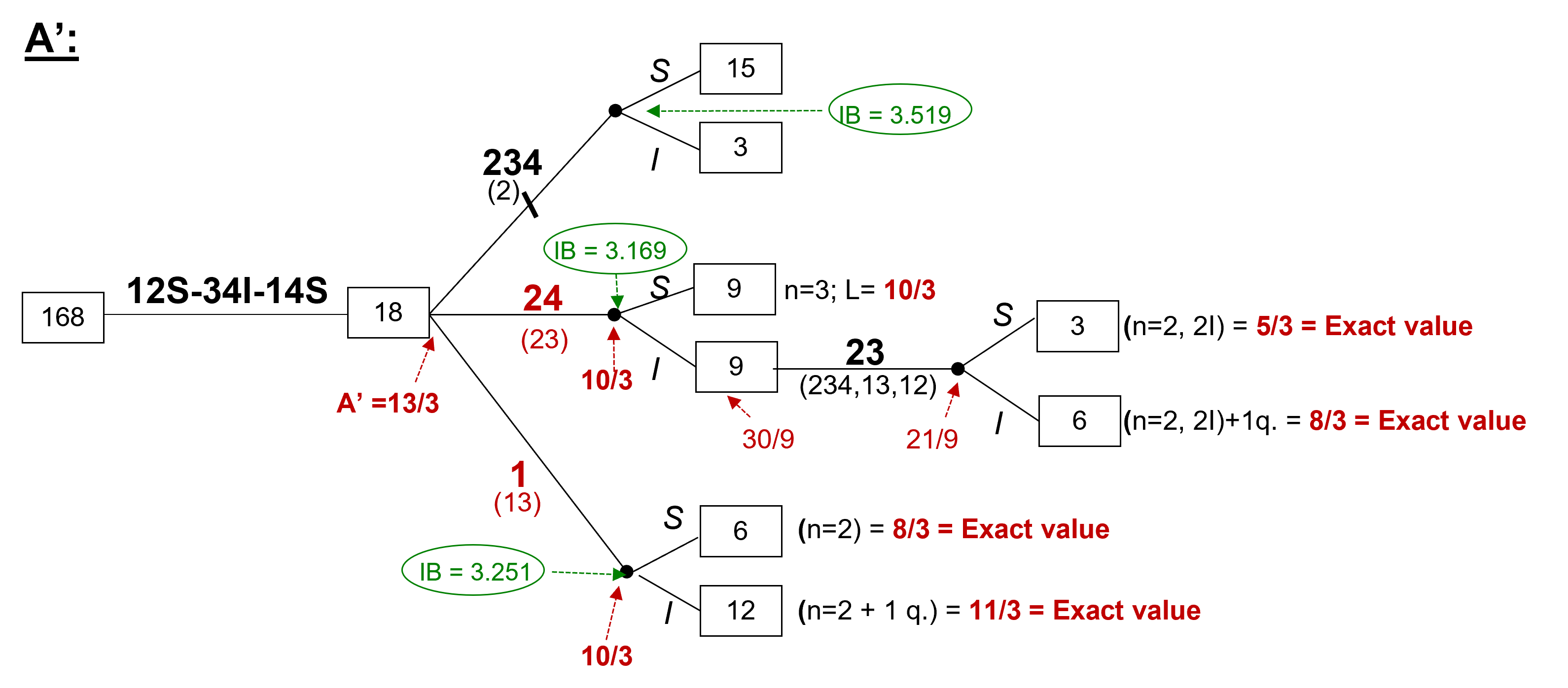}
\vspace{0.5cm}
\caption{Expected number of questions for n=4, 12S-34I-A'}
\label{fig:Expectedn=4_12S34IA'}
\end{center}
\end{figure}

\begin{figure}[h!!!]
\begin{center}
\includegraphics[width=135mm]{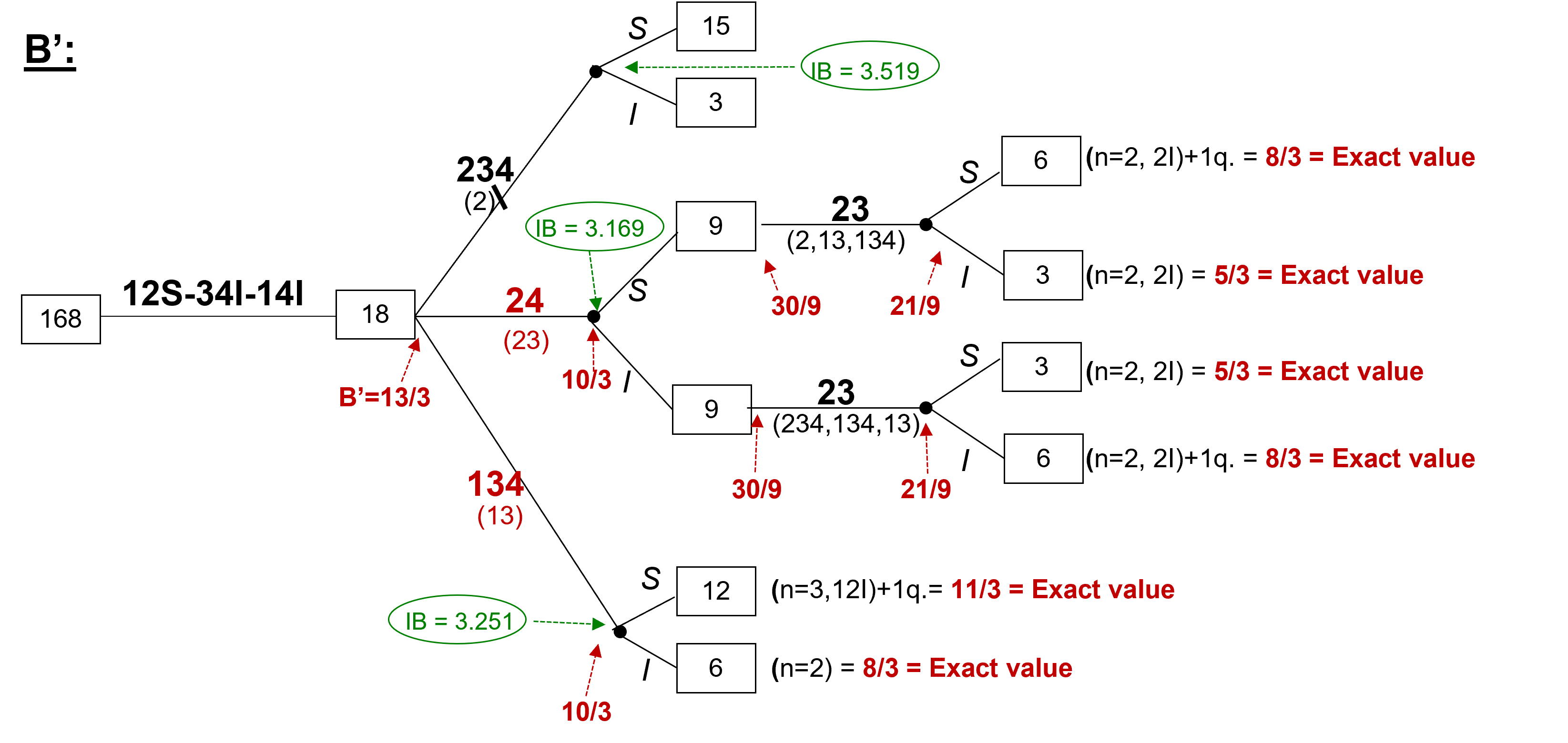}
\vspace{0.5cm}
\caption{Expected number of questions for n=4, 12S-34I-B'}
\label{fig:Expectedn=4_12S34IB'}
\end{center}
\end{figure}

\section{Appendix: Patterns used in case \texorpdfstring{$n=4$}{n=4}}\label{app:n4}
\sectionmark{Appendix: patterns used in case $n=4$}

\begin{figure}[h!!!]
\begin{center}
\includegraphics[width=12cm]{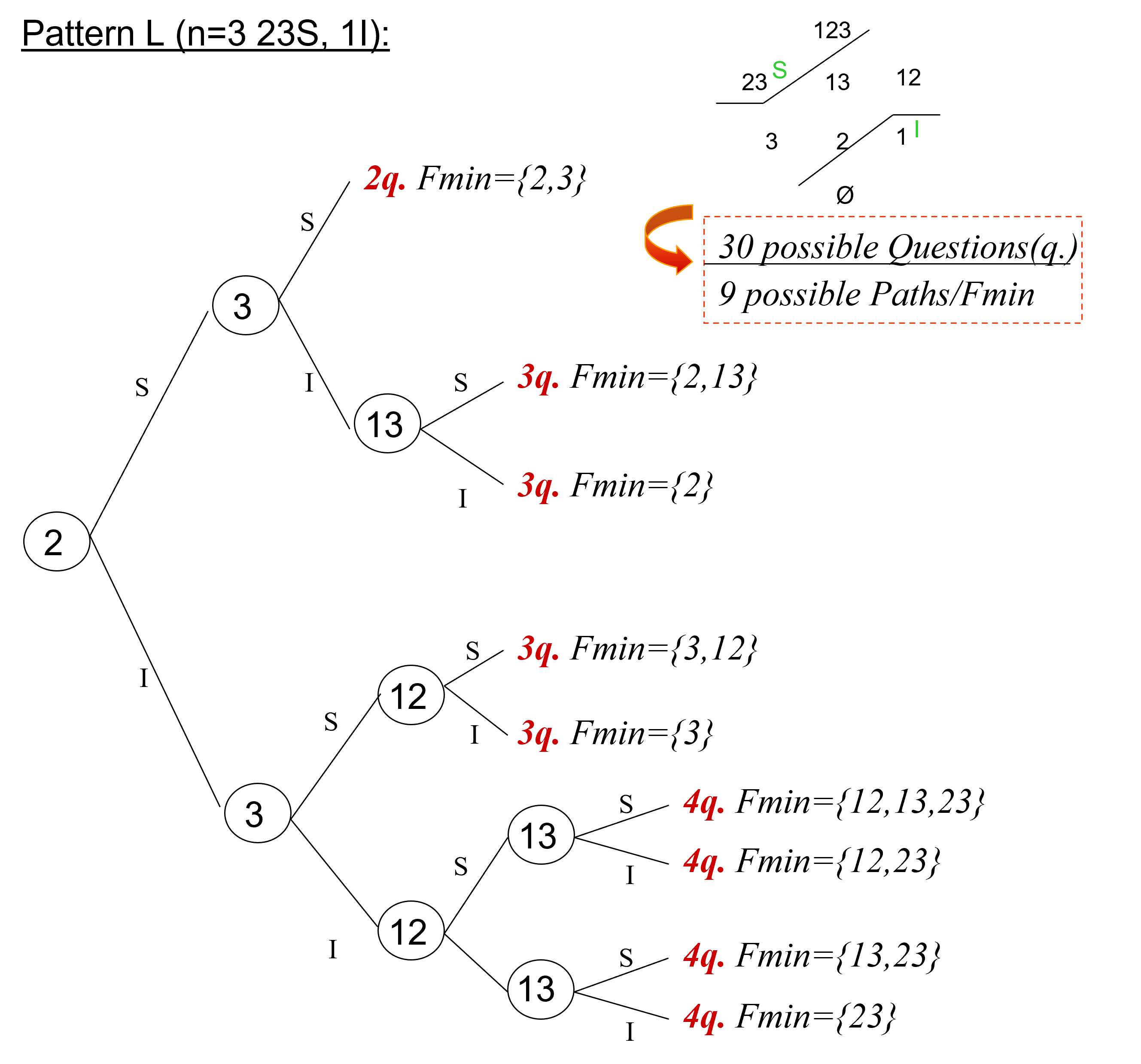}
\vspace{0.5cm}
\caption{Expected number of questions for the pattern L}
\label{fig:ExpectedL_nbq}
\end{center}
\end{figure}

\begin{figure}[h!!!]
\begin{center}
\includegraphics[width=12cm,height=12cm]{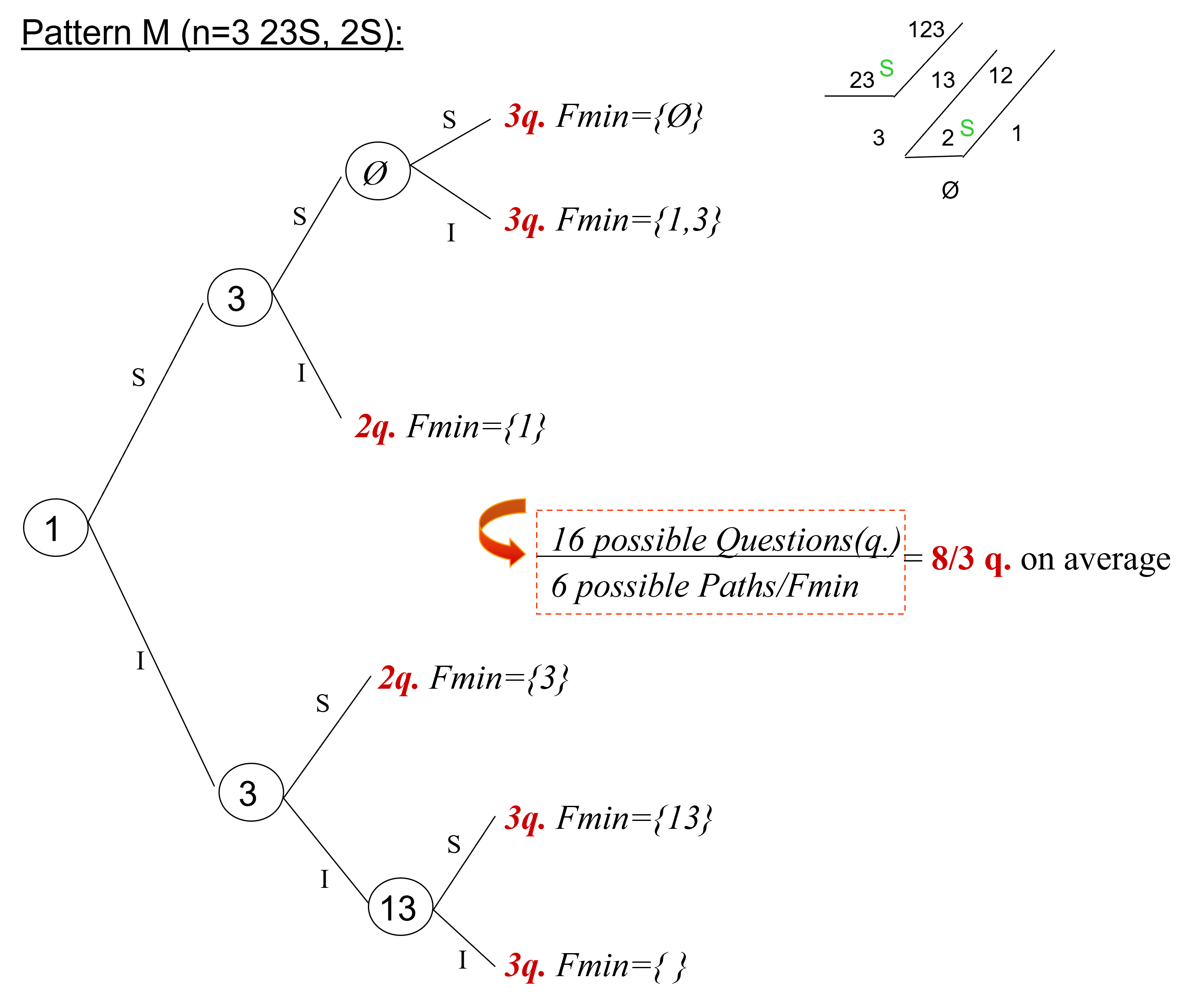}
\vspace{0.5cm}
\caption{Expected number of questions for the pattern M}
\label{fig:ExpectedM_nbq}
\end{center}
\end{figure}

\begin{figure}[h!!!]
\begin{center}
\includegraphics[width=13cm]{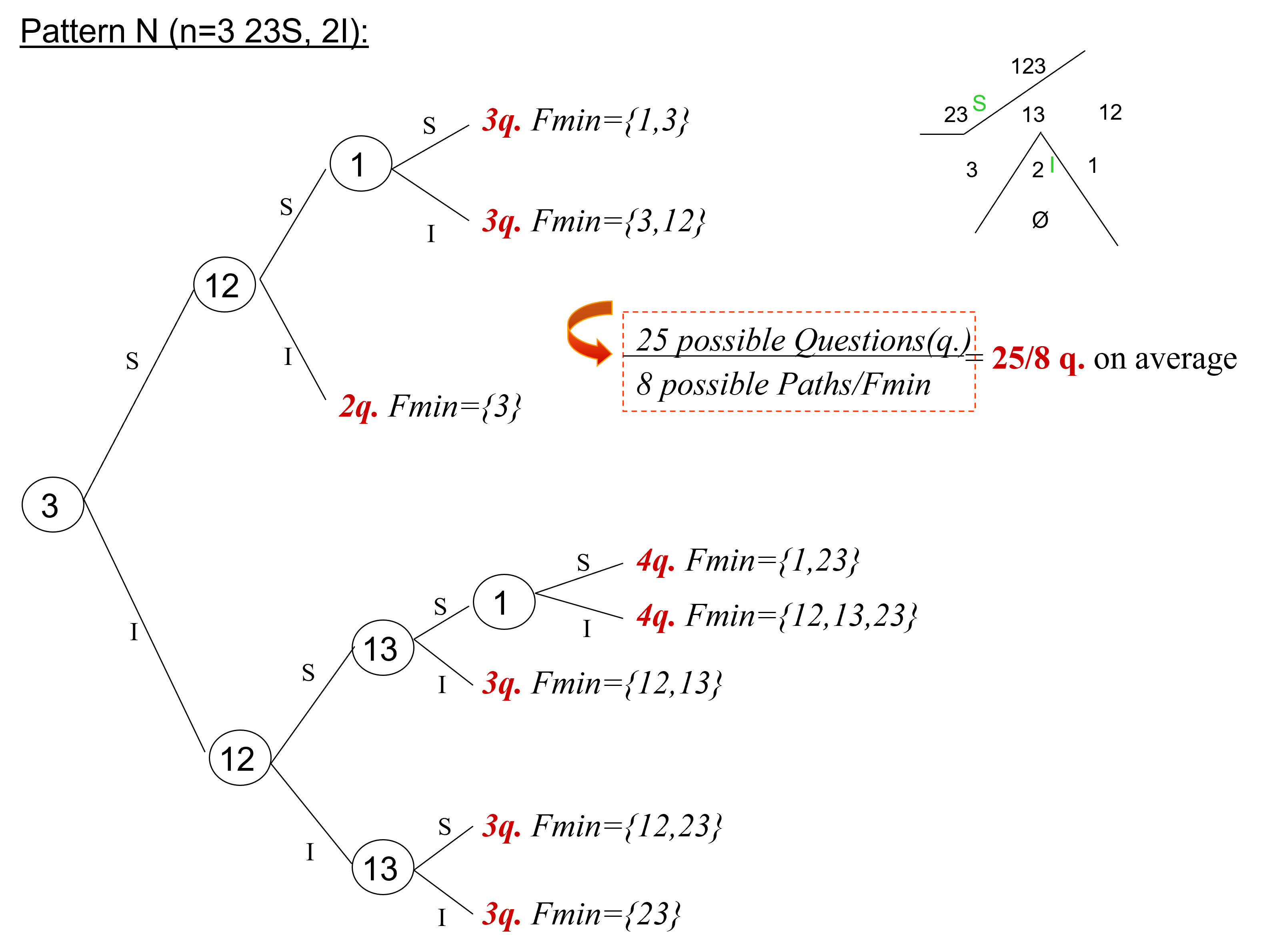}
\vspace{0.5cm}
\caption{Expected number of questions for the pattern N}
\label{fig:ExpectedN_nbq}
\end{center}
\end{figure}

\begin{figure}[h!!!]
\begin{center}
\includegraphics[width=13cm]{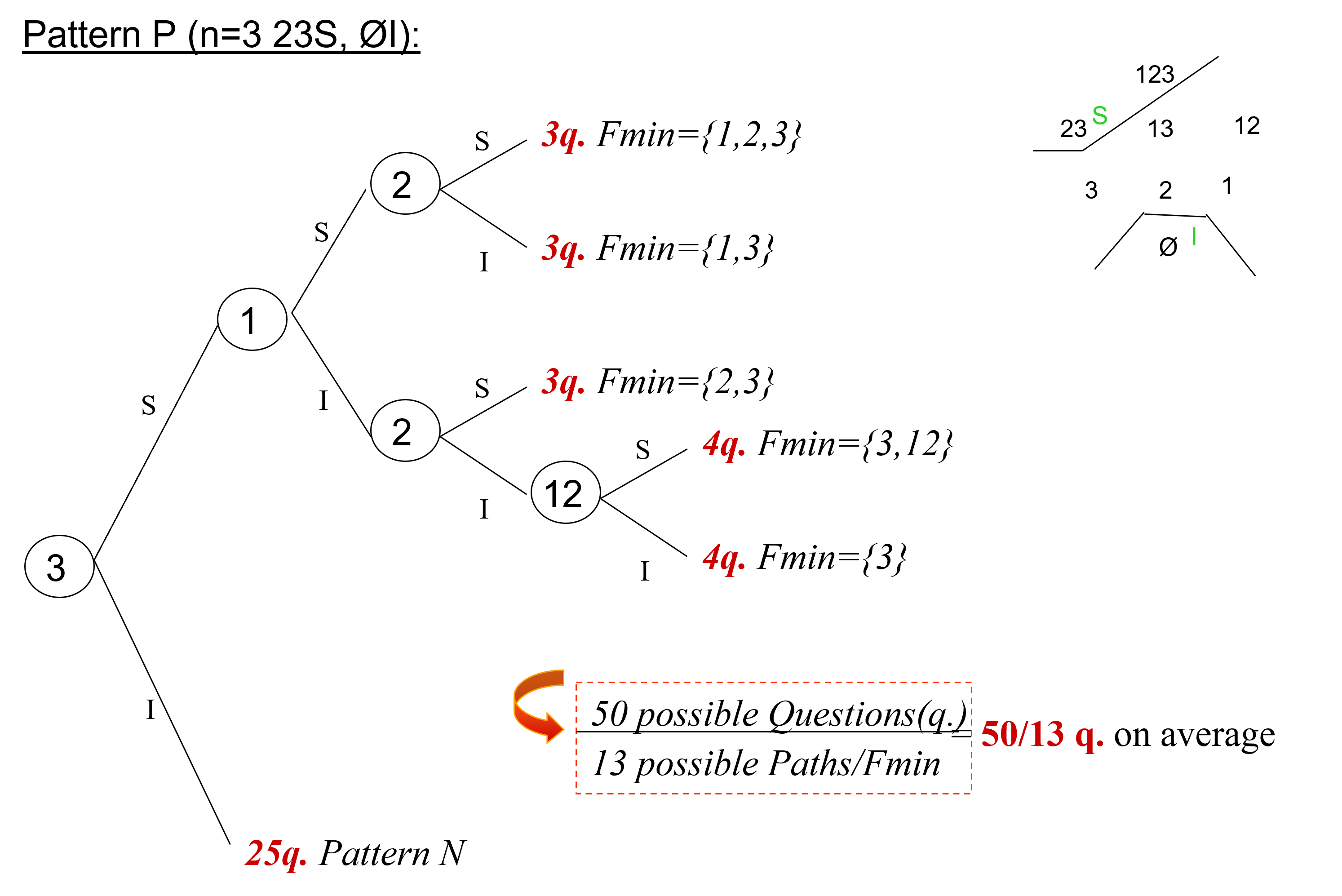}
\vspace{0.5cm}
\caption{Expected number of questions for the pattern P}
\label{fig:ExpectedP_nbq}
\end{center}
\end{figure}

\begin{figure}[h!!!]
\begin{center}
\includegraphics[width=12cm]{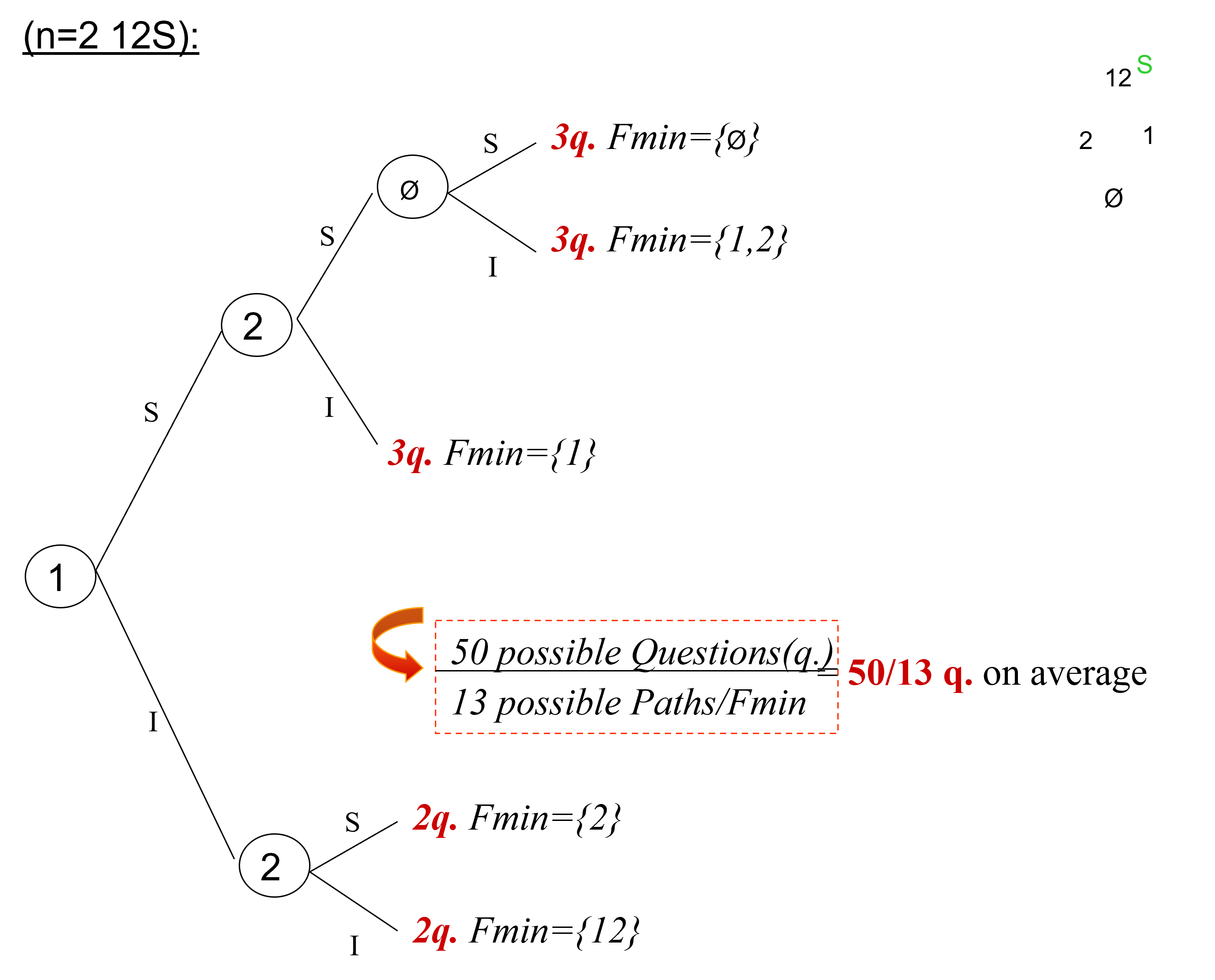}
\vspace{0.5cm}
\caption{Expected number of questions for the pattern $n=2$, 12S}
\label{fig:ExpectedN2_12S_nbq}
\end{center}
\end{figure}

\clearpage

\section{Appendix: Another optimal tree for \texorpdfstring{$n=4$}{n=4}}\label{app:otherOptimalN4}

\begin{figure}[h!!!]
   \centering
   \includegraphics[width=12cm]{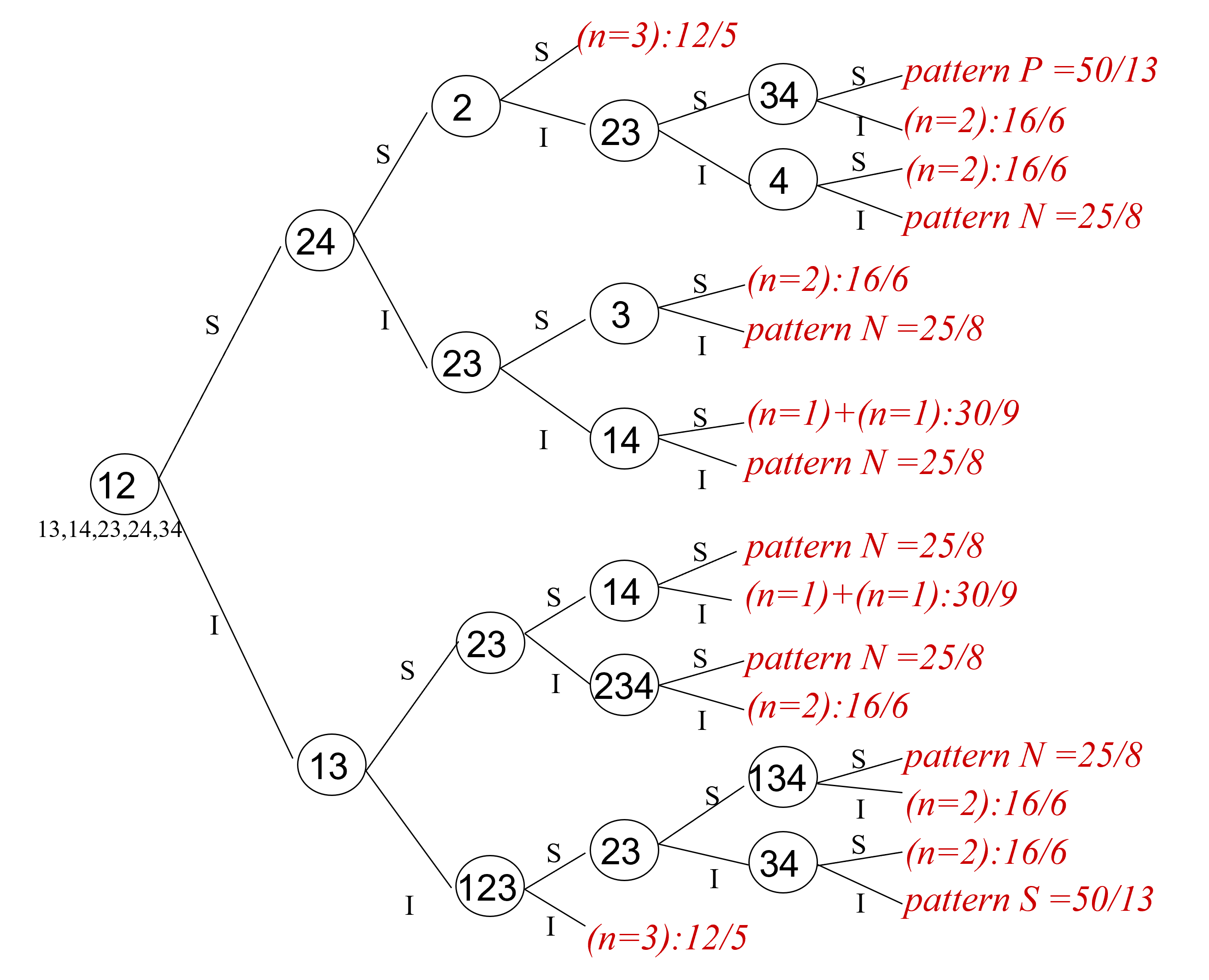}
   \caption{Another optimal questioning tree in the case of $n=4$}    \label{fig:anotherOptPath4crit}
\end{figure}
\chapter{Compact representation of sufficient coalitions}\label{chap:dam}
\chaptermark{Compact representation}

\ssk The \ncs{} model proposed and axiomatised by
\cite{bouyssoumarchant2007a,bouyssoumarchant2007b} is relatively general and can
handle criteria interactions. \mrsort which is a simplified version of the NCS model, deals only with sorting problems in which there is no interaction between criteria. In \mrsort{}, the set of sufficient coalitions is determined by means of weights attached to the criteria and a threshold. A coalition is sufficient if and only if the sum of the weights of the criteria in the coalition passes the threshold. 

Such a representation can be called \emph{compact} because it avoids to extensively list the set of sufficient coalitions (or the set of minimal sufficient coalitions). 

\ssk In this chapter we want to assess how much more general is the NCS model as compared to \mrsort. This can be seen as an analysis of the gap between \mrsort and NCS. In other words, we want to know, for a given number of criteria, how many of the families of sufficient coalitions are representable by weights and threshold.  

\ssk Due to the limitations of \mrsort{}, other compact representations of the set of minimal coalitions may be considered. A \emph{capacity} is a more general way of representing the ``strength'' of a coalition. A capacity attaches weights not only to individual criteria but also to pairs, triples, \ldots , $k$-tuples of criteria. By this way, using a threshold, one can represent more families of sufficient coalitions, at the expense of more parameters. We may consider these also as compact representations even though they are ``less compact'' than \mrsort{}. We also assess how many families of sufficient coalitions can be represented by certain categories of capacities.

\ssk The results of this chapter have been published in \citet{erseketal2014,ErsekEtAlDAM2017}.

\section{\mrsort{} and its limitations}\label{sec:MRSortLimitations}

In \mrsort{}, the set of sufficient coalitions can be represented numerically. More precisely, it is possible to represent sufficient coalitions as those verifying an inequality. 

\ssk In this study, we use \mrsort{} without veto and in the case of 2 categories $\calAU$. Thus the
assignment rule is simplified as follows:
\begin{equation}
	a \in \calA \iff
	\sum_{j: a_j \geq b_j} w_j \geq \lambda 
\label{eq-mrsort_2cat_noveto}
\end{equation}

For the situations that cannot be handled by \mrsort{}, we are looking for other compact representations for sufficient coalitions. Although the notion of capacity is less compact than \mrsort{}, it can be useful in such cases.
As we mentioned previously \mrsort{} is easy to use but has limitations. We show these limits on the example below.

\begin{example}\label{ex:nonMR2Add}
Consider a situation in which a committee of professors have to decide the successful completion of students
based on their examinations in 4 courses: French, English, Maths and Physics.

The conditions for successful completion of examinations are:
\begin{itemize}
	\item at least one mark $\geq 12/20$ in  literary subjects (French and English) and;
	\item at least one mark $\geq 12/20$ in scientific subjects (Maths and Physics).
\end{itemize}

Such an assignment rule cannot be represented in MR-Sort, \ie by weights and a threshold. Letting the four courses French, English, Maths and Physics, be labelled 1, 2, 3, 4, respectively, we have the following constraints:

\begin{equation*}
\begin{cases}
	w_1 + w_3 & \geq \lambda,\\
	w_1 + w_4 & \geq \lambda,\\
    w_2 + w_3 & \geq \lambda,\\
	w_2 + w_4 & \geq \lambda,\\
    w_1 + w_2 & < \lambda,\\
	w_3 + w_4 & < \lambda.
\end{cases}
\end{equation*}
By summing up the first four inequalities, we obtain:
\begin{equation}
	w_1 + w_2 + w_3 + w_4 \geq 2 \lambda.
\label{eq-example_impossible_1add_1}
\end{equation}

After summing up the last two inequalities, we obtain:
\begin{equation}
	w_1 + w_2 + w_3 + w_4 < 2 \lambda.
\label{eq-example_impossible_1add_2}
\end{equation}
Constraints \eqref{eq-example_impossible_1add_1} and
\eqref{eq-example_impossible_1add_2} are contradictory. This system of inequalities has no solution and 
it is not possible to use simple MR-Sort to model this assignment rule.

\end{example}

As seen in the previous example, \mrsort{} cannot deal with criteria interactions and it is not suitable to use it for such type of problems. 

\section{Capacity and \texorpdfstring{$k$-additive}{k-additive} capacity}\label{sec:capaKcapa}
All families of sufficient coalitions cannot be specified by the inequality conditions of \mrsort{}. If we are in such a case, is there another kind of inequality that can be used? In fact, any family of sufficient coalitions can be represented using a capacity and a threshold. We recall now, what is a capacity. We recall that $N$ denotes the set $\{1, \ldots, n\}$.

\begin{definition}
A (normalized) capacity is a function $\mu: 2^N \rightarrow [0, 1]$ such that:
\begin{itemize}
	\item $\mu(K) \geq \mu(L)$, for all $L \subseteq K \subseteq N$
	(monotonicity) ;
	\item $\mu(\emptyset) = 0$ and $\mu(N) = 1$ (normalization).
\end{itemize}
\label{def-capacity}
\end{definition}
A capacity can also be obtained by means of its Möbius transform $m$ defined by:
\begin{equation}\label{eq:Mobius}
    m(K) = \sum_{L \subseteq K} (-1)^{|K| - |L|} \mu(L), \ \forall K \subseteq N.
\end{equation}
We have:
\begin{equation}
	\mu(J) = \sum_{K \subseteq J} m(K), \
	\forall J \subseteq N
\label{eq:capaMobius}
\end{equation}

The term $m(K)$ is interpreted as the weight that is allocated to the set $K$ as a whole. It represents the interactions between criteria belonging to the set $K$. The capacity of a set $J$ is thus the sum of Möbius interactions of all subsets of $J$. 

\ssk A M\"{o}bius transform is a
set function $m: 2^N \to [-1, 1]$ that satisfies the following conditions:
\begin{equation}
	\sum_{j \in K \subseteq J} m(K) \geq 0, \
	\forall j \in N , \ j \in J \subseteq N \
	\text{ and }
	\sum_{K \subseteq N} m(K) = 1.
\label{eq-monotonicity_mobius}
\end{equation}
If $m$ is a M\"{o}bius interaction satisfying \eqref{eq-monotonicity_mobius}, the set function defined by
$\mu(J) = \sum_{K \subseteq J} m(K)$ is a normalized capacity. Conditions
\eqref{eq-monotonicity_mobius} guarantee that $\mu$ is monotone
\citep{chateauneufjaffray1986}.

\ssk A capacity is said to be $k$-additive when $k$ is the largest cardinality of the subsets for which $m$ is different from 0. A capacity is $1$-additive, or simply additive, if $m(K) =0$, for all $K$ with $|K| > 1$. In such a case, $\mu(J) = \sum_{j \in J} m(\{j\}) $. The term $m(\{j\})$ can be interpreted as the weight $w_j$ of criterion $j$.


\ssk Therefore, the MR-Sort assignment rule \eqref{eq-mrsort_2cat_noveto} is thus equivalent to 
\begin{equation}
   a \in \calA \ \iff \
	\ \mu(\{j: a_j \geq b_j\}) \geq \lambda, 
\label{eq-mrsort_2cat_noveto_bis}
\end{equation}
with $\mu$ a $1$-additive capacity. 

\ssk Example \ref{ex:nonMR2Add} in Section \ref{sec:MRSortLimitations} cannot be represented with a 1-additive capacity, but it can be represented with a 2-additive capacity. Let capacity $\mu$ be defined by its Möbius transform $m$ as follows:
\begin{eqnarray*}
    m(\{j\}) & = & 1/6 \ \textrm{for} \ j = 1, \ldots, 4 \\
    m(\{i,j\}) & = & 1/12 \ \textrm{for } \ \{i,j\} = \{1,3\}, \{1,4\}, \{2,3\},\{2,4\}; 
\end{eqnarray*}
all other interactions are null.
We have:
\begin{eqnarray*}
    \mu(\{j\}) & = & 1/6 \ \textrm{for} \ j = 1, \ldots, 4 \\
    \mu (\{1,3\}) & = & \mu(\{1,4\}) =  \mu(\{2,3\})  = \mu (\{2,4\}) = 1/6+1/6+1/12=5/12 \\
    \mu(\{1,2\}) & = & \mu(\{3,4\}) =2/6.
\end{eqnarray*}

By setting $\lambda = 5/12$, we reproduce the assignment rule of the example. 

\paragraph{Parameters.}
In order to define a capacity $\mu$ by means of its Möbius transform, we need to specify $2^n$
parameters.

To reduce the number of parameters, we generally use 
$k$-additive capacities, with small values of $k$ ($k=2, 3$). 
For a 2-additive capacity with $n$ criteria we need to determine $\frac{n(n+1)}{2} - 1$ parameters.
In general, a $k$-additive capacity requires the specification of 
$$
n + {n \choose 2} + \ldots + {n \choose k} - 1
$$
parameters. The ``-1'' in the formula is due to the relation $\sum_{J \subseteq N} m(J) = 1$.




\section[Numerical representation]{Numerical representation of sufficient co\-alitions in the NCS model}\label{sec:capacityNCS}
Consider a NCS model with $b$ the limit profile of category $\calA$ and $\calF$ the set of sufficient coalitions. 

\begin{proposition}
\label{proposition-any_family_of_sc}
Any family of sufficient coalitions can be represented as the set of
subsets $J \subseteq \{1, \ldots, n\}$ such that
\begin{equation*}
	\mu(J) \geq \lambda,
\end{equation*}
with some capacity $\mu$ and threshold $\lambda \geq 0$.

Reciprocally, if $\mu$ is a capacity and $\lambda$ a nonnegative number, the set of subsets $J$ which satisfy $\mu(J) \geq \lambda$ is a family of sufficient coalitions.
\end{proposition}

\bpr
A family of sufficient coalitions is a family of subsets such that any
subset that contains a subset of the family belongs to the family.

\ssk Given such a family $\calF$, we define
the nonnegative set function $\mu$ letting $\mu(J)=1$ if $J \in \calF$ and $0$ otherwise.
$\mu$ is monotone, and therefore a capacity, as a consequence of the properties of the families of sufficient coalitions.
It is also normalized.
Define the threshold $\lambda = 0.5$. Then, $\mu(J) \geq 0.5$ iff $J \in \calF$.

\ssk The reverse is also proved directly.
\epr

We define below the class $\mathcal{C}_k$ of families of sufficient coalitions representable by a $k$-additive capacity.  

\begin{definition}
A family of sufficient coalitions $\calF$ belongs to class $\mathcal{C}_k$ if
\begin{enumerate}
	\item there is a normalized $k$-additive capacity $\mu$ and a non-negative threshold  $\lambda$  such that 
$$ 
J \in \calF \ \ \textrm{iff} \ \
 \mu(J) \geq \lambda.
$$
	\item $k$ is the smallest integer for which such an inequality is valid.
\end{enumerate}
\end{definition}

We saw in Sections \ref{sec:MRSortLimitations} and \ref{sec:capaKcapa} that Example~\ref{ex:nonMR2Add} is not representable by a 1-additive capacity but is representable by a 2-additive capacity. Hence the corresponding family of sufficient coalitions belongs to class $\mathcal{C}_2$. 

\ssk Proposition~\ref{proposition-any_family_of_sc} amounts saying that any family of sufficient coalition belongs to $\bigcup_{k=1}^n \mathcal{C}_k$.


\section{Representability by \texorpdfstring{$k$-additive}{k-additive} capacity} \label{RepKaddCap}
How to examine the representability of a rule by a $k$-additive capacity and to separate the set of minimal coalitions \Fmin in families $\mathcal{C}_k$ for a given number of criteria $n$ ?


\ssk The verification procedure is as follows.
For each family of minimal sufficient coalitions, we check if it belongs to \Ck, starting from $k=1$ and increasing $k$ until the test is positive.

\ssk A linear program can test this. For this, we have to write constraints such that $\mu(A) \geq
\lambda$ for each sufficient coalition $J$ and that the same inequality is
not verified for all insufficient coalitions.

\ssk The set of minimal sufficient (\resp maximal insufficient) coalitions is denoted by \Fmin (\resp \Gmax).

\ssk If the following linear program has a solution with $\varepsilon >0$, then the set \Fmin of minimal sufficient coalitions belongs to class $\Ck$. 

\begin{equation}\label{eq:linearProg}
\left\{
\begin{array}{>{\displaystyle}rc>{\displaystyle}ll}
\max \ \varepsilon & &&\\
&&& \\
\mu(J) &\geq& \lambda & \forall J \in \Fmin \\
\mu(J) &\leq& \lambda - \varepsilon  & \forall J \in \Gmax \\
\mu(J) & = &\sum_{I \subseteq J} m(I) & \forall J \in \Fmin \cup \Gmax \\
\sum_{I : \ i \in I \subseteq J} m(I) & \geq & 0 &  \forall i \in N, \ \forall J \subseteq N\\
\sum_{I \subseteq N} m(I) & = & 1 & \\
\lambda, \varepsilon & \geq & 0 &
\end{array}
\right.
\end{equation}

\ssk The variables in the linear program are $\lambda$, $\varepsilon$ and $m(I)$ for all $I \subseteq N$. Note that the variables $m(I)$ are not necessarily positive (except for $|I|=1$). For $\mathcal{C}_k$, $m(I) = 0$ if $|I| > k$. For example, we show below how to write the third, fourth and fifth constraints in the cases $k=1$ and $k=2$.
\begin{itemize}
  \item $k=1$ (1-additive capacity)
  \begin{itemize}
    \item[.] $\mu(J) = \sum_{i \in J} m_i, \ \forall J \in \Fmin \cup \Gmax$
    \item[.] $m_i \geq 0, \ \forall i \in N$
    \item[.] $\sum_{i \in N} m_i =1$,
  \end{itemize}
  where $m_i$ stands for $m(\{i\})$
  \item $k=2$ (2-additive capacity)
  \begin{itemize}
    \item[.] $\mu(J) = \sum_{i \in J} m_i + \sum_{i,j \in J,\ i\neq j } m_{ij}, \ \forall J \in \Fmin \cup \Gmax$
    \item[.] $m_i + \sum_{j \in J,\ j\neq i} m_{ij}  \geq 0, \ \forall i \in N$ and $\forall J \ni i , \ J \subseteq N$
    \item[.] $\sum_{i \in N} m_i + \sum_{i,j \in N,\ i\neq j } m_{ij}=1$,
  \end{itemize}
  where $m_i$ stands for $m(\{i\})$ and $m_{ij}$ for $m(\{i,j\})$.
\end{itemize}
In the $3$-additive case, we need to use an additional variable $m_{ijl}$ for each subset $\{i,j,l\}$ of three different elements of $N$ in addition to the already introduced variables $m_i$ and $m_{ij}$.

\section{Inequivalent families of sufficient coalitions.} In order to count and/or list the families of sufficient coalitions that can be represented by a $k$-additive capacity, we have to apply the linear program to each of them. There is a possibility of reducing the computing effort by testing only the \emph{inequivalent} families of sufficient coalitions. 

\begin{definition}\label{def:equivalentFam}
    Two families of minimal sufficient coalitions are equivalent if one can be obtained from the other through a permutation of the criteria labels.
\end{definition}
For instance, for $n=4$, $\Fmin = \{12, 34\}$ is equivalent to $\{13,24\}$ and to $\{14,23\}$. For another example, consider the following family of minimal SCs for $n=4$: $\{2, 4\}, \{2, 3\}, \{1, 3, 4\}$. It consists of 2 subsets of 2 elements and one of 3 elements. There are 12 equivalent families that can be obtained from this one, by permuting the elements labels (the element which does not show up in the set of 3 can be chosen in 4 different ways and the two elements which distinguish the two pairs can be chosen in 3 different ways). 


\ssk A family of Minimal Sufficient Coalitions (MSCs) can be any set of subsets of $N$ such that none of them is included in another. In other words, a set of MSCs is an \emph{antichain} in the set of subsets of $N$ (partially) ordered by inclusion. It is well-known that the number of antichains in the power set of $N$ is $D(n)$, the $n$th Dedekind number (\cite{OEIS}, sequence A000372). These numbers grow extremely rapidly with $n$ and no exact closed form is known for them. These numbers have been computed up to $n=8$; their  values appear in the second column of Table \ref{ta:dedekindNumbers}. The third column shows the number $R(n)$ of inequivalent families of MSCs. The number $R(n)$ is known for $n=0$ to $n=7$ (\cite{OEIS}, sequence A003182). $R(7)$ was only recently computed by  \citet{stephenyusun2014}.
\begin{table}[h!]
\renewcommand{\arraystretch}{1}
  \centering
\begin{tabular}{crr}
  \hline
  $n$ & $D(n)$  & $R(n)$\\
\hline
0	&	2 & 2\\
1	&	3 & 3\\
2	&	6 & 5\\
3	&	20 & 10\\
4	&	168 & 30\\
5	&	7581& 210\\
6	&	7828354 & 16353 \\
7	&	2414682040998 & 490013148\\
8	&	56130437228687557907788 & ?\\
  \hline
\end{tabular}
  \caption{Known values of the Dedekind numbers, $D(n)$, and of the number of inequivalent families of sufficient coalitions, $R(n)$.} \label{ta:dedekindNumbers}\label{ta:ineqDedekindNumbers}
\end{table}

Finally we recall Sperner's theorem (\cite{CasparLeclercMonjardetBook}, pp.~116-118), a result that will be useful in the process of generating all possible families of SCs.  The maximal size of an antichain in the power set of a set of $n$ elements is ${n \choose {\lfloor n/2 \rfloor}}$. Hence the latter is the maximal number of sets in a family of minimal SCs.

\section[Inequivalent families]{Listing the families of inequivalent minimal sufficient coalitions}\label{sec:listingInequivalent}

For generating all families of Minimal Sufficient Coalitions (MSC) and selecting a representative of each class of equivalent families, we follow a strategy similar to the one used in \cite{stephenyusun2014}. We describe it below. The families of MSCs can be partitioned according to their \emph{type} (called ``profile'' in \cite{stephenyusun2014}). The type of a family of MSCs is an integer vector $(k_1, k_2, \ldots, k_n)$, where $k_i$ represents the number of coalitions of $i$ elements in the family. For instance, the family  $\{\{2, 4\}, \{2, 3\}, \{1, 3, 4\}\}$, for $n=4$, is of the type $(0,2,1,0)$, since it involves two coalitions of 2 elements and one of 3. For any feasible type, $\sum_{i=1}^n k_i \leq {n \choose {\lfloor n/2 \rfloor}}$, the maximal size of a SC (Sperner's theorem).

\bsk The listing algorithm proceeds as follows:
\begin{enumerate}
  \item generate all type vectors $(k_1, k_2, \ldots, k_n)$ in lexicographic increasing order;
  \item for each type, generate all families of subsets of $N$ having the right type and eliminate those that are not antichains, i.e. those involving a subset that is included in another subset;
  \item for each type and for each family of this type, the list of remaining families is screened for detecting families that are equivalent, counting them and eliminating them from the list of families of the type considered.
\end{enumerate}
This algorithm outputs a list containing a representative of each class of equivalent families of MSCs for each type.

\ssk Results obtained by \citet[][Lemma~2.4]{stephenyusun2014} permit to accelerate the generating process. 

\begin{lemma}\label{le:stephenYusun}
    \begin{enumerate}
        \item \label{lemSY:it:1} There is a bijection between the set of inequivalent MSCs of type: \\ $(0, \ldots, 0, k_i, $ $0, \ldots, 0)$ and the set of inequivalent MSCs of type: \\ $(0, \ldots, 0, {n \choose i} - k_i, 0, \ldots, 0)$. Hence the cardinals of those sets are equal.
        \item \label{lemSY:it:2} If $k_1>0$, then $k_n=0$ and there is a bijection between the set of MSCs of $n$ criteria of type: $(k_1, k_2, \ldots, k_{n-1},0)$ and the set of MSCs of $n-1$ criteria of type: $(k_1-1, k_2, \ldots, k_{n-1})$. Hence the cardinals of these sets are equal.
        \item \label{lemSY:it:3} There is a bijection between the set of inequivalent MSCs of type:\\ $(k_{1}, k_{2}, \ldots,$ $ k_{n-2}, k_{n-1},k_n)$ and the set of inequivalent MSCs of type:\\ $(k_{n-1}, k_{n-2}, \ldots, k_2, k_1,$ $ k_n)$.
    \end{enumerate}

\ssk \noindent  Items \ref{lemSY:it:1} and \ref{lemSY:it:3} are also valid for the sets of all kind of MSCs (inequivalent and equivalent).     
\end{lemma}

\bpr\ref{lemSY:it:1}. Type $(0, \ldots, 0, k_i, 0, \ldots, 0)$ selects $k_i$ subsets of size $i$ in $N$. The remaining ${n \choose i} - k_i$ subsets of size $i$ form a MSC of type:\\ $(0, \ldots, 0, {n \choose i} - k_i, 0, \ldots, 0)$.

\ssk \ref{lemSY:it:2}. If $k_1 > 0$, there is at least one sigleton in the MSC; therefore, the set $N$ is not part of the MSC, hence $k_n=0$. Consider one of the singletons in the MSC. Without loss of generality (for inequivalent MSCs), we may say this criterion is $n$.  The other coalitions in the MSC cannot contain this criterion. Therefore, the other coalitions can be seen as forming a MSC on $n-1$ criteria of type $(k_{1}-1, k_{2}, \ldots, k_{n-2}, k_{n-1}$).

\ssk \ref{lemSY:it:3}. For any MSC of type $(k_{1}, k_{2}, \ldots,$ $ k_{n-2}, k_{n-1},k_n)$, we obtain an MSC of type  $(k_{n-1}, k_{n-2}, \ldots, k_2, k_1,$ $ k_n)$, by replacing each coalition $A$ by its complement $N \setminus A$. Doing that we obtain a MSC. Indeed, if we had $B \subset C$ among the new coalitions, it would mean that $N\setminus B$ and $N \setminus C$ are coalitions in the original MSC and $N\setminus B \supset N\setminus C$, a contradiction.
\epr

\ssk These results are useful for sparing time while generating inequivalent MSCs. Using item~\ref{lemSY:it:1}, we need only to generate MSCs of type $(0, \ldots, 0, k_i,$ $ 0, \ldots, 0)$ for $k_i \leq \lceil {n \choose i}/2 \rceil$. Item~\ref{lemSY:it:2} allows to use what has been done for $n-1$ in case the profile type contains at least one coalition formed of a single criterion ($k_1 > 0$). Using item~\ref{lemSY:it:3}, when we find the inequivalent MSCs for a profile $(k_{1}, k_{2}, \ldots,$ $ k_{n-2}, k_{n-1},k_n)$, we automatically get those for the profile $(k_{n-1}, k_{n-2}, \ldots, k_2, k_1,$ $ k_n)$, by complementation. This divides the number of computations by 2. This is the result which has the largest impact on the computation time.  

\ssk The algorithm for listing the inequivalent MSCs was  implemented in Python. We use it for $n \leq 6$. A list of the  30  inequivalent MSCs for $n=4$ is in Appendix~\ref{appendix-ineq_msc_n4}. For $n=6$, the list of 16\,353 inequivalent MSC occupies 1.25~Gb of memory space.  

\section{Representability by \texorpdfstring{$k$-additive}{k-additive} capacity}
We are looking to check for the minimal value of $k$ such that a given family of sufficient coalitions can be separated from the corresponding family of insufficient coalitions by a $k$-additive capacity (and a threshold). All of the equivalent families of MSC belong to the same class. Thus, we can just check if one representative of each class of inequivalent families of MSCs belongs or not to $\mathcal{C}_k$. We do that first for $k=1$. Checking the separability by $k$-additive capacity is formulated by a linear program explained in \ref{RepKaddCap}.

\ssk From $n=1$ to 6 and for each family in the list of inequivalent families of
MSC, we check whether this family belongs to \Ck. We start by $k=1$ and increment its value until the check is positive.
\begin{table}[h!!!]
\renewcommand{\arraystretch}{1}
  \centering
\resizebox{\columnwidth}{!}{%
  \begin{tabular}{rrrrrrrr}
    \hline
$n$  &  \Ck[1] & &     \Ck[2] & &  \Ck[3] & & $R(n)$ \\
\hline
0  &      2 &(100.0 \%) &      0 &(00.00 \%) &   0 &(00.00 \%) & 2\\
1  &      3 &(100.0 \%) &      0 &(00.00 \%) &   0 &(00.00 \%) & 3\\
2  &      5 &(100.0 \%) &      0 &(00.00 \%) &   0 &(00.00 \%) & 5\\
3  &     10 &(100.0 \%)&      0 &(00.00 \%) &   0 &(00.00 \%) & 10\\
4  &     27 &(90.00 \%) &      3& (10.00 \%) &   0 &(00.00 \%) & 30\\
5  &    119 &(56.67 \%)&     91 &(43.33 \%) &   0 &(00.00 \%) & 210\\
6  & 1113 & (06.81 \%) & 14902 &(91.13 \%) & 338 &(02.07 \%) & 16 353\\
    \hline
  \end{tabular}
}
\caption{Number and proportion of inequivalent families of MSCs that are representable by a 1-, 2- or 3-additive capacity}\label{ta:2-3-additiveRn}
\end{table}

\begin{table}[h!!!]
\renewcommand{\arraystretch}{1}
  \centering
\resizebox{\columnwidth}{!}{%
\begin{tabular}{rrrrrrrr}
  \hline
$n$ &      \Ck[1] &       & \Ck[2] &   &      \Ck[3] &   &   $D(n)$ \\
\hline
0 &         2 &(100.0 \%) &         0 &(00.00 \%) &       0& (00.00 \%) & 2\\
1 &         3 &(100.0 \%) &         0 &(00.00 \%) &       0& (00.00 \%) & 3\\
2 &         6 &(100.0 \%) &         0 &(00.00 \%) &       0& (00.00 \%) & 6\\
3 &        20 &(100.0 \%) &         0 &(00.00 \%) &       0& (00.00 \%) & 20\\
4 &       150 &(89.29 \%) &        18 &(10.71 \%) &       0& (00.00 \%) & 168\\
5 &     3 287 &(43.36 \%) &     4 294 &(56.64 \%) &       0& (00.00 \%) & 7581\\
6 &    244 158 &(03.12 \%) & 7 438 694 &(95.02 \%) & 145 502& (01.86 \%) & 7 828 354 \\
  \hline
\end{tabular}
}
  \caption{Number and proportion of all families of MSCs that are representable by a 1-,  2- or 3-additive capacity}\label{ta:2-3-additiveDn}
\end{table}

\begin{figure}
    \centering
    \includegraphics[width=132mm]{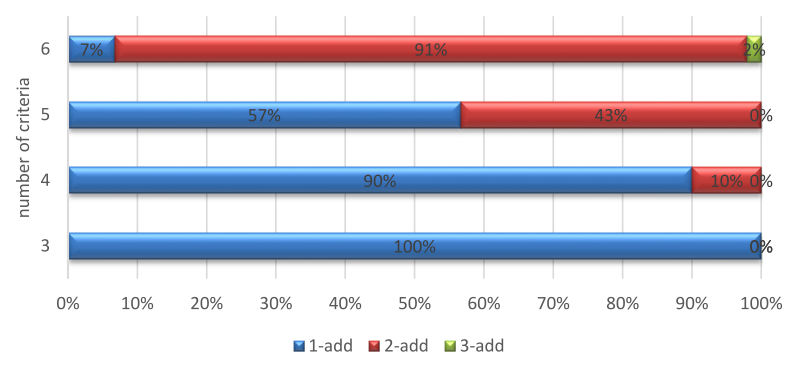}
    \caption{Percentage of inequivalent families of MSCs in $\Ck$ for $k=1, 2, 3$ }
    \label{fig:percentInequivCk}
\end{figure}

\begin{figure}
    \centering
    \includegraphics[width=132mm]{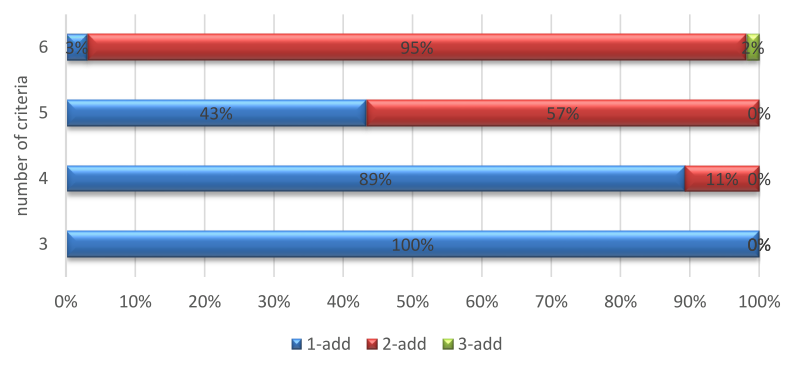}
    \caption{Percentage of all families of MSCs in $\Ck$ for $k=1, 2, 3$ }
    \label{fig:percentAllMSCs}
\end{figure}

The information displayed in Table \ref{ta:2-3-additiveRn} (\resp \ref{ta:2-3-additiveDn}) is represented in graphical form in Figure \ref{fig:percentInequivCk} (\resp \ref{fig:percentAllMSCs}). The cases of 0, 1 and 2 criteria are not represented; all families belong to class \Ck[1]. These figures clearly show that the proportion of families that can be represented by means of a 1-additive capacity, \ie by additive weights, decreases quite rapidly with the number of criteria. In contrast, the proportion of families that can be represented by a 2-additive capacity grows up to more than 91\% from $n=3$ to $n=6$. The proportions slightly differ depending on whether only inequivalent families or all families are taken into account. One can observe that the proportion of families in class \Ck[2] is a bit larger when considering all families (Table \ref{ta:2-3-additiveDn} and Figure \ref{fig:percentAllMSCs}). Class \Ck[3] appears for $n = 5$ and the proportion of MSCs in \Ck[3] grows up to around 2\% for $n=6$. One may expect that the proportion of MSCs in class \Ck[2] will eventually decrease and the share of classes $\Ck$ for $k\geq 3$ will increase when $n$ grows.

\newpage
\paragraph{Examples. } In order to illustrate, we describe some examples for $n=4$ and $n=6$. The list of all inequivalent MSCs for $n=5$, which are not representable by a $1$-additive capacity, is displayed in appendix (Section~\ref{appendix-ineq_msc_n5}). The categorization in classes $\Ck$ is available at \url{http://olivier.sobrie.be/shared/mbfs/} for $n=4,5,6$.

\ssk \noindent $\mathbf{n=4}.$  Here are the three families of MSCs on 4 elements that cannot be represented using a 1-additive capacity (they can be represented by a 2-additive capacity).
      \begin{center}
\renewcommand{\arraystretch}{1}
      \begin{tabular}{llc}
    Type & Representative & \# equivalent \\
    \hline
(0,2,0,0) & \{12, 34\} & 3 \\
(0,3,0,0) & \{13, 12, 34\} & 12\\
(0,4,0,0) & \{13, 14, 23, 24\} & 3 \\
     \end{tabular}
      \end{center}
These three inequivalent families yield, by permutations of the criteria labels, a total of 18 families that can only be represented using a 2-additive capacity.

\ssk The last inequivalent family is precisely the example that we used in Section \ref{sec:MRSortLimitations} to show that not all families of SCs can be represented by a 1-additive capacity. In contrast, it can be represented, for instance, by setting $m_1 = m_2 = m_3 = m_4 = 1/6$ and $m_{13} = m_{14} = m_{23} = m_{24} = 1/12$, while the other pairwise interactions $m_{12}$ and $m_{34}$ are set to 0. We then have: $\mu(13) = \mu(14) = \mu(23) = \mu(24) = 5/12$ while $\mu(12) = \mu(34) = 4/12$. Setting the threshold $\lambda$ to $9/24$ allows to separate the sufficient from the insufficient  coalitions. Such a representation is never unique. For example, another capacity is obtained by setting $m_1 = m_2 = m_3 = m_4 = 1/3$, $m_{12} = m_{34} = -1/6$ and $m_{13} = m_{14} = m_{23} = m_{24} = 0$. We have: $\mu(13) = \mu(14) = \mu(23) = \mu(24) = 2/3$ while $\mu(12) = \mu(34) = 1/2$. Setting the threshold $\lambda$ to $7/12$ also separates the sufficient from the insufficient coalitions.


\ssk \noindent Note that the first and the last example are \textit{complementary} in an obvious sense. These families form a bipartition of the set of all pairs of elements. This complementarity is related to the first property in \citeauthor{stephenyusun2014} Lemma~\ref{le:stephenYusun} allowing to speed up the enumeration of the families of MSCs. In view of the 2-additive capacity used in the third example, it is straightforward to build a 2-additive capacity for the first example. Therefore, both families belong to class \Ck[2].

\bsk \noindent $\mathbf{n=6}. $ We give two examples of inequivalent families of MSCs on 6 elements that are not representable by a 2-additive capacity but require a 3-additive capacity. There are 338 such inequivalent families which yield, through permutations, a total of 145\,502 families. Note that if all permutations of the elements labels were yielding different families, the total number of families would be $338\times 720 = 243\, 360$. A simple example of type (0,0,4,0,0,0) is the following family of MSCs:  $\{136, 234, 125, 456\}$. There are 30 equivalent families that can be derived from this family through permutations. Another, much more complex example is of the type (0,1,7,1,0,0). The MSCs are $\{135, 256, 345, 36, 234, 456, 1245, 146, 123\}$. There are 360 equivalent families that can be obtained through permutations.

\ssk \noindent Among the 338 families belonging to $\Ck[3]$, no MSC consists of a single criterion; none of them involves 5 elements. In other words, they are of type $(0,k_2, k_3, k_4, 0,0)$. The largest number of MSCs in such a family is 16, the maximal cardinality of a family of MSCs on 6 elements being the Sperner number 20.

\section{What about \texorpdfstring{$n>6$}{n>6} ?}
\cite{stephenyusun2014} have computed the number of inequivalent families of MSCs up to $n=7$. Categorizing the families of MSCs in classes \Ck{} for $n=7$, with the same method we used in the case $n \leq 6$ cannot be envisaged. We do not only have to generate and store the inequivalent families of MSCs. In addition, for each family, we have to compute the maximal insufficient coalitions, write and solve usually more than one linear program in order to determine the minimal value of $k$ for which the family belongs to $\Ck{}$. The order of magnitude of the time needed for categorizing each family out of the 16\,353 families for $n=6$ is about 1 second, which amounts to a total computing time of more than 4 hours. Since, the number of inequivalent families of MSCs for $n=7$, $R(7)$, is more than 30\,000 times $R(6)$, using the same method is too time-consuming. 



\section{Correspondences}\label{sec:gamesBooleanFct}

It should be noted that determining whether a partition of $2^N$ in two classes can be represented by weights and a threshold is a problem that arises in at least two other branches of science. One is the theory of monotone Boolean functions and the other is the theory of simple games.

\subsection{Monotone Boolean functions}\label{ssec:monotBoolFct}

A Boolean function (BF) is a function $f: 2^N =\{0,1\}^n \rightarrow \{0,1\}$, which associates the value 0 or 1 to each $n$-dimensional Boolean vector. The Boolean vectors assigned to 1 (\resp 0) are called \emph{true} (\resp \emph{false} points of the function $f$). 

\ssk The Boolean function is \emph{monotone} (also called \emph{positive}) if $f(x) \geq f(y)$ whenever $x \geq y$, where $x,y$ are Boolean vectors of $\{0,1\}$ and $\geq$ is the partial order on $\{0,1\}^n$ (defined by $x = (x_1, \ldots, x_n) \geq y = (y_1, \ldots, y_n)$, with $x_i, y_i \in \{0,1\}$, if $x_i \geq y_i$, for all $i =1, \ldots, n$). A monotone Boolean function (mBF) is determined by the set of its \emph{minimal} true points \citep[Theorems 1.13 and 1.26]{CramaHammer2011}. A true point $x$ is minimal if there is no other true point $y$ with $x \geq y$. A mBF is alternatively characterized by its set of maximal false points.  

\ssk Identifying the set of sufficient coalitions in a set of criteria is equivalent to identify a mBF. Subsets of criteria of the set $N$ can be represented by a Boolean vector $x=(x_1, \ldots, x_i, \ldots,  x_n)$ with $x_i = 1$ iff criterion $i$ belongs to the coalition. The partial order $\geq$ on $\{0,1\}^n$ corresponds to the inclusion $\supseteq$ between coalitions.  The set of minimal true (\resp maximal false) points of a mBF corresponds to the set of minimal sufficient (\resp maximal insufficient) coalitions. 

\ssk Since minimal true points are not comparable in the partial order $\geq$, they form antichains of the partially ordered set $(\{0,1\}^n, \geq)$. Therefore, the Dedekind numbers $D(n)$ are the numbers of mBFs in $n$ variables. 

A \emph{threshold} monotone Boolean function is a mBF for which the set of true and false points can be linearly separated, \ie, there are weights $w_i$ and a threshold $\lambda$ such that the true points are exactly those that satisfy:
$$
\sum_{i=1}^n w_i x_i \geq \lambda.
$$
For instance, if a student passes in case he succeeds in at least three subjects among French, English, Mathematics, Physics, this situation can be modelled by a threshold Boolean function $f$. The minimal true points are the four subsets of three subjects. Assigning the weight $w_i = 1/4$ to each subject and setting the threshold $\lambda = 3/4$, we have that the true points are exactly those for which 
$$
\sum_{i=1}^n \frac{1}{4} x_i \geq 3/4.
$$    
On the contrary, in the case of Example~\ref{ex:nonMR2Add}, the successfull students can not be separated linearly from the unsuccessfull ones. The corresponding mBF is not a threshold mBF. 

In the context of BFs, the categories of \emph{polynomial threshold functions of degree $k$}, for $k \in N$ generalize the threshold functions. Instead of separating the true points from the false points of a BF by a linear function, more general separating functions have been considered, namely  multilinear polynomials of degree $k$. These are functions $p(x)$ of the form
\begin{equation}\label{eq:multilinPoly}
  p(x) = \sum_{A \subseteq 2^N; |A| \leq k} c(A) \prod_{i \in A} x_i,
\end{equation}
where $c(A)$ are coefficients that vanish for subsets $A$ of $N$ of cardinality superior to $k$. For any mBF $f$, there is a degree $k \in N$ such that the true points and false points of $f$ can be separated by a multilinear polynomial $p(x)$ of degree $k$ \citep[p. 407]{CramaHammer2011}, in the following sense:
\begin{equation}\label{eq:polyThresholdBF}
  f(x)= 1 \quad \text{iff} \quad p(x) \geq 0.
\end{equation}
The BF $f$ is a \emph{polynomial threshold function of degree $k$} if $k$ is the smallest possible degree of a function $p(x)$ such that \eqref{eq:polyThresholdBF} holds (for more on this, see \cite{Anthony2010,Anthony2001,Bruck2010,Saks1993}). Polynomial threshold BFs of degree 1 are just threshold BFs. All the latter are monotone. Polynomial threshold BFs of degree $k>1$ are not necessarily monotone.

\ssk In the case $f$ is a monotone BF, its true points can be separated from its false points by a \emph{monotone nondecreasing} (pseudo-Boolean) polynomial of degree $k$, for some $k \in N$. Actually, this is equivalent to separating the true from the false points by a capacity $\mu$ and a threshold $\lambda$. We have that a positive BF $f$ is separable in the sense of \eqref{eq:polyThresholdBF} by a \emph{monotone nondecreasing} (pseudo-Boolean) polynomial $p$ of degree $k$ (for some $k \in N$) iff it is separable by a $k$-additive capacity $\mu$ and a threshold $\lambda$  in the following sense: for  $x = 1_A$, the Boolean vector corresponding to the set $A$,
\begin{equation}\label{eq:ineqCapa}
   f(x)= 1 \quad \text{iff} \quad \mu(A) \geq \lambda.
\end{equation}
This result is proved in Proposition \ref{prop:kAddBF_kAddCapa}.

\begin{definition}[$k$-additive Boolean function]\label{def:kAdditiveBF} A positive Boolean function $f$ is called $k$-additive if $k$ is the smallest integer such that the true points of $f$ can be separated from its false points by a monotone non-decreasing pseudo-Boolean polynomial of degree $k$.
\end{definition}
A $1$-additive Boolean function is a threshold function.

\begin{proposition}\label{prop:kAddBF_kAddCapa}
Let $f$ be a monotone BF. The following equivalence holds: for some $k \in N$,
\begin{enumerate}
  \item $f$ is separable by a \emph{monotone nondecreasing} (pseudo-Boolean) polynomial $p$ of degree $k$, in the sense that $f(x) =1 \ \text{iff} \ p(x) \geq 0$;
  \item $f$ is separable by a $k$-additive capacity $\mu$ and a threshold $\lambda$,  in the sense that, for $x = 1_A$,   $f(x)= 1 \ \text{iff} \ \mu(A) \geq \lambda$.
\end{enumerate}    
\end{proposition}

\bpr
Assume 2. If $m$ is the Möbius transform of the $k$ additive capacity $\mu$, we have $\mu(A) = \sum_{B \subseteq A; |B| \leq k} m(B)$. The $k$-additive capacity $\mu$ corresponds to the monotone pseudo-Boolean polynomial $p(x)$ of degree $k$ defined by $p(x) = \sum_{B\subseteq N; |B| \leq k} m(B) \prod_{i \in B} x_i$. For $x=1_A$, $p(x)= \mu(A)$ and we have $\mu(A) \geq \lambda$ iff $p(x) \geq \lambda$. It is easy to modify the polynomial $p$ into another polynomial $p'$ in order to have $p(x) \geq \lambda$ iff $p'(x) \geq 0$. To do this, we translate $p$ by adding $-\lambda$. Actually, we change the value $m(\emptyset) = 0$ into $m'(\emptyset) = -\lambda$. The obtained polynomial $p'$ is also monotone nondecreasing.

\ssk Starting from hypothesis 1., \ie that $f(x) =1 \ \text{iff} \ p(x) \geq 0$, with $p$ a monotone nondecreasing polynomial of degree $k$, we add $\lambda= - p(0)$, the value of the characteristic function of the emptyset, to both sides of the latter inequality, getting $p(x) - p(0) \geq \lambda$. The righthand side is a monotone non-decreasing pseudo-Boolean function which vanishes on the null vector. The set function $\mu$ defined by $\mu(A) = p(1_A) -p(0)$ is a capacity. Since $p(x) - p(0)$ has a representation as a multilinear polynomial of degree $k$, the corresponding capacity is $k$-additive.

\ssk Note that the trivial cases in which all coalitions are sufficient or none of them is are separable by a polynomial of degree 0 (i.e., a constant) or a 0-additive capacity (i.e., the null constant).
\epr

\subsection{Simple games theory}\label{ssec:simpleGames}

In the theory of simple games, $N$ is a set of players (or voters, or DMs). They can form coalitions in order to achieve some goal. The power of a coalition represents the ability of the set of players to achieve this goal. For instance, at the general assembly of a company shareholders, decisions are made through votes. Each shareholder has a voting power proportional to her share. A decision is adopted if the sum of the shares of the shareholders supporting the decision reaches 50\%.  

\ssk In general, a simple game $v$ (also called 0-1 capacity \cite[p.~42]{GrabischBook2016}) is a monotone nondecreasing set function on $N$, which only takes the values 0 or 1 and is such that $v(\emptyset) = 0$.

\ssk Simple games correspond to mBFs which take value 0 on the null binary vector. The sets on which the game $v$ has value 1 (which correspond to the true points of the associated positive BF), are called the \emph{winning coalitions} of the game. They form an upset of $2^N$. The set of sufficient coalitions of a NCSM is the set of winning coalitions of a simple game (except in the trivial case of a NCSM for which all coalitions are sufficient). \emph{Weighted} simple games correspond to \mrsort{} models and to threshold BFs. \cite{Grabisch1997} has introduced the notion of $k$-additive games (see also \cite[p. 73]{GrabischBook2016}). The winning coalitions of $k$-additive simple games are the coalitions that fulfill $\mu(A) \geq \lambda$ for some $k$-additive capacity $\mu$ and a threshold $\lambda$.

\ssk In game theory, MSCs correspond to minimal \emph{winning} coalitions. The number of simple games with $n$ players in minimal winning form \citep{vonNeumanMorgenstern43,KurzTautenham2013} equals $D(n)-1$ (the empty set is excluded as a winning coalition by definition of a simple game).

\section{Chapter conclusion}\label{sec:concluDAM}
The main goal of this chapter is to assess the expressivity gap between the \mrsort{} and the NCS models. Up to $n=3$ criteria, the two models are equivalent since all families of MSCs can be represented by criteria weights and a threshold. This is no longer the case from $n=4$ on. Moreover, the gap between these two models increases very rapidly with $n$. For $n=6$, only a small percentage of NCS models are \mrsort models: 6.81 \% (if we consider the inequivalent families of MSCs) or 3.12 \% (if we consider all families of MSCs) (see Tables \ref{ta:2-3-additiveRn} or \ref{ta:2-3-additiveDn}). In order to increase the descriptive strength of the model, it is interesting to use 2 and 3-additive capacities. However, for larger (than 6) values of $n$, it is likely that 2- and 3-additive capacities will not be sufficient to describe a large part of the cases.  

\ssk  Possible extensions of model \mrsort{} (or a restriction of model NCS) are suggested by the above study, \ie working with 2-additive or 3-additive NCs models. 

\ssk  In this study, the families of minimal sufficient coalitions are explored as they appear in sorting models such as \mrsort{} and NCS. One of the objective is to give a detailed picture of the possible families of SCs for as many criteria as we could. This information could help further investigations related to the characterization of the families of SCs that can be separated from the insufficient ones by an inequality involving a $k$-additive capacity. In other words, this could help in the study of $k$-additive BFs. Our second objective is to make a list of the possible sorting rules in the \ncs{} model, in order to enable to draw a rule at random according to any specified probability distribution and use it in simulations. By retaining only inequivalent rules we can reduce the space to store these lists and the scanning time. 

\ssk  After that, we can try to list inequivalent families of SCs for $n=7$, which will probably require a theoretical study of the different classes \Ck. Different approaches to subdividing the set of all families of SCs could also be of interest.

\section[Appendix: Inequivalent families for \texorpdfstring{$n=4, 5$}{n=4 or 5}]{Appendix: Inequivalent families of minimal sufficient coalitions for \texorpdfstring{$n=4, 5$}{n=4 or 5}}\label{appChap7:inequivFamN4-5}

\paragraph{List of
inequivalent families of monotone sufficient coalitions for $n=4$.}
\label{appendix-ineq_msc_n4}

The families are grouped by type. There are 25 possible types, 29 inequivalent families of \msc{} (plus the trivial case in which all coalitions are sufficient) and 167 families of \msc{} (plus the same trivial case).
Each inequivalent family in the list is associated the size of its equivalence class. All inequivalent families, except three of them, can be represented by a $1$-additive capacity. The three other families can be represented by a $2$-additive capacity. They are marked in the last column by \Ck[2].

\begin{table}
\centering
\caption{List of inequivalent MSCs of $n=4$ \& number of
equivalence for each class. Inequivalent MSC that are not representable by a 1-additive capacity (weighted sum) are pointed out in the last column.
}
\renewcommand{\arraystretch}{1}
\begin{tabular}{llll}
	\toprule
	Type & Family of \msc{} & \# eq. & \Ck \\
	\midrule
	(0,0,0,0) & \{\} &1 & \\
	(0,0,0,1) & \{1234\} & 1 & \\
	(0,0,1,0) & \{124\} & 4 & \\
	(0,0,2,0) & \{234, 124\} & 6 & \\
	(0,0,3,0) & \{134, 123, 124\} & 4 & \\
	(0,0,4,0) & \{134, 123, 234, 124\} & 1 & \\
	(0,1,0,0) & \{24\} & 6 & \\
	(0,1,1,0) & \{14, 123\} & 12 & \\
	(0,1,2,0) & \{24, 134, 123\} & 6 & \\
	(0,2,0,0) & \{12, 23\} & 12 & \\
	          & \{23, 14\} & 3 & \Ck[2] \\
	(0,2,1,0) & \{24, 134, 23\} & 12 & \\
	(0,3,0,0) & \{13, 12, 34\} & 12 & \Ck[2]\\
	          & \{24, 12, 14\} & 4 & \\
	          & \{24, 34, 14\} & 4 & \\
	(0,3,1,0) & \{13, 34, 23, 124\} & 4 & \\
	(0,4,0,0) & \{24, 12, 13, 34\} & 3 & \Ck[2] \\
	          & \{24, 12, 14, 23\} & 12 & \\
	(0,5,0,0) & \{24, 12, 14, 13, 34\} & 6 & \\
	(0,6,0,0) & \{24, 12, 14, 34, 23, 13\} & 1 & \\
	(1,0,0,0) & \{1\} & 4 & \\
	(1,0,1,0) & \{234, 1\} & 4 & \\
	(1,1,0,0) & \{14, 2\} & 12 & \\
	(1,2,0,0) & \{13, 34, 2\} & 12 & \\
	(1,3,0,0) & \{24, 34, 23, 1\} & 4 & \\
	(2,0,0,0) & \{4, 3\} & 6 & \\
	(2,1,0,0) & \{4, 23, 1\} & 6 & \\
	(3,0,0,0) & \{4, 2, 1\} & 4 & \\
	(4,0,0,0) & \{4, 2, 3, 1\} & 1 & \\
	\bottomrule
\end{tabular}
\end{table}
\FloatBarrier

\paragraph{List of inequivalent families of
monotone sufficient coalitions for class \Ck[2] for $n=5$.}
\label{appendix-ineq_msc_n5}

We list below the 91 inequivalent families of MSC that cannot be
represented by a $1$-additive capacity.
They can all be represented using a $2$-additive capacity.
The families are grouped by type. Each inequivalent family in the list is associated the size of its equivalence class.

\renewcommand{\arraystretch}{1}
\begin{longtable}{llr}
\caption{List of inequivalent families of MSC in class \Ck[2] for
$n=5$ and number of equivalence for each class.}\\
\toprule
Type & Family of \msc{} & \# eq.\\
\midrule
\endfirsthead
\multicolumn{3}{c}{\tablename\ \thetable\ --
\textit{Continued from previous page}} \\
\midrule
Type & Family of \msc{} & \# eq.\\
\midrule
\endhead
\bottomrule
\multicolumn{3}{r}{\textit{Continued on next page}} \\
\endfoot
\bottomrule
\endlastfoot
	(0,0,2,0,0)
	& \{135, 234\} & 15  \\

	(0,0,2,1,0)
	& \{234, 125, 1345\} & 15 \\

	(0,0,3,0,0)
	& \{145, 123, 345\} & 30 \\
	& \{235, 234, 125\} & 60 \\

	(0,0,3,1,0)
	& \{134, 135, 2345, 124\} & 60 \\

	(0,0,4,0,0)
	& \{145, 234, 345, 124\} & 15 \\
	& \{135, 245, 234, 125\} & 60 \\
	& \{235, 145, 135, 123\} & 60 \\
	& \{134, 345, 234, 125\} & 10 \\

	(0,0,4,1,0)
	& \{245, 123, 234, 125, 1345\} & 15 \\

	(0,0,5,0,0)
	& \{235, 134, 135, 345, 125\} & 60 \\
	& \{235, 134, 135, 245, 124\} & 12 \\
	& \{235, 145, 134, 245, 124\} & 60 \\
	& \{145, 134, 123, 234, 125\} & 60 \\

	(0,0,6,0,0)
	& \{135, 235, 234, 125, 145, 123\} & 15 \\
	& \{135, 345, 234, 125, 245, 123\} & 10 \\
	& \{345, 235, 234, 125, 124, 134\} & 60 \\
	& \{135, 345, 235, 125, 124, 145\} & 60 \\

	(0,0,7,0,0)
	& \{345, 234, 125, 145, 134, 245, 123\} & 30 \\
	& \{135, 235, 125, 124, 145, 134, 245\} & 60 \\

	(0,0,8,0,0)
	& \{135, 345, 234, 125, 124, 145, 245, 123\} & 15 \\

	(0,1,1,0,0)
	& \{123, 45\} & 10 \\

	(0,1,2,0,0)
	& \{15, 123, 345\} & 60 \\
	& \{12, 134, 345\} & 60 \\

	(0,1,3,0,0)
	& \{235, 14, 123, 125\} & 60 \\
	& \{13, 235, 145, 124\} & 60 \\
	& \{235, 14, 123, 245\} & 60 \\
	& \{24, 134, 135, 123\} & 30 \\

	(0,1,4,0,0)
	& \{235, 15, 245, 123, 234\} & 120 \\
	& \{135, 123, 25, 345, 124\} & 60 \\
	& \{235, 34, 145, 125, 124\} & 60 \\
	& \{24, 235, 135, 123, 125\} & 20 \\

	(0,1,5,0,0)
	& \{345, 235, 15, 234, 134, 123\} & 30 \\
	& \{235, 125, 124, 145, 34, 123\} & 60 \\
	& \{24, 135, 345, 235, 125, 123\} & 60 \\

	(0,1,6,0,0)
	& \{24, 135, 345, 235, 145, 134, 123\} & 60 \\

	(0,2,0,0,0)
	& \{34, 15\} & 15 \\

	(0,2,1,0,0)
	& \{12, 35, 234\} & 60 \\
	& \{145, 23, 25\} & 60 \\

	(0,2,2,0,0)
	& \{24, 13, 125, 345\} & 30 \\
	& \{24, 12, 135, 345\} & 30 \\
	& \{134, 23, 35, 124\} & 60 \\
	& \{13, 12, 245, 234\} & 120 \\
	& \{12, 245, 35, 234\} & 60 \\

	(0,2,3,0,0)
	& \{15, 23, 134, 345, 124\} & 60 \\
	& \{45, 134, 135, 234, 25\} & 120 \\
	& \{135, 123, 45, 125, 14\} & 60 \\
	& \{24, 235, 14, 345, 135\} & 30 \\
	& \{24, 34, 135, 123, 125\} & 60 \\

	(0,2,4,0,0)
	& \{135, 235, 14, 234, 123, 45\} & 60 \\
	& \{14, 35, 234, 125, 245, 123\} & 15 \\
	& \{24, 135, 235, 125, 34, 123\} & 30 \\

	(0,3,0,0,0)
	& \{12, 14, 45\} & 60 \\
	& \{12, 34, 45\} & 30 \\

	(0,3,1,0,0)
	& \{24, 145, 23, 25\} & 60 \\
	& \{34, 14, 35, 125\} & 60 \\
	& \{34, 245, 23, 14\} & 120 \\
	& \{34, 14, 123, 25\} & 60 \\

	(0,3,2,0,0) & \{15, 14, 123, 25, 345\} & 60 \\
	& \{24, 12, 134, 35, 145\} & 30 \\
	& \{13, 23, 245, 125, 14\} & 120 \\
	& \{15, 45, 123, 234, 25\} & 60 \\

	(0,3,3,0,0)
	& \{24, 135, 145, 134, 23, 25\} & 20 \\
	& \{12, 35, 234, 145, 13, 245\} & 60 \\

	(0,4,0,0,0)
	& \{34, 15, 14, 35\} & 15 \\
	& \{24, 15, 23, 25\} & 60 \\
	& \{24, 34, 15, 23\} & 10 \\
	& \{24, 34, 15, 35\} & 60 \\

	(0,4,1,0,0)
	& \{13, 34, 35, 25, 145\} & 60 \\
	& \{24, 13, 15, 25, 345\} & 60 \\
	& \{13, 15, 23, 25, 345\} & 30 \\
	& \{34, 14, 45, 125, 23\} & 60 \\

	(0,4,2,0,0)
	& \{24, 12, 35, 145, 134, 23\} & 60 \\
	& \{24, 35, 145, 34, 25, 123\} & 15 \\

	(0,5,0,0,0)
	& \{24, 13, 15, 23, 14\} & 60 \\
	& \{24, 12, 15, 35, 25\} & 60 \\
	& \{24, 12, 15, 35, 34\} & 12 \\
	& \{12, 15, 34, 25, 45\} & 60 \\

	(0,5,1,0,0)
	& \{135, 12, 14, 34, 23, 25\} & 60 \\
	& \{15, 35, 124, 23, 13, 45\} & 60 \\

	(0,6,0,0,0)
	& \{24, 12, 23, 25, 13, 45\} & 15 \\
	& \{24, 12, 35, 34, 25, 13\} & 10 \\
	& \{24, 12, 34, 23, 13, 45\} & 60 \\
	& \{15, 14, 34, 23, 25, 45\} & 60 \\

	(0,6,1,0,0)
	& \{24, 12, 35, 145, 34, 25, 13\} & 10 \\

	(0,7,0,0,0)
	& \{12, 14, 34, 23, 25, 13, 45\} & 30 \\
	& \{24, 12, 15, 14, 35, 34, 45\} & 60 \\

	(0,8,0,0,0)
	& \{24, 12, 15, 34, 23, 25, 13, 45\} & 15 \\

	(1,2,0,0,0)
	& \{34, 15, 2\} & 15 \\

	(1,3,0,0,0)
	& \{24, 15, 3, 25\} & 60 \\

	(1,4,0,0,0)
	& \{13, 2, 14, 35, 45\} & 15 \\
\end{longtable}




\chapter{Taking a step back 
}\label{chap:recul}

This work has concentrated on designing direct rigorous methods for eliciting the parameters of a NCS model. We furthermore limited ourselves to NCS models with two categories and no veto. In this chapter, we try to open up the picture in two directions. One is to see how the rigorous methods, often inapplicable in practice for different reasons, could help in real decision aiding processes. In another direction, we try to outline how the basic methods developed can be integrated in more complex settings, considering \eg vetoes and multiple categories. 

\ssk The chapter has three sections. First of all, a didactic example (a car selection example) is taken in order to show that the rigorous procedures can serve as guidelines to build a sorting model. Then we address the issues of insufficient, contradictory and imprecise information on the model's parameters in the model construction process. Finally, we suggest how the methods can be adapted to integrate vetoes and multiple categories. We also mention an adaptation to hierarchically structured criteria.
All these aspects are illustrated by using the car selection example. 

\section{How to use what has been done}
In most practical modelling contexts, a completely rigorous elicitation procedure cannot be followed. There are many possible causes for that. An important one is the availability of the Decision-Maker (DM). Rigorous procedures tend to use a large number of questions. Even though the DM's commitment may be total, tiredness inevitably hinders the process in case too much cognitive effort is requested.

\ssk With a model like NCS, the main source of strain is related to the number of criteria. At this stage, we would like to discuss the limits of using the rigorous techniques proposed and analyzed in this work. And in case it is not possible to use such methods, can we get inspired by rigorous procedures to propose ``near-rigorous'' ones? By this, we mean departing as little as possible from the rigorous procedures or making some simplifying hypotheses that allow to reduce the size of the problem and make it compatible with rigorous methods. We first focus on the elicitation of the sufficient coalitions, which is the most complex question when using the NCS model.

\subsection{Cases in which rigorous methods are applicable}
For several situations in real-world problems, simplifying tools can be mixed with some available information in order to base the problem on rigorous methods.
\begin{itemize}
    \item Up to 6 criteria, using DFSHansel can be envisaged. The number of questions may grow up to 35 in the worst case, which can be considered close to an upper limit of a DM's patience. 
    \item If the presence of an order on the criteria importance can be assumed (and if criteria coalitions can be assumed to be monotone \wrt this order), starting with asking for such an order (or partial information comparing coalitions importance) reduces the number of needed questions. Such a possibility is implemented in the DFS3 questioning tool. An example to the functioning of the DFS3 with a reduction in the number of questions for $n=10$ criteria is shown below. A situation having the additional information $10>9>8>7>6>5>4>3>2>1$ and for \Fmin $= \{ 345678910, 1245678910 \}$, DFS3 asks $7$ questions to the DM which is an acceptable number of questions for ten criteria.
\end{itemize}

\begin{verbatim}
C:\12nis10\bin>java eda.main
Enter the number of criteria
10
Are there any known importance
y
Enter Importance like: '1>2>3>4' or '2 3>1 3' or '0' to end
10>9>8>7>6>5>4>3>2>1
Enter Importance like: '1>2>3>4' or '2 3>1 3' or '0' to end
0
is 6 7 8 9 10 sufficient?
n
is 5 6 7 8 9 10 sufficient?
n
is 4 5 6 7 8 9 10 sufficient?
n
is 3 4 5 6 7 8 9 10 sufficient?
y
is 2 4 5 6 7 8 9 10 sufficient?
n
is 1 2 4 5 6 7 8 9 10 sufficient?
y
is 1 2 3 5 6 7 8 9 10 sufficient?
n
Total Number of Questions:7
Q1:6 7 8 9 10==>answer:N
Q2:5 6 7 8 9 10==>answer:N
Q3:4 5 6 7 8 9 10==>answer:N
Q4:3 4 5 6 7 8 9 10==>answer:Y
Q5:2 4 5 6 7 8 9 10==>answer:N
Q6:1 2 4 5 6 7 8 9 10==>answer:Y
Q7:1 2 3 5 6 7 8 9 10==>answer:N
Fmin:{3 4 5 6 7 8 9 10 , 1 2 4 5 6 7 8 9 10}
\end{verbatim}

\subsection{Simplifying hypotheses and tools}\label{ssec:simplifyTools}
In order to reduce the length of the questioning procedures, general questions can be used. 

\ssk Assume that the DM feels comfortable with stating satisfactory levels on each criterion \ie levels that are characteristic of alternatives in the ``Good'' category. Similarly, we assume that the DM is ready to state unsatisfactory levels on each criterion, \ie levels that are not characteristic of ``Good'' alternatives. We can assume therefrom the lower-limit profile of the ``Good'' category lies between the vectors of satisfactory and unsatisfactory levels. An alternative that reaches the satisfactory levels on all the criteria can be considered as ``Good''. An alternative that has unsatisfactory levels on all criteria is ``Bad''. We may then start asking questions such as:
\begin{itemize}
    \item Is an alternative that reaches satisfactory levels on all but one criterion (\ie $n-1$ criteria) also ``Good'' ?
    \item If an alternative has satisfactory evaluations on less than half the number of criteria, can it be ``Good''?
    \item Are there criteria on which it is absolutely necessary to reach the satisfactory levels in order to be assigned to the ``Good'' category?
\end{itemize}
One cannot expect that such questions will always be answered by ``yes'' or ``no''. Sometimes, perhaps oftentimes, the answer will be ``It depends'' or ``I don't know''. However, if the answer to the first question is ``yes'', we know that the minimal sufficient coalitions have no more than $n-1$ criteria. We can then  try to ask the same question, with $n-2$ criteria. If the answer to the first question is ``no'', we can at least deduce that reaching the satisfactory level is required on some criteria. The same conclusion is drawn if the answer is ``It depends''. Regarding the second question, if the answer is ``no'', it means that sufficient coalitions involve at least $\lceil n/2 \rceil$ criteria. We can then ask, for instance, whether alternatives well below the satisfactory level on a quarter of the criteria can be ``Good''.  In case the answer to the second question is ``yes'' or ``it depends'', it means that some coalitions with less than $\lceil n/2 \rceil$ criteria are sufficient. 
For the third question, if the answer is ``yes'' and the DM lists the criteria on which a satisfactory level must be attained, we can conclude that these criteria belong to all sufficient coalitions. Let $l$ denote the number of these criteria. In this case, eliciting the sufficient coalitions is tantamount to elicit sufficient coalitions in a set of $n-l$ criteria. If the answer to the third question si ``no'', for each criterion $i$, there exists a sufficient coalition (hence a coalition in \Fmin{}) which doesn't contain the criterion $i$. 

\ssk The answers to such questions may determine the status, sufficient or insufficient, of a \emph{set} of coalitions, thereby simplifying the subsequent questioning procedure. 

\subsection{Example: the car selection problem} \label{ssec:carSelection}
Let us consider the problem of the car selection analyzed in \citet[][Chapter~6]{bouyssouetal2000}. We use this example to illustrate the use of simplifying tools, although the problem is small in size and could be tackled with rigorous methods.
The DM is a student who is fond of car rallies. Therefore he is mainly interested in performance and safety, while he wants to keep the cost as low as possible. He has 5 criteria which are respectively $c_1:$ cost, $c_2:$  acceleration, $c_3:$ pick-up, $c_4:$ brakes, $c_5:$ road-holding. Criterion $c_1:$ is the cost of a second hand car together with its usage cost (in \euro). The second criterion $c_2$ named as ``acceleration'' is the time (in seconds) required to make one kilometer starting from rest. The third criterion $c_3$ named as ``pick-up'', measures the time (in seconds) of pick up in traffic. The quality of braking and the quality of road-holding (assessed by experts on a qualitative scale) are represented as fourth and fifth criteria, respectively $c_4$ and $c_5$.

\ssk The decision problem is to assign cars to category ``Good'' or ``Bad'' (while in \citet{bouyssouetal2000} it is presented as a ranking problem). We note that, criteria 1, 2 and 3 are to be minimised while 4 and 5 have to be maximised. 

\begin{figure}[h!]
\centering
\includegraphics[width=130mm]{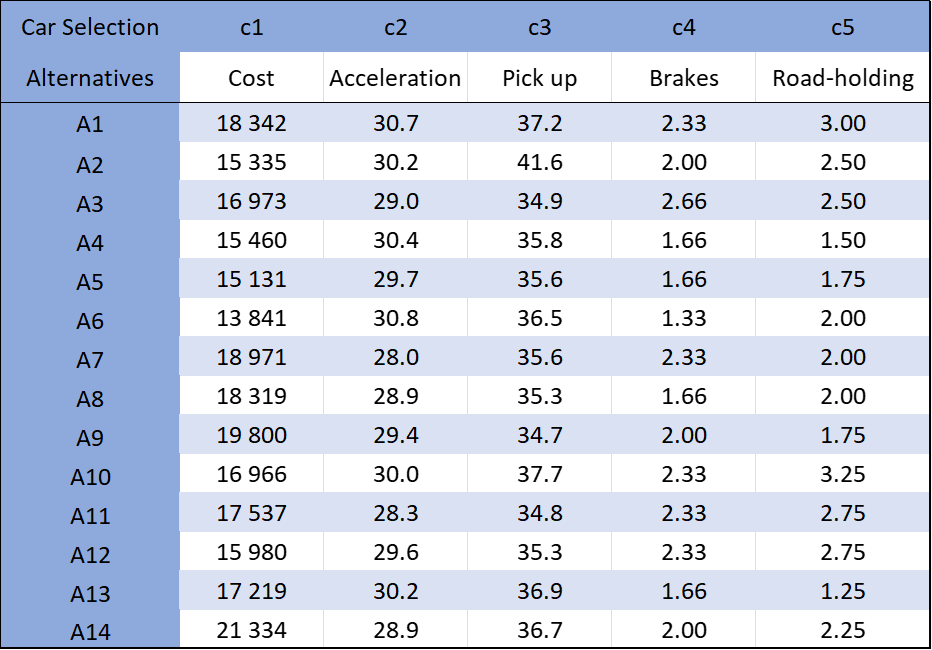}\caption{ Evaluation of the selected cars \citep[][Table 6.2, p. 94]{bouyssouetal2000}\label{fig:EvalSelectedCars}}
\end{figure}

The DM has selected 14 alternatives as potential buys. Their evaluations are displayed in Figure~\ref{fig:EvalSelectedCars}

\ssk The DM is able to provide satisfactory and unsatisfactory levels for each criterion. These values are in Table~\ref{ta:SatisUnsatisLevels}. 
\begin{table}[h!!!]
    \centering
\begin{tabular}{llllll}
\hline
    & $c_1$ & $c_2$ & $c_3$ & $c_4$ & $c_5$\\
    \hline
Satisfactory   & 16000 & 28.5  & 35 & 2.33 & 2.5 \\
Unsatisfactory & 19000 & 30  & 38 & 1.33 & 1.25\\ 
\hline
\end{tabular}
    \caption{Satisfactory and unsatisfactory values for all criteria}
\label{ta:SatisUnsatisLevels}
\end{table}

\ssk This means that a car with evaluations $(16000, 28.5, 35, 2.33, 2.5)$ or better is certainly ``Good'' while a car whose evaluations are $(19000, 30, 38,$\\$ 1.33, 1.25)$ is certainly ``Bad''.  The limit profile is thus in between the satisfactory and the unsatisfactory evaluations vectors (it must strictly dominate the unsatisfactory vector but it can share values with the satisfactory vector, or even be equal to the latter). We shall illustrate the procedure for eliciting the limit profile later. 

\subsection{Car selection problem: eliciting the sufficient coalitions}\label{ssec:carSuffCoal}
Since the problem is of small size, we could proceed with a questioning algorithm such as DFS3 or DFSHansel to obtain the list of sufficient coalitions (by asking questions in terms of cars performances). Instead, we use more general questions as suggested in Section~\ref{ssec:simplifyTools}.

\ssk We ask the previously suggested questions to the DM:
\begin{itemize}
    \item Is an alternative that reaches satisfactory levels on all but one criterion (\ie $5-1=4$ criteria) also ``Good'' ? We assume that the answer is ``YES''. There is thus no need to ask questions to the 4th and 5th levels of the decision tree. In other words, a car with satisfactory evaluations on 4 out of 5 criteria and an unsatisfactory evaluation on the other criterion is ``Good''. For instance, we know that $(16000, 28.5, 35, 2.33, 1.25)$ is ``Good''.  
    \item If an alternative has satisfactory evaluations on less than half the number of criteria (thus having satisfactory evaluations on at most $2$ criteria), can it be ``Good''? Assume that the DM's answer is``NO'', then coalitions involving less than 3 criteria are insufficient. At this stage the only remaining uncertainty about sufficient coalitions is : which coalitions of 3 criteria are sufficient, which are insufficient. We proceed with the third type of questions.
    
    \item Are there criteria on which it is absolutely necessary to reach the satisfactory levels in order to be assigned to the ``Good'' category?

    Assume that the DM says ``No; except the cost  $c_1$''. He might add: ``And there should be at least one satisfactory value on the car performance criteria $c_2, c_3$, and one on the safety criteria $c_4, c_5$''.  Such an answer would solve the problem. The minimal sufficient coalitions would then be: $\{124, 125, 134, 135\}$. The insufficient coalitions of 3 criteria would thus be $\{123, 145, 234, 235, 245, 345\}$. These cases, interpreted in terms of car can be presented to the DM for confirmation. 
\end{itemize}

\ssk In this simple case, the type of questions suggested in Section~\ref{ssec:simplifyTools} may suffice to solve the problem. In more complex cases or when the DM's answers are different, such questions may be insufficient but could contribute to reduce the set of undetermined coalitions. 

\ssk Note that in case we had used DFS3 instead of simplifying questions, the sequence of questions would have been as follows : 

\begin{verbatim}
 C:\12nis10\bin>java eda.main
Enter the number of criteria
5
Are there any known importance
n
is 3 4 5 sufficient?
n
is 2 4 5 sufficient?
n
is 2 3 5 sufficient?
n
is 2 3 4 5 sufficient?
n
is 1 4 5 sufficient?
n
is 1 3 5 sufficient?
y
is 1 3 sufficient?
n
is 1 3 4 sufficient?
y
is 1 2 5 sufficient?
y
is 1 2 sufficient?
n
is 1 2 4 sufficient?
y
is 1 2 3 sufficient?
n
Total Number of Questions:12
Q1:3 4 5==>answer:N
Q2:2 4 5==>answer:N
Q3:2 3 5==>answer:N
Q4:2 3 4 5==>answer:N
Q5:1 4 5==>answer:N
Q6:1 3 5==>answer:Y
Q7:1 3==>answer:N
Q8:1 3 4==>answer:Y
Q9:1 2 5==>answer:Y
Q10:1 2==>answer:N
Q11:1 2 4==>answer:Y
Q12:1 2 3==>answer:N
Fmin:{1 3 5 , 1 3 4 , 1 2 5 , 1 2 4}   
\end{verbatim}

\ssk The DFS3 questioning algorithm would require 12 questions to determine $\Fmin{}$. Note that the questions should not be asked in terms of coalitions but in terms of cars showing satisfactory or unsatisfactory evaluations on the appropriate criteria.

\ssk Using DFSHansel requires 20 questions, \ie the worst case, as shown below. The questions concentrate on the coalitions ``in the middle'' of the chains: 9 questions about coalitions of 2 criteria, 10 about coalitions of 3 criteria (\ie all coalitions of 3 criteria), and 1 question about a coalition of 4 criteria. In this example, DFSHansel is much less efficient than DFS3 (although it is more efficient on average under the hypothesis that all \Fmin{} are equally likely, see Figure~\ref{fig:comparisonExpected}). 

\begin{verbatim}
Enter number of criteria: 5
Is 10100 (S) or (I)? I
Is 10101 (S) or (I)? S
Is 10010 (S) or (I)? I
Is 10110 (S) or (I)? S
Is 11000 (S) or (I)? I
Is 11001 (S) or (I)? S
Is 01010 (S) or (I)? I
Is 11010 (S) or (I)? S
Is 01100 (S) or (I)? I
Is 11100 (S) or (I)? I
Is 10001 (S) or (I)? I
Is 10011 (S) or (I)? I
Is 01001 (S) or (I)? I
Is 01011 (S) or (I)? I
Is 00101 (S) or (I)? I
Is 01101 (S) or (I)? I
Is 00110 (S) or (I)? I
Is 01110 (S) or (I)? I
Is 00111 (S) or (I)? I
Is 01111 (S) or (I)? I

Question Count: 20
Fmin:  ['11010', '11001', '10110', '10101']
\end{verbatim}

\subsection{Car selection problem : eliciting the profiles}\label{ssec:carProfiles}
We elicited the sufficient coalitions; therefrom it is obvious to obtain the insufficient ones. It remains to elicit the limit profile.

Note that some cars can already be assigned to a category without precisely knowing the limit profile. Cars A9 and A14 (see Figure~\ref{fig:EvalSelectedCars}) can be directly assigned to the category \emph{Bad} because of their prices higher, \ie worse, than the unsatisfactory level $19\,000$ euros. Being unsatisfactory on the cost criterion entails assignment to category \emph{Bad}. Car A2 is also assigned to the category \emph{Bad}, because of its values on $c_2$ and $c_3$ are $30.2$ and $41.6$ while $30$ and $38$ are unsatisfactory values respectively. Being unsatisfactory on both criteria $c_2$ and $c_3$ entails assignment to category \emph{Bad}.

Assigning the other alternatives requires eliciting the limit profile. This can be done by the dichotomic procedure described in Section~\ref{sec:limitProfile}. We may apply it to the interval between the satisfactory and the unsatisfactory levels identified on each criterion (under the hypothesis that these values were reliably positioned above and below the limit profile). To elicit the value of the profile on criterion $i$, it is sufficient to choose an element $I$ of the set \Fmin{} containing criterion $i$ and according to this coalition of criteria construct the questions. For constructing the questioning procedure, we build an alternative which has satisfactory evaluations on the criteria in $I \setminus\{i\}$, the unsatistactory levels on the complement of $I$;  we then change the value of the criterion $i$ by dichotomy. We recall that the minimal sufficient coalitions are $\{124, 125, 134, 135\}$. 

 
\ssk For the first criterion $c_1$ which is the cost of the car, the profile value is situated between $16\,000$ and $19\,000$ euros. We decide to stop the dichotomic questioning when the uncertainty gap is less than $500$ euros or if there is not any car in this range. We choose to work with the minimal sufficient coalition $\{124\}$. The middle value of the $[16\,000, 19\,000]$ interval is $17\,500$. Thus, the first question for the chosen coalition of criteria is: ``Is a car evaluated by (17\,500, 28.5, 38, 2.33, 1.25) Good ?''. Assume the answer is ``yes''. The new uncertainty interval is $17\,500$ and $19\,000$ and we continue to the dichotomy for the first criterion. Then the DM is asked for the coalition of criteria with values of (18\,250, 28.5, 38, 2.33, 1.25). The DM finds $18\,250$ euros too expensive and says ``no''. The dichotomy now has to be done for the range $[17\,500, 18\,250]$ and the middle is $17\,875$. According to the DM, (17\,875, 28.5, 38, 2.33, 1.25) is ``Good'' and as the uncertainty interval $[17\,875, 18\,250]$ is lower than $500$, thus the procedure stops. Furthermore, we observe that there is no car cost situated in this range, we discuss with the DM and define the profile limit for the cost criterion $c_1$ as $18\,000$ euros (The choice of any particular value in the interval $[17\,875, 18\,250]$ has no impact of the assignment of the 14 cars).

\ssk The second criterion is the acceleration $c_2$ and it is also to be minimised. We decide to stop the dichotomy questioning as soon as the uncertainty is less than .2. The satisfactory and unsatisfactory levels were given by the DM as $28.5$ and $30$, respectively.  Consequently, we make a dichotomy for this interval and find $29.25$ as the middle value, which we round up to $29.3$. 
The first question for $c_2$ is: ``Is a car evaluated by (16\,000, 29.3, 38, 2.33, 1.25) Good ?''. The answer is ``No''. The new interval is $[28.5, 29.3]$ and the next question is asked for a car evaluated by (16\,000, 28.9, 38, 2.33, 1.25). The DM says such a car is not satisfactory. We now ask whether (16\,000, 28.7, 38, 2.33, 1.25) is satisfactory and the answer is ``Yes''. The uncertainty interval is $[28.7, 28.9]$. As the desired precision limit is reached, we stop the dichotomy and fix the profile value to 28.8. 

\ssk Working similarly for $c_3$, the road-holding criterion, let's assume that we finish with 35.5 for the profile value on $c_3$. 

\ssk The case of criteria $c_4$ and $c_5$ is a bit particular. These criteria scales are discrete. Criterion $c_4$, for instance, only takes values that are multiples of 1/3 (between 1.33 and 2.66). So there are only 5  possible values for the limit profile on $c_4$. The initial uncertainty interval is $[1.33, 2.33]$. The first question is thus: ``Is a car evaluated by (16\,000, 28.5, 38, 2, 1.25) Good ?''. The answer is ``No''. Then the only possibility for the limit profile value of the 4th criterion is 2.33. The case of $c_5$ is similar. We take 2.25 as the value of the limit profile on $c_5$.  

\begin{figure}[h!]
\centering
\includegraphics[width=130mm]{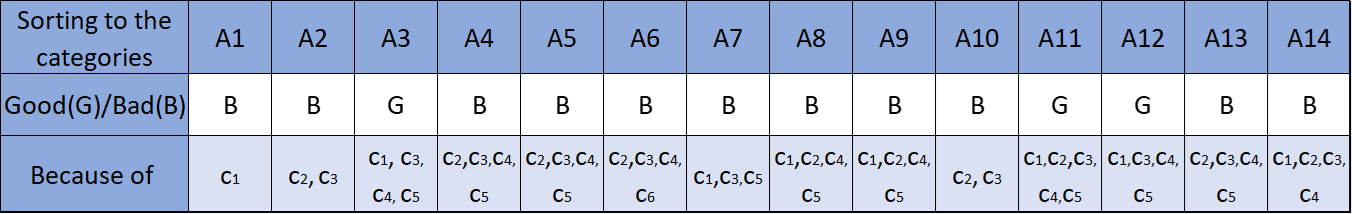}\caption{ Sorting the selected cars into the categories ``Good'' or ``Bad''; cars are assigned to the``Good'' or ``Bad'' category because the coalition of criteria in the last row is sufficient or not  \label{fig:SortingSelectedCars}}
\end{figure}

\ssk Finally the limit profile can be described by the vector (18\,000, 28.8, 35.5, 2.33, 2.25). According to this limit values of criteria, we can now assign the 14 alternatives to their corresponding categories. As shown in Figure~\ref{fig:SortingSelectedCars}, 14 selected cars are sorted according to their values on the 5 criteria. The criteria that were the reason of assignments are also indicated in Figure~\ref{fig:SortingSelectedCars}.  We conclude that 3 cars (A3, A11 and A12) are assigned to the category of `` Good '' and 11 cars (A1, A2, A4, A5, A6, A7, A8, A9, A10, A13 and A14)  are assigned to the category of `` Bad ''.

\section{Addressing imperfect information}\label{sec:imperfectInfo} 
Up to this point, we postulate that the DM's answers to questions are reliable, non contradictory and not subject to revision. Moreover, the DM has been assumed to answer all questions and never gets tired.  Of course, such difficulties often occur in practice. We shall only address two particular situations here. One is the case in which the DM is unable to answer some questions or hesitates in giving answers (incomplete information). The other is a form of contradiction in which the questioning precess reveals that the chosen model (here the NCS model) does not allow to represent the DM's preferences. 

\subsection{The case of incomplete information}\label{ssec:hesitation}

When questioning on sufficient coalitions, it may happen that the DM is unable to decide whether some coalitions are sufficient or not, especially when the number of criteria is large, the questioning process may become too long given the limited availability of the DM. At the end, coalitions with undetermined (S or I) character may remain. In this case, a software can be used to find a model compatible with the available data. 

\ssk Actually, works done in the framework of indirect elicitation or learning can be used. In particular, techniques based on  Linear Programming such as in \citet{sobrieetal2015b} can be used to find weights or 2-additive capacities compatible with the available information on sufficient and insufficient coalitions (formulated in terms of Good and Bad alternatives, respectively). We can also use a satisfiability problem formulation and a SAT solver to find a set $\calF$ of sufficient coalitions compatible with the available information as in \citet{TliliBKMO22}. Such approaches will also yield a compatible limit profile. Using the compatible model, it is possible to generate assignment examples that will be submitted to the DM for validation. Some examples may be validated, some others rejected. Another round of indirect elicitation may begin using the newly validated and unvalidated assignments. This approach may help to reduce the cognitive load on the DM and to make progress in the construction process of a model that represents the DM's preferences.  

\ssk A similar procedure can be applied also for profiles. In the example, we start from an object that is certainly satisfactory (\resp unsatisfactory) because it achieves satisfactory (\resp unsatisfactory) levels on all criteria. The limit profile lies between the certainly satisfactory and the certainly unsatisfactory objects. At a moment of the elicitation of the limit profile by dichotomy, the DM may not know how to specify further. In this case, we may take an ``average'' value in the uncertainty interval and show to the DM the consequences of this default choice. This can lead her/him to refine and to specify more precisely the value of the profile. 

\ssk Alternatively, one can keep the upper and the lower approximations of the profile and use the two corresponding models to produce assignments (actually, the lower approximation is known to be an unsatisfactory value; one should thus consider a slightly greater value instead).  Showing to the DM alternatives that are assigned differently by the two models, may cause the DM to choose one of the two models or specify further the limit profile.

\subsection{Model inadequacy}\label{ssec:inadequacy}
Before deciding to use model NCS (without veto), one has to check whether the DM's preferences are compatible with the NCS model logic. This can be informally discussed by explaining the principles of NCS. Namely, there are acceptable levels on each criterion and in order to be considered acceptable, an alternative must have acceptable levels on a sufficient subset of criteria. No compensation occurs. If an alternative is unacceptable while it has good values on some criteria, it cannot become acceptable by improving further the already good values. 

\ssk Conversely, an alternative that is acceptable while having bad values on some criteria cannot become unacceptable by worsening its already bad values. The latter principle is only true for the NCS model without veto. One should not insist on checking this condition from the beginning because it is not true in the NCS model with veto and such vetoes can be introduced later on if necessary (see below Section~\ref{ssec:veto}).

\ssk Assuming that the NCS model was chosen, in the course of the elicitation process, some tests may be done to verify that the NCS model under construction is adequate for representing the DM's preferences. 

\ssk In the example, for instance, assume that certainly satisfactory and certainly unsatisfactory levels were obtained for all criteria (see Table~\ref{ta:SatisUnsatisLevels} and that the sufficient coalitions have been elicited ($c_1$ and ($c_2$ or $c_3$) and ($c_4$ or $c_5$)). The limit profile is not yet known. Consider an alternative that is certainly Bad such as A2 : it is not acceptable because it is worse than the unsatisfactory levels on $c_2$ and $c_3$. The performances of A2 are certainly satisfactory on criteria $c_1$ and $c_5$. On criterion $c_4$ it is unknown at this stage whether its performance is satisfactory or not. A question challenging the principle of the NCS model is the following. If the cost of A2 would be lowered for instance to that of A6 ($= 13841$ \EUR{}), would the car become Good ? Similarly, if the road-holding of A2 would be for instance like A1 ($=3$), would the car become Good? In case the DM answers such a question by ``Yes'', then the NCS model is put into question, because the latter model is non-compensatory. 

Of course, similar questions can be asked at the end of the process, when all the parameters have been elicited. 

\subsection{Making a step towards \etri}\label{ssec:towardsEtri}
In the elicitation of the profile process, it will sometimes be the case that the remaining uncertainty of the exact position of the profile is not negligible. Let $[\ \underline{b_i}, \overline{b_i}\ ]$ denote the uncertainty interval on $b_i$ remaining when the elicitation process ends up. There are several ways of dealing with such an uncertainty. One is to choose an arbitrary value for $b_i$ inside the interval, for instance, in the middle (if the scale is not discrete). Some sensitivity analysis can thus be performed by randomly choosing other values of the limit profile in the uncertainty intervals on each criterion. In this way, we can see whether the precise position of the limit profile has an impact on the classification and see also which assignments are robust, \ie remain invariant for all values of the .  

\ssk Another approach brings us close to \etri{}. Let us model the uncertainty on $b$ by introducing graduality in modeling the fact that an alternative is at least as good as the profile. We introduce a partial concordance degree function $\mathcal{C}_i(a,b)$ on criterion $i$ defined as in equation \eqref{eq-etri_partial_concordance} in Chapter~\ref{chap:SortingMCDA}, \ie
\begin{equation}\label{eq:partialConcUncertainty}
	\mathcal{C}_i(a, b) =
	\begin{cases}
		0 & \text{if } a_i \leq_i \underline{b_i},\\
		1 & \text{if } a_i \geq_i \overline{b_i},\\
		\frac{a_i - \underline{b_i}}{\overline{b_i} - \underline{b_i}} & \text{if } \underline{b_i} \leq_i a_i \leq_i \overline{b_i}.
	\end{cases}
\end{equation}
The partial concordance degree uses a linear interpolation in case $a_i$ falls into the uncertainty interval of the profile on criterion $i$. We may adapt the assignment rule of \etri{} (without discordance) as follows:
\begin{equation}\label{eq:etriRuleAdapted}
    a \in \calA \ \textrm{if } \ \sum_{i \in N} w_i \mathcal{C}_i(a,b) \geq \lambda,
\end{equation}
where $w_i$ are criteria weights and $\lambda$, a credibility threshold. The values of these parameters can be set through an indirect elicitation method. Since we start the process with a direct elicitation of the sufficient coalitions of criteria, we have at disposal a set of assignment examples. The elicitation of the sufficient coalitions indeed used alternatives whose evaluations are certainly satisfactory or certainly unsatisfactory. These evaluations are thus outside the uncertainty interval of the profile. We may hence use linear programming to find values of the parameters $w_i$ and $\lambda$ that assign as many as possible of the assignment examples to the right category. Note that it will not be possible in general to assign all examples to the right category since sets of sufficient coalitions are rarely representable by additive weights and a threshold (as observed in Chapter~\ref{chap:dam}). However, it has been shown experimentally \citep{sobrieetal2015b} that MR-Sort models  provide in general good approximations of NCS models, in the sense that they may reproduce a large proportion of the assignements made by NCS models. The obtained values of $w_i$ and $\lambda$ allow to use formula~\eqref{eq:etriRuleAdapted} to assign all alternatives, including those that are close to the limit profiles on some criteria.

\ssk The approach just outlined could offer an interesting way of eliciting the parameters of an \etri{} model (without discordance). It mixes direct and indirect elicitation methods. The set of assignment examples is constructed using alternatives that are clearly above or below the profile. The uncertainty on the position of the profile is reduced. Then a simple indirect method (based on linear programming) can be used to obtain an \etri{} model and assign all alternatives.  

\section{General NCS models}\label{ssec:NCSgeneral}
In this work we focused on eliciting the parameters of a basic version of NCS models, \ie the sufficient coalitions and the limit profile delimiting two categories. We briefly discuss how more general NCS models could be elicited, in particular how veto could be introduced and how to deal with more than two categories. 

\subsection{Eliciting vetoes}\label{ssec:veto}
The necessity of introducing veto thresholds may appear in the decision aiding process even if the DM did not consider it at the beginning. As briefly described in Section~\ref{ssec:ncs}, vetoes in the NCS model with two categories are sets of values on each criterion below a veto level (itself below the limit profile value). If an alternative has a evaluation below the veto level on any criterion, it cannot be assigned to the acceptable category \calA.    

\ssk Consider again the car selection example \ref{ssec:carSelection}. With the values of the NCS model that were elicited, there are three cars assigned to the Good category: A3, A11 and A12. Looking at these alternatives, the DM may feel that the performance of A12 on criterion $c_2$ is too weak. An acceleration time of 29.6 seconds may be unacceptable given his/her needs. A veto threshold of 29.5 seconds may for instance be imposed on $c_2$. This would eliminate A12 from the Good cars, without any other consequence in this example. Here, there is no need of introducing veto thresholds on the other criteria.   

\subsection{The case of more than 2 categories}\label{ssec:moreThan2Cat}
When the decision problem contains multiple categories $C^h, h=1, \ldots p$, the NCS model will have multiple limit profiles $b^h$ and multiple sets of sufficient coalitions $\calF$. A strategy for using the questioning procedures developed in the case of two categories is the following. We can first consider separating the alternatives in $C_1$, the least preferred category from those in any of the more preferred categories. The resulting bipartition would be $\langle C^1, C^{\geq 2} \rangle$, where $C^{\geq h}$ is the union of the categories at least as preferred as $C^h$. 

\ssk Using a NCS model to separate these two categories requires to elicit a set $\calF^1$ of sufficient coalitions and a limit profile $b^1$. If an alternative is at least as good as $b^1$ on a set of criteria belonging to $\calF^1$, it is not assigned to $C^1$ but to one of the categories in $C^{\geq 2}$. In order to elicit the sufficient coalitions and the limit profile, we have to identify levels on each criterion that are certainly satisfactory for an alternative to be assigned to $C_2$ or better and levels that correspond to alternatives that are certainly assigned to $C_1$. Once these sufficient coalitions and profile have been elicited, we may build a model separating the alternatives in $C^{\leq 2}$ from those in $C^{\geq 3}$ (where $C^{\leq h}$ denotes the union of categories not more preferred than $C^h$). The sufficient coalitions in $\calF^2$ can be elicited more easily because they form a subset of $\calF^1$. We thus have to check which coalitions from $\calF^1$ remain sufficient. To do that efficiently, one may start by asking whether the minimal sufficient coalitions are still sufficient. For eliciting $\calF^2$ and the second limit profile $b^2$, we have to identify levels that are typical of alternatives assigned to $C_1$ or $C_2$ and levels typical of alternatives assigned to $C_3$ or better. Then we may apply the dichotomic procedure to find $b^2$. 

\ssk Instead of starting from the bottom categories the elicitation of the model's parameters, we may also start from the best category and separate it from the less preferred categories. In such a case, the set of sufficient coalitions has to be gradually enriched, since the sufficient coalitions for higher categories are included in the set of sufficient coalitions for the lower categories.

\ssk We illustrate a gradual introduction of an additional category on the car selection example. Suppose that the classification of the 14 cars in Good or Bad categories appears insufficiently refined to the DM. In particular, the fact that only three cars are considered good gives him/her too restrictive possibility of choice. The DM wants that the Bad category be split in two categories: one of really unacceptable cars and the other of cars that could possibly be considered. We thus have to work with three categories : Good, Acceptable and Unacceptable, the reunion of the latter two being the original Bad category and the Good category being unchanged. If the DM maker is agree that the original good cars are indeed good, we may lean on the work already done, consider the original limit profile as $b^2$ and the original set of sufficient coalitions as $\calF^2$. We thus need to elicit the limit profile $b^1$ separating the Acceptable from the Unacceptable cars and the corresponding set of sufficient coalitions $\calF^1 \supseteq \calF^2$. 

\ssk There is no reason for changing the set of sufficient coalitions. The assignment to Acceptable rather than to Unacceptable will be decided by comparing alternatives to  less demanding standards reflected in $b^1$. Hence $\calF^1 = \calF^2$ and the elicitation of $b^1$ is done by dichotomy between certainly satisfactory and Unsatisfactory levels, the latter typically corresponding to the new Unacceptable. We can keep the certainly satisfactory levels from Table~\ref{ta:SatisUnsatisLevels}. In view of applying dichotomy to shorter value intervals, we may instead use the values in profile $b_2 = (18000, 28.8, 35.5, 2.33, 2.25)$. In contrast, since the Unsatisfactory levels in Table~\ref{ta:SatisUnsatisLevels} are typical of the original Bad category, we don't know whether they are ``sufficiently bad'' for being typical of the new Unacceptable category. It is thus prudent to use worse evaluations on each criterion than those in the table. Let us define certainly Unacceptable levels on each criterion as the vector: $21000, 30.4, 41.5, 1.33, 1.25$. 

\ssk In view of determining the first coordinate of profile $b_1$, we choose a sufficient coalition containing the first criterion $c_1$, say $\{c_1, c_2, c_4\}$. An alternative taking values from $b^2$ on this coalition and values certainly unacceptable on criteria $C_3$ and $c_5$ is a fortiori better than $b^1$ on a sufficient coalition of criteria and therefore assigned to a category at least Acceptable. If the value on $c_1$ is substituted by the typically unacceptable value (=21000), then the alternative is assigned to Unacceptable. To determine the value of $b^1$ on criterion $c_1$, the first question in the dichotomic procedure would be : Is a car having evaluations (19500, 28.8, 41.5, 2.33 1.25) at least Acceptable ? Assume that 19500 is the value finally retained for $b_1^1$ and that at the end of the elicitation process, we find that $b^1 = (19500, 29.2, 37.5, 2.0, 1.75)$, then the set of cars assigned into the Acceptable category is : $\{A_1, A_7\}$.   

\ssk Note that vetoes can be introduced when there are more than two categories. A veto profile $v^h$ has to be associated with each limit profile $b^h$, thus with each category $C^h$. A veto profile associated with a more preferred category must dominate one associated with a less preferred category. 

\subsection{Structuring criteria into a hierarchy}\label{ssec:hierarchy}
 
The suggested algorithms for eliciting the parameters are restricted in terms of the number of criteria. They cannot deal with decision problems having a large set of criteria because of an excessive number of questions to be asked. In order to find solution to these cases, some simplifications can be performed by working with criteria hierarchies, if they exist. Between several criteria, if the DM provides some information enabling to construct hierarchies of criteria and to aggregate the sub-criteria, the number of criteria can be reduced through such grouping. Such an approach has been widely developed in literature \citep[see, \eg ][]{keeneyraiffa1976,saaty1980analytic,DeSmet14HierarchClust,CorrenteDGSZ2015UTADISHierarch,CorrenteGSHierarchyTRI16,CostaEtAL_EJOR2020NominalHierarchicalRobustClassif}.

\ssk Consider the car selection example in Section~\ref{ssec:carSelection}. There is a natural criteria hierarchy that is implicitly used when searching for the sufficient coalitions. Criteria $c_2$ and $c_3$ can be grouped under the heading ``performance'' while criteria $c_4$ and $c_5$ represent the ``safety'' dimension in the car selection problem. We could thus restructure the problem into a three criteria problem: selecting good cars on the basis of their evaluation on three super-criteria: cost, performance, safety. The issue is to aggregate the evaluations on the sub-criteria in order to get an evaluation on the two regrouped criteria. Assuming that we keep the logic of the NCS model for the aggregation of the super-criteria, it suffices to determine the set of cars that can be considered ``good'' \wrt performance and safety. For the cost, we keep the condition previously established: a car is satisfactory if it is not more expensive than 18000 \EUR{}. 

\begin{figure}[h!]
\centering
\includegraphics[width=130mm]{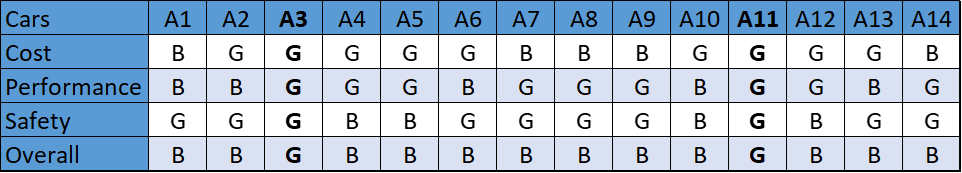}\caption{Sorting by NCS using a hierarchy of criteria} \label{fig:SortingByHierarchy}
\end{figure}

\ssk In order to model what is a good performance, we might use a kind of weighted sum of appropriate normalizations of the evaluations on $c_2$ and $c_3$ and use a threshold. But this would not be in line with the logic of a non-compensatory method such as \ncs{}. We could alternatively elicit good levels on each of the two criteria and combine them to define overall good performance. Starting with the certainly Satisfactory and unsatisfactory levels in Table~\ref{ta:SatisUnsatisLevels} and questioning the DM by using dichotomy, we might come to the values of the limit profile as before. If we choose to combine these values disjunctively, this leads to the condition : Performance is considered good if 
\begin{equation}\label{eq:disjunctiveRule}
    c_2 \leq 28.8 \ \textrm{or } \ c_3 \leq 35.5,
\end{equation}
and similarly for the safety criterion ($c_4 \geq 2.33$ and $c_5 \geq 2.25$). Then, considering that a car, to be good, must be good on all three super-criteria, we would obtain exactly the same bipartition of the cars as with the initial \ncs{} model. Alternatively, we may consider to model super-criteria in a conjunctive way, in order to keep some control on the values on both sub-criteria (which is not the case with the disjunctive rule). Substituting ``or'' by ``and'' in \eqref{eq:disjunctiveRule} would probably too selective. Some weakening of the requested level  on the sub-criteria could be envisaged. For instance, a conjunctive rule could be : Performance is considered good if 
\begin{equation}\label{eq:conjunctiveRulePerf}
    c_2 \leq 29 \ \textrm{and } \ c_3 \leq 36,
\end{equation}    
and Safety is considered good if 
\begin{equation}\label{eq:conjunctiveRuleSafety}
    c_4 \geq 2.0 \ \textrm{and} \ c_5 \geq 2.0.
\end{equation}
Using these rules for defining good performance and good safety, together with $c_1 \leq 18000$ for defining good cost, and considering that only the coalition of the three super-criteria is sufficient, leads to Figure~\ref{fig:SortingByHierarchy}.

The good cars are $A_3, A_11$, a subset of the set of good cars obtained with the initial model. The modelling using hierarchy leads here to a more severe selection, but consistent with the output of the initial model. 

\ssk In case a hierarchy of criteria is used in the framework of the NCS model, the main issue is how to aggregate sub-criteria evaluations into a super-criterion in a way that is consistent with the logic of the NCS model. This question deserves more investigation.  

\section{Chapter conclusion }\label{sec:concluChap8}
This chapter emphasized several aspects of the application in real decision aiding processes of the rigorous methods developed in this work. The constraints of real-world situations exclude using methods in an absolutely rigorous way. Adaptability and flexibility are required. However, we believe that keeping in mind a rigorous way of doing things, provides a guarantee not to betray the models principles and significance. Clearly, there would be much more to do on this topic but we hope to have done enough to show the interest of paying attention to rigorous methods. 
\chapter{Conclusion and Perspectives}\label{chap:conclusion}
The goal of this work was to propose rigorous methods for the direct elicitation of the  parameters used in sorting methods based on outranking relations. Due to the fact that the only one among such methods that has been characterized by an axiomatic analysis is the Non-Compensatory Sorting (NCS) method, we focused on this method. Our aim was to use only information provided by the assignment of well-structured alternatives in order to efficiently determine the model's parameters. 

\ssk We wanted to design an elicitation procedure for outranking based sorting models in the spirit of those used for rigorously eliciting an additive value function. Of course, due to the constraints of real-world decision processes, it is usually not possible to apply exactly the theoretically rigorous methods. Nevertheless, we believe that such approaches may provide valuable guidelines to the analyst working in decision aiding processes. In particular, they help the analyst in formulating questions to the DM that efficiently and reliably determine the value of the model's parameters. 

\ssk This chapter first gives an overview of the thesis from which we present what we consider as the main contributions. It closes with some perspectives of further research.

\section{Summary}\label{sec:summary}
We focused on a basic version of the NCS model, \ie the case of sorting in two categories without veto effects. In this setting, the parameters to be determined are the set of sufficient coalitions of criteria and the profile delimiting the categories. We started by developing questioning methods for eliciting the sufficient coalitions, which is the most delicate part, because the number of coalitions grows exponentially with the number of criteria. Our first idea was to engraft questions on classical graphs exploration algorithms such as depth-first search. We proposed different strategies for choosing the nodes in which questions are asked. Our favorite one is DFS3, in which questions are asked in the middle of the branches of the depth-first search arborescence. This was done in Chapter~\ref{chap:elicitNCS}, where we also described a procedure for determining the limit profile with the desired precision.  

\ssk Chapter~\ref{chap:questioner} is devoted to the implementation of the DFS3 questioning algorithm. This algorithm could be used interactively by an analyst in a decision process to ask questions to the decision maker, in terms of alternatives assignments. After each answer is encoded in the program, a new question is generated untill all sufficient coalitions of criteria are known. In order to assess the performance of this algorithm in terms of the number of questions to be asked, we used it as a simulation tool, running it over all possible families of sufficient coalitions for up to 6 criteria. From this, we know the average number of questions needed for up to 6 criteria and, also, the number of questions in the worst case. The DFS3 questioning algorithm can take into account additional information, namely, an existing importance order on the criteria or a partial order on some coalitions. 

\ssk Optimal ways of questioning about sufficient coalitions can be defined in at least two manners: in the worst case or on average. An optimal algorithm in the worst case has been proposed in a different but equivalent setting, that of monotone Boolean functions. The algorithm is based on a decomposition of the partially ordered set of subsets of the set of criteria into Hansel chains. It guarantees that the number of asked questions never surpasses a lower bound. We implemented this algorithm as a variant of the depth-first search approaches that were investigated in Chapter~\ref{chap:elicitNCS}. To do that, we had to find a way of generating the Hansel chains directly for $n$ criteria instead of recursively as indicated in the original paper by Hansel and in the subsequent literature. 

\ssk Chapter~\ref{chap-average} investigates the other optimality paradigm, \ie minimizing the number of questions on average for a given number of criteria. We explicitly developed and evaluated decision trees for up to 4 criteria. We identified optimal questioning trees in these cases under the hypothesis that each set of sufficient coalitions has the same probability of occurrence. Dealing with more than four criteria by hand is almost impossible. Even for $n=4$ criteria, we had to simplify the problem by pruning the tree, using symmetries, already known patterns (obtained for smaller values of $n$) and  branch and bound techniques. A heuristic was proposed (actually re-discovered) which may be used with more than 4 criteria. 

\ssk The MR-Sort model is a particular case of the NCS model in which the sufficient coalitions of criteria can be represented additively by weights associated to the criteria and a threshold. In Chapter~\ref{chap:dam}, we try to assess how more general is NCS as compared to MR-Sort. In other words, we try to estimate the proportion of NCS models that can be viewed as MR-Sort models, \ie the proportion of sets of sufficient coalitions that can be represented additively. We do this by (exhaustive) simulation for up to 6 criteria. We also compute the proportion of the sets of sufficient coalitions that can be represented by 2-additive and 3-additive models, \ie more general models in which additional weights model interactions between pairs or triples of criteria. It turns out that the proportion of sets of sufficient coalitions representable by weights and a threshold quickly decreases with $n$. The same seems to happen with 2-additive and 3-additive representations, although for larger values of $n$. 

\ssk We said above that we followed a theoretically oriented approach. Our aim with the last chapter (Chapter~\ref{chap:recul}, is to show that the developed theoretical approaches are useful in different ways. The main difficulty, in real situations, is the number and the difficulty of the questions asked to DM. Using a simple example, we showed that it may be possible to reduce the number of possible sets of sufficient coalitions by asking few simple, general, questions. Then it becomes possible to finish elicitation more quickly and rigorously. In another direction, we showed that less basic versions of the sorting model may be introduced incrementally. For instance, vetoes and multiple categories can be considered, starting from a simple NCS model whose parameters were firstly elicited. It is also possible to combine an interrupted questioning procedure with indirect elicitation or learning techniques.

\section{Contributions}\label{sec:contributions}
Our main contribution is, in line with our objective, the development of a rigorous methodology for determining the set of sufficient coalitions and the limit profile of a NCS model with two categories and no veto. The application of this method as such is limited by the number of criteria but it can be adapted or imbedded in indirect elicitation techniques to deal with real-world problems.

\ssk The more particular issues and contributions that we would like to pinpoint are:
\begin{itemize}
    \item the exploration of the coalitions graph following a depth-first search arborescence and the idea of questioning in the middle of the branches (DFS3 algorithm);
    \item the reformulation of Hansel's algorithm as a questioning algorithm following a graph arborescence that is a variant of depth-first search;
    \item the simplifying tools (symmetries, patterns, branch and bound) that help to find optimal questioning strategies on average for $n=4$;
    \item the comparison of the NCS and MR-Sort models, which is published in \citep{ErsekEtAlDAM2017}.    
\end{itemize}

\section{Perspectives}\label{sec:perspectives}
Several research issues related to the topic of this work require additional efforts. Some issues may appear as hopelessly difficult as, \eg identifying the optimal questioning paths on average even only for $n$ up to 7.   
Among the questions that we think are interesting and workable, the following seem promising:

\paragraph{DFS3 with additional information } Implementing and testing the DFS3 algorithm in case there is additional information on the importance of criteria. This is already possible with the current implementation, but the latter uses the depth-first search arborescence of the original coalitions graph. Instead, it would be beneficial in terms of the number of questions, to apply DFS3 to the Hasse diagram of the partial order on the coalitions enriched by the additional information. An example of such a Hasse diagram is shown in Figure~\ref{fig:additionalInfoGraph} for $n=3$ and $n=4$. The resulting graph is ``stretched''. Therefore, the paths produced by  depth-first search are longer (at least the initial ones) and dichotomy will be more efficient. One may hence expect further reductions in the number of questions. An experimental  evaluation of the obtained gain in number of questions is an interesting research issue.

\paragraph{Efficient strategy on average } Tackling the question of the optimal questioning paths on average for $n=5$ is a further challenge. Solving this question requires the support of a computer program since there are 7581 possible sets of minimal sufficient coalitions. It is probably possible to develop the decision tree manually if a computer program can be used to compute the remaining number of antichains after each answer (Is this coalition sufficient? Yes/No). Using such a tool together with the simplifying tricks described in Section~\ref{ssec:simplifyTools} could allow to prune the decision tree and keep it to manageable size. This would provide an additional exact optimal value allowing to assess the performance of algorithms such as DFSHansel and variants of the heuristic algorithm proposed in Section~\ref{sec:heuristic}.  
    
\paragraph{Implementing the suggested heuristic } The implementation of the heuristic described in Section~\ref{sec:heuristic} and to assess its performance by simulation for $n \geq 4$ can be interesting. Different versions of the question selection rule can be experimentally tested. Exhaustive simulation is in principle possible up to $n=6$, since the list of all antichains is available up to this number of criteria. For more criteria, on can only sample the antichains. Sampling the antichains according to an a priori distribution on the set of all antichains is in itself a research issue. \citet{torvikEtAl2002} analyzed this question and proposed a way of improving the uniformity of a natural sampling procedure. While their proposal does improve the uniformity of the sampling, it does not guarantee that all antichains have equal probability of being sampled. It would be useful and interesting to investigate other sampling ideas.       
   
\paragraph{Using the questioning algorithms in real problems} In Chapter~\ref{chap:recul}, we try to support the idea that the rigorous approach developed may be useful tools in practice. It would be interesting to try to use questioning algorithms in real decision aiding situations. This would certainly enrich the tricks that enable to reduce the number of questions about sufficient coalitions. It would also be interesting to see how to combine direct and indirect elicitation procedures whenever the direct questioning is interrupted. Other issues could also benefit from tackling real decision situations and being challenged by real DMs. Some of these issues are: 
\begin{itemize}
\item How to deal with evolutions of the DM's preferences or judgments in the course of the decision aiding process?
\item Taking into account the cognitive difficulty of the questions for the DM, instead of the number of questions only.
\item Dealing with uncertainty in the DM's answers: the DM may for instance hesitate between two possible answers, be unable to provide an answer, assign plausibilities to different answers, etc.
\item Dealing with contradictions in successive answers, possibly implying that the chosen model is inadequate. The algorithms proposed in this thesis aim to minimize the number of questions; they avoid redundancy. Therefore, they are not designed to detect inconsistencies. In Section~\ref{ssec:inadequacy}, we recommend to validate the model through asking additional questions that challenge it. These questions are asked at the end of the elicitation process. It might be useful to ask such questions in the course of the elicitation in order to save time and reorient the process by using another model if necessary. 
\end{itemize}

\paragraph{ Towards \etri{} } In Section~\ref{ssec:towardsEtri}, we outline an incremental way of eliciting a simple version of \etri{}, starting from the elicitation of a NCS model. Upper and lower approximations of the limit profile could be used to define the preference and indifference thresholds used in \etri{}.  Such an interpretation of the imprecision in the determination of the profile deserves further  investigation given that direct (and even indirect) elicitation of the parameters of \etri{} is a challenging task.

\paragraph{NCS with more than 2 categories } In Section~\ref{ssec:moreThan2Cat} we suggested a way of eliciting the profiles and sufficient coalitions of a NCS model with more than two categories by repeated application of the questioning algorithms used in the case of two categories. Alternative questioning methods could be investigated aiming to be more parsimonious in the number of asked questions. In particular, one could think of asking the DM to assign alternatives to a precise category in a set of more than two categories.

\paragraph{}


%

\backmatter
\bibliographystyle{plainnat}
\phantomsection
\addcontentsline{toc}{chapter}{References}
\bibliography{biblio/me,biblio/mcda,biblio/eda,biblio/CoalSuff}

@ARTICLE{Hansel1966,
  AUTHOR =       {Hansel, G.},
  TITLE =        {Sur le nombre de fonctions booléennes de $n$ variables},
  JOURNAL =      {Comptes-Rendus de l'Académie des Sciences de Paris (Série A)},
  YEAR =         {1966},
  volume =       {262},
  number =       {20},
  pages =        {1088-1090},
}

@ARTICLE{torvikEtAl2002,
  AUTHOR =       {Torvik, B. and Triantaphyllou, E.},
  title=         {Minimizing the average query complexity of learning monotone {B}oolean functions},
  JOURNAL =      {{INFORMS} {J}ournal on Computing},
  YEAR =         {2002},
  volume =       {14},
  number =       "2",
  pages =        {142-172},
}

@ARTICLE{torvikEtAl2003,
  AUTHOR =       {Torvik, B. and Triantaphyllou, E.},
  title=         {Guided inference of nested monotone {B}oolean functions},
  JOURNAL =      {Information Sciences},
  YEAR =         {2003},
  volume =       {151},
  pages =        {171-200},
}

@ARTICLE{Korobkov1965,
  AUTHOR =       {Korobkov, B.K.},
  TITLE =        {O monotonnych funkcijach algebry logiki},
  JOURNAL =      {Problemy Kibernetiki},
  YEAR =         {1965},
  volume =       {13},
  pages =        {5--28},
}

@article{kovalerchuketal1996,
  author    = {B. Kovalerchuk and
               E. Triantaphyllou and
               A. S. Deshpande and
               E. Vityaev},
  title     = {Interactive Learning of Monotone Boolean Functions},
  journal   = {Inf. Sci.},
  volume    = {94},
  number    = {1-4},
  pages     = {87--118},
  year      = {1996},
}

@MISC{OEIS,
  author =       {{The {OEIS} Foundation}},
  title =        {On-Line Encyclopedia of Integer Sequences},
  howpublished = {\url{https://oeis.org}},
  note =         {[Online; accessed November 31, 2022}
}

@Book{CasparLeclercMonjardetBook,
author = {N. Caspard and B. Leclerc and B. Monjardet},
title = {Finite Ordered Sets : Concepts, results and uses},
publisher = {Cambridge University Press},
year = {2012},
series = {Encyclopedia of Mathematics and its applications}
}

@book{CramaHammer2011,
  author    = {Y. Crama and
               P. L. Hammer},
  title     = {Boolean Functions - Theory, Algorithms, and Applications},
  series    = {Encyclopedia of mathematics and its applications},
  year      = {2011},
  volume    = {142},
  publisher = {Cambridge University Press}
}

@book{vonWinterfeldEdwards1986,
  title={Decision Analysis and Behavioral Research},
  author={Von Winterfeldt, D. and Edwards, W.},
  year={1986},
  publisher={Cambridge University Press}
}

@book{raiffa1968decision,
  title={Decision Analysis: Introductory Lectures on Choices Under Uncertainty},
  author={Raiffa, H.},
  series={Behavioral science, quantitative methods},
  year={1968},
  publisher={Random House}
}

@BOOK{CormenEtAl1990,
  title     = "Introduction to Algorithms",
  author    = "Cormen, T. H. and Leiserson, C. E. and
               Rivest, R. L.",
  publisher = "MIT Press",
  series    = "MIT Electrical Engineering and Computer Science",
  year      =  "1990",
  address   = "London, England"
}

@BOOK{vonNeumanMorgenstern43,
  AUTHOR =       {J. {Von Neumann} and O. Morgenstern},
  TITLE =        {Theory of Games and Economic Behavior},
  PUBLISHER =    {Princeton University Press, NJ, USA},
  YEAR =         {1972},
  edition =      {3rd},
}

@article{KurzTautenham2013,
year={2013},
journal={International Journal of Game Theory},
volume={42},
number={2},
title={On {Dedekind\'{}}s problem for complete simple games},
author={Kurz, S. and Tautenhahn, N.},
pages={411-437},
}

@BOOK{Kuntzmann1965,
  author =       {Kuntzmann, J.},
  title =        {Alg\`{e}bre de {B}oole},
 publisher =    {Dunod},
  year =         {1965},
}

@BOOK{Anthony2001,
  AUTHOR =       {M. Anthony},
  TITLE =        {Discrete mathematics of neural networks: selected topics},
  PUBLISHER =    {Society for industrial and applied mathematics, New York},
  YEAR =         {2001},
  series =       {SIAM monographs on discrete mathematics and applications},
  isbn =         {9780898714807},
}

@inbook{Anthony2010,
series={Encyclopedia of Mathematics and its Applications},
volume={134},
title={Probabilistic Learning and Boolean Functions},
booktitle={Boolean Models and Methods in Mathematics, Computer Science, and Engineering},
publisher={Cambridge University Press, New York, USA},
author={Anthony, M.},
year={2010},
pages={197-220},
}

@INPROCEEDINGS{Bruck2010,
  AUTHOR =       {J. Bruck},
  TITLE =        {{F}ourier transforms and threshold circuit complexity},
  BOOKTITLE =    {Boolean Models and Methods in Mathematics, Computer Science, and Engineering},
  YEAR =         {2010},
  editor =       {Crama, Y. and Hammer, Peter L.},
  volume =       {134},
  series =       {Encyclopedia of mathematics and its applications},
  pages =        {531-553},
  publisher =    {Cambridge University Press, New York, USA},
}

@INproceedings{Saks1993,
 author = {Saks, M. E.},
 title = {Slicing the Hypercube},
 booktitle = {Surveys in Combinatorics},
 editor = {Walker, Keith},
 year = {1993},
 pages = {211--255},
 publisher = {Cambridge University Press, New York, USA}
}

@article{Grabisch1997,
author = {Grabisch, M.},
title = "$k$-order additive discrete fuzzy measures and their representation ",
journal = "Fuzzy Sets and Systems ",
volume = "92",
number = "2",
pages = "167 - 189",
year = "1997"
}

@BOOK{GrabischBook2016,
   AUTHOR =       {Grabisch, M.},
   TITLE =        {Set Functions, Games and Capacities in Decision Making},
   PUBLISHER =    {Springer},
   YEAR =         {2016},
   isbn =         {978-3-319-30690-2},
}

@article{stephenyusun2014,
	title = {Counting inequivalent monotone Boolean functions},
	author = {Stephen, T. and Yusun, T.},
	journal = {Discrete Applied Mathematics},
	volume = {167},
	number = {0},
	pages = {15--24},
	year = {2014},
}

@BOOK{Wolsey1998,
  author =       {L.A. Wolsey},
  title =        {Integer Programming},
  publisher =    {Wiley},
  year =         {1998},
}

@article{BelahceneMOPS23,
  author    = {K. Belahc{\`{e}}ne and
               V. Mousseau and
               W. Ouerdane and
               M. Pirlot and
               O. Sobrie},
  title     = {Multiple criteria sorting models and methods - Part {I:} survey of
               the literature},
  journal   = {4OR},
  volume    = {21},
  number    = {1},
  pages     = {1--46},
  year      = {2023},
}

@article{PODINOVSKII1994237,
title = {Criteria importance theory},
journal = {Mathematical Social Sciences},
volume = {27},
number = {3},
pages = {237-252},
year = {1994},
author = {V.V. Podinovskii},
}

@article{BRVM96,
author="B. Roy and V. Mousseau",
year=1996,
title="A theoretical framework for analysing the notion of relative importance of criteria",
journal="Journal of Multi-Criteria Decision Analysis",
volume=5,
pages="145-159",
}

@article{Keeney2002,
 author = {R. L. Keeney},
 journal = {Operations Research},
 number = {6},
 pages = {935--945},
 publisher = {INFORMS},
 title = {Common Mistakes in Making Value Trade-Offs},
 volume = {50},
 year = {2002}
}

@ARTICLE{RoyMousseau96,
  author       = "Roy, B. and Mousseau, V.",
  year         = "1996",
  title        = "A theoretical framework for analysing the notion of relative importance
                  of criteria",
  journal      = "Journal of Multiple Criteria Decision Analysis",
  volume       = "5",
  number       = "2",
  pages        = "145--159",
}

@article{diasmoussfig2002,
	title = {An aggregation/disaggregation approach to obtain robust
		 conclusions with {ELECTRE TRI}},
	author = {Dias, L. and Mousseau, V. and Figueira, J. R.
		  and Cl\'imaco, J.},
	journal = {European Journal of Operational Research},
	volume = {138},
	number = {1},
	pages = {332--348},
	year = {2002},
	publisher = {Elsevier}
}

@article{KadzinskiTervonen2013,
title = {Stochastic ordinal regression for multiple criteria sorting problems},
journal = {Decision Support Systems},
volume = {55},
number = {1},
pages = {55-66},
year = {2013},
author = {M. Kadzi\'{n}ski and T. Tervonen}
}

@article{KadzinskiM21EnrichedPrefModel,
  author    = {M. Kadziński and
               M. Martyn},
  title     = {Enriched preference modeling and robustness analysis for the {ELECTRE Tri-B} method},
  journal   = {Annals of Operations Research},
  volume    = {306},
  number    = {1},
  pages     = {173--207},
  year      = {2021}
}

@article{KadzinskiCiomek21ActiveLearnOutrankSorting,
  author    = {M. Kadziński and
               K. Ciomek},
  title     = {Active learning strategies for interactive elicitation of assignment
               examples for threshold-based multiple criteria sorting},
  journal   = {European Journal of Operational Research},
  volume    = {293},
  number    = {2},
  pages     = {658--680},
  year      = {2021}
}

@ARTICLE{BouyssouPirlot2015AOR,
  author       = "Bouyssou, D. and Pirlot, M.",
  year         = "2015",
  title        = "A consolidated approach to the axiomatization of outranking relations: {A} survey and new results",
  journal      = "Annals of Operations Research",
  volume       = "229",
  number       = "1",
  pages        = "159--212",
}

@ARTICLE{MousseauSlowinski98,
  author       = "Mousseau, V. and S{\l }owi{\'n}ski, R.",
  year         = "1998",
  title        = "Inferring an {ELECTRE TRI} model from assignment examples",
  journal      = "Journal of Global Optimization",
  volume       = "12",
  number       = "2",
  pages        = "157--174",
}

@article{TervonenFigueiraJMCDA2008,
author = {Tervonen, T. and Figueira, J. Rui},
title = {A survey on stochastic multicriteria acceptability analysis methods},
journal = {Journal of Multi-Criteria Decision Analysis},
volume = {15},
number = {1-2},
pages = {1-14},
year = {2008}
}

@article{tervonenEtAl_EJOR2009,
  title={A stochastic method for robustness analysis in sorting problems},
  author={Tervonen, T. and Figueira, J. and Lahdelma, R. and Dias, J.A. and Salminen, P.},
  journal={European Journal of Operational Research},
  volume={192},
  number={1},
  pages={236--242},
  year={2009},
}

@article{RochaDias08JGlobOpt,
  author    = {C. Rocha and
               L. C. Dias},
  title     = {An algorithm for ordinal sorting based on {ELECTRE} with categories defined by examples},
  journal   = {Journal of Global Optimization},
  volume    = {42},
  number    = {2},
  pages     = {255--277},
  year      = {2008}
}

@article{LahdelmaHS1998,
title = {{SMAA} - Stochastic multiobjective acceptability analysis},
journal = {European Journal of Operational Research},
volume = {106},
number = {1},
pages = {137-143},
year = {1998},
author = {R. Lahdelma and J. Hokkanen and P. Salminen}
}

@article{KadzinskiCS2015SortAddValPair,
title = {Modeling assignment-based pairwise comparisons within integrated framework for value-driven multiple criteria sorting},
journal = {European Journal of Operational Research},
volume = {241},
number = {3},
pages = {830-841},
year = {2015},
author = {M. Kadzi\'{n}ski and K. Ciomek and R. Słowiński}
}

@inproceedings{SobrieMP17,
  author    = {O. Sobrie and
               V. Mousseau and
               M. Pirlot},
  editor    = {Heike Trautmann and
               G{\"{u}}nter Rudolph and
               Kathrin Klamroth and
               Oliver Sch{\"{u}}tze and
               Margaret M. Wiecek and
               Yaochu Jin and
               Christian Grimme},
  title     = {A Population-Based Algorithm for Learning a Majority Rule Sorting
               Model with Coalitional Veto},
  booktitle = {Evolutionary Multi-Criterion Optimization - 9th International Conference,
               {EMO} 2017, M{\"{u}}nster, Germany, March 19-22, 2017, Proceedings},
  series    = {Lecture Notes in Computer Science},
  volume    = {10173},
  pages     = {575--589},
  publisher = {Springer},
  year      = {2017},
}

@inproceedings{OlteanuMeyer2014,
author = {Olteanu, A. L. and Meyer, P.},
title = {{Inferring the parameters of a majority rule sorting model with vetoes on large datasets}},
booktitle = {DA2PL 2014 : From Multicriteria Decision Aid to Preference Learning},
year = {2014},
pages = {87-94} }

@article{sobrieetal2016ASA,	
author = {Sobrie, O. and Lazouni, M. E. A. and Mahmoudi, S. and Mousseau, V. and Pirlot, M.},
	title = {A new decision support model for preanesthetic evaluation},
	journal = {Computer Methods and Programs in Biomedicine},
	year = {2016},
	volume = {133},
	pages = {183--193},
}

@article{TliliBKMO22,
  author    = {A. Tlili and
               K. Belahc{\`{e}}ne and
               O. Khaled and
               V. Mousseau and
               W. Ouerdane},
  title     = {Learning non-compensatory sorting models using efficient SAT/MaxSAT
               formulations},
  journal   = {Eur. J. Oper. Res.},
  volume    = {298},
  number    = {3},
  pages     = {979--1006},
  year      = {2022},
}

@book{saaty1980analytic,
  title={The Analytic Hierarchy Process: Planning, Priority Setting, Resource Allocation},
  author={Saaty, T.L.},
  publisher={McGraw-Hill International Book Company},
year={1980}
}

@article{BisdorffMeyerRoubens08Rubis,
  author       = {R. Bisdorff and
                  P. Meyer and
                  M. Roubens},
  title        = {R {UBIS} : a bipolar-valued outranking method for the choice problem},
  journal      = {4OR},
  volume       = {6},
  number       = {2},
  pages        = {143--165},
  year         = {2008},
}

@INCOLLECTION{Roy81Trichotomic,
  author       = "Roy, B.",
  editor       = "Nijkamp, P. and Spronk, J.",
  year         = "1981",
  title        = "A Multicriteria Analysis for Trichotomic Segmentation Problems",
  booktitle    = "Multiple Criteria Analysis: {O}perational methods",
  pages        = "245--257",
  publisher    = "Gower Publishing Company",
  address      = "Aldershot, England",
}

@inproceedings{DeSmet14HierarchClust,
  author       = {Y. {De Smet}},
  title        = {An extension of {PROMETHEE} to divisive hierarchical multicriteria
                  clustering},
  booktitle    = {2014 {IEEE} International Conference on Industrial Engineering and
                  Engineering Management, {IEEM} 2014, Selangor Darul Ehsan, Malaysia,
                  December 9-12, 2014},
  pages        = {555--558},
  publisher    = {{IEEE}},
  year         = {2014},
  url          = {https://doi.org/10.1109/IEEM.2014.7058699},
  doi          = {10.1109/IEEM.2014.7058699},
  bibsource    = {dblp computer science bibliography, https://dblp.org}
}

@article{CorrenteDGSZ2015UTADISHierarch,
  year = {2015},
  month = may,
  publisher = {Springer},
  volume = {251},
  number = {1-2},
  pages = {117--139},
  author = {S. Corrente and M. Doumpos and S. Greco and R. S{\l}owi{\'{n}}ski and C. Zopounidis},
  title = {Multiple criteria hierarchy process for sorting problems based on ordinal regression with additive value functions},
  journal = {Annals of Operations Research}
}

@article{CostaEtAL_EJOR2020NominalHierarchicalRobustClassif,
title = {A robust hierarchical nominal multicriteria classification method based on similarity and dissimilarity},
journal = {European Journal of Operational Research},
volume = {286},
number = {3},
pages = {986-1001},
year = {2020},
author = {A. S. Costa and S. Corrente and S. Greco and J. R. Figueira and J. Borbinha}
}

@article{CorrenteGSHierarchyTRI16,
  author    = {S. Corrente and
               S. Greco and
               R. S{\l}owi\'{n}ski},
  title     = {Multiple Criteria Hierarchy Process for {ELECTRE} {T}ri methods},
  journal   = {European Journal of Operational Research},
  volume    = {252},
  number    = {1},
  pages     = {191--203},
  year      = {2016}
}

@book{bouyssouetal2000,
	title = {Evaluation and decision models: a critical perspective},
	author = {Bouyssou, D. and Marchant, T. and Pirlot, M.
		  and Perny, P. and Tsouki\`as, A. and Vincke, P.},
	publisher = {Kluwer Academic Publishers},
	address = {Dordrecht},
	year = {2000}
}

@book{bouyssouetal2006,
	title = {Evaluation and decision models with multiple criteria:
		 Stepping stones for the analyst},
	author = {Bouyssou, D. and Marchant, T. and Pirlot, M.
		  and Tsouki\`as, A. and Vincke, P.},
	year = {2006},
	publisher = {Springer},
	series = {International Series in Operations Research and
		  Management Science, Volume 86},
	address = {Boston},
	edition = {1st},
	isbn = {0-387-31098-3},
	url = {http://www.springer.com/sgw/cda/frontpage/0,11855,4-40521-22-116132747-0,00.html}
}

@article{bouyssoumarchant2007a,
	title = {An axiomatic approach to noncompensatory sorting methods
		 in {MCDM}, {I}: The case of two categories},
	author = {Bouyssou, D. and Marchant, T.},
	journal = {European Journal of Operational Research},
	volume = {178},
	number = {1},
	pages = {217--245},
	year = {2007},
	publisher = {Elsevier}
}

@article{bouyssoumarchant2007b,
	title = {An axiomatic approach to noncompensatory sorting methods
		 in {MCDM}, {II}: More than two categories},
	author = {Bouyssou, D. and Marchant, T.},
	journal = {European Journal of Operational Research},
	volume = {178},
	number = {1},
	pages = {246--276},
	year = {2007},
	publisher = {Elsevier}
}

@article{bouyssoupirlot2009,
	author = {Bouyssou, D. and Pirlot, M.},
	year = {2009},
	title = {An axiomatic analysis of concordance-discordance
		 relations},
	journal = {European Journal of Operational Research},
	volume = {199},
	pages = {468--477},
}

@incollection{bransetal1984,
	author = {Brans, J.-P. and Mareschal, B. and Vincke, P.},
	title = {{PROMETHEE}: a new family of outranking methods in
		 multicriteria analysis},
	publisher = {North Holland, Amsterdam},
	editor = {J.-P. Brans},
	booktitle = {Operational Research},
	year = {1984},
	pages = {477--490},
}

@inproceedings{chateauneufjaffray1986,
	author = {Chateauneuf, A. and Jaffray, J.-Y.},
	title = {Derivation of some results on monotone capacities by
		 {M}{\"{o}}bius inversion},
	booktitle = {Uncertainty in Knowledge-Based Systems, International
		     Conference on Information Processing and Management of
		     Uncertainty in Knowledge-Based Systems,
		     {IPMU} '86, Paris, France, June 30 - July 4, 1986,
		     Selected and Extended Contributions},
	year = {1986},
	pages = {95--102},
	crossref  = {ipmu1986},
	url = {http://dx.doi.org/10.1007/3-540-18579-8_8},
	doi = {10.1007/3-540-18579-8_8},
	bibsource = {dblp computer science bibliography, http://dblp.org}
}

@article{desmetetal2012,
	author = {{De Smet}, Y. and Nemery, P. and Selvaraj, R.},
	title = {An exact algorithm for the multicriteria ordered clustering
		 problem},
	journal = {Omega},
	volume = {40},
	number = {6},
	pages = {861 -- 869},
	year = {2012},
	note = {Special Issue on Forecasting in Management Science},
	issn = {0305-0483},
	doi = {http://dx.doi.org/10.1016/j.omega.2012.01.007},
	url = {http://www.sciencedirect.com/science/article/pii/S0305048312000187},
}

@inproceedings{erseketal2014,
	author = {Ersek Uyan{\i}k, E. and Sobrie, O. and Mousseau, V.
		  and Pirlot, M.},
	title = {Listing the families of Sufficient Coalitions of criteria
		 involved in Sorting procedures},
	booktitle = {DA2PL 2014 Workshop From Multiple Criteria Decision
		     Aid to Preference Learning },
	year = {2014},
	pages = {60--70},
	url = {http://www.lgi.ecp.fr/~mousseau/DA2PL-2014},
	note = {Paris, France},
}

@proceedings{ipmu1986,
	editor= {Bouchon{-}Meunier, B. and
	         Yager, R. R.},
	title = {Uncertainty in Knowledge-Based Systems, International
		 Conference on Information Processing and Management of
		 Uncertainty in Knowledge-Based Systems, {IPMU} '86,
		 Paris, France, June 30 - July 4, 1986, Selected and
		 Extended Contributions},
	series = {Lecture Notes in Computer Science},
	year = {1987},
	volume = {286},
	publisher = {Springer},
	isbn = {3-540-18579-8},
	bibsource = {dblp computer science bibliography, http://dblp.org}
}

@article{diasmousseau2006,
	title = {Inferring {ELECTRE}'s veto-related parameters from
		 outranking examples},
	author = {Dias, L. and Mousseau, V.},
	journal = {European Journal of Operational Research},
	volume = {170},
	number = {1},
	pages = {172--191},
	year = {2006},
	publisher = {Elsevier}
}

@article{doumposetal2009,
	title = {An evolutionary approach to construction of outranking
		 models for multicriteria classification: The case
		 of the {ELECTRE TRI} method},
	author = {Doumpos, M. and Marinakis, Y. and Marinaki, M.
		  and Zopounidis, C.},
	journal = {European Journal of Operational Research},
	volume = {199},
	number = {2},
	year = {2009},
	pages = {496--505},
	ee = {http://dx.doi.org/10.1016/j.ejor.2008.11.035},
	bibsource = {DBLP, http://dblp.uni-trier.de}
}

@article{figueiraroy2002,
	title = {Determining the weights of criteria in the {ELECTRE} type
		 methods with a revised {S}imos' procedure},
	author = {Figueira, J. and Roy, B.},
	journal = {European Journal of Operational Research},
	volume = {139},
	number = {2},
	pages = {317--326},
	year = {2002},
	note = {{EURO} XVI: O.R. for Innovation and Quality of Life},
	issn = {0377-2217},
	doi = {http://dx.doi.org/10.1016/S0377-2217(01)00370-8},
	url = {http://www.sciencedirect.com/science/article/pii/S0377221701003708},
}

@article{grecoetal2008,
	author = {Greco, S. and Mousseau, V. and S{\l}owi\'nski, R.},
	title = {Ordinal regression revisited: Multiple criteria ranking
		 using a set of additive value functions},
	journal = {European Journal of Operational Research},
	volume = {191},
	number = {2},
	pages = {416--436},
	year = {2008},
	issn = {0377-2217},
	doi = {http://dx.doi.org/10.1016/j.ejor.2007.08.013},
	url = {http://www.sciencedirect.com/science/article/pii/S0377221707008752},
}

@article{grecoetal2010,
	title = {Multiple criteria sorting with a set of additive
		 value functions},
	author = {Greco, S. and Mousseau, V. and S{\l}owi\'nski, R.},
	journal = {European Journal of Operational Research},
	year = {2010},
	volume = {207},
	number = {3},
	pages = {1455--1470},
	month = {December},
}

@article{jaquetlsiskos1982,
	author = {{Jacquet-Lagr\`eze}, E. and Siskos, Y.},
	title = {Assessing a set of additive utility functions for
		 multicriteria decision making: the {UTA} method},
	journal = {European Journal of Operational Research},
	volume = {10},
	year = {1982},
	pages = {151--164}
}

@article{siskostsotsolas2015,
	author = {Siskos, E. and Tsotsolas, N.},
	title = {Elicitation of criteria importance weights through the {Simos}
		 method: a robustness concern},
	journal = {European Journal of Operational Research},
	volume = {246},
	number = {2},
	pages = {543--553},
	year = {2015},
	url = {http://dx.doi.org/10.1016/j.ejor.2015.04.037},
	doi = {10.1016/j.ejor.2015.04.037},
}

@book{keeneyraiffa1976,
	author = {Keeney, R. L. and Raiffa, H.},
	title = {Decisions with multiple objectives: Preferences and
		 value tradeoffs},
	publisher = {John Wiley \& Sons},
	year = {1976},
}

@article{koksalanetal2009,
	title = {A new outranking-based approach for assigning alternatives
		 to ordered classes},
	author = {K\"oksalan, M. and Mousseau, V. and \"Ozpeynirci, \"O.
		  and \"Ozpeynirci, S. B.},
	year = {2009},
	journal = {Naval Research Logistics},
	pages = {74--85},
	publisher = {John Wiley \& Sons},
}

@incollection{leroyetal2011,
	title = {Learning the Parameters of a Multiple Criteria Sorting
		 Method},
	author = {Leroy, A. and Mousseau, V. and Pirlot, M.},
	booktitle = {Algorithmic Decision Theory},
	series = {Lecture Notes in Artificial Intelligence},
	editor = {Brafman, R. and Roberts, F. and Tsouki\`as, A.},
	publisher = {Springer},
	pages = {219--233},
	volume = {6992},
	year = {2011}
}

@incollection{massagliaostanello1991,
	author = {Massaglia, R. and Ostanello, A.},
	title = {N-tomic: a support system for multicriteria segmentation
		 problems},
	booktitle = {Multiple Criteria Decision Support},
	editor = {Korhonen, P. and Lewandowski, A. and Wallenius, J.},
	publisher = {Springer},
	year = {1991},
	pages = {167--174},
}

@article{meyerolteanu2013,
	author = {Meyer, P. and Olteanu, A.-L.},
	title = {Formalizing and solving the problem of clustering in {MCDA}},
	journal = {European Journal of Operational Research},
	volume = {227},
	number = {3},
	pages = {494--502},
	year = {2013},
	url = {http://dx.doi.org/10.1016/j.ejor.2013.01.016},
	doi = {10.1016/j.ejor.2013.01.016},
}

@article{moscarolaroy1977,
	title = {Proc\'edure automatique d'examen de dossiers fond\'ee sur
		 une segmentation trichotomique en pr\'esence de crit\`eres
		 multiples},
	author = {Moscarola, J. and Roy, B.},
	journal = {RAIRO Recherche Op\'erationnelle},
	year = {1977},
	volume = {11},
	number = {2},
	pages = {145--173},
}

@article{mousseauetal2001,
	title = {Using assignment examples to infer weights for
		 {ELECTRE TRI} method: Some experimental results},
	author = {Mousseau, V. and Figueira, J. R. and Naux, J.-Ph.},
	journal = {European Journal of Operational Research},
	volume = {130},
	number = {1},
	pages = {263--275},
	year = {2001},
	publisher = {Elsevier}
}

@article{mousseauetal2006,
	author = {Mousseau, V. and Dias, L. C. and Figueira, J. R.},
	title = {Dealing with inconsistent judgments in multiple criteria
		 sorting models},
	journal = {4OR},
	volume = {4},
	number = {2},
	pages = {145--158},
	year = {2006},
	url = {http://dx.doi.org/10.1007/s10288-005-0076-8},
}

@article{ngothemousseau2002,
	title = {Using assignment examples to infer category limits for
		 the {ELECTRE TRI} method},
	author = {Ngo The, A. and Mousseau, V.},
	journal = {Journal of Multi-criteria Decision Analysis},
	volume = {11},
	number = {1},
	pages = {29--43},
	year = {2002},
	publisher = {John Wiley \& Sons}
}

@article{rogersbruen1998,
	title = {Choosing realistic values of indifference, preference and
		 veto thresholds for use with environmental criteria within
		 {ELECTRE}},
	author = {Rogers, M. and Bruen, M.},
	journal = {European Journal of Operational Research},
	volume = {107},
	number = {3},
	pages = {542--551},
	year = {1998},
	note = {},
	issn = {0377-2217},
	doi = {http://dx.doi.org/10.1016/S0377-2217(97)00175-6},
}

@inproceedings{roybertier1973,
	author = {Roy, B. and Bertier, P.},
	year = {1973},
	title = {La m\'ethode {ELECTRE II} - Une application au
		 m\'edia-planning},
	editor = {Ross M.},
	booktitle = {OR'72},
	publisher = {North-Holland Publishing Company},
	pages = {291--302},
}

@article{roy1968,
	author = {Roy, B.},
	title = {Classement et choix en pr\'{e}sence de points de vue
		 multiples: La m\'ethode {ELECTRE}},
	journal = {Revue Francaise d'Informatique et de Recherche
		   Op\'{e}rationnelle},
	volume = {8},
	year = {1968},
	pages = {57--75},
}

@book{roy1985,
	title = {M\'ethodologie multicrit\`ere d'aide \`a la d\'ecision},
	author = {Roy, B.},
	year = {1985},
	publisher = {Economica}
}

@article{roy1991,
	title = {The outranking approach and the foundations of {ELECTRE}
		 methods},
	author = {Roy, B.},
	year = {1991},
	journal = {Theory and Decision},
	volume = {31},
	pages = {49--73}
}

@book{roy1996,
	title = {Multicriteria methodology for decision aiding},
	author = {Roy, B.},
	year = {1996},
	publisher = {Springer}
}

@book{roybouyssou1993,
	title = {Aide multicrit\`ere \`a la d\'ecision: m\'ethodes et cas},
	author = {Roy, B. and Bouyssou, D.},
	year = {1993},
	publisher = {Economica Paris}
}

@book{simosmaystre1990,
	title = {\'Evaluer l’impact sur l’environnement. Une approche originale
		 par l’analyse multicritère et la négotiation},
	author = {Simos, J. and Maystre, L. Y.},
	publisher = {Presses Polytechniques et Universitaires Romandes},
	year = {1990},
}

@book{vincke1992,
	title = {Multicriteria decision-aid},
	author = {Vincke, P.},
	year = {1992},
	publisher = {John Wiley \& Sons}
}

@phdthesis{yu1992,
	author = {W. Yu},
	title = {Aide multicrit\`ere \`a la d\'ecision dans le cadre de la
		 probl\'ematique du tri: m\'ethodes et applications},
	school = {LAMSADE, Universit\'e Paris Dauphine, Paris},
	year = {1992}
}

@inproceedings{sobrieetal2013d,
	title = {Learning a majority rule model from large sets
		 of assignment examples},
	author = {Sobrie, O. and Mousseau, V. and Pirlot, M.},
	booktitle = {Algorithmic Decision Theory},
	series = {Lecture Notes in Artificial Intelligence},
	year = {2013},
	pages = {336--350},
	editor = {Perny, P. and Pirlot, M. and Tsouki\'as, A.},
	publisher = {Springer},
	address = {Brussels, Belgium},
}

@inproceedings{sobrieetal2015b,
	title = {Learning the Parameters of a Non Compensatory Sorting Model},
	author = {Sobrie, O. and Mousseau, V. and Pirlot, M.},
	booktitle = {Algorithmic Decision Theory},
	series = {Lecture Notes in Artificial Intelligence},
	year = {2015},
	pages = {153--170},
	editor = {Walsh, T.},
	publisher = {Springer},
	address = {Lexington, KY, USA},
}

@article{ErsekEtAlDAM2017,
  author    = {E. {Ersek Uyan{\i}k} and
               O. Sobrie and
               V. Mousseau and
               M. Pirlot},
  title     = {Enumerating and categorizing positive Boolean functions separable by a $k$-additive capacity},
  journal   = {Discrete Applied Mathematics},
  volume    = {229},
  pages     = {17--30},
  year      = {2017}
}

@ARTICLE{Sobrie2018,
  author       = "Sobrie, O. and Mousseau, V. and Pirlot, M.",
  title        = "Learning monotone preferences using a majority rule sorting model",
  journal      = "International Transactions in Operational Research",
  year         = "2019",
  volume       = "26",
  number       = "5",
  pages        = "1786--1809",
}

\end{document}